\documentclass[10pt,twocolumn,twoside]{IEEEtran}

\usepackage[T1]{fontenc}
\usepackage[utf8]{inputenc}

\usepackage{relsize} %

\usepackage{amsmath}
\usepackage{amssymb, subfigure, url, doi, xspace}

\usepackage{mathtools}

\usepackage{xr}
\usepackage{ifpdf}
\ifpdf
  \usepackage[final]{graphicx}
  \DeclareGraphicsExtensions{.pdf}
\else
  \usepackage[final]{graphicx}
  \DeclareGraphicsExtensions{.pstex,.ps,.eps}
\fi
\usepackage{color}

\graphicspath{{Figures/}}

\hypersetup{bookmarks=false,hyperfigures=false,hidelinks=true}

\newcommand{\ie}{i.e.\xspace}
\newcommand{\eg}{e.g.\xspace}
\newcommand{\etal}{et al.\xspace}
\newcommand{\eq}[1]{Eq.~(\ref{eq:#1})}
\newcommand{\eqc}[1]{(\ref{eq:#1})}
\newcommand{\fig}[1]{Fig.~\ref{fig:#1}}
\newcommand{\figs}[1]{Figs.~\ref{fig:#1}}
\newcommand{\fign}[1]{\ref{fig:#1}}
\newcommand{\tbl}[1]{Table~\ref{tab:#1}}
\newcommand{\sctn}[1]{Sec.~\ref{sec:#1}}
\newcommand{\sctnc}[1]{~\ref{sec:#1}}

\newcommand{\mb}[1]{\mathbf{#1}}

\newcommand{\mbb}[1]{\mathbb{#1}}
\newcommand{\mc}[1]{\mathcal{#1}}

\newcommand{\sign}{\mathop{\mathrm{sign}}}

\newcommand{\prox}{{\mathrm{prox}}}
\providecommand{\abs}[1]{\left\lvert#1\right\rvert}
\providecommand{\norm}[1]{\left\lVert#1\right\rVert}
\DeclarePairedDelimiterX{\normsz}[1]{\lVert}{\rVert}{#1}

\DeclareMathOperator*{\argmin}{arg\,min}

\DeclareRobustCommand{\orderof}{\ensuremath{\mathcal{O}}}

\def \co {\mathcal{O}}

\date{\today}

\begin{document}

\title{Convolutional Dictionary Learning: A Comparative Review and New Algorithms}

\author{Cristina Garcia-Cardona
  and Brendt Wohlberg
  \thanks{C. Garcia-Cardona is with CCS Division, Los Alamos National Laboratory,
  Los Alamos, NM 87545, USA. Email:
  \texttt{cgarciac@lanl.gov}
}%
  \thanks{B. Wohlberg is with Theoretical Division, Los Alamos
    National Laboratory, Los Alamos, NM 87545, USA. Email:
    \texttt{brendt@ieee.org}
}%
  \thanks{This research was supported by the U.S. Department
    of Energy through the LANL/LDRD Program.}%
}

\maketitle

\begin{abstract}
Convolutional sparse representations are a form of sparse representation with a dictionary that has a structure that is equivalent to convolution with a set of linear filters. While effective algorithms have recently been developed for the convolutional sparse coding problem, the corresponding dictionary learning problem is substantially more challenging. Furthermore, although a number of different approaches have been proposed, the absence of thorough comparisons between them makes it difficult to determine which of them represents the current state of the art. The present work both addresses this deficiency and proposes some new approaches that outperform existing ones in certain contexts.  A thorough set of performance comparisons indicates a very wide range of performance differences among the existing and proposed methods, and clearly identifies those that are the most effective.
\end{abstract}

\begin{IEEEkeywords}
Sparse Representation, Sparse Coding, Dictionary Learning,
Convolutional Sparse Representation
\end{IEEEkeywords}

\IEEEpeerreviewmaketitle

\section{Introduction}
\label{sec:intro}

Sparse representations~\cite{mairal-2014-sparse} have become one of the most widely used and successful models for inverse problems in signal processing, image processing, and computational imaging. The reconstruction of a signal $\mb{s}$ from a sparse representation $\mb{x}$ with respect to \emph{dictionary} matrix $D$ is linear, i.e. $\mb{s} \approx D \mb{x}$, but computing the sparse representation given the signal, referred to as \emph{sparse coding}, usually involves solving an optimization problem\footnote{We do not consider the analysis form~\cite{figueiredo-2017-synthesis} of sparse representations in this work, focusing instead on the more common synthesis form.}. When solving problems involving images of any significant size, these representations are typically independently applied to sets of overlapping image patches due to the intractability of learning an unstructured dictionary matrix $D$ mapping to a vector space with the dimensionality of the number of pixels in an entire image.

The convolutional form of sparse representations replaces the unstructured dictionary $D$ with a set of linear filters $\{\mb{d}_m\}$. In this case the reconstruction of $\mb{s}$ from representation $\{\mb{x}_m\}$ is $\mb{s} \approx \sum_m \mb{d}_m \ast \mb{x}_m$, where $\mb{s}$ can be an entire image instead of a small image patch. This form of representation was first introduced some time ago under the label \emph{translation-invariant sparse representations}~\cite{lewicki-1999-coding}, but has recently enjoyed a revival of interest as \emph{convolutional sparse representations}, inspired by \emph{deconvolutional networks}~\cite{zeiler-2010-deconvolutional} (see~\cite[Sec. II]{wohlberg-2016-efficient}). This interest was spurred by the development of more efficient methods for the computationally-expensive \emph{convolutional sparse coding} (CSC) problem~\cite{chalasani-2013-fast, bristow-2013-fast, wohlberg-2014-efficient, heide-2015-fast}, and has led to a number of applications in which the convolutional form provides state-of-the-art performance~\cite{gu-2015-convolutional, liu-2016-image, zhang-2016-convolutional, quan-2016-compressed, serrano-2016-convolutional, zhang-2017-convolutional}.

The current leading CSC algorithms~\cite{wohlberg-2014-efficient, heide-2015-fast, wohlberg-2016-boundary} are all based on the Alternating Direction Method of Multipliers (ADMM)~\cite{boyd-2010-distributed}, which decomposes the problem into two subproblems, one of which is solved by soft-thresholding, and the other having a very efficient non-iterative solution in the DFT domain~\cite{wohlberg-2014-efficient}. The design of convolutional dictionary learning (CDL) algorithms is less straightforward. These algorithms adopt the usual approach for standard dictionary learning, alternating between a sparse coding step that updates the sparse representation of the training data given the current dictionary, and a dictionary update step that updates the current dictionary given the new sparse representation. It is the inherent computational cost of the latter update that makes the CDL problem more difficult than the CSC problem.

Most recent batch-mode\footnote{We do not consider the very recent online CDL algorithms~\cite{liu-2017-online, degraux-2017-online, wang-2017-online, liu-2018-online} in this work.} CDL algorithms share the structure introduced in~\cite{bristow-2013-fast} (and described in more detail in~\cite{kong-2014-fast}), the primary features of which are the use of Augmented Lagrangian methods and the solution of the most computationally expensive subproblems in the frequency domain. Earlier algorithms exist (see~\cite[Sec. II.D]{wohlberg-2016-efficient} for a thorough literature review), but since they are less effective, we do not consider them here, focusing on subsequent methods:
\begin{IEEEitemize}
  \item[\cite{wohlberg-2016-efficient}] Proposed a number of improvements on the algorithm of~\cite{bristow-2013-fast}, including more efficient sparse representation and dictionary updates, and a different Augmented Lagrangian structure with better convergence properties (examined in more detail in~\cite{garcia-2017-subproblem}).
  \item[\cite{sorel-2016-fast}] Proposed a number of dictionary update methods that lead to CDL algorithms with better performance than that of~\cite{bristow-2013-fast}.
  \item[\cite{heide-2015-fast}] Proposed a CDL algorithm that allows the inclusion of a spatial mask in the data fidelity term by exploiting the mask decoupling technique~\cite{almedia-2013-deconvolving}.
  \item[\cite{wohlberg-2016-boundary}] Proposed an alternative masked CDL algorithm that has much lower memory requirements than that of~\cite{heide-2015-fast}, and that converges faster in some contexts.
\end{IEEEitemize}
Unfortunately, due to the absence of any thorough performance comparisons between all of them (for example,~\cite{sorel-2016-fast} provides comparisons with~\cite{bristow-2013-fast} but not~\cite{wohlberg-2016-efficient}), as well as due to the absence of a careful exploration of the optimum choice of algorithm parameters in most of these works, it is difficult to determine which of these methods truly represents the state of the art in CDL.

Three other very recent methods do not receive the same thorough attention as those listed above. The algorithm of \cite{jas-2017-learning} addresses a variant of the CDL problem that is customized for neural signal processing and not relevant to most imaging applications, and \cite{papayan-2017-convolutional, chun-2017-convolutional} appeared while we were finalizing this paper, so that it was not feasible to include them in our analysis or our main set of experimental comparisons. However, since the authors of~\cite{papayan-2017-convolutional} have made an implementation of their method publicly available, we do include this method in some additional performance comparisons in~\sctn{largecompare} to~\sctnc{otheralgcompare} of the Supplementary Material.

The main contributions of the present paper are:
\begin{IEEEitemize}
\item Providing a thorough performance comparison among the different methods proposed in~\cite{wohlberg-2016-efficient, sorel-2016-fast, heide-2015-fast, wohlberg-2016-boundary}, allowing reliable identification of the most effective algorithms.
\item Demonstrating that two of the algorithms proposed in~\cite{sorel-2016-fast}, with very different derivations, are in fact closely related and fall within the same class of algorithm.
\item Proposing a new approach for the CDL problem without a spatial mask that outperforms all existing methods in a serial processing context.
\item Proposing new approaches for the CDL problem with a spatial mask that respectively outperform existing methods in serial and parallel processing contexts.
\item Carefully examining the sensitivity of the considered CDL algorithms to their parameters, and proposing simple heuristics for parameter selection that provide good performance.
\end{IEEEitemize}

\section{Convolutional Dictionary Learning}
\label{sec:cdl}

CDL is usually posed in the form of the problem
\begin{align}
  \argmin_{\{\mb{d}_{m}\},\{\mb{x}_{m,k}\}} & \frac{1}{2} \sum_k
  \normsz[\Big]{\sum_m \mb{d}_m \! \ast \mb{x}_{m,k} \!- \mb{s}_k}_2^2 \!\!+
  \lambda \sum_{m,k}\norm{\mb{x}_{m,k}}_1 \nonumber \\
  & \text{ such that } \norm{\mb{d}_{m}}_2 = 1 \; \forall m \;,
\label{eq:cdl}
\end{align}
where the constraint on the norms of filters $\mb{d}_{m}$ is required to avoid the scaling ambiguity between filters and coefficients\footnote{The constraint $\norm{\mb{d}_{m}}_2 \leq 1$ is frequently used instead of $\norm{\mb{d}_{m}}_2 = 1$. In practice this does not appear to make a significant difference to the solution.}. The training images $\mb{s}_k$ are considered to be $N$ dimensional vectors, where $N$ is the number of pixels in each image, and we denote the number of filters and the number of training images by $M$ and $K$ respectively. This problem is non-convex in both variables $\{\mb{d}_{m}\}$ and $\{\mb{x}_{m,k}\}$, but is convex in $\{\mb{x}_{m,k}\}$ with $\{\mb{d}_{m}\}$ constant, and vice versa. As in standard (non-convolutional) dictionary learning, the usual approach to minimizing this functional is to alternate between updates of the sparse representation and the dictionary. The design of a CDL algorithm can therefore be decomposed into three components: the choice of sparse coding algorithm, the choice of dictionary update algorithm, and the choice of coupling mechanism, including how many iterations of each update should be performed before alternating, and which of their internal variables should be transferred across when alternating.

\subsection{Sparse Coding}
\label{sec:mmvcsc}

While a number of greedy matching pursuit type algorithms were developed for translation-invariant sparse representations~\cite[Sec. II.C]{wohlberg-2016-efficient}, recent algorithms have largely concentrated on a convolutional form of the standard Basis Pursuit DeNoising (BPDN)~\cite{chen-1998-atomic} problem
\begin{equation}
  \argmin_{\mb{x}}  \; (1/2) \norm{ D \mb{x}  - \mb{s} }_2^2 + \lambda
  \norm{\mb{x}}_1 \;.
\label{eq:bpdn}
\end{equation}
This form, which we will refer to as Convolutional BPDN (CBPDN), can be written as
\begin{equation}
  \argmin_{\{\mb{x}_m\}} \frac{1}{2} \normsz[\Big]{\sum_m
    \mb{d}_m \ast \mb{x}_m - \mb{s}}_2^2 + \lambda \sum_m \norm{\mb{x}_m}_1 \;.
\label{eq:cbpdn}
\end{equation}
If we define $D_m$ such that $D_m \mb{x}_m = \mb{d}_m \ast \mb{x}_m$, and
\begin{equation}
  D = \left( \begin{array}{ccc}D_0 & D_1 &
     \ldots \end{array} \right) \qquad  \mb{x} =
 \left( \begin{array}{c}  \mb{x}_0\\
          \mb{x}_1\\
          \vdots \end{array} \right) \;,
\end{equation}
we can rewrite the CBPDN problem in standard BPDN form~\eq{bpdn}. The Multiple Measurement Vector (MMV) version of CBPDN, for multiple images, can be written as
\begin{align}
  \argmin_{\{\mb{x}_{m,k}\}} & \frac{1}{2} \sum_k
  \normsz[\Big]{\sum_m \mb{d}_m \!\ast\! \mb{x}_{m,k} - \mb{s}_k}_2^2 +
  \lambda \sum_{m,k} \norm{\mb{x}_{m,k}}_1 \;,
  \label{eq:cbpdnmmv}
\end{align}
where $\mb{s}_k$ is the $k^{\text{th}}$ image, and $\mb{x}_{m,k}$ is the coefficient map corresponding to the  $m^{\text{th}}$ dictionary filter and the $k^{\text{th}}$ image. By defining
\begin{equation}
  X = \left( \begin{array}{ccc}  \mb{x}_{0,0} & \mb{x}_{0,1} & \ldots \\
         \mb{x}_{1,0} &  \mb{x}_{1,1} & \ldots \\
         \vdots & \vdots & \ddots \end{array} \right)
\quad
     S = \left( \begin{array}{ccc}
                  \mb{s}_0 & \mb{s}_1 & \ldots
                \end{array} \right)  \;,
\end{equation}
we can rewrite~\eq{cbpdnmmv} in the standard BPDN MMV form,
\begin{equation}
  \argmin_{X}  \; (1/2) \norm{ D X - S }_F^2 + \lambda \norm{X}_1 \;.
  \label{eq:bpdnmmv}
\end{equation}
Where possible, we will work with this form of the problem instead of~\eq{cbpdnmmv} since it simplifies the notation, but the reader should keep in mind that $D$, $X$, and $S$ denote the specific block-structured matrices defined above.

The most effective solution for solving~\eq{cbpdnmmv} is currently based on ADMM\footnote{It is worth noting, however, that a solution based on FISTA with the gradient computed in the frequency domain, while generally less effective than the ADMM solution, exhibits a relatively small performance difference for the larger $\lambda$ values typically used for CDL~\cite[Sec. IV.B]{wohlberg-2016-efficient}.}~\cite{boyd-2010-distributed}, which solves problems of the form
\begin{equation}
\argmin_{\mb{x},\mb{y}} f(\mb{x}) + g(\mb{y}) \; \text{ such that } \;
A \mb{x} + B \mb{y} = \mb{c}
  \label{eq:admmform}
\end{equation}
by iterating over the steps
{\small
  \setlength{\abovedisplayskip}{6pt}
  \setlength{\belowdisplayskip}{\abovedisplayskip}
  \setlength{\abovedisplayshortskip}{0pt}
  \setlength{\belowdisplayshortskip}{3pt}
\begin{align}
\mb{x}^{(i+1)} &= \argmin_{\mb{x}} f(\mb{x}) + \frac{\rho}{2} \norm{A
  \mb{x} + B \mb{y}^{(i)} - \mb{c} + \mb{u}^{(i)}}_2^2 \label{eq:admmxup} \\
\mb{y}^{(i+1)} &= \argmin_{\mb{y}} g(\mb{y}) + \frac{\rho}{2} \norm{A
  \mb{x}^{(i+1)} + B \mb{y} - \mb{c} + \mb{u}^{(i)}}_2^2 \label{eq:admmyup} \\
\mb{u}^{(i+1)} &=  \mb{u}^{(i)} + A \mb{x}^{(i+1)} + B  \mb{y}^{(i+1)}
- \mb{c}  \label{eq:admmuup} \;,
\end{align}
}%
where \emph{penalty parameter} $\rho$ is an algorithm parameter that plays an important role in determining the convergence rate of the iterations, and $\mb{u}$ is the \emph{dual variable} corresponding to the constraint $A \mb{x} + B \mb{y} = \mb{c}$.  We can apply ADMM to problem~\eq{bpdnmmv} by \emph{variable splitting}, introducing an auxiliary variable $Y$ that is constrained to be equal to the primary variable $X$, leading to the equivalent problem \vspace{-1mm}
\begin{equation}
  \argmin_{X, Y}  \; (1/2) \norm{ D X - S }_F^2 + \lambda
  \norm{Y}_1 \;\; \text{ s.t. } \;\; X = Y \;,
  \label{eq:bpdnmmvsplit}
\end{equation}
for which we have the ADMM iterations
{\small
  \setlength{\abovedisplayskip}{6pt}
  \setlength{\belowdisplayskip}{\abovedisplayskip}
  \setlength{\abovedisplayshortskip}{0pt}
  \setlength{\belowdisplayshortskip}{3pt}
\begin{align}
  X^{(i+1)} &= \argmin_{X} \frac{1}{2}
  \normsz[\big]{D X \!-\! S}_F^2 +  \frac{\rho}{2} \norm{
    X - Y^{(i)} \! + U^{(i)}}_F^2 \label{eq:bpdnxprob} \\
  Y^{(i+1)} &= \argmin_{Y} \lambda \norm{Y}_1 +  \frac{\rho}{2}
 \norm{ X^{(i+1)} - Y + U^{(i)}}_F^2 \label{eq:bpdnyprob}  \\
  U^{(i+1)} &= U^{(i)} + X^{(i+1)} \! - Y^{(i+1)} \;. \label{eq:bpdnuprob}
\end{align}
}%

Step~\eq{bpdnuprob} involves simple arithmetic, and step~\eq{bpdnyprob} has a closed-form solution
\begin{equation}
 Y^{(i+1)} = \mc{S}_{\lambda/\rho}\left( X^{(i+1)} + U^{(i)} \right) \;,
\label{eq:bpdnysoln}
\end{equation}
where $\mc{S}_{\gamma}(\cdot)$ is the soft-thresholding function~\cite[Sec. 6.5.2]{parikh-2014-proximal}
\begin{equation}
\mc{S}_{\gamma}(V) = \sign(V) \odot \max(0, \abs{V} - \gamma) \;,
\end{equation}
with $\sign(\cdot)$ and $\abs{\cdot}$ of a vector considered to
be applied element-wise, and $\odot$ denoting element-wise
multiplication. The most computationally expensive step is~\eq{bpdnxprob}, which requires solving the linear system
\begin{equation}
(D^T D + \rho I) X = D^T S + \rho (Y - U) \;.
\label{eq:xlinsys}
\end{equation}
Since $D^T D$ is a very large matrix, it is impractical to solve this linear system using the approaches that are effective when $D$ is not a convolutional dictionary. It is possible, however, to exploit the FFT for efficient
implementation of the convolution via the DFT convolution theorem. Transforming~\eq{xlinsys} into the DFT domain gives
\begin{equation}
(\hat{D}^H \hat{D} + \rho I) \hat{X} = \hat{D}^H \hat{S} + \rho (\hat{Y} - \hat{U}) \;,
\label{eq:xlinsysdft}
\end{equation}
where $\hat{Z}$ denotes the DFT of variable $Z$. Due to the structure of $\hat{D}$, which consists of concatenated diagonal matrices $\hat{D}_m$, linear system~\eq{xlinsysdft} can be decomposed into a set of $N K$ independent linear systems~\cite{bristow-2013-fast}, each of which has a left hand side consisting of a diagonal matrix plus a rank-one component, which can be solved very efficiently by exploiting the Sherman-Morrison formula~\cite{wohlberg-2014-efficient}.

\subsection{Dictionary Update}
\label{sec:dictup}

In developing the dictionary update, it is convenient to switch the indexing of the coefficient map from $\mb{x}_{m,k}$ to $\mb{x}_{k,m}$, writing the problem as
\begin{align}
  \argmin_{\{\mb{d}_m\}} \frac{1}{2}  \sum_k \normsz[\Big]{\sum_m
    \mb{x}_{k,m} \!\ast\! \mb{d}_m \!-\!
    \mb{s}_k}_2^2  \text{ s.t. }  \norm{\mb{d}_{m}}_2 = 1 \;,
\end{align}
which is a convolutional form of Method of Optimal Directions (MOD)~\cite{engan-1999-method} with a constraint on the filter normalization. As for CSC, we will develop the algorithms for solving this problem in the spatial domain, but will solve the critical sub-problems in the frequency domain. We want to solve for $\{\mb{d}_m\}$ with a relatively small support, but when computing convolutions in the frequency domain, we need to work with $\mb{d}_m$ that have been zero-padded to the common spatial dimensions of $\mb{x}_{k,m}$ and $\mb{s}_k$. The most straightforward way of dealing with this complication is to consider the $\mb{d}_m$ to be zero-padded and add a constraint that requires that they be zero outside of the desired support. If we denote the projection operator that zeros the regions of the filters outside of the desired support by $P$, we can write a constraint set that combines this support constraint with the normalization constraint as
\begin{equation}
  \label{eq:csetpn}
  C_{\text{PN}} = \{ \mb{x} \in \mbb{R}^N : (I - P) \mb{x} = 0, \;
  \norm{\mb{x}}_2 = 1\} \;,
\end{equation}
and write the dictionary update as
\begin{align}
  \argmin_{\{\mb{d}_m\}} \frac{1}{2}  \sum_k \normsz[\Big]{\sum_m
    \mb{x}_{k,m} \ast \mb{d}_m -
    \mb{s}_k}_2^2 \; \text{ s.t. } \; \mb{d}_{m} \in C_{\text{PN}} \; \forall m
   \;.
  \label{eq:ccmodcpn}
\end{align}
Introducing the indicator function $\iota_{C_{\text{PN}}}$ of the constraint set $C_{\text{PN}}$, where the indicator function of a set $S$ is defined as
\begin{equation}
\iota_S(X) = \left\{ \begin{array}{ll}
    0 & \text{ if } X \in S\\
    \infty & \text{ if } X \notin S
    \end{array} \right. \;,
\end{equation}
allows~\eq{ccmodcpn} to be written in unconstrained form~\cite{afonso-2011-augmented}
\begin{align}
  \argmin_{\{\mb{d}_m\}} \frac{1}{2}  \sum_k \normsz[\Big]{\sum_m
    \mb{x}_{k,m} \!\ast\! \mb{d}_m -
    \mb{s}_k}_2^2 + \sum_m \iota_{C_{\text{PN}}} (\mb{d}_m) \;.
\end{align}

Defining $X_{k,m}$ such that $X_{k,m} \mb{d}_m = \mb{x}_{k,m} \ast \mb{d}_m$ and
\begin{equation}
\addtolength{\arraycolsep}{-0.6mm}
X_k = \left( \begin{array}{ccc}
    X_{k,0} & X_{k,1} & \ldots
  \end{array} \right)
\quad
\mb{d} = \left( \begin{array}{c}  \mb{d}_0\\ \mb{d}_1\\
    \vdots  \end{array} \right) \;,
\label{eq:xddef}
\end{equation}
this problem can be expressed as
\begin{align}
  \argmin_{\mb{d}} \; (1/2)  \sum_k \normsz[\big]{X_k \mb{d} -
    \mb{s}_k}_2^2 + \iota_{C_{\text{PN}}} (\mb{d}) \;,
\end{align}
or, by defining
\begin{equation}
\addtolength{\arraycolsep}{-0.6mm}
X = \left( \begin{array}{ccc}
             X_{0,0} & X_{0,1} & \ldots \\
             X_{1,0} & X_{1,1} & \ldots \\
             \vdots & \vdots & \ddots
  \end{array} \right)
\quad
\mb{s} = \left( \begin{array}{c}  \mb{s}_0\\ \mb{s}_1\\
    \vdots  \end{array} \right) \;,
\label{eq:duxs}
\end{equation}
as
\begin{align}
  \argmin_{\mb{d}} \; (1/2)  \normsz[\big]{X \mb{d} -
    \mb{s}}_2^2 + \iota_{C_{\text{PN}}} (\mb{d}) \;.
\label{eq:dupprob}
\end{align}

Algorithms for solving this problem will be discussed in~\sctn{dctup}. A common feature of most of these methods is the need to solve a linear system that includes the data fidelity term $(1/2) \norm{X \mb{d} - \mb{s}}_2^2$. As in the case of the $X$ step~\eq{bpdnxprob} for CSC, this problem can be solved in the frequency domain, but there is a critical difference: $\hat{X}^H \hat{X}$ is composed of independent components of rank $K$ instead of rank 1, so that the very efficient Sherman Morrison solution cannot be directly exploited. It is this property that makes the dictionary update inherently more computationally expensive than the sparse coding stage, complicating the design of algorithms, and leading to the present situation in which there is far less clarity as to the best choice of dictionary learning algorithm than there is for the choice of the sparse coding algorithm.

\subsection{Update Coupling}

Both the sparse coding and dictionary update stages are typically solved via iterative algorithms, and many of these algorithms have more than one working variable that can be used to represent the current solution. The major design choices in coupling the alternating optimization of these two stages are therefore:
\begin{IEEEenumerate}
  \item how many iterations of each subproblem to perform before switching to the other subproblem, and
  \item which working variable from each subproblem to pass across to the other subproblem.
\end{IEEEenumerate}

Since these issues are addressed in detail in~\cite{garcia-2017-subproblem}, we only summarize the conclusions here:
\begin{IEEEitemize}
  \item When both subproblems are solved by ADMM algorithms, most authors have  coupled the subproblems via the primary variables (corresponding, for example, to $X$ in~\eq{bpdnmmvsplit}) of each ADMM algorithm.
  \item This choice tends to be rather unstable, and requires either multiple iterations of each subproblem before alternating, or very large penalty parameters, which can lead to slow convergence.
  \item The alternative strategy of coupling the subproblems via the auxiliary variables (corresponding, for example, to $Y$ in~\eq{bpdnmmvsplit}) of each ADMM algorithm tends to be more stable, not requiring multiple iterations before alternating, and converging faster.
\end{IEEEitemize}

\section{Dictionary Update Algorithms}
\label{sec:dctup}

Since the choice of the best CSC algorithm is not in serious dispute, the focus of this work is on the choice of dictionary update algorithm.

\subsection{ADMM with Equality Constraint}
\label{sec:admmeq}

The simplest approach to solving~\eq{dupprob} via an ADMM algorithm is to apply the variable splitting
\begin{align}
  \argmin_{\mb{d}, \mb{g}} \; (1/2)  \normsz[\big]{X \mb{d} -
    \mb{s}}_2^2 + \iota_{C_{\text{PN}}} (\mb{g}) \; \text{ s.t. } \; \mb{d} = \mb{g} \;,
\label{eq:ccmodeqsplit}
\end{align}
for which the corresponding ADMM iterations are
{\small
  \setlength{\abovedisplayskip}{6pt}
  \setlength{\belowdisplayskip}{\abovedisplayskip}
  \setlength{\abovedisplayshortskip}{0pt}
  \setlength{\belowdisplayshortskip}{3pt}
\begin{align}
  \mb{d}^{(i+1)} &= \argmin_{\mb{d}} \frac{1}{2}
  \normsz[\big]{X \mb{d} - \mb{s}}_2^2 +  \frac{\sigma}{2} \norm{
    \mb{d} - \mb{g}^{(i)} + \mb{h}^{(i)}}_2^2 \label{eq:ccmoddprob} \\
  \mb{g}^{(i+1)} &= \argmin_{\mb{g}} \iota_{C_{\text{PN}}} (\mb{g}) +  \frac{\sigma}{2}
 \norm{ \mb{d}^{(i+1)} - \mb{g} + \mb{h}^{(i)}}_2^2 \label{eq:ccmodgprob}  \\
  \mb{h}^{(i+1)} &= \mb{h}^{(i)} + \mb{d}^{(i+1)} - \mb{g}^{(i+1)} \;. \label{eq:ccmodhprob}
\end{align}
}%

Step~\eq{ccmodgprob} is of the form
\begin{equation}
\argmin_{\mb{x}} \; (1/2)
\norm{\mb{x} - \mb{y}}_2^2 + \iota_{C_{\text{PN}}}(\mb{x}) =
\prox_{\iota_{C_{\text{PN}}}}(\mb{y}) \;.
\end{equation}
It is clear from the geometry of the problem that
\begin{equation}
  \prox_{\iota_{C_{\text{PN}}}}(\mb{y}) = \frac{P P^T \mb{y}}{\norm{P
      P^T \mb{y}}_2} \;,
\label{eq:proxpn1}
\end{equation}
or, if the normalization $\norm{\mb{d}_m}_2 \leq 1$ is desired
instead,
\begin{equation}
  \prox_{\iota_{C_{\text{PN}}}}(\mb{y}) =
\begin{cases}
P P^T \mb{y}  \;\;\;\;\;\;\;\, \text{ if } \norm{P P^T \mb{y}}_2 \leq 1\\
\frac{P P^T\mb{y}}{\norm{P P^T \mb{y}}_2} \;\;\;\; \text{ if }
\norm{P P^T \mb{y}}_2 > 1
\end{cases}  \;.
\label{eq:proxpn2}
\end{equation}

Step~\eq{ccmoddprob} involves solving the linear system
\begin{equation}
(X^T X + \sigma I) \mb{d} = X^T \mb{s} + \sigma (\mb{g} - \mb{h}) \;,
\end{equation}
which can be expressed in the DFT domain as
\begin{equation}
(\hat{X}^H \hat{X} + \sigma I) \hat{\mb{d}} = \hat{X}^H \hat{\mb{s}} + \sigma (\hat{\mb{g}} - \hat{\mb{h}}) \;.
\label{eq:ccmodxstepdft}
\end{equation}
This linear system can be decomposed into a set of $N$ independent linear systems, but in contrast to~\eq{xlinsysdft}, each of these has a left hand side consisting of a diagonal matrix plus a rank $K$ component, which precludes direct use of the Sherman-Morrison formula~\cite{wohlberg-2016-efficient}.

We consider three different approaches to solving these linear systems:

\subsubsection{Conjugate Gradient}
\label{sec:ducg}

An obvious approach to solving~\eq{ccmodxstepdft} without having to explicitly construct the matrix $\hat{X}^H \hat{X} + \sigma I$ is to apply an iterative method such as Conjugate Gradient (CG). The experiments reported in~\cite{wohlberg-2016-efficient} indicated that solving this system to a relative residual tolerance of $10^{-3}$ or better is sufficient for the dictionary learning algorithm to converge reliably. The number of CG iterations required can be substantially reduced by using the solution from the previous outer iteration as an initial value.

\subsubsection{Iterated Sherman-Morrison}
\label{sec:duits}

Since the independent linear systems into which~\eq{ccmodxstepdft} can be decomposed have a left hand side consisting of a diagonal matrix plus a rank $K$ component, one can iteratively apply the Sherman-Morrison formula to obtain a solution~\cite{wohlberg-2016-efficient}. This approach is very effective for small to moderate $K$, but performs poorly for large $K$ since the computational cost is $\orderof(K^2)$.

\subsubsection{Spatial Tiling}
\label{sec:dust}

When $K=1$ in~\eq{ccmodxstepdft}, the very efficient solution via the Sherman-Morrison formula is possible. As pointed out in~\cite{sorel-2016-fast}, a larger set of training images can be spatially tiled to form a single large image, so that the problem is solved with $K'=1$.

\subsection{Consensus Framework}
\label{sec:ducns}

In this section it is convenient to introduce different block-matrix and vector notation for the coefficient maps and dictionary, but we overload the usual symbols to emphasize their corresponding roles. We define $X_k$ as in~\eq{xddef}, but define
\begin{equation}
\addtolength{\arraycolsep}{-0.7mm}
X = \left( \begin{array}{ccc}
             X_0 & 0 & \ldots \\
             0 & X_1 & \ldots \\
             \vdots & \vdots & \ddots
  \end{array} \right) \;
\mb{d}_k = \left( \begin{array}{c}  \mb{d}_{0,k}\\ \mb{d}_{1,k}\\
    \vdots  \end{array} \right) \;
\mb{d} = \left( \begin{array}{c}  \mb{d}_{0}\\ \mb{d}_{1}\\
    \vdots  \end{array} \right)
\end{equation}
where $\mb{d}_{m,k}$ is distinct copy of dictionary filter $m$ corresponding to training image $k$.

As proposed in~\cite{sorel-2016-fast}, we can pose problem~\eq{dupprob} in the form of an ADMM consensus problem~\cite[Ch. 7]{boyd-2010-distributed}
\begin{align}
  \argmin_{\mb{d}_k} \;(1/2) & \sum_k \normsz[\big]{X_k \mb{d}_k -
    \mb{s}_k}_2^2 + \iota_{C_{\text{PN}}} (\mb{g}) \nonumber \\ & \text{s.t.} \;\;
   \mb{g} = \mb{d}_{k} \; \forall k  \;,
\label{eq:dupprobcns}
\end{align}
which can be written in standard ADMM form as
\begin{align}
  \argmin_{\mb{d}} \; \frac{1}{2} \normsz[\big]{X \mb{d} -
    \mb{s}}_2^2 + \iota_{C_{\text{PN}}} (\mb{g}) \;\;\; \text{s.t.} \;\;\;
  \mb{d} - E \mb{g} = 0 \;,
\label{eq:dupprobcnsblk}
\end{align}
where $E = \left( \begin{array}{ccc}  I & I & \hdots  \end{array} \right)^T$.

The corresponding ADMM iterations are
{\small
  \setlength{\abovedisplayskip}{6pt}
  \setlength{\belowdisplayskip}{\abovedisplayskip}
  \setlength{\abovedisplayshortskip}{0pt}
  \setlength{\belowdisplayshortskip}{3pt}
\begin{align}
  \mb{d}^{(i+1)} &= \argmin_{\mb{d}} \frac{1}{2}
  \normsz[\big]{X \mb{d} - \mb{s}}_2^2 +  \frac{\sigma}{2} \norm{
    \mb{d} - E \mb{g}^{(i)} + \mb{h}^{(i)}}_2^2 \label{eq:ccmoddprobcns} \\
  \mb{g}^{(i+1)} &= \argmin_{\mb{g}} \iota_{C_{\text{PN}}} (\mb{g}) +  \frac{\sigma}{2}
 \norm{ \mb{d}^{(i+1)} - E \mb{g} + \mb{h}^{(i)}}_2^2 \label{eq:ccmodgprobcns}  \\
  \mb{h}^{(i+1)} &= \mb{h}^{(i)} + \mb{d}^{(i+1)} - E \mb{g}^{(i+1)} \;.  \label{eq:ccmodhprobcns}
\end{align}
}%
Since $X$ is block diagonal,~\eq{ccmoddprobcns} can be solved as the $K$ independent problems
{\small
  \setlength{\abovedisplayskip}{6pt}
  \setlength{\belowdisplayskip}{\abovedisplayskip}
  \setlength{\abovedisplayshortskip}{0pt}
  \setlength{\belowdisplayshortskip}{3pt}
\begin{equation}
\mb{d}_k^{(i+1)} = \argmin_{\mb{d}_k} \frac{1}{2}
  \normsz[\Big]{X_k \mb{d}_k \!-\! \mb{s}_k}_2^2 +  \frac{\sigma}{2} \norm{
    \mb{d}_k \!-\! \mb{g}^{(i)} \!+\! \mb{h}_k^{(i)}}_2^2 \;,
\end{equation}
}%
each of which can be solved via the same efficient DFT-domain Sherman-Morrison method used for~\eq{bpdnxprob}. Subproblem~\eq{ccmodgprobcns} can be expressed as~\cite[Sec. 7.1.1]{boyd-2010-distributed}
{\small
  \setlength{\abovedisplayskip}{6pt}
  \setlength{\belowdisplayskip}{\abovedisplayskip}
  \setlength{\abovedisplayshortskip}{0pt}
  \setlength{\belowdisplayshortskip}{3pt}
\begin{align}
 \mb{g}^{(i+1)} =& \argmin_{\mb{g}} \iota_{C_{\text{PN}}} (\mb{g}) \,+ \nonumber \\ &  \frac{K \sigma}{2}
 \normsz[\bigg]{\mb{g}  - K^{-1} \bigg( \sum_{k=0}^{K-1} \mb{d}_k^{(i+1)} + \sum_{k=0}^{K-1} \mb{h}_k^{(i)} \bigg) }_2^2 \;,
\end{align}
}%
which has the closed-form solution
\begin{equation}
\mb{g}^{(i+1)} = \prox_{\iota_{C_{\text{PN}}}}\bigg(K^{-1} \Big( \sum_{k=0}^{K-1} \mb{d}_k^{(i+1)} + \sum_{k=0}^{K-1} \mb{h}_k^{(i)} \Big)\bigg) \;.
\label{eq:cnsgstepsoln}
\end{equation}

\subsection{3D / Frequency Domain Consensus}
\label{sec:dufdcn}

Like spatial tiling (see~\sctn{dust}), the ``3D'' method proposed in~\cite{sorel-2016-fast} maps the dictionary update problem with $K > 1$ to an equivalent problem for which $K' = 1$. The ``3D'' method achieves this by considering an array of $K$ 2D training images as a single 3D training volume.  The corresponding dictionary filters are also inherently 3D, but the constraint is modified to require that they are zero other than in the first 3D slice (this can be viewed as an extension of the constraint that the spatially-padded filters are zero except on their desired support) so that the final results is a set of 2D filters, as desired.

While ADMM consensus and ``3D'' were proposed as two entirely distinct methods~\cite{sorel-2016-fast}, it turns out they are closely related: the ``3D'' method is ADMM consensus with the data fidelity term and constraint expressed in the DFT domain. Since the notation is a bit cumbersome, the point will be illustrated for the $K=2$ case, but the argument is easily generalized to arbitrary $K$.

When $K=2$, the dictionary update problem can be expressed as
\begin{align}
  \argmin_{\mb{d}} \frac{1}{2}  \norm{
  \left( \begin{array}{c}  X_0\\ X_1\\
     \end{array} \right) \mb{d} -
  \left( \begin{array}{c}  \mb{s}_0\\ \mb{s}_1\\
    \end{array} \right)
  }_2^2 + \iota_{C_{\text{PN}}} (\mb{d})
   \;\;,
\end{align}
which can be rewritten as the equivalent problem\footnote{Equivalence when the constraints are satisfied is easily verified by multiplying out the matrix-vector product in the data fidelity term in~\eq{cns3dk2}.}
\begin{align}
  \addtolength{\arraycolsep}{-1.2mm}
  \argmin_{\mb{d}_0,\mb{d}_1} \frac{1}{2} &  \norm{
  \left( \begin{array}{cc}  X_0 & X_1 \\ X_1 & X_0\\
         \end{array} \right)
\left( \begin{array}{c}  \mb{d}_0\\ \mb{d}_1\\
     \end{array} \right)
  -
  \left( \begin{array}{c}  \mb{s}_0\\ \mb{s}_1\\
    \end{array} \right)
  }_2^2 + \iota_{C_{\text{PN}}} (\mb{g})
  \nonumber \\
 & \text{ s.t. } \;\; \mb{d}_0 = \mb{g} \quad \mb{d}_1 = \mb{0}
   \;,
\label{eq:cns3dk2}
\end{align}
where the constraint can also be written as
\begin{align}
\left( \begin{array}{c}  \mb{d}_0\\ \mb{d}_1\\
     \end{array} \right) = \left( \begin{array}{c}   I \\ 0 \\
     \end{array} \right) \mb{g} \;\;.
\end{align}
The general form of the matrix in~\eq{cns3dk2} is a block-circulant matrix constructed from the blocks $X_k$. Since the multiplication of the dictionary block vector by the block-circulant matrix is equivalent to convolution in an additional dimension, this equivalent problem represents the ``3D'' method.

Now, define the un-normalized $2 \times 2$ block DFT matrix operating in this extra dimension as
\begin{align}
F = \left( \begin{array}{rr}  I & I \\ I & -I
         \end{array} \right) \;,
\end{align}
and apply it to the objective function and constraint, giving
{
  \addtolength{\arraycolsep}{-1.3mm}
\begin{align}
  \argmin_{\mb{d}_0,\mb{d}_1} \frac{1}{2} &  \norm{
F  \left( \begin{array}{cc}  X_0 & X_1 \\ X_1 & X_0
         \end{array} \right) F^{-1} F
\left( \begin{array}{c}  \mb{d}_0\\ \mb{d}_1
     \end{array} \right)
  - F
  \left( \begin{array}{c}  \mb{s}_0\\ \mb{s}_1
    \end{array} \right)
  }_2^2 \nonumber \\ & + \iota_{C_{\text{PN}}} (\mb{g})
  \;\;\; \text{ s.t. } \;\;\;
   F   \left( \begin{array}{c}  \mb{d}_0\\ \mb{d}_1
     \end{array} \right) = F \left( \begin{array}{c}   I \\ 0 \\
     \end{array} \right) \mb{g}  \;.
\end{align}
}
Since the DFT diagonalises a circulant matrix, this is
{
\addtolength{\arraycolsep}{-1.5mm}
\begin{align}
  \argmin_{\mb{d}_0,\mb{d}_1} \frac{1}{2} &  \norm{
  \left( \begin{array}{cc}  X_0 \!+\! X_1 & 0 \\ 0 & X_0 \!-\! X_1
         \end{array} \right)
\left( \begin{array}{c}  \mb{d}_0 \!+\! \mb{d}_1 \\ \mb{d}_0 \!-\! \mb{d}_1
     \end{array} \right)
  -
  \left( \begin{array}{c}  \mb{s}_0 \!+\! \mb{s}_1 \\ \mb{s}_0 \!-\! \mb{s}_1
    \end{array} \right)
  }_2^2 \nonumber \\ & + \iota_{C_{\text{PN}}} (\mb{g})
      \;\;\; \text{ s.t. } \;\;\;
     \left( \begin{array}{c}  \mb{d}_0 + \mb{d}_1 \\ \mb{d}_0 - \mb{d}_1\\
     \end{array} \right) =  \left( \begin{array}{c}  \mb{g}  \\  \mb{g}\\
     \end{array} \right) \;.
\end{align}
}
In this form the problem is an ADMM consensus problem in variables
\begin{align}
X'_0 &= X_0 + X_1 & \mb{d}'_0 &= \mb{d}_0 + \mb{d}_1 &
  \mb{s}'_0 &= \mb{s}_0 + \mb{s}_1 \nonumber \\
X'_1 &= X_0 - X_1 & \mb{d}'_1 &= \mb{d}_0 - \mb{d}_1 &
  \mb{s}'_1 &= \mb{s}_0 - \mb{s}_1 \;.
\end{align}

\subsection{FISTA}
\label{sec:dufista}

The Fast Iterative Shrinkage-Thresholding Algorithm (FISTA)~\cite{beck-2009-fast}, an accelerated proximal gradient method, has been used for CSC~\cite{chalasani-2013-fast, wohlberg-2016-efficient, degraux-2017-online}, and in a recent online CDL algorithm~\cite{liu-2017-online}, but has not previously been considered for the dictionary update of a batch-mode dictionary learning algorithm.

The FISTA iterations for solving~\eq{dupprob} are
{\small
  \setlength{\abovedisplayskip}{6pt}
  \setlength{\belowdisplayskip}{\abovedisplayskip}
  \setlength{\abovedisplayshortskip}{0pt}
  \setlength{\belowdisplayshortskip}{3pt}
\begin{align}
  \mb{y}^{(i+1)} &= \prox_{\iota_{C_{\text{PN}}}}\bigg( \mb{d}^{(i)} - \frac{1}{L} \nabla_{\mb{d}} \Big(\frac{1}{2}
  \norm{X \mb{d} - \mb{s}}_2^2 \Big) \bigg) \label{eq:ccmodyfista}  \\
  t^{(i+1)} & = \frac{1}{2} \bigg(1 + \sqrt{1 + 4 \, (t^{(i)})^2} \bigg) \label{eq:ccmodtfista} \\
  \mb{d}^{(i+1)} &= \mb{y}^{(i+1)} +  \frac{t^{(i)} - 1}{t^{(i+1)}} \Big(\mb{y}^{(i+1)} - \mb{d}^{(i)} \Big) \;, \label{eq:ccmoddfista}
\end{align}
}%
where $t^0 = 1$, and $L > 0$ is a parameter controlling the gradient descent step size. Parameter $L$ can be computed adaptively by using a backtracking step size rule~\cite{beck-2009-fast}, but in the experiments reported here we used a constant $L$ for simplicity. The gradient of the data fidelity term $(1/2) \norm{X \mb{d} - \mb{s}}_2^2$ in~\eq{ccmodyfista} is computed in the DFT domain
\begin{equation}
 \nabla_{\hat{\mb{d}}} \Big(\frac{1}{2} \normsz[\big]{\hat{X} \hat{\mb{d}} - \hat{\mb{s}}}_2^2 \Big) = \hat{X}^H \big(\hat{X} \hat{\mb{d}} - \hat{\mb{s}} \big) \;,
\label{eq:gradf}
\end{equation}
as advocated in~\cite{wohlberg-2016-efficient} for the FISTA solution of the CSC problem, and the $\mb{y}^{(i+1)}$ variable is taken as the result of the dictionary update.

\section{Masked Convolutional Dictionary Learning}
\label{sec:mskcdl}

When we wish to learn a dictionary from data with missing samples, or have reason to be concerned about the possibility of boundary artifacts resulting from the circular boundary conditions associated with the computation of the convolutions in the DFT domain, it is useful to introduce a variant of~\eq{cdl} that includes a spatial mask~\cite{heide-2015-fast}, which can be represented by a diagonal matrix $W$
\begin{align}
  \argmin_{\{\mb{d}_{m}\},\{\mb{x}_{m,k}\}} & \frac{1}{2} \sum_k
  \normsz[\Big]{W \Big(\sum_m \mb{d}_m \ast \mb{x}_{m,k} - \mb{s}_k \Big)}_2^2 +
  \nonumber \\ & \lambda \sum_{m,k} \norm{\mb{x}_{m,k}}_1
  \text{ s.t. } \norm{\mb{d}_{m}}_2 = 1 \; \forall m \; .
\label{eq:mcdl}
\end{align}
As in~\sctn{cdl}, we separately consider the minimization of this functional with respect to $\{\mb{x}_{m,k}\}$ (sparse coding) and $\{\mb{d}_{m}\}$ (dictionary update).

\subsection{Sparse Coding}

A masked form of the MMV CBPDN problem~\eq{bpdnmmv} can be expressed as the problem\footnote{For simplicity, the notation presented here assumes a fixed mask $W$ across all columns of $D X$ and $S$, but the algorithm is easily extended to handle a different $W_k$ for each column $k$.}
\begin{align}
  \argmin_{X} \; (1/2)  \normsz[\big]{W (D X - S)}_F^2 + \lambda \| S \|_1
   \;.
  \label{eq:cbpdnmsk}
\end{align}
There are two different methods for solving this problem. The one, proposed in~\cite{heide-2015-fast}, exploits the mask decoupling technique~\cite{almedia-2013-deconvolving}, involving applying an alternative variable splitting to give the ADMM problem\footnote{This is a variant of the earlier problem formulation~\cite{heide-2015-fast}, which was
\begin{align*}
  \argmin_{X} \; (1/2)  \norm{W Y_1 - W S}_F^2 + \lambda \|Y_0 \|_1
   \;\text{ s.t. }\; Y_0 = X \;,\; Y_1 = D X \;.
\end{align*}
}
\begin{align}
  \argmin_{X} \; (1/2)  \norm{W Y_1}_F^2 + \lambda \|Y_0 \|_1 \nonumber \\
   \text{ s.t. } Y_0 = X \quad Y_1 = D X - S\;,
\label{eq:mdcsc}
\end{align}
where the constraint can also be written as
\begin{align}
\left( \begin{array}{c}  Y_0\\ Y_1\\
     \end{array} \right) = \left( \begin{array}{c}   I \\ D \\
     \end{array} \right) X - \left( \begin{array}{c}   0 \\ S \\
     \end{array} \right) \;.
\end{align}

The corresponding ADMM iterations are
{\small
  \setlength{\abovedisplayskip}{6pt}
  \setlength{\belowdisplayskip}{\abovedisplayskip}
  \setlength{\abovedisplayshortskip}{0pt}
  \setlength{\belowdisplayshortskip}{3pt}
\addtolength{\arraycolsep}{-1.0mm}
\begin{align}
  X^{(i+1)} &= \argmin_{X}
  \frac{\rho}{2} \norm{  D X - (Y_1^{(i)} + S - U_1^{(i)}) }_F^2 +
  \nonumber \\
 & \qquad \qquad \;\;\; \frac{\rho}{2} \norm{ X - (Y_0^{(i)} - U_0^{(i)}) }_F^2
 \label{eq:cbpdnwsxprob} \\
  Y_0^{(i+1)} &= \argmin_{Y_0} \lambda
  \norm{Y_0}_1 +
\frac{\rho}{2} \norm{Y_0 - (X^{(i+1)}   + U_0^{(i)}) }_F^2
 \label{eq:cbpdnwsy0prob}  \\
  Y_1^{(i+1)} &= \argmin_{Y_1} \frac{1}{2}
\normsz[\big]{W Y_1}_F^2 +  \nonumber \\ & \hspace{4em} \;\;\;
 \frac{\rho}{2} \norm{Y_1 - (D X^{(i+1)} - S  + U_1^{(i)})
 }_F^2
 \label{eq:cbpdnwsy1prob}  \\
  U_0^{(i+1)} &= U_0^{(i)} + X^{(i+1)} -
  Y_0^{(i+1)}  \label{eq:cbpdnwsu0prob} \\
U_1^{(i+1)} &= U_1^{(i)} +  D X^{(i+1)} -
  Y_1^{(i+1)} - S\;.   \label{eq:cbpdnwsu1prob}
\end{align}
}%
The functional minimized in~\eq{cbpdnwsxprob}  is of the same form as~\eq{bpdnxprob}, and can be solved via the
same frequency domain method, the solution to~\eq{cbpdnwsy0prob} is as in~\eq{bpdnysoln}, and the solution to~\eq{cbpdnwsy1prob} is given by
\begin{align}
(W^T W + \rho I) Y_1^{(i+1)} = \rho (D X^{(i+1)} - S + U_1^{(i)}) \;.
\end{align}

The other method for solving~\eq{cbpdnmsk} involves appending an impulse filter to the dictionary and solving the problem in a way that constrains the coefficient map corresponding to this filter to be zero where the mask is unity, and to be unconstrained where the mask is zero~\cite{wohlberg-2014-endogenous, wohlberg-2016-boundary}. Both approaches provide very similar performance~\cite{wohlberg-2016-boundary}, the major difference being that the former is a bit more complicated to implement, while the latter is restricted to addressing problems where $W$ has only zero or one entries. We will use the mask decoupling approach for the experiments reported here since it does not require any restrictions on $W$.

\subsection{Dictionary Update}
\label{sec:mskdctup}

The dictionary update requires solving the problem
\begin{align}
  \argmin_{\mb{d}} \; (1/2)  \normsz[\big]{W (X \mb{d} -
    \mb{s})}_2^2 + \iota_{C_{\text{PN}}} (\mb{d})  \;.
  \label{eq:ccmodmsk}
\end{align}
Algorithms for solving this problem are discussed in the following section.

\section{Masked Dictionary Update Algorithms}
\label{sec:mskdu}

\subsection{Block-Constraint ADMM}
\label{sec:blkadmmmdu}

 Problem~\eq{ccmodmsk} can be solved via the splitting~\cite{heide-2015-fast}
\begin{align}
  \argmin_{\mb{d}} \; (1/2)  \norm{W \mb{g}_1}_2^2 + \iota_{C_{\text{PN}}} (\mb{g}_0) \nonumber \\
   \text{ s.t. } \mb{g}_0 = \mb{d} \quad \mb{g}_1 = X \mb{d} - \mb{s}\;,
   \label{eq:ccmodmsk_block}
\end{align}
where the constraint can also be written as
\begin{align}
\left( \begin{array}{c}  \mb{g}_0\\ \mb{g}_1\\
     \end{array} \right) = \left( \begin{array}{c}   I \\ X \\
     \end{array} \right) \mb{d} - \left( \begin{array}{c}   0 \\ \mb{s} \\
     \end{array} \right) \;.
\end{align}

This problem has the same structure as~\eq{mdcsc}, the only difference being the replacement of the $\ell_1$ norm with the indicator function of the constraint set. The ADMM iterations are thus largely the same as~\eq{cbpdnwsxprob} -- \eqc{cbpdnwsu1prob}, the differences being that the $\ell_1$ norm in~\eq{cbpdnwsy0prob} is replaced with the  indicator function of the constraint set, and that the step corresponding to~\eq{cbpdnwsxprob} is more computationally expensive to solve, just as~\eq{ccmoddprob} is more expensive than~\eq{bpdnxprob}.

\subsection{Extended Consensus Framework}
\label{sec:mskcns}

In this section we re-use the variant notation introduced in~\sctn{ducns}. The  masked dictionary update~\eq{ccmodmsk} can be solved via a hybrid of the mask decoupling and ADMM consensus approaches, which can be formulated as
\begin{align}
  \argmin_{\mb{d}} \; (1/2)  \norm{W \mb{g}_1}_2^2 + \iota_{C_{\text{PN}}} (\mb{g}_0) \nonumber \\
   \text{ s.t. } E \mb{g}_0 = \mb{d} \quad \mb{g}_1 = X \mb{d} - \mb{s}  \;,
   \label{eq:ccmodmsk_cns}
\end{align}
where the constraint can also be written as
\begin{align}
\left( \begin{array}{c}   I \\ X \\
     \end{array} \right) \mb{d} +
\left( \begin{array}{cc}   -E & 0 \\ 0 & -I
     \end{array} \right)
\left( \begin{array}{c}  \mb{g}_0\\ \mb{g}_1\\
     \end{array} \right) =
 \left( \begin{array}{c}  0 \\ \mb{s}\\
     \end{array} \right)  \;,
\end{align}
or, expanding the block components of $\mb{d}$, $\mb{g}_1$, and $\mb{s}$,
\begin{align}
\addtolength{\arraycolsep}{-0.5mm}
\left( \begin{array}{ccc}
             I & 0 & \ldots \\
             0 & I & \ldots \\
             \vdots & \vdots & \ddots \\
             X_0 & 0 & \ldots \\
             0 & X_1 & \ldots \\
             \vdots & \vdots & \ddots
     \end{array} \right)
\left( \begin{array}{c}  \mb{d}_{0}\\ \mb{d}_{1}\\
    \vdots  \end{array} \right) -
\left( \begin{array}{c}  \mb{g}_0\\ \mb{g}_0\\ \vdots \\ \mb{g}_{1,0}\\
   \mb{g}_{1,1}\\ \vdots
     \end{array} \right) =
 \left( \begin{array}{c}  0 \\ 0 \\ \vdots \\ \mb{s}_0\\ \mb{s}_1 \\ \vdots
     \end{array} \right)  \;.
\end{align}

The corresponding ADMM iterations are
{\small
  \setlength{\abovedisplayskip}{6pt}
  \setlength{\belowdisplayskip}{\abovedisplayskip}
  \setlength{\abovedisplayshortskip}{0pt}
  \setlength{\belowdisplayshortskip}{3pt}
\addtolength{\arraycolsep}{-1.0mm}
\begin{align}
  \mb{d}^{(i+1)} &= \argmin_{\mb{d}}
  \frac{\rho}{2} \norm{  X \mb{d} - (\mb{g}_1^{(i)} + \mb{s} - \mb{h}_1^{(i)}) }_2^2 +
  \nonumber \\
 & \qquad \qquad \;\;\; \frac{\rho}{2} \norm{ \mb{d} - (E \mb{g}_0^{(i)} - \mb{h}_0^{(i)}) }_2^2
 \label{eq:ccmodcnswsxprob} \\
  \mb{g}_0^{(i+1)} &= \argmin_{\mb{g}_0} \iota_{C_{\text{PN}}} (\mb{g}_0)  +
\frac{\rho}{2} \norm{E \mb{g}_0 - (\mb{d}^{(i+1)}   + \mb{h}_0^{(i)}) }_2^2
 \label{eq:ccmodcnswsy0prob}  \\
  \mb{g}_1^{(i+1)} &= \argmin_{\mb{g}_1} \frac{1}{2}
\normsz[\big]{W \mb{g}_1}_2^2 +  \nonumber \\ & \hspace{4em} \;\;\;
 \frac{\rho}{2} \norm{\mb{g}_1 - (X \mb{d}^{(i+1)} - \mb{s}  + \mb{h}_1^{(i)})
 }_2^2
 \label{eq:ccmodcnswsy1prob}  \\
  \mb{h}_0^{(i+1)} &= \mb{h}_0^{(i)} + \mb{d}^{(i+1)} -
  E \mb{g}_0^{(i+1)}  \label{eq:ccmodcnswsu0prob} \\
\mb{h}_1^{(i+1)} &= \mb{h}_1^{(i)} +  X \mb{d}^{(i+1)} -
  \mb{g}_1^{(i+1)} - \mb{s} \;. \label{eq:ccmodcnswsu1prob}
\end{align}
}%
Steps~\eq{ccmodcnswsxprob}, \eqc{ccmodcnswsy0prob}, and~\eqc{ccmodcnswsu0prob} have the same form, and can be solved in the same way, as steps~\eq{ccmoddprobcns}, \eqc{ccmodgprobcns}, and~\eqc{ccmodhprobcns} respectively of the ADMM algorithm in~\sctn{ducns}, and steps~\eq{ccmodcnswsy1prob} and~\eqc{ccmodcnswsu1prob} have the same form, and can be solved in the same way, as the corresponding steps in the ADMM algorithm of~\sctn{blkadmmmdu}.

\subsection{FISTA}
\label{sec:mdufista}

Problem~\eq{ccmodmsk} can be solved via FISTA as described in~\sctn{dufista}, but the calculation of the gradient term is complicated by the presence of the spatial mask. This difficulty can be handled by transforming back and forth between spatial and frequency domains so that the convolution operations are computed efficiently in the frequency domain, while the masking operation is computed in the spatial domain, \ie

{\small
  \setlength{\abovedisplayskip}{6pt}
  \setlength{\belowdisplayskip}{\abovedisplayskip}
  \setlength{\abovedisplayshortskip}{0pt}
  \setlength{\belowdisplayshortskip}{3pt}
\addtolength{\arraycolsep}{-1.0mm}
\begin{equation}
 F \bigg(\nabla_{\mb{d}} \Big(\frac{1}{2} \norm{W (X \mb{d} - \mb{s})}_2^2\Big)\bigg) = \hat{X}^H F \Big(W^T W F^{-1} \big(\hat{X} \hat{\mb{d}} - \hat{\mb{s}} \big) \Big) \;, \label{eq:gradf_mask}
\end{equation}
}%
where $F$ and $F^{-1}$ represent the DFT and inverse DFT transform operators, respectively.

\section{Multi-Channel CDL}
\label{sec:mltchn}

As discussed in~\cite{wohlberg-2016-convolutional}, there are two distinct ways of defining a convolutional representation of multi-channel data: a single-channel dictionary together with a distinct set of coefficient maps for each channel, or a multi-channel dictionary together with a shared set of coefficient maps\footnote{One might also use a mixture of these two approaches, with the channels partitioned into subsets, each of which is assigned a distinct dictionary channel, but this option is not considered further here.}. Since the dictionary learning problem for the former case is a straightforward extension of the single-channel problems discussed above, here we focus on the latter case, which can be expressed as\footnote{Multi-channel CDL is presented in this section as an extension of the CDL framework of~\sctn{cdl} and~\sctnc{dctup}. Application of the same extension to the masked CDL framework of~\sctn{mskcdl} is straightforward, and is supported in our software implementations~\cite{wohlberg-2016-sporco}.}
\begin{align}
  \argmin_{\{\mb{d}_{c,m}\},\{\mb{x}_{m,k}\}} \; & \frac{1}{2} \sum_{c,k}
  \normsz[\Big]{\sum_m \mb{d}_{c,m}  \ast \mb{x}_{m,k} - \mb{s}_{c,k}}_2^2 + \nonumber \\ & \lambda \sum_{m,k} \norm{\mb{x}_{m,k}}_1
 \text{ s.t. } \norm{\mb{d}_{c,m}}_2 = 1 \; \forall c, m \;,
\label{eq:cdlmchn}
\end{align}
where $\mb{d}_{c,m}$ is channel $c$ of the $m^{\text{th}}$ dictionary filter, and $\mb{s}_{c,k}$ is channel $c$ of the  $k^{\text{th}}$ training signal. We will denote the number of channels by $C$. As before, we separately consider the sparse coding and dictionary updates for alternating minimization of this functional.

\subsection{Sparse Coding}

Defining $D_{c,m}$ such that $D_{c,m} \mb{x}_{m,k} = \mb{d}_{c,m} \ast \mb{x}_{m,k}$, and
\begin{equation}
  D_c = \left( \begin{array}{ccc}D_{c,0} & D_{c,1} &
     \ldots \end{array} \right) \qquad  \mb{x}_k =
 \left( \begin{array}{c}  \mb{x}_{0,k}\\
          \mb{x}_{1,k}\\
          \vdots \end{array} \right) \;,
\end{equation}
we can write the sparse coding component of~\eq{cdlmchn} as
\begin{align}
  \argmin_{\{\mb{x}_{k}\}} \; & (1/2) \sum_{c,k}
  \normsz[\big]{D_c \mb{x}_k - \mb{s}_{c,k}}_2^2 + \lambda \sum_{m,k} \norm{\mb{x}_{k}}_1  \;,
\label{eq:cscmchn}
\end{align}
or by defining
\begin{equation}
  D = \left( \begin{array}{ccc}
         D_{0,0} & D_{0,1} & \ldots \\
         D_{1,0} & D_{1,1} & \ldots \\
         \vdots & \vdots & \ddots
\end{array} \right)
\end{equation}
and
\begin{equation}
\arraycolsep=3pt
X = \left( \begin{array}{ccc}
          \mb{x}_{0,0} & \mb{x}_{0,1} & \dots \\
          \mb{x}_{1,0} & \mb{x}_{1,1} & \dots \\
          \vdots & \vdots & \ddots
 \end{array} \right)
\quad
S = \left( \begin{array}{ccc}
          \mb{s}_{0,0} & \mb{s}_{0,1} & \dots \\
          \mb{s}_{1,0} & \mb{s}_{1,1} & \dots \\
          \vdots & \vdots & \ddots
 \end{array} \right) \;,
\end{equation}
as
\begin{align}
  \argmin_{X} \; & (1/2) \norm{D X - S}_2^2 + \lambda \norm{X}_1  \;.
\label{eq:cdlmchn2}
\end{align}
This has the same form as the single-channel MMV problem~\eq{bpdnmmv}, and the iterations for an ADMM algorithm to solve it are the same as~\eq{admmxup} -- \eqc{admmuup}. The only significant difference is that $D$ in~\sctn{mmvcsc} is a matrix with a $1 \times M$ block structure, whereas here it has a $C \times M$ block structure. The corresponding frequency domain matrix $\hat{D}^H \hat{D}$ can be decomposed into a set of $N$ components of rank $C$, just as $\hat{X}^H \hat{X}$ with $X$ as in~\eq{duxs} can be decomposed into a set of $N$ components of rank $K$. Consequently, all of the dictionary update algorithms discussed in~\sctn{dctup} can also be applied to the multi-channel CSC problem, with the $\mb{g}$ step corresponding to the projection onto the dictionary constraint set, \eg~\eq{ccmodgprob}, replaced with a $Y$ step corresponding to the proximal operator of the $\ell_1$ norm, \eg~\eq{bpdnyprob}. The Iterated Sherman-Morrison method is very effective for RGB images with only three channels\footnote{This is the only multi-channel CSC approach that is currently supported in the SPORCO package~\cite{wohlberg-2017-sporco}.}, but for a significantly larger number of channels the best choices would be the ADMM consensus or FISTA methods.

For the FISTA solution, we compute the gradient of the data fidelity term $(1/2) \sum_{c,k} \normsz[\big]{D_c \mb{x}_k - \mb{s}_{c,k}}_2^2$ in~\eq{cscmchn} in the DFT domain
\begin{equation}
 \nabla_{\hat{\mb{x}}_k} \Big(\frac{1}{2} \sum_{c} \normsz[\big]{\hat{D}_c \hat{\mb{x}}_k - \hat{\mb{s}}_{c,k}}_2^2 \Big) =  \sum_{c} \hat{D}_c^H \big(\hat{D}_c \hat{\mb{x}}_k - \hat{\mb{s}}_{c,k} \big) \;.
\label{eq:gradf_cscmchn}
\end{equation}
In contrast to the ADMM methods, the multi-channel problem is not significantly more challenging than the single channel case, since it simply involves an additional sum over the $C$ channels.

\subsection{Dictionary Update}

In developing the dictionary update it is convenient to re-index the variables in~\eq{cdlmchn}, writing the problem as
\begin{align}
  \argmin_{\{\mb{d}_{m,c}\}} \; & \frac{1}{2} \sum_{k,c}
  \normsz[\Big]{\sum_m \mb{x}_{k,m} \ast \mb{d}_{m,c} - \mb{s}_{k,c}}_2^2 \nonumber \\ &
 \text{ s.t. } \norm{\mb{d}_{m,c}}_2 = 1 \; \forall m, c \;.
\label{eq:cdumchn}
\end{align}
Defining $X_{k,m}$, $X_k$, $X$ and $C_{\text{PN}}$ as in~\sctn{dictup}, and
\begin{equation}
\mb{d}_c = \left( \begin{array}{c}
          \mb{d}_{0,c}\\ \mb{d}_{1,c}\\
          \vdots \end{array} \right)
\quad
D = \left( \begin{array}{ccc}
          \mb{d}_{0,0} & \mb{d}_{0,1} & \dots \\
          \mb{d}_{1,0} & \mb{d}_{1,1} & \dots \\
          \vdots & \vdots & \ddots
 \end{array} \right) \;,
\end{equation}
we can write~\eq{cdumchn} as
\begin{align}
  \argmin_{\{\mb{d}_{c}\}} \; & (1/2) \sum_{k,c}
  \normsz[\big]{X_k \mb{d}_c - \mb{s}_{k, c}}_2^2 + \sum_c \iota_{C_{\text{PN}}}(\mb{d}_c) \;,
\end{align}
or in simpler form\footnote{The definition of $\iota_{C_{\text{PN}}}(\cdot)$ is overloaded here in that the specific projection from which $C_{\text{PN}}$ is defined depends on the matrix structure of its argument.}
\begin{align}
  \argmin_{D} \; &  (1/2) \norm{X D - S}_2^2 + \iota_{C_{\text{PN}}}(D) \;.
\end{align}
It is clear that the structure of $X$ is the same as in the single-channel case and that the solutions for the different channel dictionaries $\mb{d}_c$ are independent, so that the dictionary update in the multi-channel case is no more computationally challenging than in the single channel case.

\subsection{Relationship between $K$ and $C$}

The above discussion reveals an interesting dual relationship between the number of images, $K$, in coefficient map set $X$, and the number of channels, $C$, in dictionary $D$. When solving the CDL problem via proximal algorithms such as ADMM or FISTA, $C$ controls the rank of the most expensive subproblem of the convolutional sparse coding stage in the same way that $K$ controls the rank of the main subproblem of the convolutional dictionary update. In addition, algorithms that are appropriate for the large $K$ case of the dictionary update are also suitable for the large $C$ case of sparse coding, and vice versa.

\section{Results}
\label{sec:rslt}

In this section we compare the computational performance of the various approaches that have been discussed, carefully selecting optimal parameters for each algorithm to ensure a fair comparison.

\subsection{Dictionary Learning Algorithms}
\label{sec:cdlalg}

Before proceeding to the results of the computational experiments, we summarize the dictionary learning algorithms that will be compared. Instead of using the complete dictionary learning algorithm proposed in each prior work, we consider the primary contribution of these works to be in the dictionary update method, which is incorporated into the CDL algorithm structure that was demonstrated in~\cite{garcia-2017-subproblem} to be most effective: auxiliary variable coupling with a single iteration for each subproblem\footnote{In some cases, slightly better time performance can be obtained by performing a few iterations of the sparse coding update followed by a single dictionary update, but we do not consider this complication here.} before alternating. Since the sparse coding stages are the same, the algorithm naming is based on the dictionary update algorithms.

The following CDL algorithms are considered for problem~\eq{cdl} without a spatial mask
\begin{LaTeXdescription}
  \item[Conjugate Gradient (CG)] The CDL algorithm is as proposed in~\cite{wohlberg-2016-efficient}.
  \item[Iterated Sherman-Morrison (ISM)] The CDL algorithm is as proposed in~\cite{wohlberg-2016-efficient}.
  \item[Spatial Tiling (Tiled)] The CDL algorithm uses the dictionary update proposed in~\cite{sorel-2016-fast}, but the more effective variable coupling and alternation strategy discussed in~\cite{garcia-2017-subproblem}.
  \item[ADMM Consensus (Cns)] The CDL algorithm uses the dictionary update technique proposed in~\cite{sorel-2016-fast}, but the substantially more effective variable coupling and alternation strategy discussed in~\cite{garcia-2017-subproblem}.
\item[ADMM Consensus in Parallel (Cns-P)] The algorithm is the same as Cns, but with a parallel implementation of both the sparse coding and dictionary update stages\footnote{\v{S}orel and \v{S}roubek~\cite{sorel-2016-fast} observe that the ADMM consensus problem is inherently parallelizable~\cite[Ch. 7]{boyd-2010-distributed}, but do not actually implement the corresponding CDL algorithm in parallel form to allow the resulting computational gain to be quantified empirically.}. All steps of the CSC stage are completely parallelizable in the training image index $k$, as are the $\mb{d}$ and $\mb{h}$ steps of the dictionary update, the only synchronization point being in the $\mb{g}$ step,~\eq{ccmodgprobcns}, where all the independent dictionary estimates are averaged and projected (see~\eq{cnsgstepsoln}) to update the consensus variable that all the processes share.
  \item[3D (3D)] The CDL algorithm uses the dictionary update proposed in~\cite{sorel-2016-fast}, but the more effective variable coupling and alternation strategy discussed in~\cite{garcia-2017-subproblem}.
  \item[FISTA (FISTA)] Not previously considered for this problem.
\end{LaTeXdescription}

The following dictionary learning algorithms are considered for problem~\eq{mcdl} with a spatial mask
\begin{LaTeXdescription}
  \item[Conjugate Gradient (M-CG)] Not previously considered for this problem.
  \item[Iterated Sherman-Morrison (M-ISM)] The CDL algorithm is as proposed in~\cite{wohlberg-2016-boundary}.
  \item[Extended Consensus (M-Cns)] The CDL algorithm is based on a new dictionary update constructed as a hybrid of the dictionary update methods proposed in~\cite{heide-2015-fast} and~\cite{sorel-2016-fast}, with the effective variable coupling and alternation strategy discussed in~\cite{garcia-2017-subproblem}.
  \item[Extended Consensus in Parallel (M-Cns-P)] The algorithm is the same as M-Cns, but with a parallel implementation of both the sparse coding and dictionary update. All steps of the CSC stage and the $\mb{d}$, $\mb{g}_1$, $\mb{h}_0$, and $\mb{h}_1$ steps of the dictionary update are completely parallelizable in the training image index $k$, the only synchronization point being in the $\mb{g}_0$ step,~\eq{ccmodcnswsy0prob}, where all the independent dictionary estimates are averaged and projected to update the consensus variable that all the processes share.
  \item[FISTA (M-FISTA)] Not previously considered for this problem.
\end{LaTeXdescription}

In addition to the algorithms listed above, we investigated Stochastic Averaging ADMM (SA-ADMM)~\cite{zhong-2014-fast}, as proposed for CDL in~\cite{gu-2015-convolutional}. Our implementation of a CDL algorithm based on this method was found to have promising computational cost per iteration, but its convergence was not competitive with some of the other methods considered here. However, since there are a number of algorithm details that are not provided in~\cite{gu-2015-convolutional} (CDL is not the primary topic of that work), it is possible that our implementation omits some critical components. These results are therefore not included here in order to avoid making an unfair comparison.

We do not compare with the dictionary learning algorithm in~\cite{bristow-2013-fast} because the algorithms of both~\cite{heide-2015-fast} and~\cite{sorel-2016-fast} were both reported to be substantially faster. We do not include the algorithms of either~\cite{heide-2015-fast} and~\cite{sorel-2016-fast} in our main set of experiments because we do not have implementations that are practical to run over the large number of different training image sets and parameter choices that are used in these experiments, but we do include these algorithms in some additional performance comparisons in~\sctn{otheralgcompare} of the Supplementary Material. Multi-channel CDL problems are not included in our main set of experiments due to space limitations, but some relevant experiments are provided in~\sctn{colorcompare} of the Supplementary Material.

\subsection{Computational Complexity}

\begin{table}[htb]
\renewcommand{\arraystretch}{1.15}
  \caption{Computational complexities for a single iteration of the CDL algorithms, broken down into complexities for the sparse coding (CSC and M-CSC) and dictionary update steps, which are themselves decomposed into complexities for the frequency-domain solutions (FFT), the solution of the frequency-domain linear systems (Linear), the projection corresponding to the proximal map of the indicator function $\iota_{C_{\text{PN}}}$ (Prox), and additional operations due to a spatial mask (Mask). The number of pixels in the training images, the number of dictionary filters, and the number of training images are denoted by $N$, $M$, and $K$ respectively, and $\co_{CG}$ denotes the complexity of solving a linear system by the conjugate gradient method.}
\label{tab:complexity}
\centering
\setlength{\tabcolsep}{3.5pt}
\begin{tabular}{ | l | l  | l  | l  | l |}
\hline
\multicolumn{1}{|c|}{ {\bf Algorithm} } & \multicolumn{4}{|c|}{ {\bf Complexity} } \\
\cline{2-5}
& \multicolumn{1}{|c|}{ {\bf FFT} } & \multicolumn{1}{|c|}{ {\bf Linear} } & \multicolumn{1}{|c|}{ {\bf Prox} } & \multicolumn{1}{|c|}{ {\bf Mask} } \\
\hline
\hline
{\bf CSC} & $\co(KMN \log N)$ &  $\co(KMN)$ & $\co(KMN)$ &
\\
\hline
{\bf CG} & $\co(KMN \log N)$ &  $\co_{CG}$ & $\co(MN)$ &
\\
\hline
{\bf ISM} & $\co(KMN \log N)$ & $\co(K^2MN)$ & $\co(MN)$ &
\\
\hline
{\bf Tiled}, {\bf 3D} & $\co \left (KMN \left (\log N \right . \right . $ & $\co(KMN)$ & $\co(MN)$ &
\\
& $\left . \left . + \log K \right ) \right )$ & & & \\
\hline
{\bf Cns}, &  &  & &  \\
{\bf Cns-P}, & $\co(KMN \log N)$ & $\co(KMN)$ & $\co(MN)$  &
\\
{\bf FISTA} &  &  & &  \\
\hline \hline
{\bf M-CSC} & $\co(KMN \log N)$ &  $\co(KMN)$ & $\co(KMN)$ & $\co(KMN)$ \\
\hline
{\bf M-CG} & $\co(KMN \log N)$ & $\co_{CG}$ & $\co(MN)$ & $\co(KN)$ \\
& & + $\co(KMN)$ & & \\
\hline
{\bf M-ISM} & $\co(KMN \log N)$ & $\co(K^2MN)$ & $\co(MN)$ & $\co(KN)$ \\
& & + $\co(KMN)$ & & \\
\hline
{\bf M-Cns} &  &  & &  \\
{\bf M-Cns-P} & $\co(KMN \log N)$ & $\co(KMN)$ & $\co(MN)$ & $\co(KN)$ \\
{\bf M-FISTA} &  &  & &  \\
\hline
\end{tabular}
\vspace{-2mm}
\end{table}

The per-iteration computational complexities of the methods are summarized in~\tbl{complexity}. Instead of just specifying the dominant terms, we include all major contributing terms to provide a more detailed picture of the computational cost. All methods scale linearly with the number of filters, $M$, and with the number of images, $K$, except for the ISM variants, which scale as $\co(K^2)$. The inclusion of the dependency on $K$ for the parallel algorithms provides a very conservative view of their behavior. In practice, there is either no scaling or very weak scaling with $K$ when the number of available cores exceeds $K$, and weak scaling with $K$ when it exceeds the number of available cores. Memory usage depends on the method and implementation, but all the methods have an $\co(KMN)$ memory requirement for their main variables.

\subsection{Experiments}
\label{sec:cdlexp}

We used training sets of 5, 10, 20, and 40 images. These sets were nested in the sense that all images in a set were also present in all of the larger sets. The parent set of 40 images consisted of greyscale images of size 256 $\times$ 256 pixels, derived from the MIRFLICKR-1M dataset\footnote{The image data directly included in the MIRFLICKR-1M dataset is of very low resolution since the dataset is primarily targeted at image classification tasks. We therefore identified and downloaded the original images that were used to construct the MIRFLICKR-1M dataset.}~\cite{huiskes-2008-new} by cropping, rescaling, and conversion to greyscale. An additional set of 20 images, of the same size and from the same source, was used as a test set to allow comparison of generalization performance, taking into account possible differences in overfitting effects between the different methods.

The 8 bit greyscale images were divided by 255 so that pixel values were within the interval [0,1], and were highpass filtered (a common approach for convolutional sparse representations~\cite{kavukcuoglu-2010-hierarchies, zeiler-2011-adaptive, wohlberg-2016-efficient}\cite[Sec. 3]{wohlberg-2016-convolutional2}) by subtracting a lowpass component computed by Tikhonov regularization with a gradient term~\cite[pg. 3]{wohlberg-2017-sporco}, with regularization parameter $\lambda = 5.0$.

The results reported here were computed using the Python implementation of the SPORCO library~\cite{wohlberg-2016-sporco, wohlberg-2017-sporco} on a Linux workstation equipped with two Xeon E5-2690V4 CPUs.

\subsection{Optimal Penalty Parameters}
\label{sec:optimal_parameters}

\begin{table}[htb]
\renewcommand{\arraystretch}{1.15}
\caption{Dictionary learning: optimal parameters found by grid search.}
\label{tab:optimalParam}
\centering
\begin{tabular}{| c || c | r | r | c || r | r |}
\cline{3-4} \cline{6-7}
  \multicolumn{1}{c}{} & \multicolumn{1}{c}{} & \multicolumn{2}{|c|}{ {\bf Parameter} }  & \multicolumn{1}{c}{}  & \multicolumn{2}{|c|}{ {\bf Parameter} } \\
\cline{3-4}
 \hline
 \multicolumn{1}{|c||}{ {\bf Method} } &  \multicolumn{1}{|c|}{ $K$ } & \multicolumn{1}{|c|}{$\mathbf{\rho}$} & \multicolumn{1}{|c|}{$\mathbf{\sigma}$} &
 \multicolumn{1}{|c||}{ {\bf Method} } & \multicolumn{1}{|c|}{$\mathbf{\rho}$} & \multicolumn{1}{|c|}{$\mathbf{\sigma}$} \\
 \hline
\hline
 & 5 & 3.59	& 4.08 & & 3.59 & 5.99 \\
\cline{2-4} \cline{6-7}
{\bf CG} & 10 & 3.59	& 12.91 & {\bf M-CG} & 3.59 & 7.74 \\
\cline{2-4} \cline{6-7}
& 20 & 2.15 &	24.48 & & 2.15 & 7.74 \\
\cline{2-4} \cline{6-7}
& 40 & 2.56	& 62.85 & & 2.49 & 11.96 \\
 \hline
 & 5 & 3.59	& 4.08 & & 3.59 & 5.99 \\
 \cline{2-4} \cline{6-7}
{\bf ISM} & 10 & 3.59 & 12.91 & {\bf M-ISM} & 3.59 & 7.74 \\
\cline{2-4} \cline{6-7}
& 20 & 2.15	& 24.48 & & 2.15 & 7.74 \\
\cline{2-4} \cline{6-7}
& 40 & 2.56	& 62.85 & & 2.49 & 11.96 \\
\hline
& 5 & 3.59      & 7.74 & \multicolumn{3}{c}{ {} } \\
 \cline{2-4}
{\bf Tiled} & 10 & 3.59	& 12.91 & \multicolumn{3}{c}{ {} } \\
 \cline{2-4}
& 20 & 3.59	& 40.84 & \multicolumn{3}{c}{ {} } \\
 \cline{2-4}
& 40 & 3.59	& 72.29 & \multicolumn{3}{c}{ {} } \\
\hline
& 5 & 3.59	& 1.29 & & 3.59 & 1.13 \\
 \cline{2-4} \cline{6-7}
{\bf Cns} & 10 & 3.59	& 1.29 & {\bf M-Cns} & 3.59 & 0.68 \\
\cline{2-4} \cline{6-7}
& 20 & 3.59	& 2.15 & & 3.59 & 1.13 \\
\cline{2-4} \cline{6-7}
& 40 & 3.59	& 1.08 & & 3.59 & 1.01 \\
\hline
& 5 & 3.59	& 7.74 & \multicolumn{3}{c}{ {} } \\
\cline{2-4}
{\bf 3D} & 10 & 3.59	& 12.91 & \multicolumn{3}{c}{ {} } \\
\cline{2-4}
& 20 & 3.59	& 40.84 & \multicolumn{3}{c}{ {} } \\
 \cline{2-4}
& 40 & 3.59	& 72.29 & \multicolumn{3}{c}{ {} } \\
 \cline{1-4}
\multicolumn{3}{c}{} & \multicolumn{1}{|c|}{$L$} & \multicolumn{3}{c}{ {} } \\
 \cline{1-4}
& 5 & 3.59      & 48.14 & \multicolumn{3}{c}{ {} } \\
 \cline{2-4}
{\bf FISTA} & 10 & 3.59	& 92.95 & \multicolumn{3}{c}{ {} } \\
 \cline{2-4}
& 20 & 3.59     & 207.71 & \multicolumn{3}{c}{ {} } \\
 \cline{2-4}
& 40 & 3.59	& 400.00 & \multicolumn{3}{c}{ {} } \\
 \cline{1-4}
\end{tabular}
\end{table}

To ensure a fair comparison between the methods, the optimal penalty parameters for each method and training image set were selected via a grid search, of CDL functional values obtained after 100 iterations, over $(\rho, \sigma)$ values for the ADMM dictionary updates, and over $(\rho, L)$ values for the FISTA dictionary updates. The grid resolutions were
\begin{IEEEdescription}
  \item[$\rho$] 10 logarithmically spaced points in $[10^{-1}, 10^{4}]$
  \item[$\sigma$] 15 logarithmically spaced points in $[10^{-2}, 10^{5}]$
  \item[$L$] 15 logarithmically spaced points in $[10^{1}, 10^{5}]$
\end{IEEEdescription}
The best set of $(\rho, \sigma)$ or $(\rho, L)$ for each method \ie the ones yielding the lowest value of the CDL functional at 100 iterations, was selected as a center for a finer grid search, of CDL functional values obtained after 200 iterations, with 10 logarithmically spaced points in $[0.1 \rho_{\mathrm{center}}, 10 \rho_{\mathrm{center}}]$ and 10 logarithmically spaced points in $[0.1 \sigma_{\mathrm{center}}, 10 \sigma_{\mathrm{center}}]$ or 10 logarithmically spaced points in $[0.1 L_{\mathrm{center}}, 10 L_{\mathrm{center}}]$. The optimal parameters for each method were taken as those yielding the lowest value of the CDL functional at 200 iterations in this finer grid.  This procedure was repeated for sets of 5, 10, 20 and 40 images.  As an indication of the sensitivities of the different methods to their parameters, results for the coarse grid search for the 20 image set can be found in~\sctn{grids_sm} in the Supplementary Material. The optimal parameters determined via these grid searches are summarized in~\tbl{optimalParam}.

\subsection{Performance Comparisons}
\label{sec:rsltdl}

\begin{figure}[htb]
        \center
        \scalebox{0.26}{\includegraphics{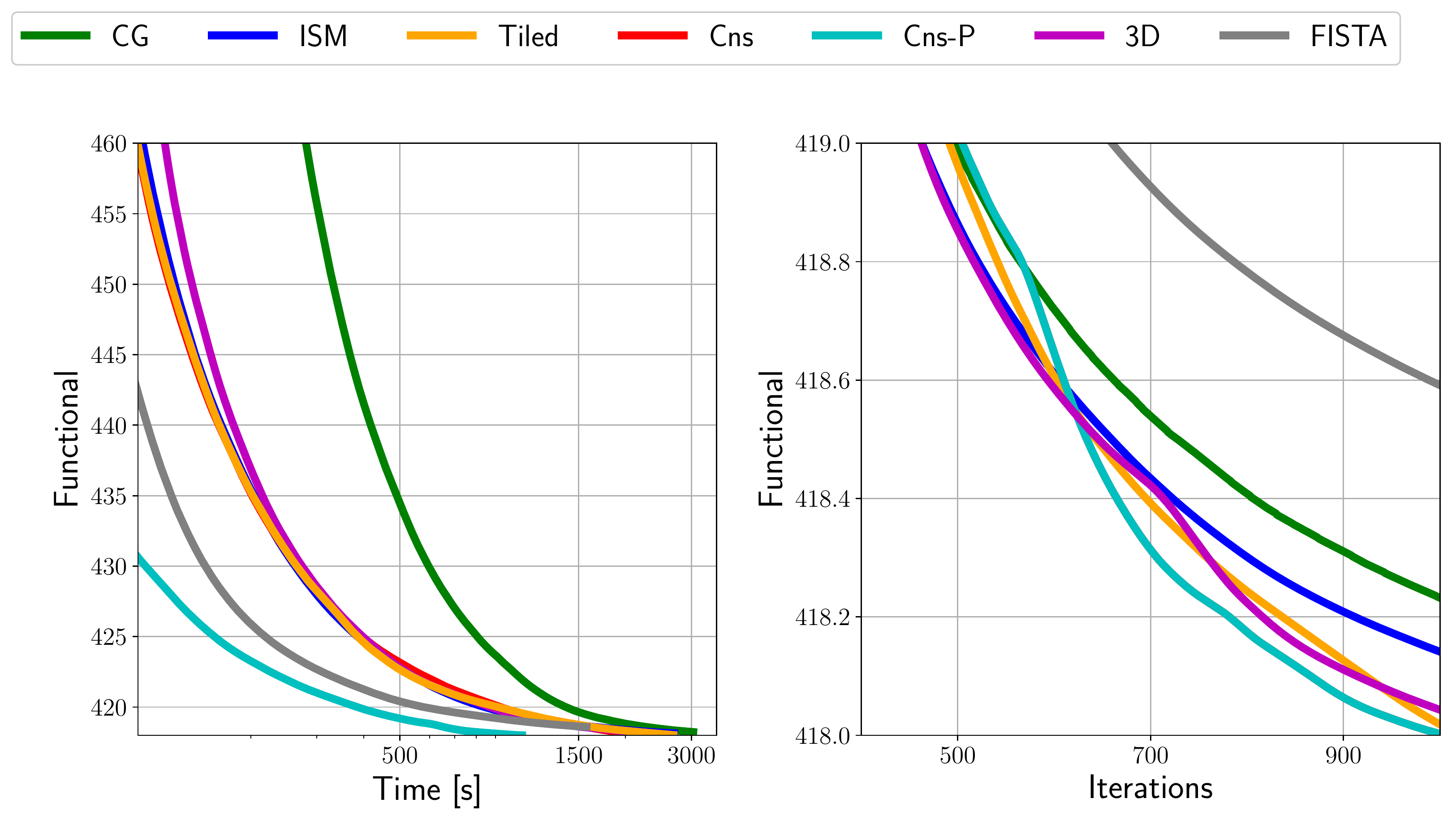}}
        \caption{Dictionary Learning ($K=5$): A comparison on a set of $K=5$ images of the decay of the value of the CBPDN functional~\eq{cbpdnmmv} with respect to run time and iterations. ISM, Tiled, Cns and 3D overlap in the time plot, and Cns and Cns-P overlap in the iterations plot.}
        \label{fig:fObj_k5}
\end{figure}

\begin{figure}[htb]
        \center
        \scalebox{0.26}{\includegraphics{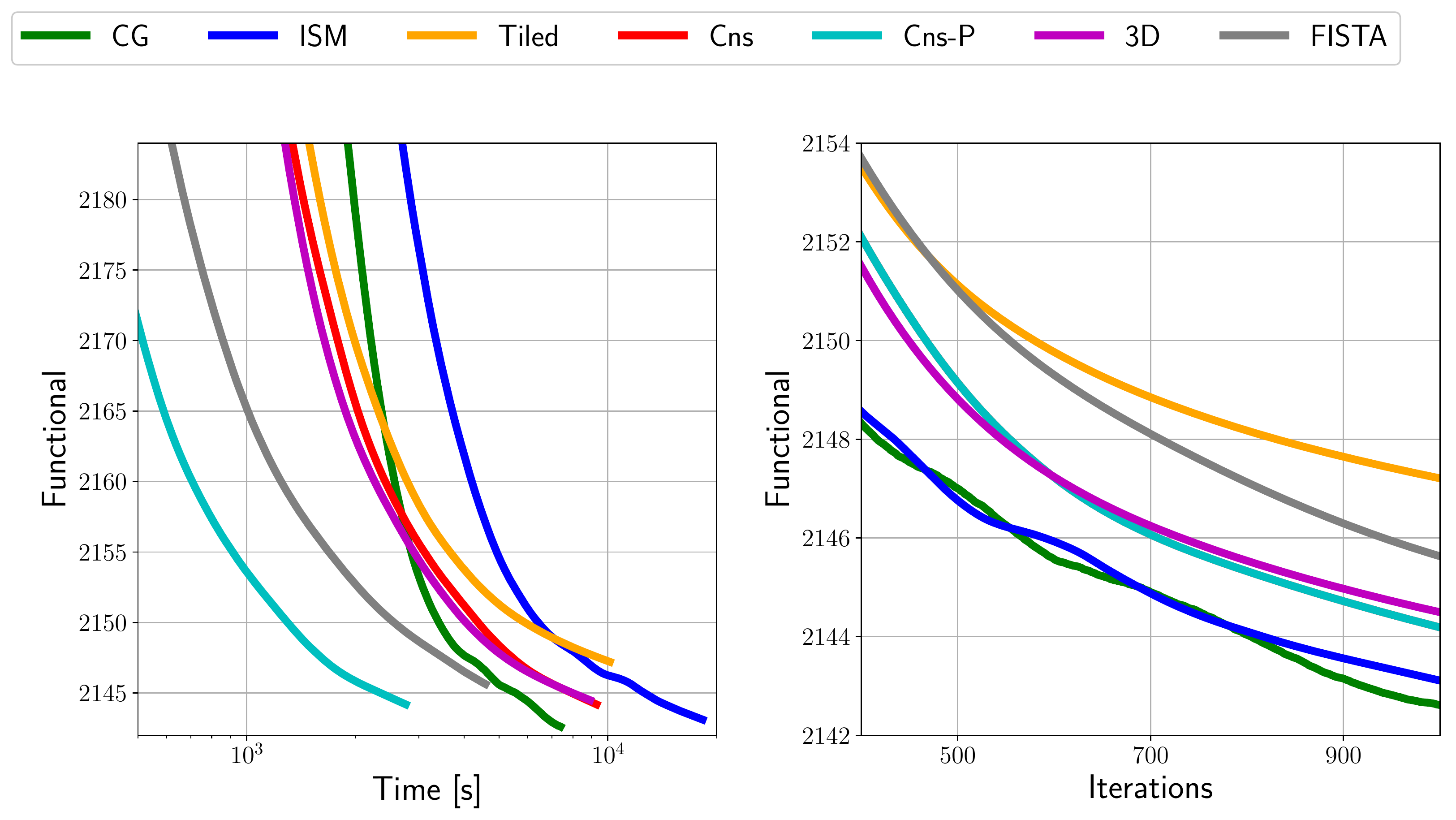}}
        \caption{Dictionary Learning ($K=20$): A comparison on a set of $K=20$ images of the decay of the value of the CBPDN functional~\eq{cbpdnmmv} with respect to run time and iterations. Cns and 3D overlap in the time plot, and Cns, Cns-P and 3D overlap in the iterations plot.}
        \label{fig:fObj_k20}
\end{figure}

\begin{figure}[htb]
        \center
        \scalebox{0.26}{\includegraphics{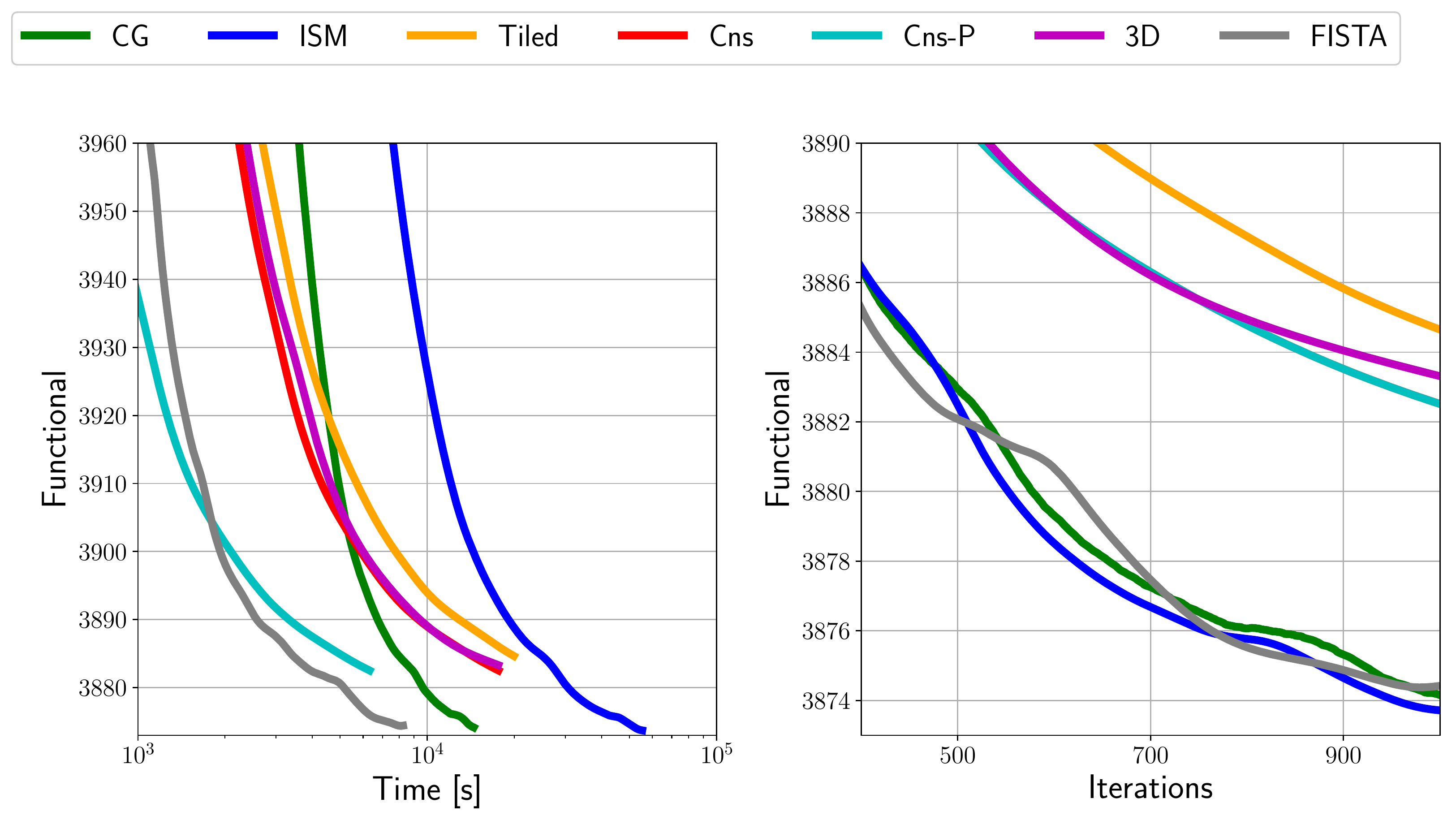}}
        \caption{Dictionary Learning ($K=40$): A comparison on a set of $K=40$ images of the decay of the value of the CPBDN functional~\eq{cbpdnmmv} with respect to run time and iterations. Cns and 3D overlap in the time plot, and Cns, Cns-P and 3D overlap in the iterations plot.}
        \label{fig:fObj_k40}
\end{figure}

\begin{figure}[htb]
\centerline{
        \subfigure[Without Spatial Mask]{\scalebox{0.24}{\includegraphics{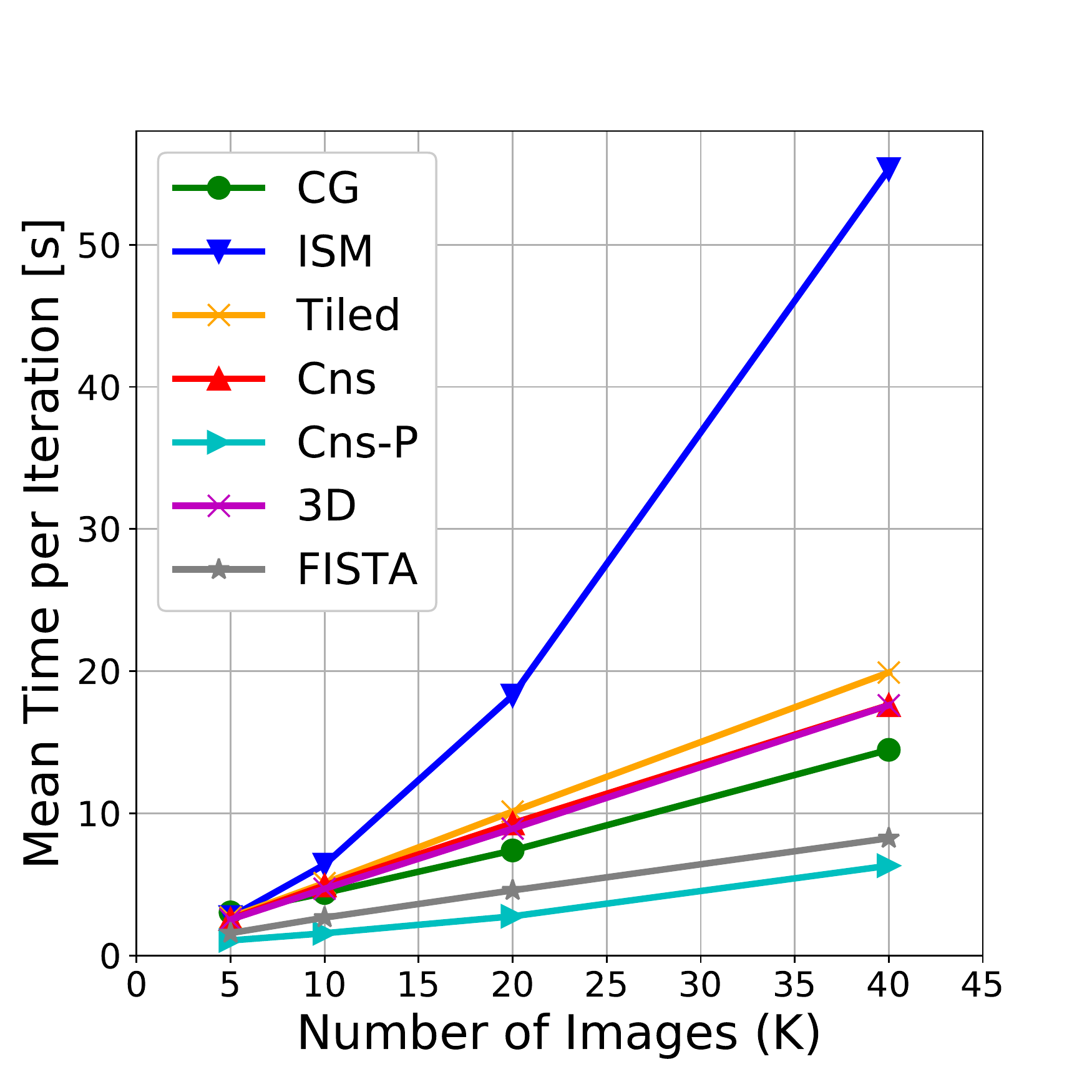}}
        \label{fig:t_per_it_scaling_regular}}
        \hfil
        \subfigure[With Spatial Mask]{\scalebox{0.24}{\includegraphics{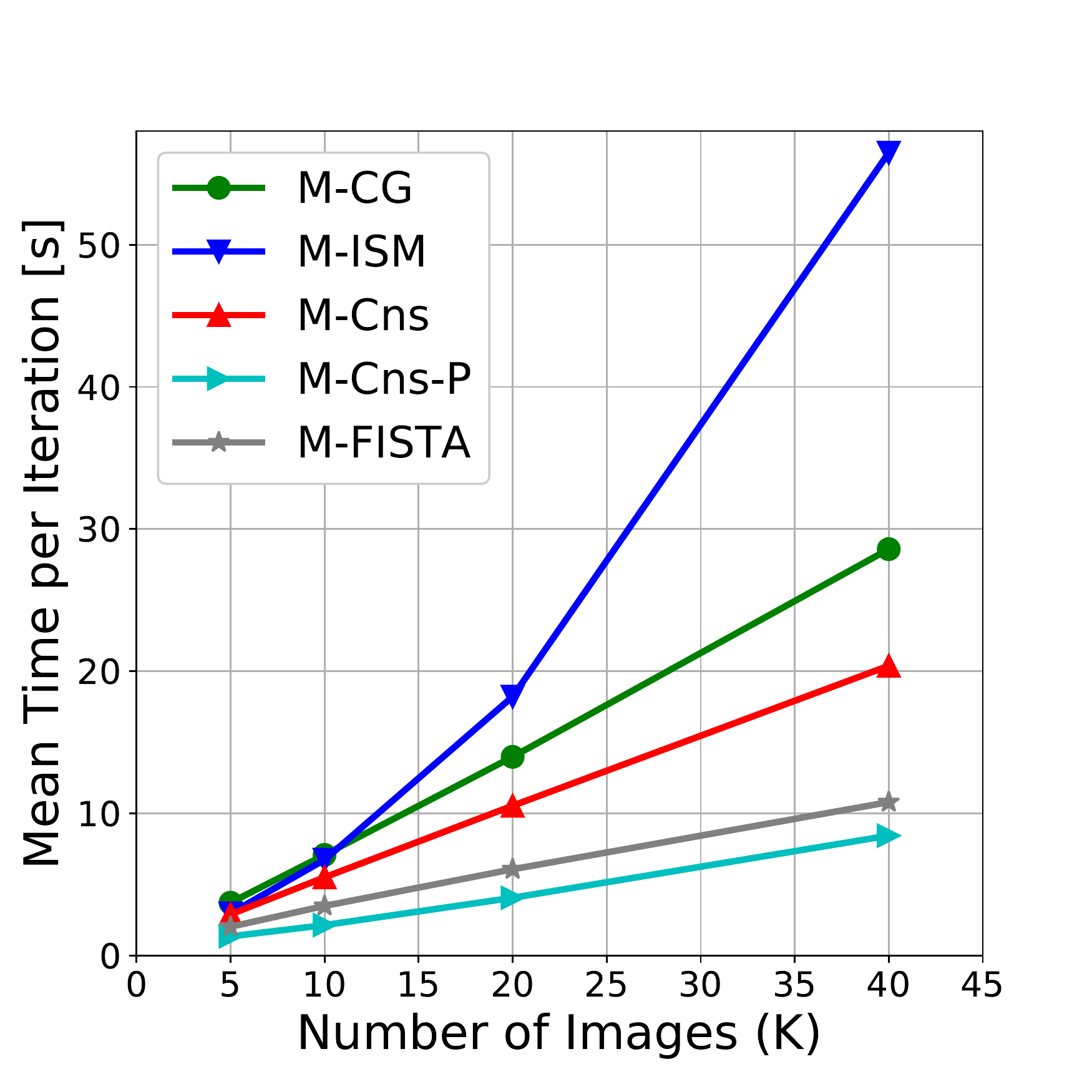}}
        \label{fig:t_per_it_scaling_mask}}
}
        \caption{Comparison of time per iteration for the dictionary learning methods for sets of 5, 10, 20 and 40 images.}
        \label{fig:t_per_it_scaling}
\end{figure}

We compare the performance of the methods in learning a dictionary of 64 filters of size 8 $\times$ 8 for sets of 5, 10, 20 and 40 images, setting the sparsity parameter $\lambda = 0.1$, and using the parameters determined by the grid searches for each method. To avoid complicating the comparisons, we used fixed penalty parameters $\rho$ and $\sigma$, without any adaptation methods~\cite[Sec. III.D]{wohlberg-2016-efficient}\cite{wohlberg-2017-admm}, and did not apply relaxation methods~\cite[Sec. 3.4.3]{boyd-2010-distributed}\cite[Sec. III.D]{wohlberg-2016-efficient} in any of the ADMM algorithms. Similarly, we used a fixed $L$ for FISTA, without applying any backtracking step-size adaptation rule. Performance in terms of the convergence rate of the CDL functional, with respect to both iterations and computation time, is compared in \figs{fObj_k5}~--~\fign{fObj_k40}. The time scaling with $K$ of all the methods is summarized in~\fig{t_per_it_scaling_regular}.

For the $K=5$ case, all the methods have quite similar performance in terms of functional value convergence with respect to iterations. For the larger training set sizes, CG and ISM have somewhat better performance with respect to iterations, but ISM has very poor performance with respect to time.  CG has substantially better time scaling, depending on the relative residual tolerance. We ran our experiments for CG with a fixed tolerance of $10^{-3}$, resulting in computation times that are comparable with those of the other methods. A smaller tolerance leads to better convergence with respect to iterations, but substantially worse time performance.

The ``3D'' method behaves similarly to ADMM consensus, as expected from the relationship established in~\sctn{dufdcn}, but has a larger memory footprint. The spatial tiling method (Tiled), on the other hand, tends to have slower convergence with respect to both iterations and time than the other methods. We do not further explore the performance of these methods since they do not provide substantial advantages over the others.

Both parallel (Cns-P) and regular consensus (Cns) have the same evolution of the CBPDN functional,~\eq{cbpdnmmv}, with respect to iterations, but the former requires much less computation time, and is the fastest method overall.  Moreover, parallel consensus exhibits almost ideal parallelizability, with some overhead for $K=5$, but scaling linearly for $K \in [10, 40]$, and with very competitive computation times.  FISTA is also very competitive, achieving good results in less time than any of the other serial methods, and even outperforming the time performance of Cns-P for the $K=40$ case shown in~\fig{fObj_k40}. We believe that this variation of relative performance with $K$ is due to the unstable dependence of the CDL functional on $L$ that is illustrated, for example, in~\fig{fista_par_selection_K5_L} in the Supplementary Material. This functional decreases slowly as $L$ is decreased, but then increases very rapidly after the minimum is reached, due to the constraint on $L$ discussed in~\sctn{parameter_sensitivity}.

\begin{figure}[htb]
        \center
        \scalebox{0.26}{\includegraphics{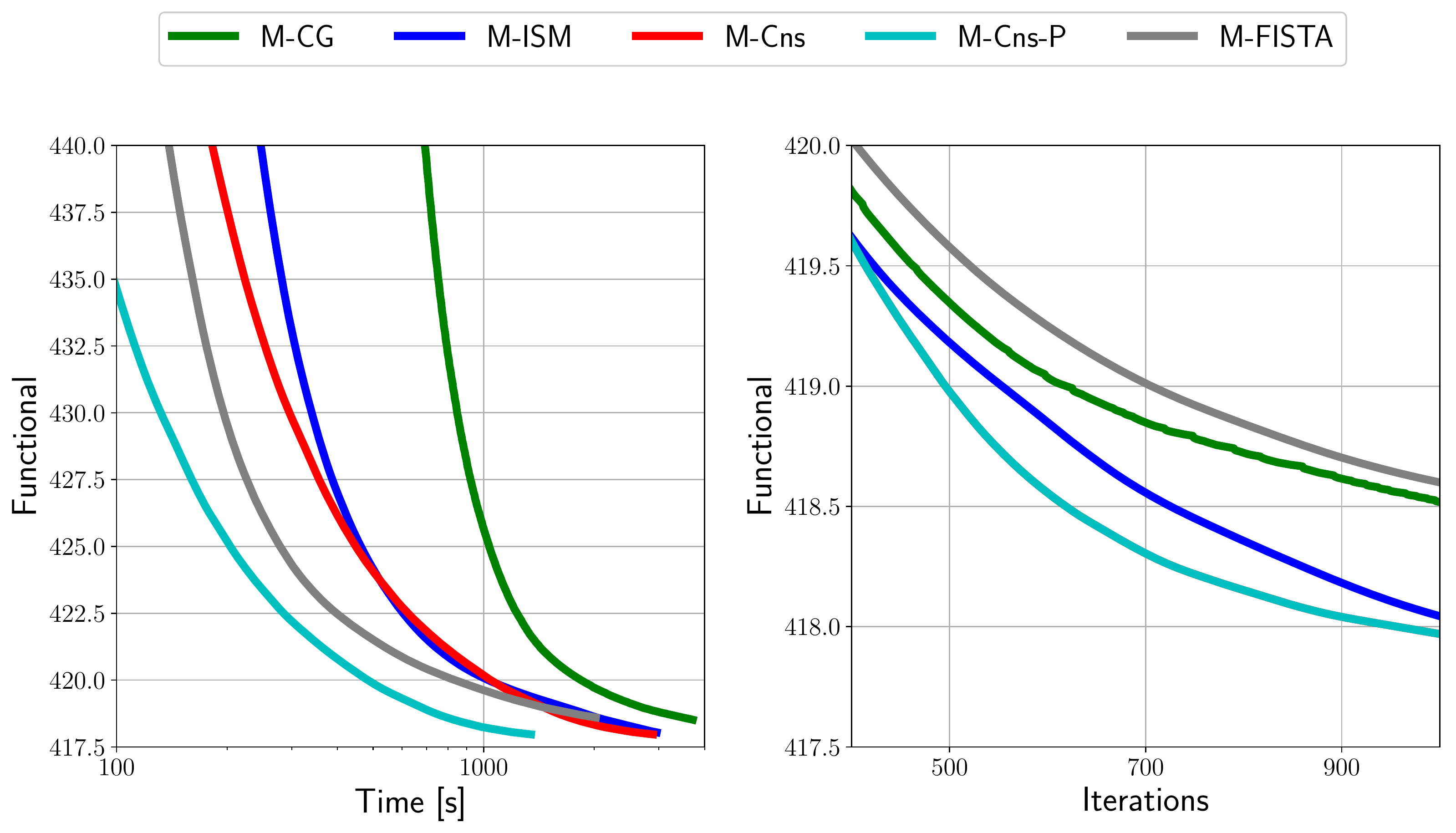}}
        \caption{Dictionary Learning with Spatial Mask ($K=5$): A comparison on a set of $K=5$ images of the decay of the value of the masked CBPDN functional~\eq{cbpdnmsk} with respect to run time and iterations for masked versions of the algorithms. M-Cns and M-Cns-P overlap in the iterations plot.}
        \label{fig:fObj_k5_mask}
\end{figure}

\begin{figure}[htb]
        \center
        \scalebox{0.26}{\includegraphics{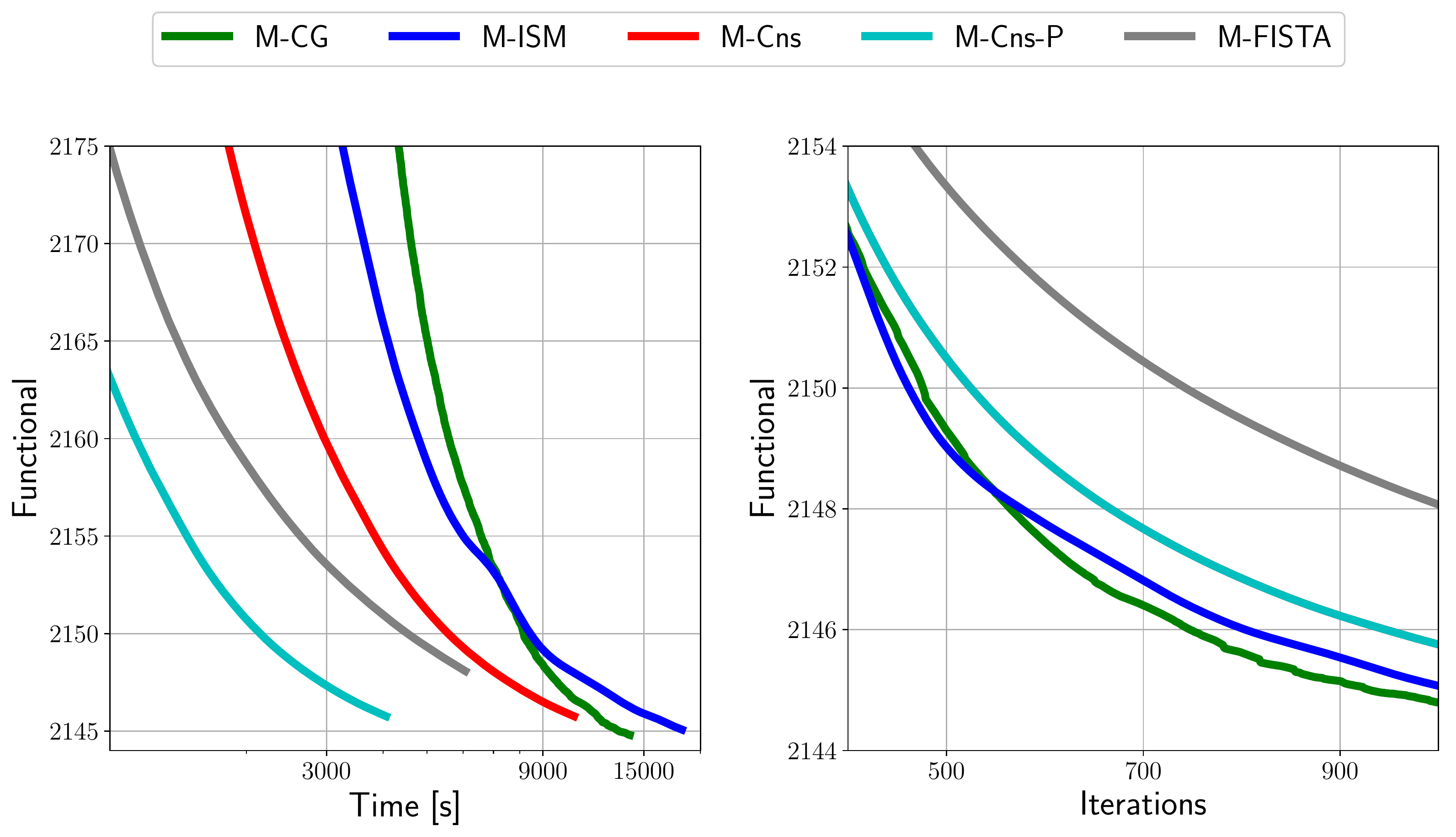}}
        \caption{Dictionary Learning with Spatial Mask ($K=20$): A comparison on a set of $K=20$ images of the decay of the value of the masked CBPDN functional~\eq{cbpdnmsk} with respect to run time and iterations for masked versions of the algorithms. M-Cns and M-Cns-P overlap in the iterations plot.}
        \label{fig:fObj_k20_mask}
\end{figure}

\begin{figure}[htb]
        \center
        \scalebox{0.26}{\includegraphics{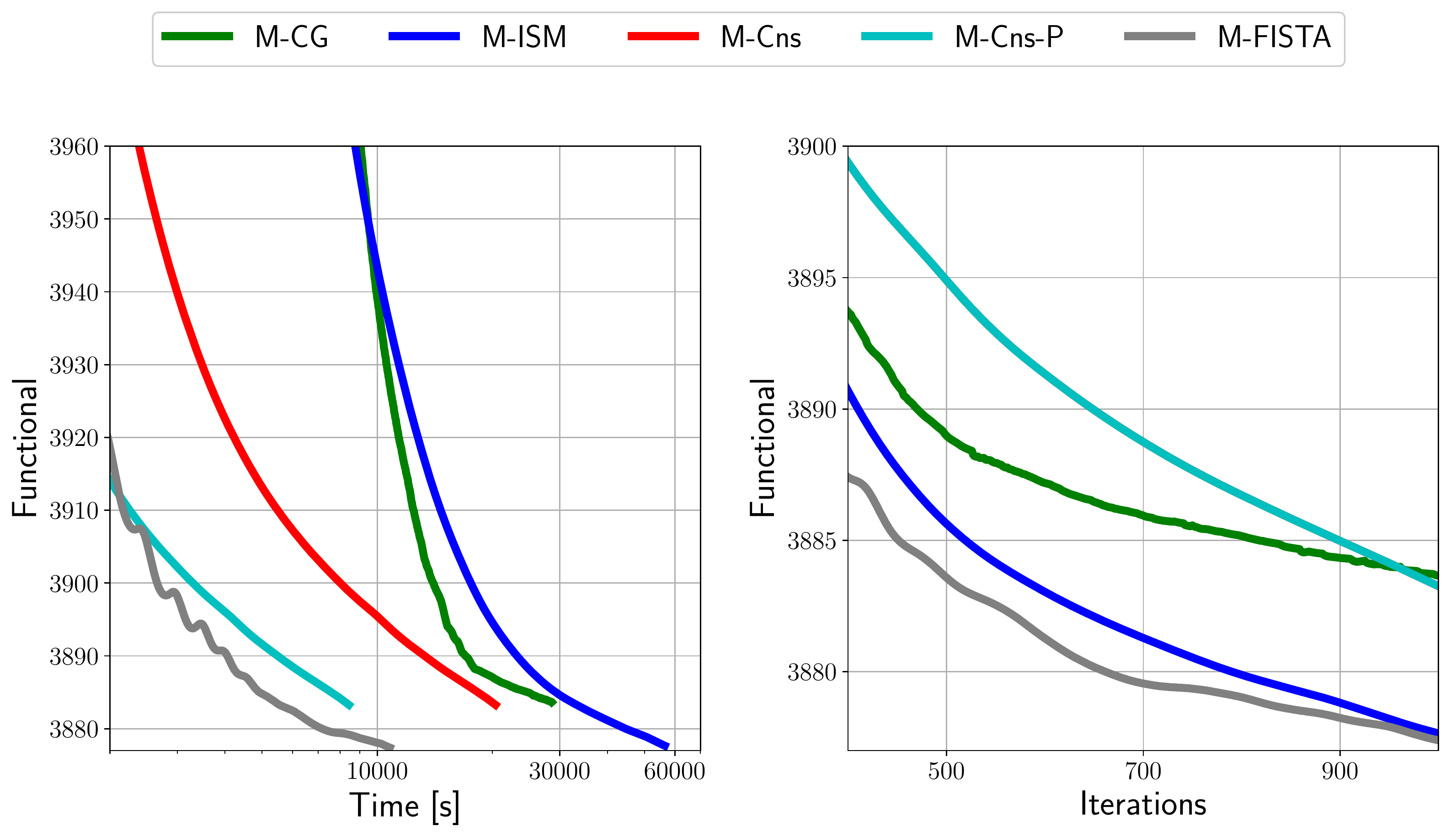}}
        \caption{Dictionary Learning with Spatial Mask ($K=40$): A comparison on a set of $K=40$ images of the decay of the value of the masked CBPDN functional~\eq{cbpdnmsk} with respect to run time and iterations for masked versions of the algorithms. M-Cns and M-Cns-P overlap in the iterations plot.}
        \label{fig:fObj_k40_mask}
\end{figure}

All experiments with algorithms that include a spatial mask set the mask to the identity ($W = I$) to allow comparison with the performance of the algorithms without a spatial mask.  Plots comparing the evolution of the masked CBPDN functional,~\eq{cbpdnmsk}, over 1000 iterations and problem sizes of $K \in \{5, 20, 40\}$ are displayed in~\figs{fObj_k5_mask}~--~\fign{fObj_k40_mask}, respectively. The time scaling of all the masked methods is summarized in~\fig{t_per_it_scaling_mask}.

While the convergence performance with iterations of the masked version of the FISTA algorithm, M-FISTA, is mixed (providing the worst performance for $K=5$ and $K=20$, but the best performance for $K=40$), it consistently provides good performance in terms of convergence with respect to computation time, despite the additional FFTs discussed in~\sctn{mdufista}. The parallel hybrid mask decoupling/consensus method, M-Cns-P, is the other competitive approach for this problem, providing the best time performance for $K=5$ and $K=20$, while lagging slightly behind M-FISTA for $K=40$.

In contrast with the corresponding mask-free variants, M-CG and M-ISM have worse performance in terms of both time and iterations. This suggests that M-CG requires a value for the relative residual tolerance smaller than $10^{-3}$ to produce good results, but this would be at the expense of much longer computation times. With the exception of CG, for which the cost of computing the masked version increases for $K \geq 20$, the computation time for the masked versions is only slightly worse than the mask-free variants (\fig{t_per_it_scaling}). In general, using the masked versions leads to a marginal decrease in convergence rate with respect to iterations, and a small increase in computation time.

\subsection{Evaluation on the Test Set}
\label{sec:rsltdlvld}

\begin{figure}[htb]
        \center
        \scalebox{0.26}{\includegraphics{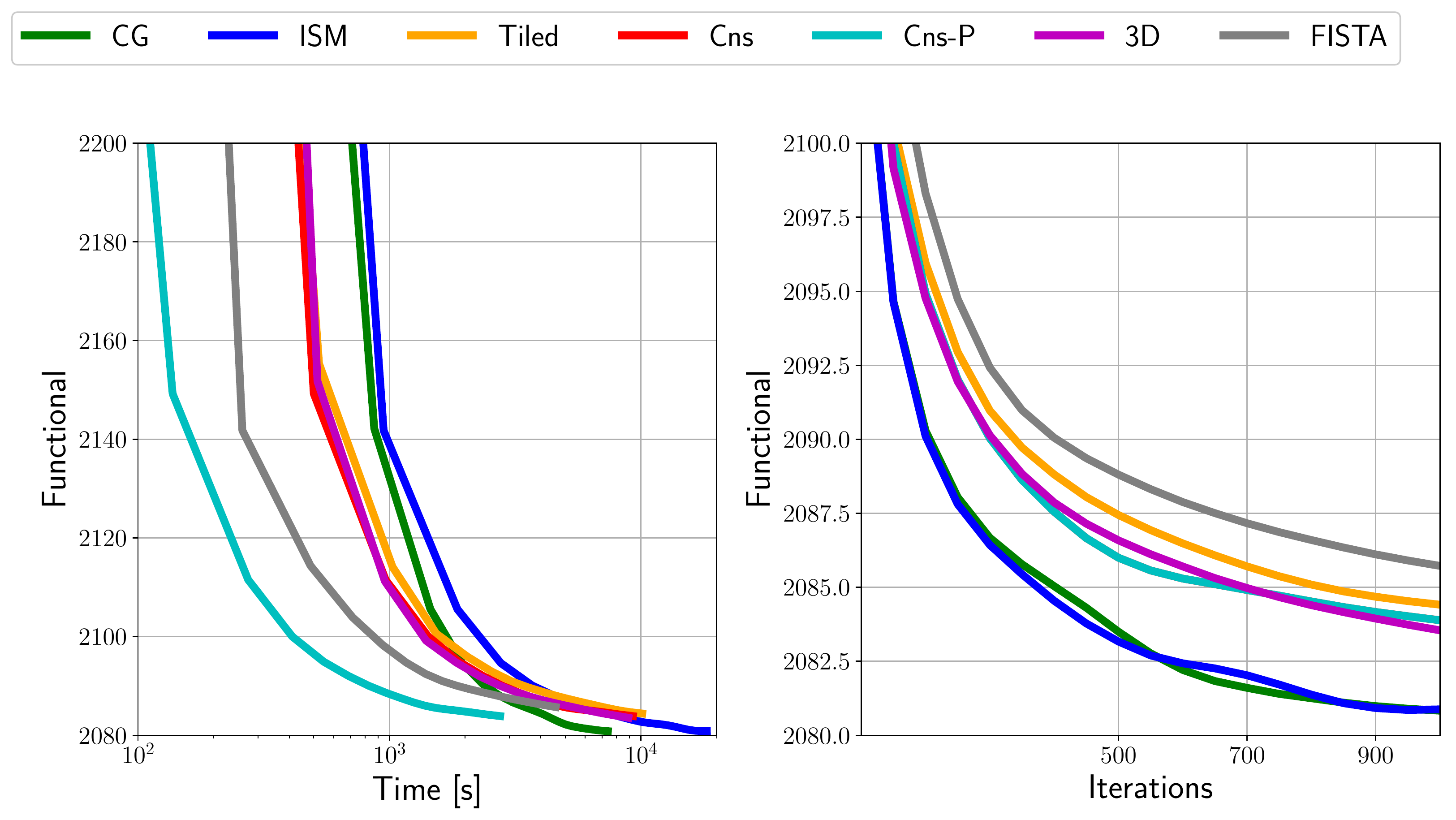}}
        \caption{Evolution of the CBPDN functional~\eq{cbpdnmmv} for the test set using the partial dictionaries obtained when training for $K=20$ images. Tiled, Cns and 3D overlap in the time plot, and Cns and Cns-P overlap in the iterations plot.}
        \label{fig:val_k20}
\end{figure}

\begin{figure}[htb]
        \center
        \scalebox{0.26}{\includegraphics{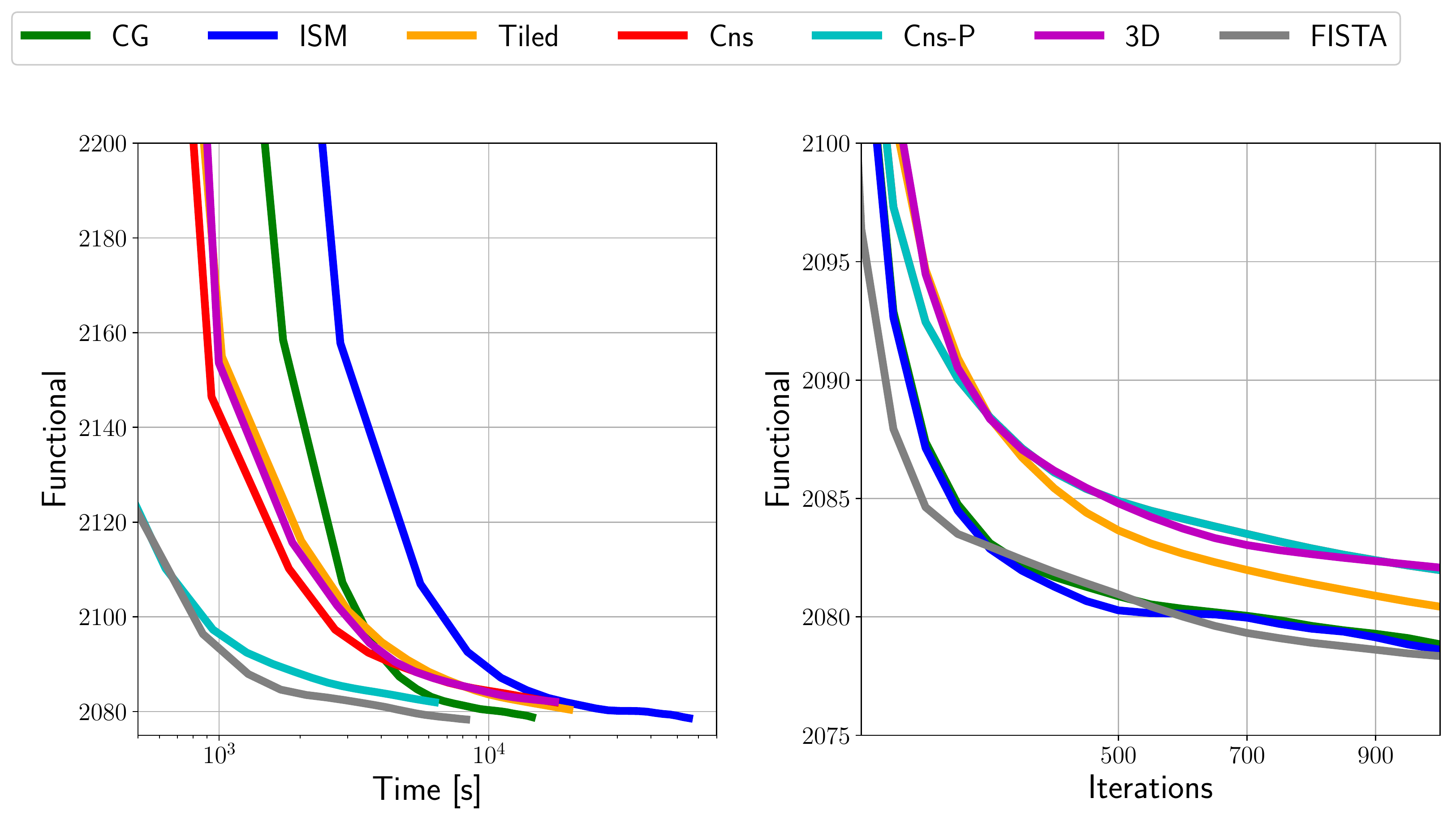}}
        \caption{Evolution of the CBPDN functional~\eq{cbpdnmmv} for the test set using the partial dictionaries obtained when training for $K=40$ images. Tiled, Cns and 3D have a large overlap in the time plot, and Cns and Cns-P overlap in the iterations plot.}
        \label{fig:val_k40}
\end{figure}

\begin{figure}[htb]
        \center
        \scalebox{0.26}{\includegraphics{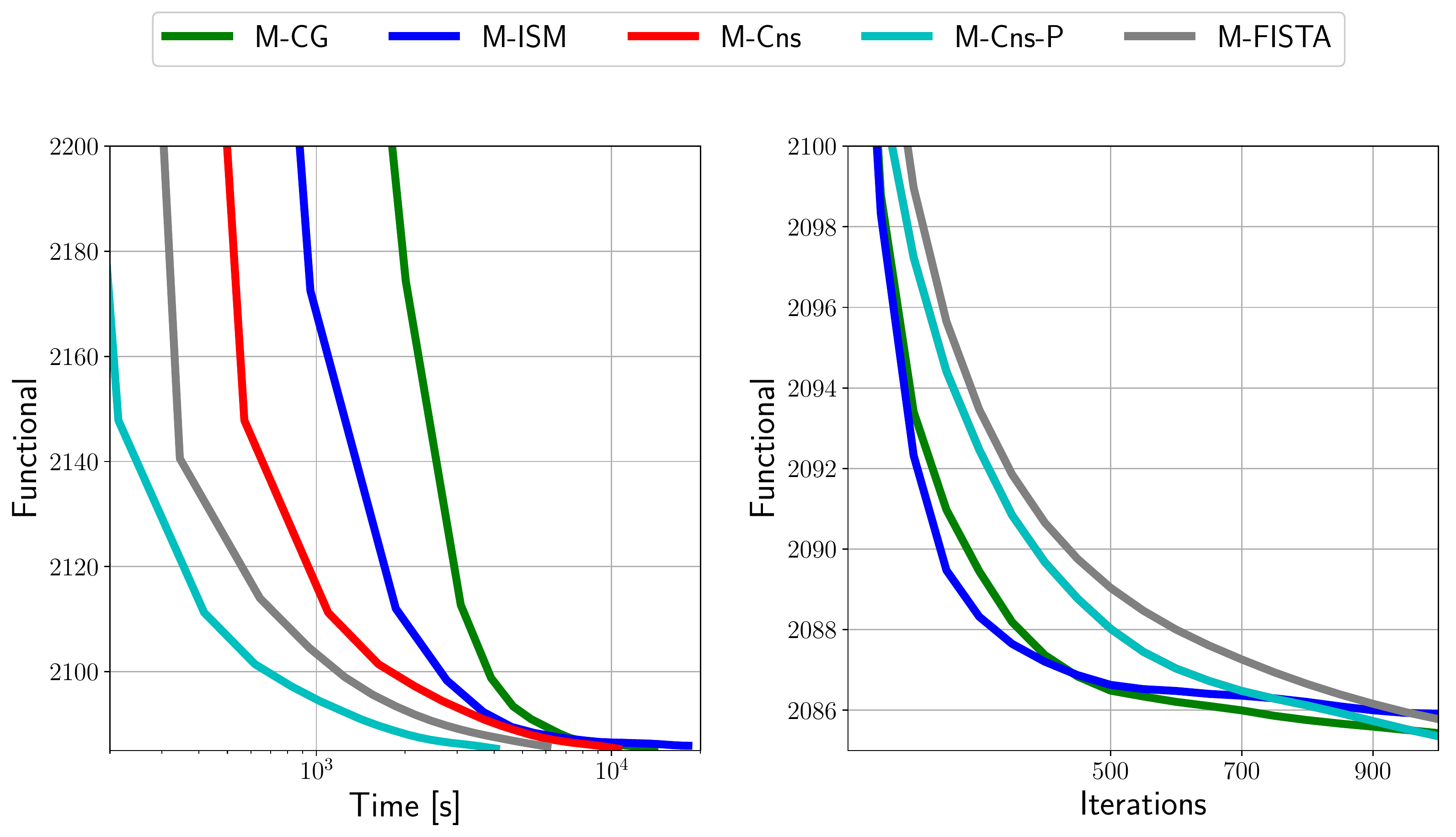}}
        \caption{Evolution of the CBPDN functional~\eq{cbpdnmmv} for the test set using the partial dictionaries obtained when training for $K=20$ images for masked versions of the algorithms. M-Cns and M-Cns-P overlap in the iterations plot.}
        \label{fig:val_k20_mask}
\end{figure}

\begin{figure}[htb]
        \center
        \scalebox{0.26}{\includegraphics{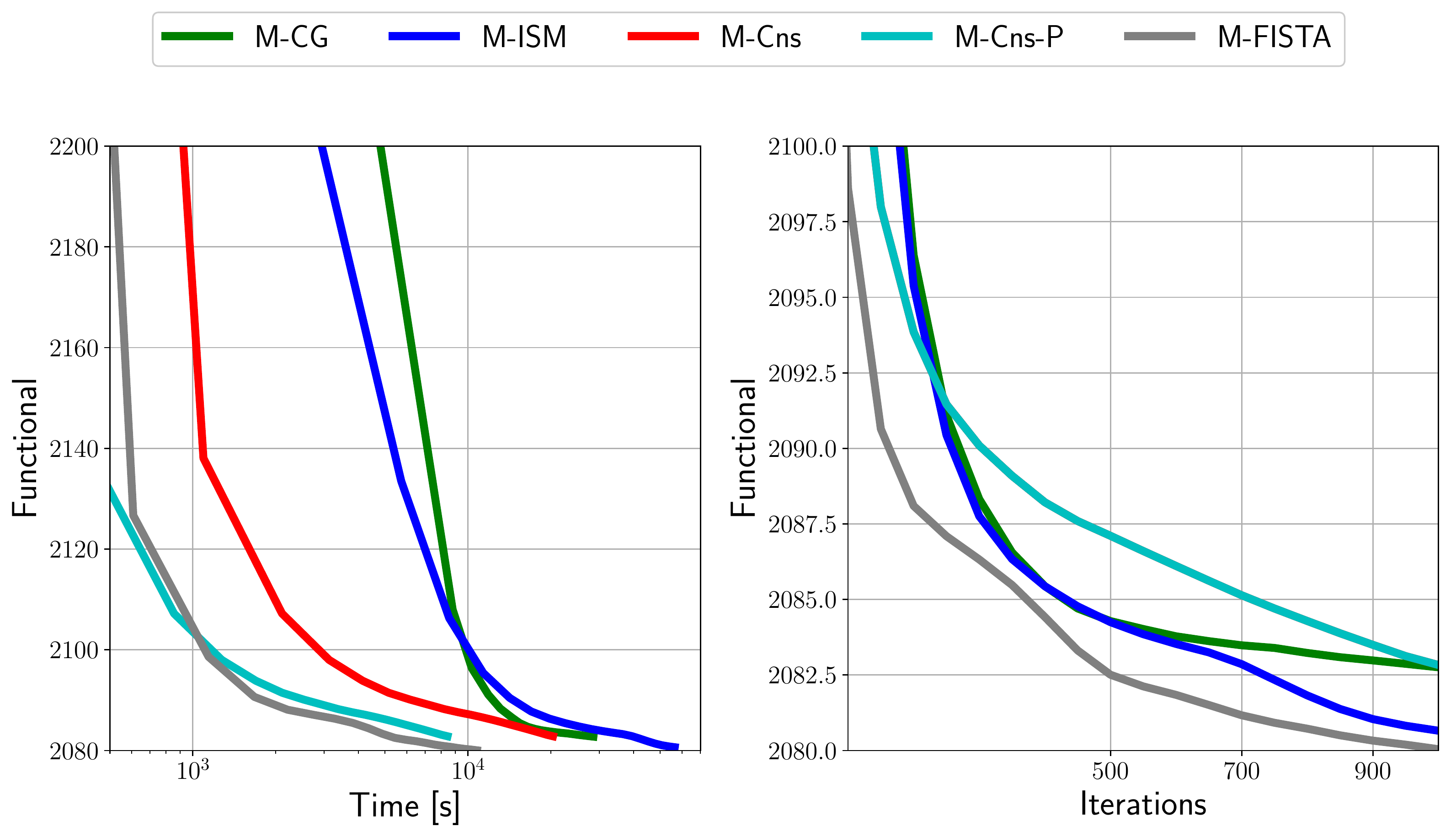}}
        \caption{Evolution of the CBPDN functional~\eq{cbpdnmmv} for the test set using the partial dictionaries obtained when training for $K=40$ images for masked versions of the algorithms. M-Cns and M-Cns-P overlap in the iterations plot.}
        \label{fig:val_k40_mask}
\end{figure}

To provide a comparison that takes into account any possible differences in overfitting and generalization properties of the dictionaries learned by the different methods, we ran experiments over a 20 image test set that is not used during learning.  For all the methods discussed, we saved the dictionaries at 50 iteration intervals (including the final one obtained at 1000 iterations) while training. These dictionaries were used to sparse code the images in the test set with $\lambda=0.1$, allowing evaluation of the evolution of the test set CBPDN functional as the dictionaries change during training. Results for the dictionaries learned while training with $K=20$ and $K=40$ images are shown in~\figs{val_k20} and~\fign{val_k40} respectively, and corresponding results for the algorithms with a spatial mask are shown in~\figs{val_k20_mask} and~\fign{val_k40_mask} respectively. Note that the time axis in these plots refers to the run time of the dictionary learning code used to generate the relevant dictionary, and \emph{not} to the run time of the sparse coding on the test set.

As expected, independent of the method, the dictionaries obtained for training with 40 images exhibit better performance than the ones trained with 20 images.
Overall, performance on training is a good predictor of performance in testing, which suggests that the functional value on a sufficiently large training set is a reliable indicator of dictionary quality.

\subsection{Penalty Parameter Selection}
\label{sec:ppselect}

The grid searches performed for determining optimal parameters ensure a fair comparison between the methods, but they are not convenient as a general approach to parameter selection. In this section we show that it is possible to construct heuristics that allow reliable parameter selection for the best performing CDL methods considered here.

\subsubsection{Parameter Scaling Properties}

Estimates of parameter scaling properties with respect to $K$ are derived in~\sctn{parameters_analytic_sm} in the Supplementary Material. For the CDL problem without a spatial mask, these scaling properties are derived for the sparse coding problem, and for the dictionary updates based on ADMM with an equality constraint, ADMM consensus, and FISTA. These estimates indicate that the scaling of the penalty parameter $\rho$ for the convolutional sparse coding is $\orderof(1)$, the scaling of the penalty parameter $\sigma$ for the dictionary update is $\orderof(K)$ for the ADMM with equality constraint and $\orderof(1)$ for ADMM consensus, and the scaling of the step size $L$ for FISTA is $\orderof(K)$. Derivations for the Tiled and 3D methods do not lead to a simple scaling relationship, and are not included.

For the CDL problem with a spatial mask, these scaling properties are derived for the sparse coding problem, and for the dictionary updates based on ADMM with a block-constraint, and extended ADMM consensus. The scaling of the penalty parameter $\rho$ for the masked version of convolutional sparse coding is $\orderof(1)$, the scaling of the penalty parameter $\sigma$ for the dictionary update in the extended consensus framework is $\orderof(1)$, while there is no simple rule of the $\sigma$ scaling in the block-constraint ADMM of \sctn{blkadmmmdu}.

\subsubsection{Parameter Selection Guidelines}
\label{sec:parameter_sensitivity}

The derivations discussed above indicate that the optimal algorithm parameters should be expected to be either constant or linear in $K$. For the parameters of the most effective CDL algorithms, \ie CG, Cns, FISTA, and M-Cns, we performed additional computational experiments to estimate the constants in these scaling relationships. Cns-P and M-Cns-P have the same parameter dependence as their serial counterparts, and are therefore not evaluated separately. Similarly, M-FISTA is not included in these experiments because it has the same functional evolution as FISTA for the identity mask $W = I$.

\begin{table}[htb]
\caption{Grid Search Ranges}
\label{tab:paramGrid}
\renewcommand{\arraystretch}{1.4}
\centering
\begin{tabular}{| c | c | l |}
\hline
  \multicolumn{1}{|c|}{{\bf Parameter}} & \multicolumn{1}{c}{{\bf Method}} & \multicolumn{1}{|c|}{ {\bf Range} }  \\
 \hline
  & {CG} & $[10^{0.1}, 10^{1.1}]$ \\
 \cline{2-3}
 $\rho$ & {Cns} & $[10^{0.25}, 10^{1.2}]$ \\
 \cline{2-3}
  & {M-Cns} & $[10^{0.33}, 10]$ \\
 \cline{2-3}
  & {FISTA} & $[10^{0.14}, 10]$ \\
  \hline
  \hline
 $\sigma$  & {CG} & $[1, 10^{2.5}]$ \\
 \cline{2-3}
   & {{Cns}, {M-Cns}} & $[10^{-1}, 10]$ \\
 \hline
  \hline
 $L$ & {FISTA} & $[10, 10^{2.9}]$ \\
 \hline
\end{tabular}
\end{table}

For each training set size $K \in \{5, 10, 20\}$, we constructed an ensemble of 20 training sets of that size by random selection from the 40 image training set. For each CDL algorithm and each $K$, the dependence of the convergence behavior on the algorithm parameters was evaluated by computing 500 iterations of the CDL algorithm for all 20 members of the ensemble of size $K$, and over grids of $(\rho, \sigma)$ values for the ADMM dictionary updates, and $(\rho, L)$ values for the FISTA dictionary updates. The parameter grids consisted of 10 logarithmically spaced points in the ranges specified in~\tbl{paramGrid}. These parameter ranges were set such that the corresponding functional values remained within 0.1\% to 1\% of their optimal values.

\begin{figure}[htb]
\centerline{
\subfigure[CG $\rho$]{\scalebox{0.18}{\includegraphics{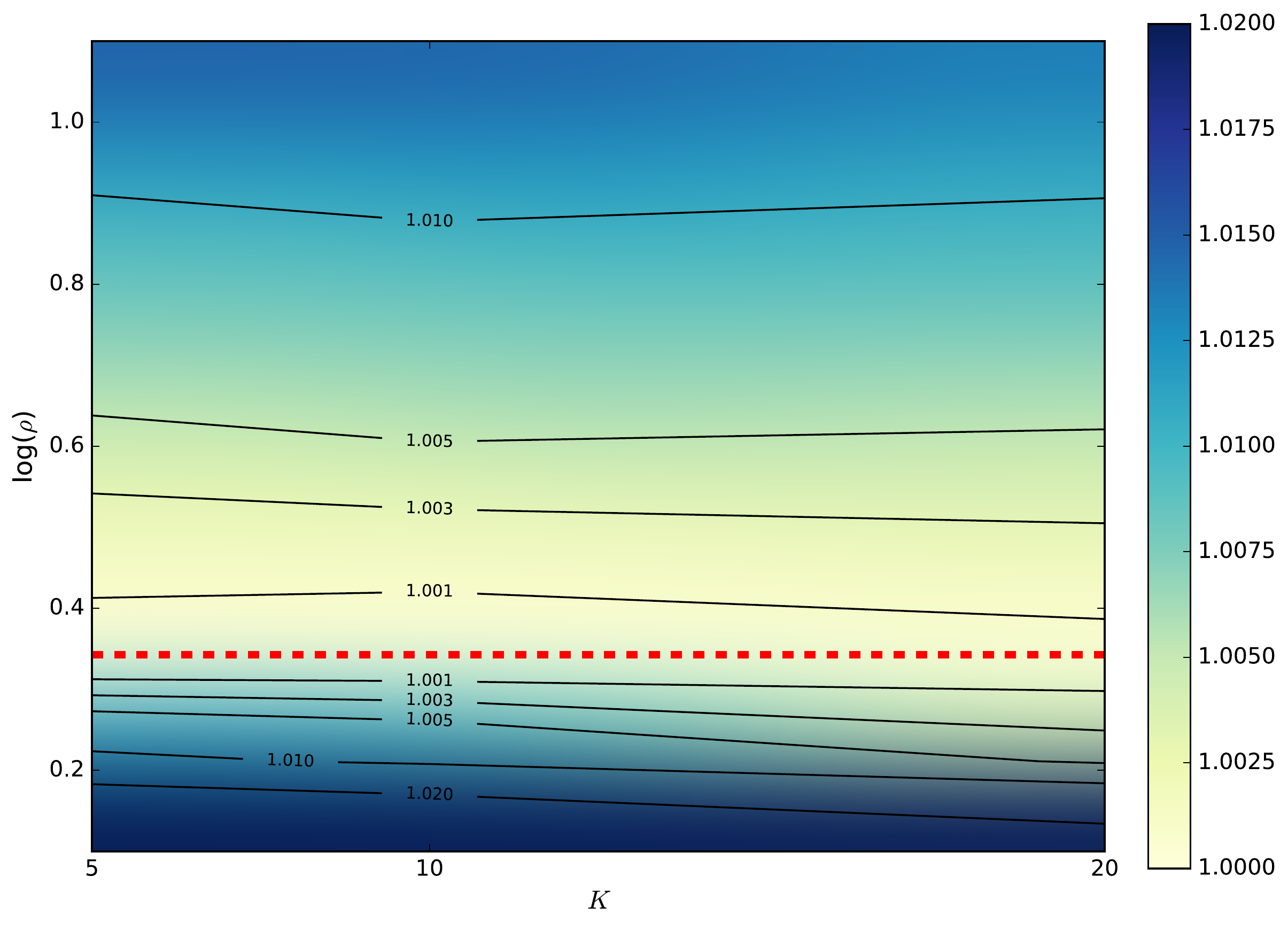}}}
\hfil
\subfigure[CG $\sigma$]{\scalebox{0.18}{\includegraphics{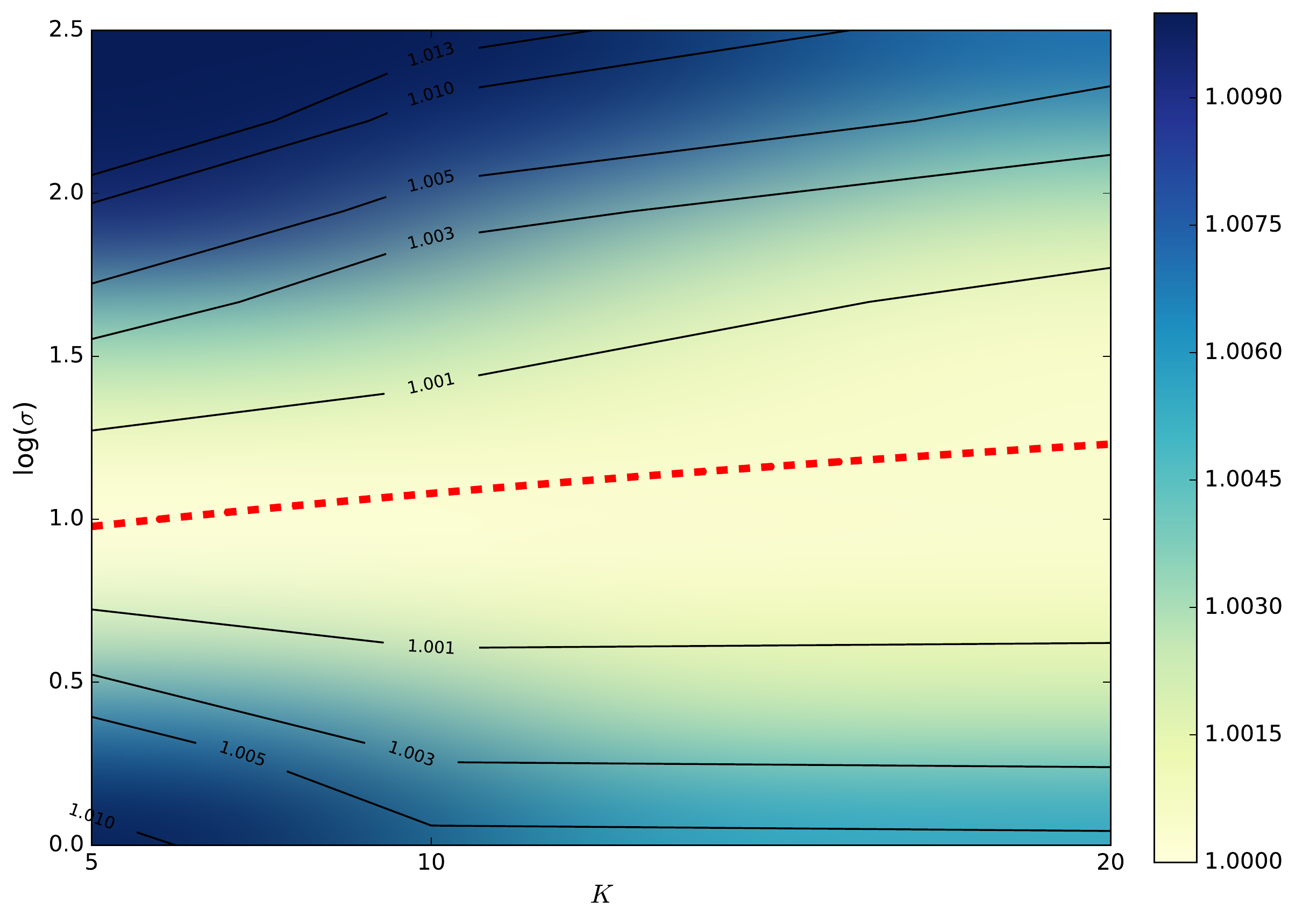}}}
}
\vfil
\centerline{
\subfigure[Cns $\rho$]{\scalebox{0.18}{\includegraphics{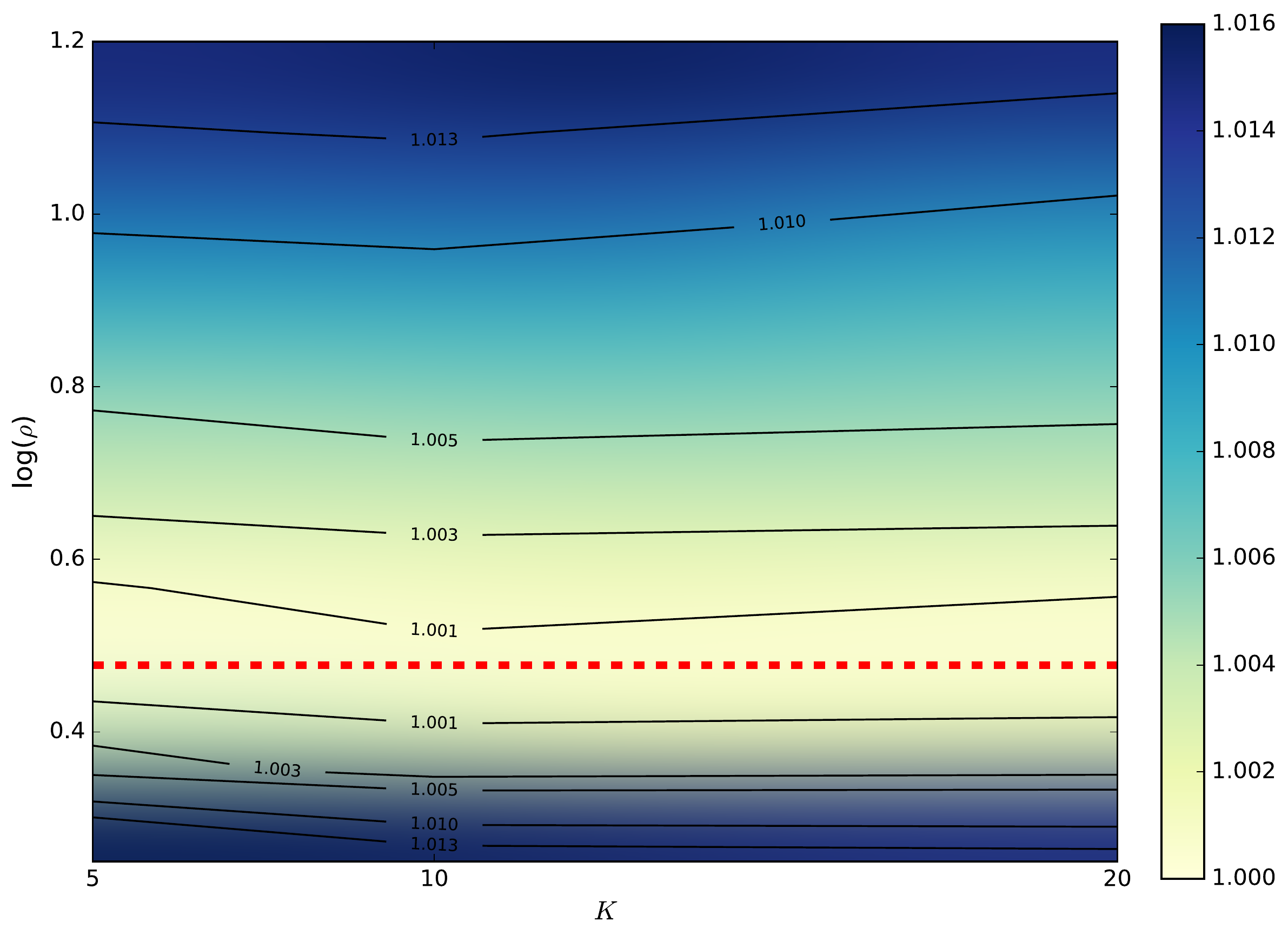}}}
\hfil
\subfigure[Cns $\sigma$]{\scalebox{0.18}{\includegraphics{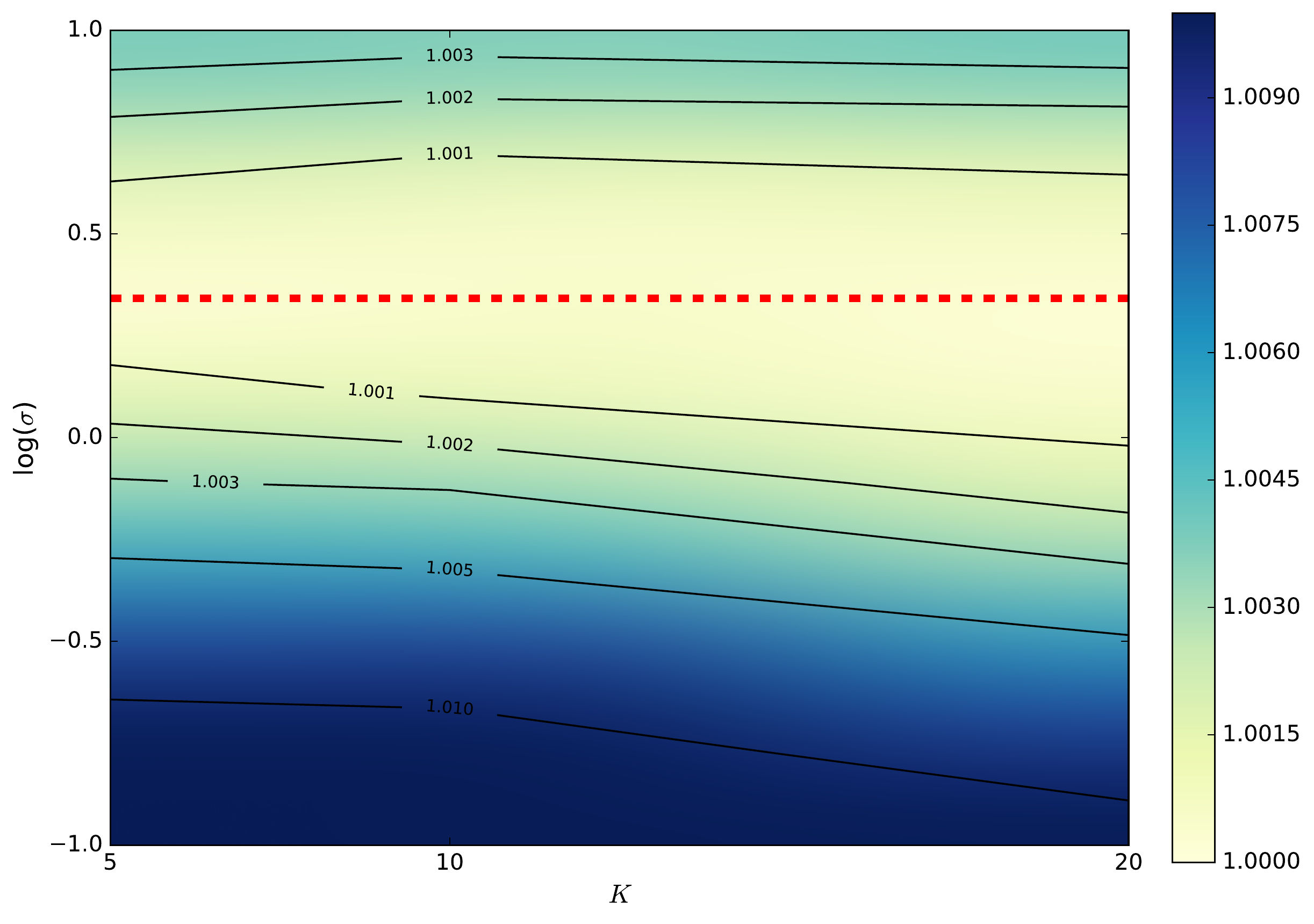}}}
}
\vfil
\centerline{
\subfigure[M-Cns $\rho$]{\scalebox{0.18}{\includegraphics{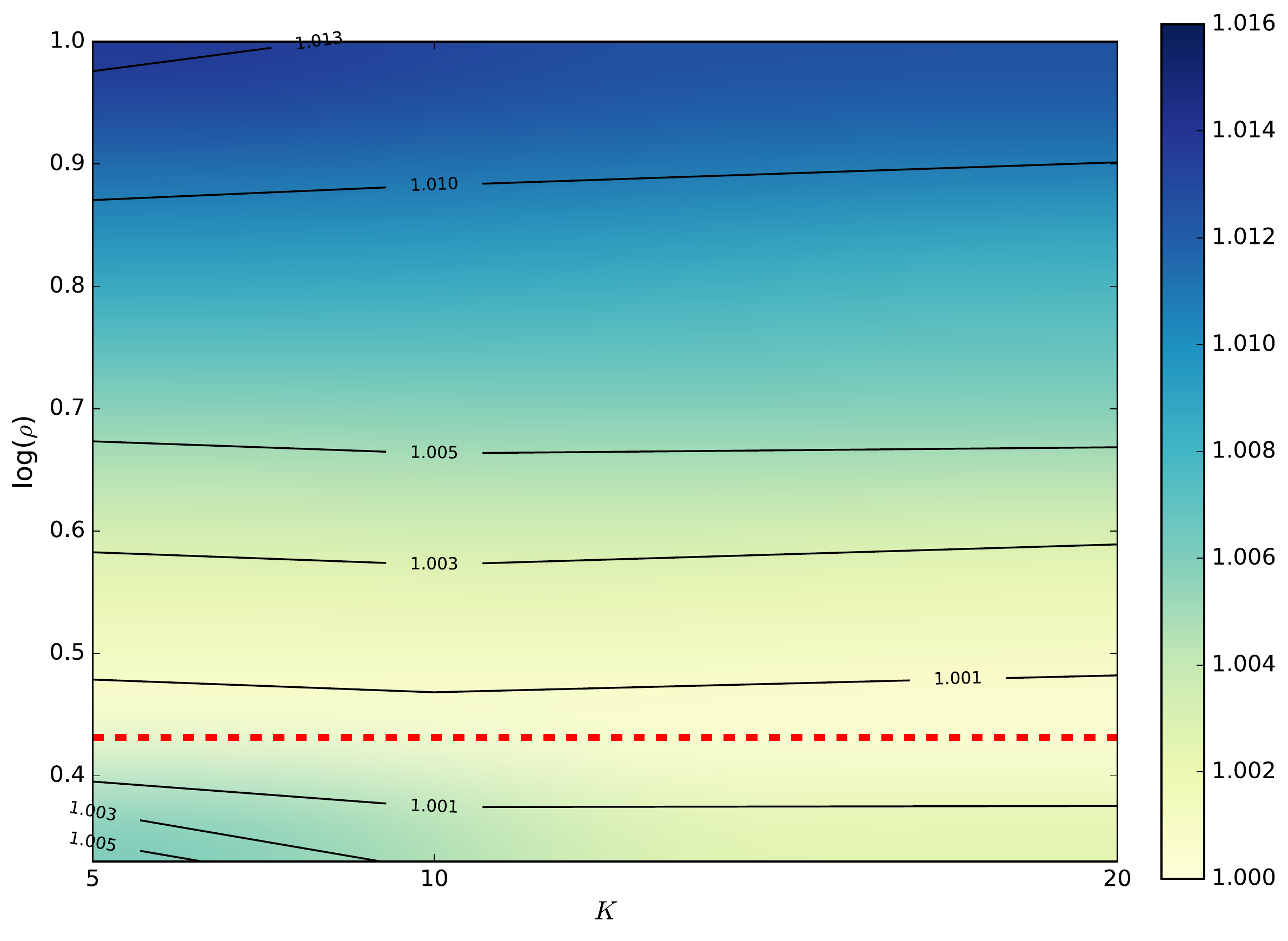}}}
\hfil
\subfigure[M-Cns $\sigma$]{\scalebox{0.18}{\includegraphics{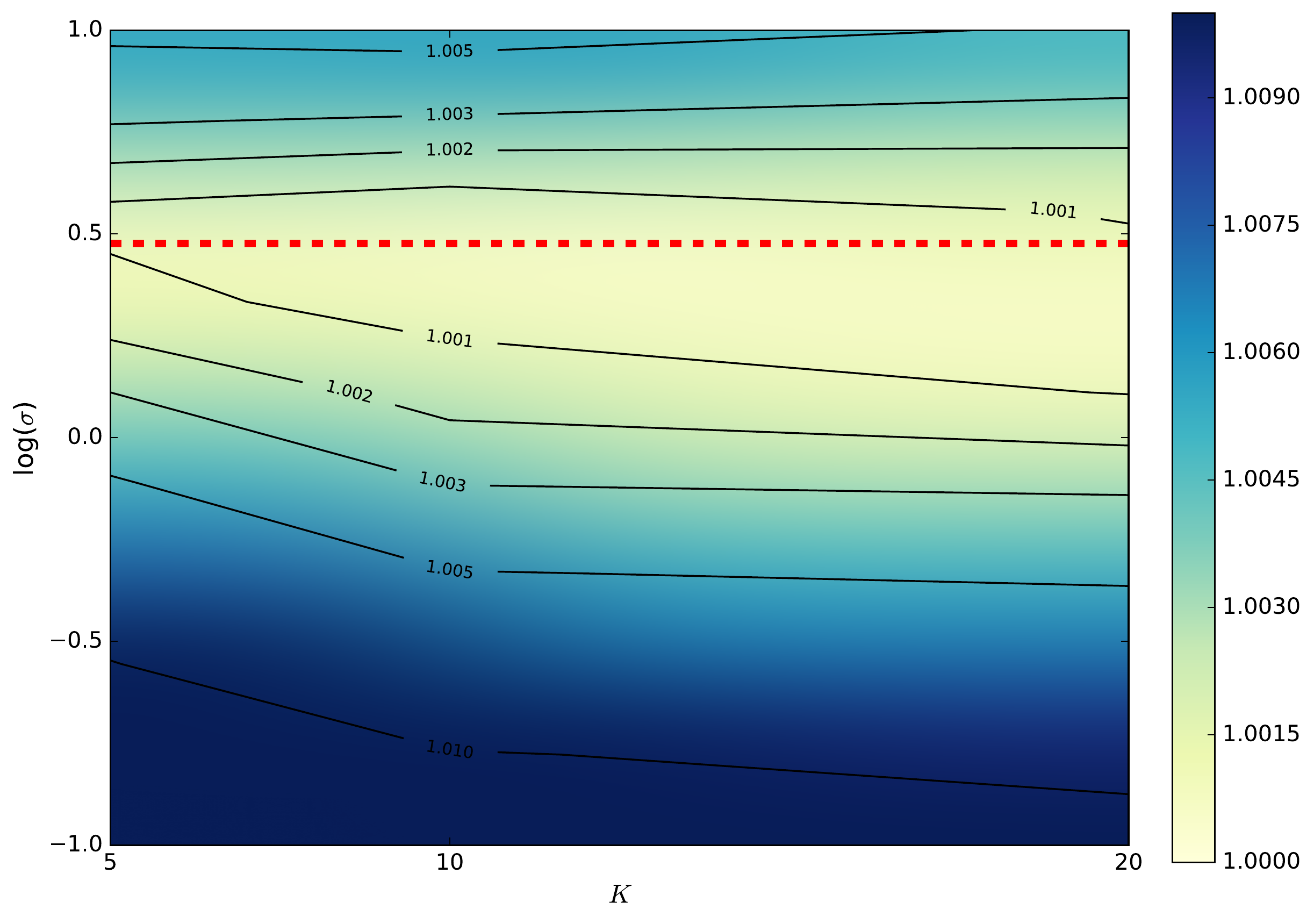}}
\label{fig:Mcns_sigma}}
}
\vfil
\centerline{
\subfigure[FISTA $\rho$]{\scalebox{0.18}{\includegraphics{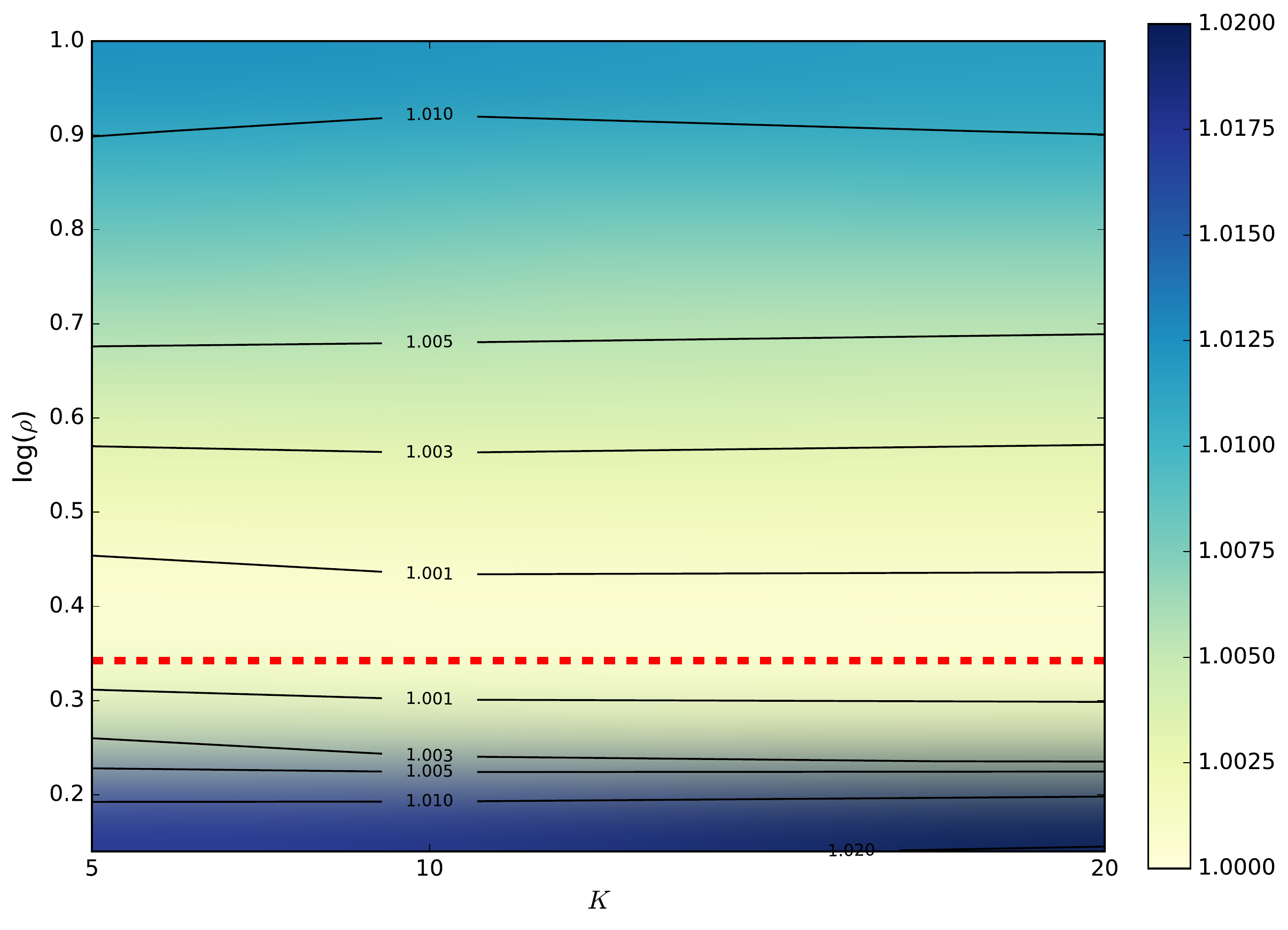}}}
\hfil
\subfigure[FISTA $L$]{\scalebox{0.18}{\includegraphics{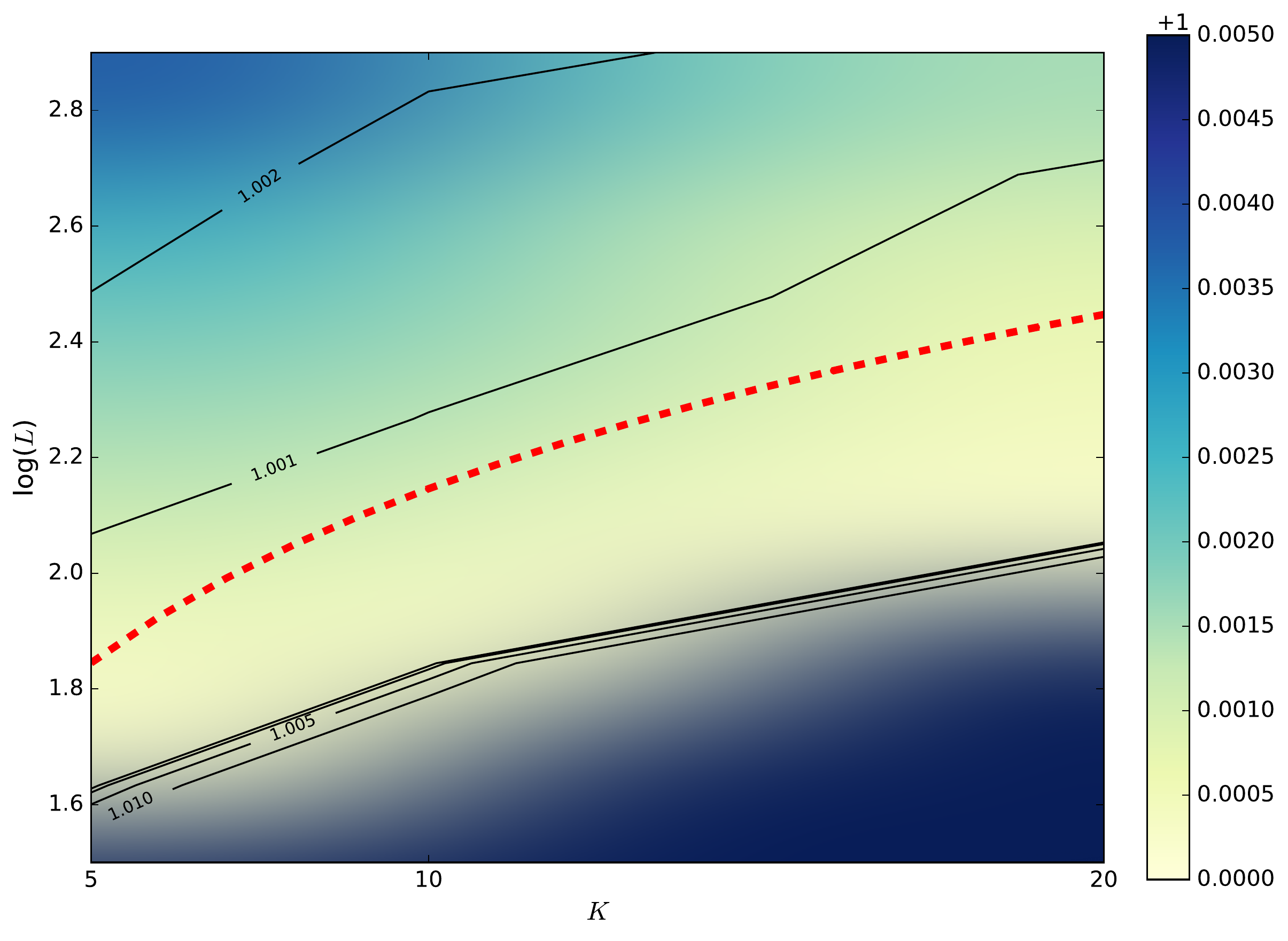}}
\label{fig:FISTA_L}}
}
\caption{Contour plots of the ensemble median of the normalized CDL functional values for different algorithm parameters. The black lines correspond to level curves at the indicated values of the plotted surfaces, and the dashed red lines represent parameter selection guidelines that combine the analytic derivations with the empirical behavior of the plotted surfaces.}
        \label{fig:parameters_scaling}
\end{figure}

\begin{table}[htb]
\caption{Penalty Parameter Selection Guidelines}
\label{tab:paramRules}
\centering
\renewcommand{\arraystretch}{1.2}
\begin{tabular}{| c | c | l |}
\hline
  \multicolumn{1}{|c|}{{\bf Parameter}} & \multicolumn{1}{c}{{\bf Method}} & \multicolumn{1}{|c|}{ {\bf Rule} }  \\
 \hline
  & {CG, ISM, FISTA} & $\rho = 2.2$ \\
 \cline{2-3}
 $\rho$ & {Cns} & $\rho = 3.0$ \\
 \cline{2-3}
  & {M-Cns} & $\rho = 2.7$ \\
  \hline
  \hline
  & {{ CG},  { ISM}} & $\sigma = 0.5 K + 7.0$ \\
 \cline{2-3}
 $\sigma$ &  {Cns} & $\sigma = 2.2$ \\
  \cline{2-3}
   & {M-Cns} & $\sigma = 3.0$ \\
 \hline
  \hline
 $L$ & {FISTA} & $L = 14.0 K$ \\
 \hline
\end{tabular}
\end{table}

We normalized the results for each training set by dividing by the minimum of the functional for that set, and computed statistics over these normalized values for all sets of the same size, $K$. These statistics, which are reported as box plots in~\sctn{sensitivity_sm} of the Supplementary Material, were also aggregated into contour plots of the median (across the ensemble of training images sets of the same size) of the normalized CDL functional values, displayed in~\fig{parameters_scaling}. (Results for ISM are the same as for CG and are not shown.) In each of these contour plots, the horizontal axis corresponds to the number of training images, $K$, and the vertical axis corresponds to the parameter of interest. The scaling behavior of the optimal parameter with $K$ can clearly be seen in the direction of the valley in the contour plots. Parameter selection guidelines obtained by manual fitting of the constant or linear scaling behavior to these contour plots are plotted in red, and are also summarized in~\tbl{paramRules}.

In~\fig{Mcns_sigma}, the guideline for $\sigma$ for M-Cns does not appear to follow the path of the 1.001 level curves. We did not select the guideline to follow this path because (i) the theoretical estimate of the scaling properties of this parameter with $K$ in~\sctn{mcnskscale} of the Supplementary Material is that it is constant, and (ii) the path suggested by the 1.001 level curves leads to a logarithmically decreasing curve that would reach negative parameter values for sufficiently large $K$. We do not have a reliable explanation for the unexpected behavior of the 1.001 level curves, but suspect that it may be related to the loss of diversity of training image sets for $K=20$, since each of these sets of 20 images was chosen from a fixed set of 40 images. It is also worth noting that the upper level curves for larger functional values, e.g. 1.002, do not follow the same unexpected decreasing path.

To guarantee convergence of FISTA, the inverse of the gradient step size, $L$, has to be greater than or equal to the Lipschitz constant of the gradient of the functional~\cite{beck-2009-fast}. In~\fig{FISTA_L}, the level curves below the guideline correspond to this potentially unstable regime where the functional value surface has a large gradient. The gradient of the surface is much smaller above the guideline, indicating that convergence is not very sensitive to the parameter value in this region. We chose the guideline precisely to be more biased towards the stable regime.

The parameter selection guidelines presented in this section should only be expected to be reliable for training data with similar characteristics to those used in our experiments, \ie natural images pre-processed as described in~\sctn{cdlexp}, and for the same or similar sparsity parameter, \ie $\lambda = 0.1$. Nevertheless, since the scaling properties derived in~\sctn{parameters_analytic_sm} of the Supplementary Material remain valid, it is reasonable to expect that similar heuristics, albeit with different constants, would hold for different training data or sparsity parameter settings.

\section{Conclusions}
\label{sec:concl}

Our results indicate that two distinct approaches to the dictionary update problem provide the leading CDL algorithms. In a serial processing context, the FISTA dictionary update proposed here outperforms all other methods, including consensus, for CDL with and without a spatial mask. This may seem surprising when considering that ADMM outperforms FISTA on the CSC problem, but is easily understood when taking into account the critical difference between the linear systems that need to be solved when tackling the CSC and convolutional dictionary update problems via proximal methods such as ADMM and FISTA. In the case of CSC, the major linear system to be solved has a frequency domain structure that allows very efficient solution via the Sherman-Morrison formula, providing an advantage to ADMM. In contrast, except for the $K=1$ case, there is no such highly efficient solution for the convolutional dictionary update, giving an advantage to methods such as FISTA that employ gradient descent steps rather than solving the linear system.

In a parallel processing context, the consensus dictionary update proposed in~\cite{sorel-2016-fast} used together with the alternative CDL algorithm structure proposed in~\cite{garcia-2017-subproblem} leads to the CDL algorithm with the best time performance for the mask-free CDL problem, and the hybrid mask decoupling/consensus dictionary update proposed here provides the best time performance for the masked CDL problem. It is interesting to note that, despite the clear suitability of the ADMM consensus framework for the convolutional dictionary update problem, a parallel implementation is essential to outperforming other methods; in a serial processing context it is significantly outperformed by the FISTA dictionary update, and even the CG method is competitive with it.

We have also demonstrated that the optimal algorithm parameters for the leading methods considered here tend to be quite stable across different training sets of similar type, and have provided reliable heuristics for selecting parameters that provide good performance. It should be noted, however, that FISTA appears to be more sensitive to the $L$ parameter than the ADMM methods are to the penalty parameter.

The additional experiments reported in the Supplementary Material indicate that the FISTA and parallel consensus methods are scalable to relatively large training sets, \eg 100 images of $512 \times 512$ pixels. The computation time exhibits linear scaling in the number of training images, $K$, and the number of dictionary filters, $M$, and close to linear scaling in the number of pixels in each image, $N$. The limited experiments involving color dictionary learning indicate that the additional computational cost compared with greyscale dictionary learning is moderate. Comparisons with the publicly available implementations of complete CDL methods by other authors indicate that:
\begin{IEEEitemize}
  \item The method of Heide \etal~\cite{heide-2015-fast} does not scale well to training images sets of even moderate size, exhibiting very slow convergence with respect to computation time.
  \item While the consensus CDL method proposed here gives very good performance, the consensus method of \v{S}orel and \v{S}roubek~\cite{sorel-2016-fast} converges much more slowly, and does not learn dictionaries with properly normalized filters\footnote{It is not clear whether this is due to weaknesses in the algorithm, or to errors in the implementation.}.
  \item The method of Papyan \etal~\cite{papayan-2017-convolutional} converges rapidly with respect to the number of iterations, and appears to scale well with training set size, but is slower than the FISTA and parallel consensus methods with respect to time, and the resulting dictionaries do not offer competitive performance to the leading methods proposed here in terms of performance on testing image sets.
\end{IEEEitemize}

In the interest of reproducible research, software implementations
of the algorithms considered here have been made publicly available as part of the SPORCO library~\cite{wohlberg-2016-sporco, wohlberg-2017-sporco}.

\bibliographystyle{IEEEtranD}
\bibliography{bcdl}

\clearpage

\begin{center}
\textbf{\large Convolutional Dictionary Learning: A Comparative Review and New Algorithms (Supplementary Material)}
\end{center}
\setcounter{equation}{0}
\setcounter{figure}{0}
\setcounter{table}{0}
\setcounter{page}{1}
\setcounter{section}{0}
\makeatletter
\renewcommand{\theequation}{S\arabic{equation}}
\renewcommand{\thefigure}{S\arabic{figure}}
\renewcommand*{\thesection}{S\Roman{section}}

\section{Introduction}
\label{sec:smintro}

This document provides additional detail and results that were omitted
from the main document due to space restrictions. All citations refer to the References section of the main document.

\section{Penalty Parameter Grid Search}
\label{sec:grids_sm}

\begin{figure}[htb]
\centerline{
\subfigure[CG]{\scalebox{0.18}{\includegraphics{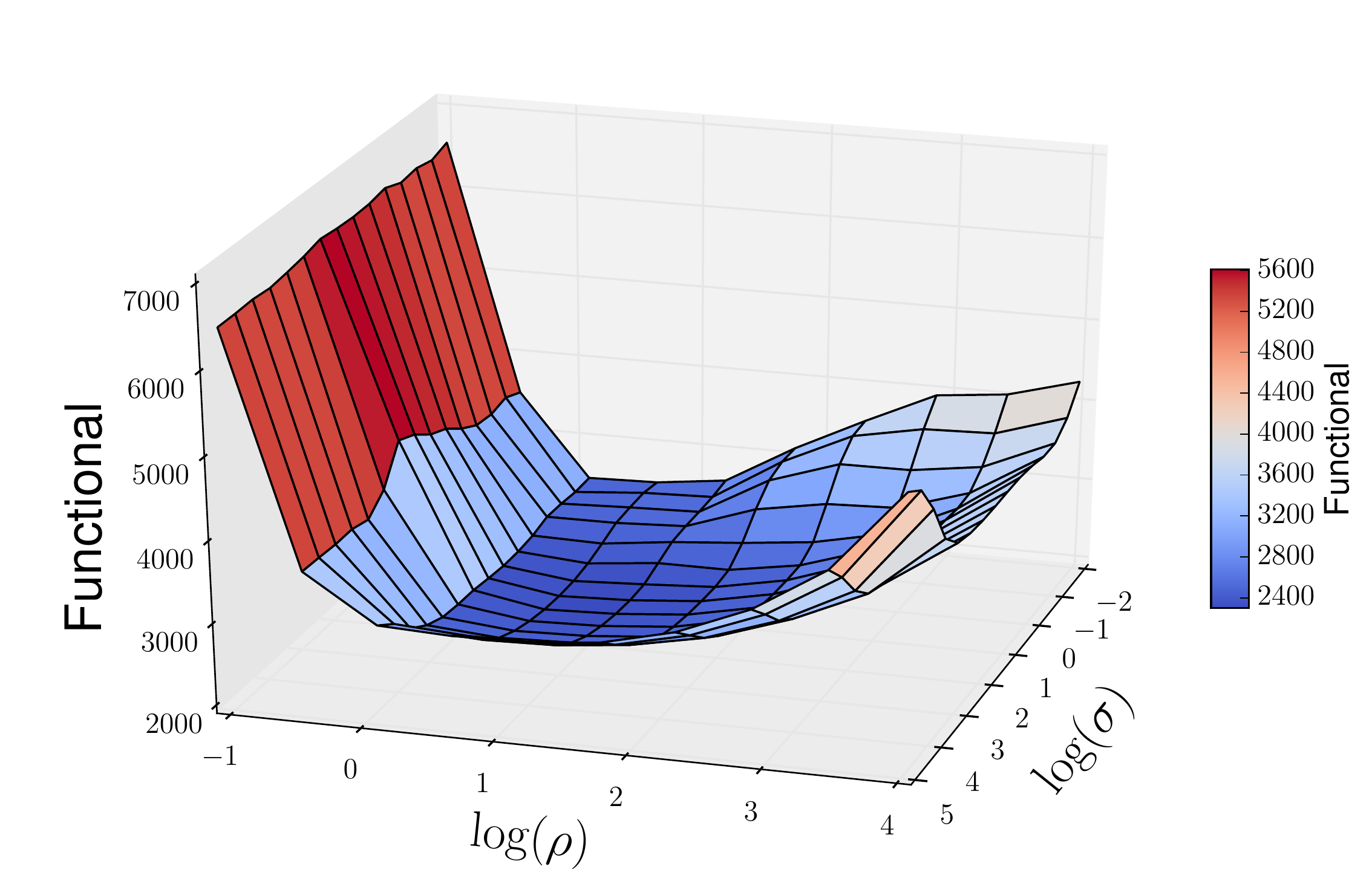}}}
\hfil
\subfigure[ISM]{\scalebox{0.18}{\includegraphics{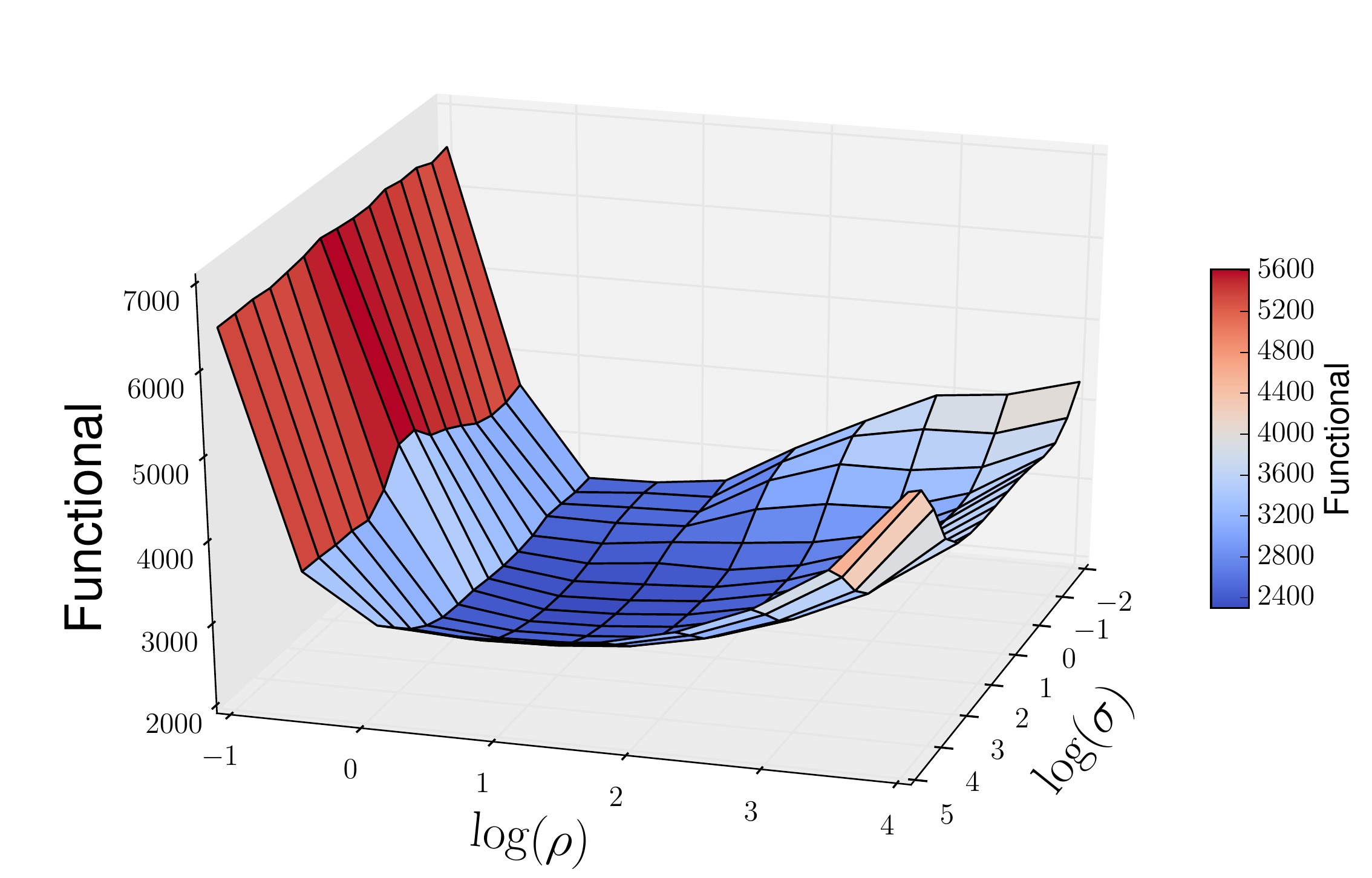}}}
}
        \caption{Grid search surfaces for conjugate gradient (CG) and Iterated Sherman-Morrison (ISM) algorithms with $K=20$. Each surface represents the value of the CBPDN functional (\eq{cbpdnmmv} in the main document) after 100 iterations, for different parameters $\rho$ and $\sigma$.}
        \label{fig:grid_a}
\end{figure}

\begin{figure}[htb]
\centerline{
\subfigure[Tiled]{\scalebox{0.18}{\includegraphics{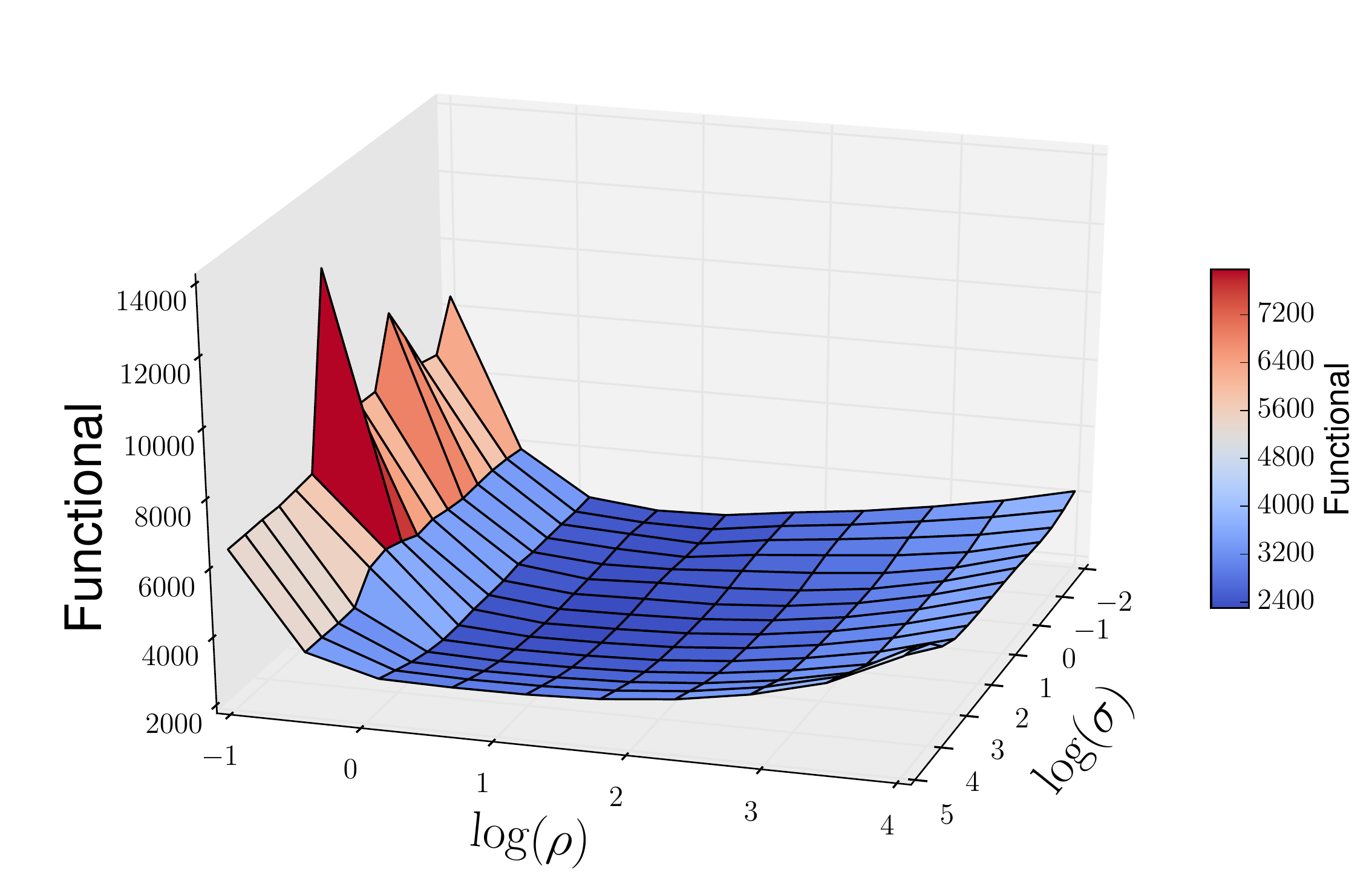}}}
\hfil
\subfigure[Cns]{\scalebox{0.18}{\includegraphics{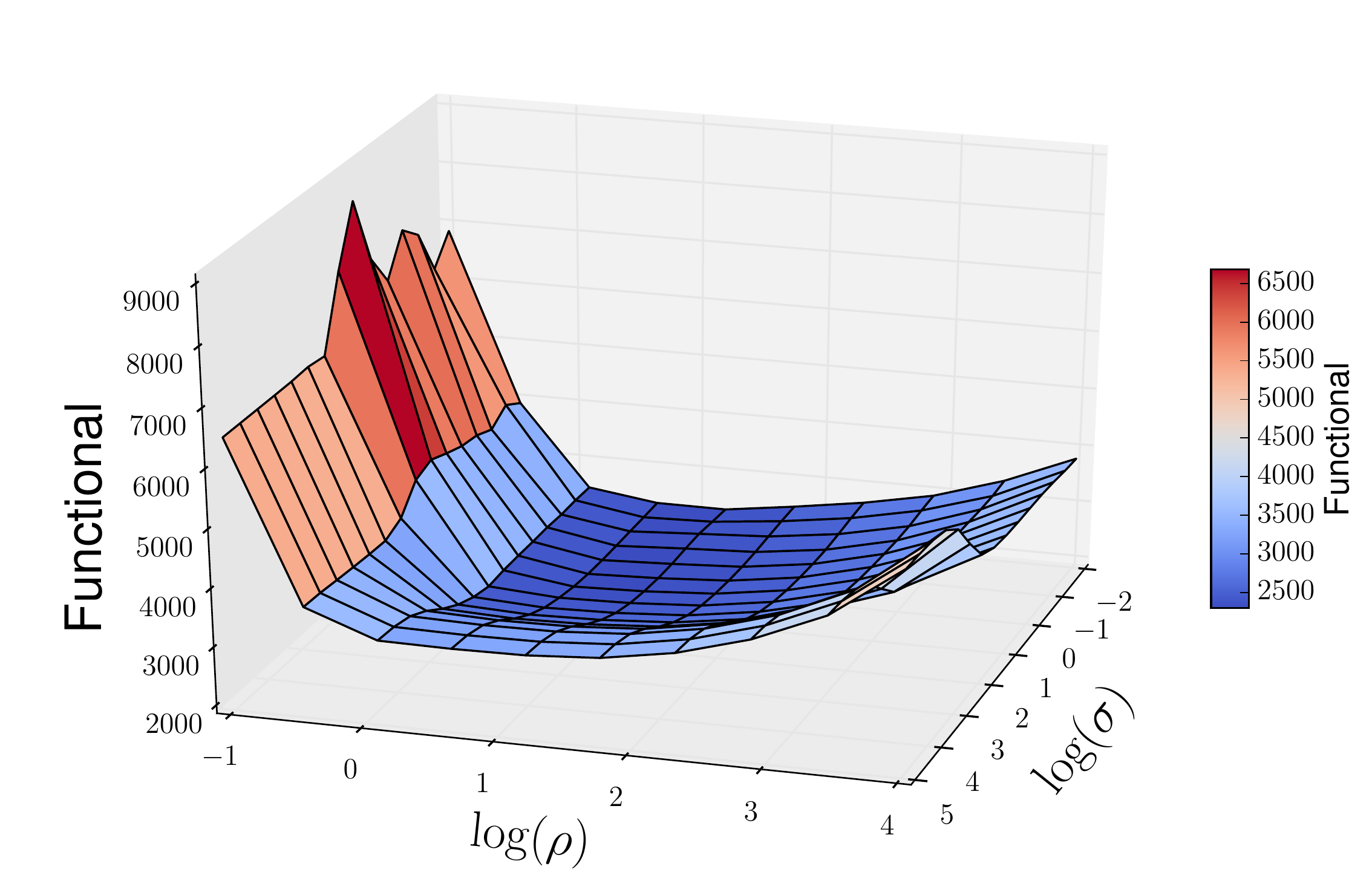}}}
}
\vfil
\centerline{
\subfigure[3D]{\scalebox{0.18}{\includegraphics{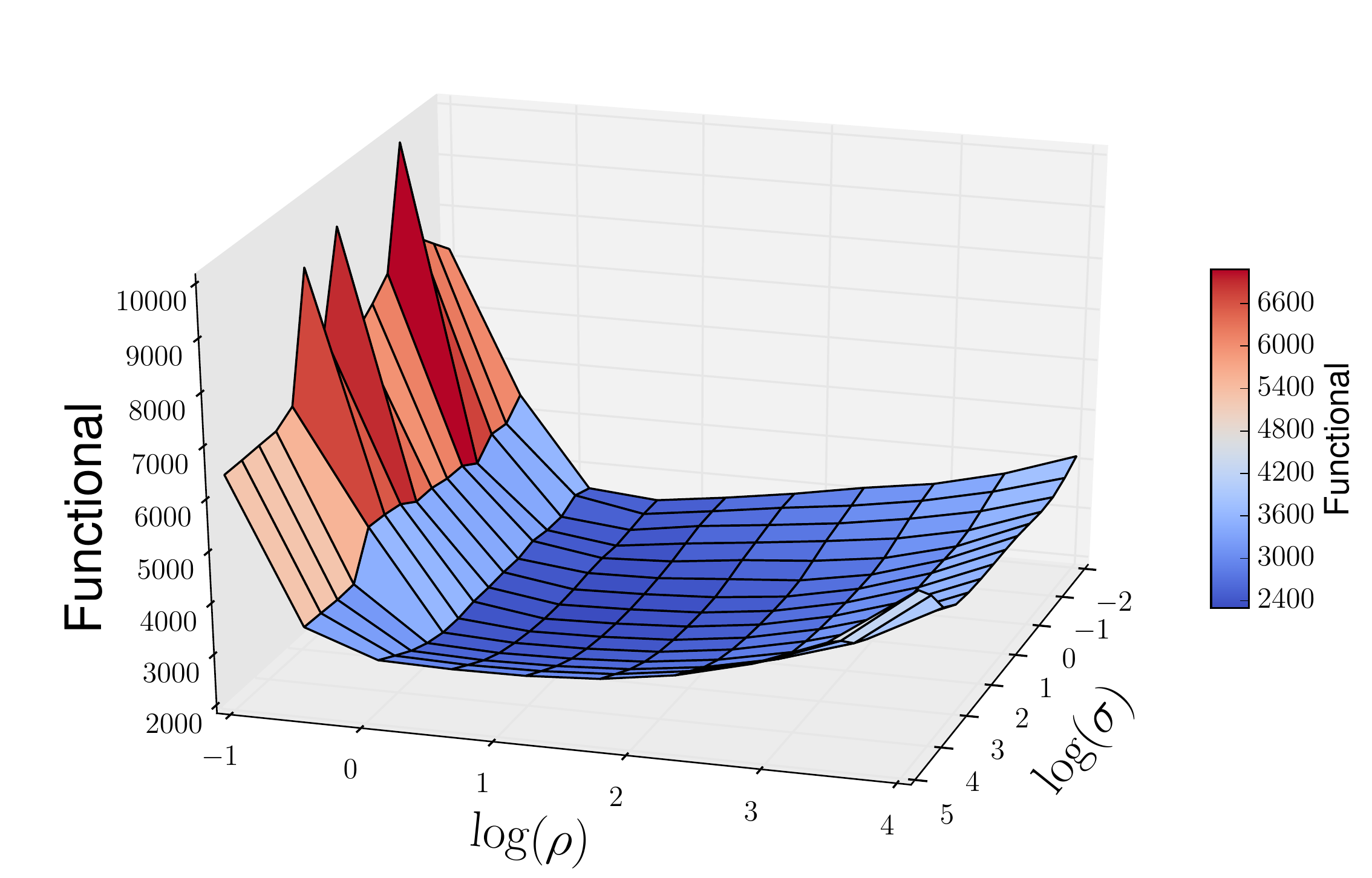}}}
\hfil
\subfigure[FISTA]{\scalebox{0.18}{\includegraphics{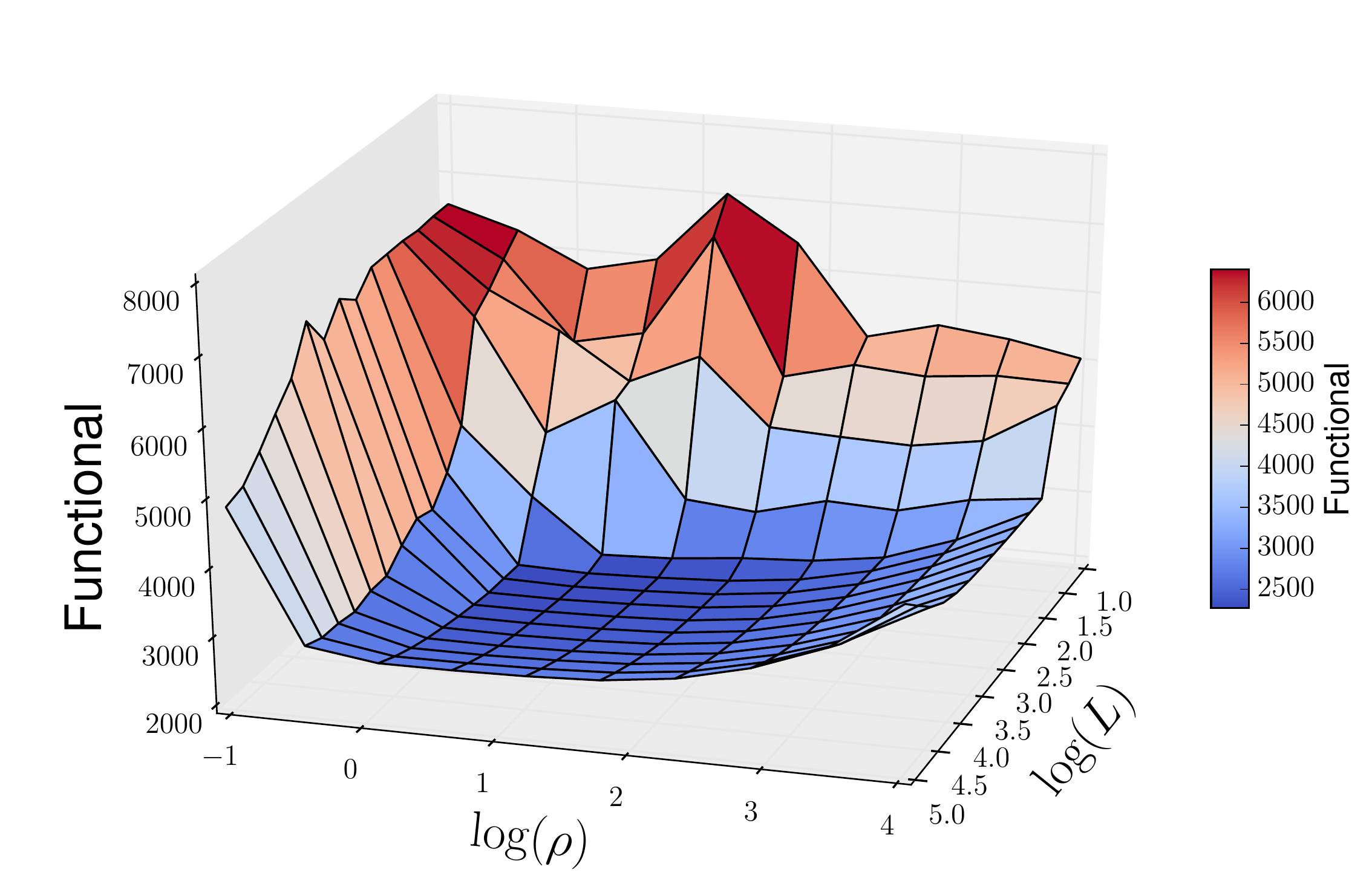}}}
}
        \caption{Grid search surfaces for spatial tiling (Tiled), consensus (Cns), frequency domain consensus (3D) and FISTA algorithms with $K=20$. Each surface represents the value of the CBPDN functional (\eq{cbpdnmmv} in the main document) after 100 iterations, for different parameters $\rho$, and $\sigma$ or $L$.}
        \label{fig:grid_b}
\end{figure}

\begin{figure}[htb]
\centerline{
\subfigure[M-CG]{\scalebox{0.18}{\includegraphics{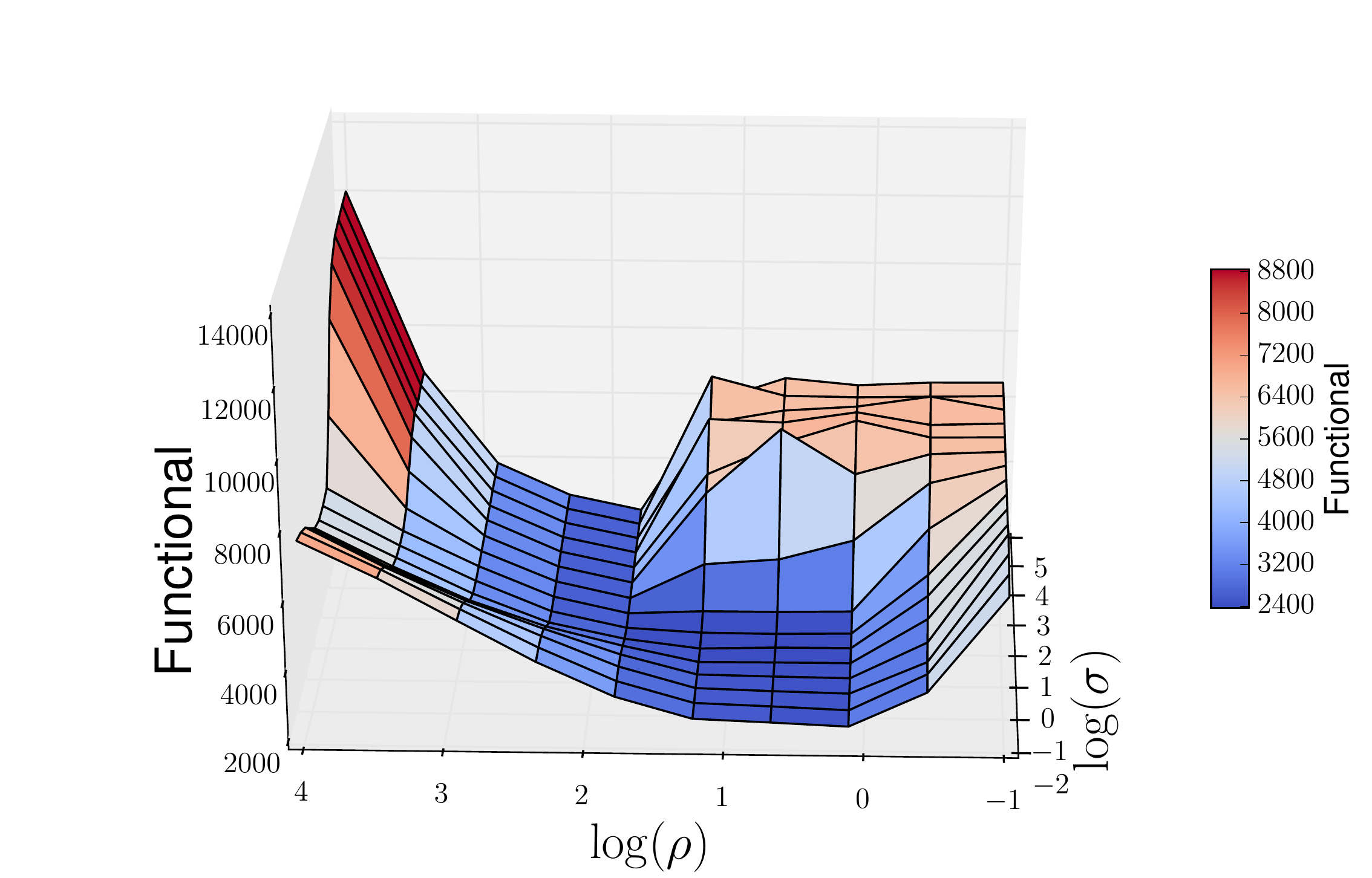}}}
\hfil
\subfigure[M-ISM]{\scalebox{0.18}{\includegraphics{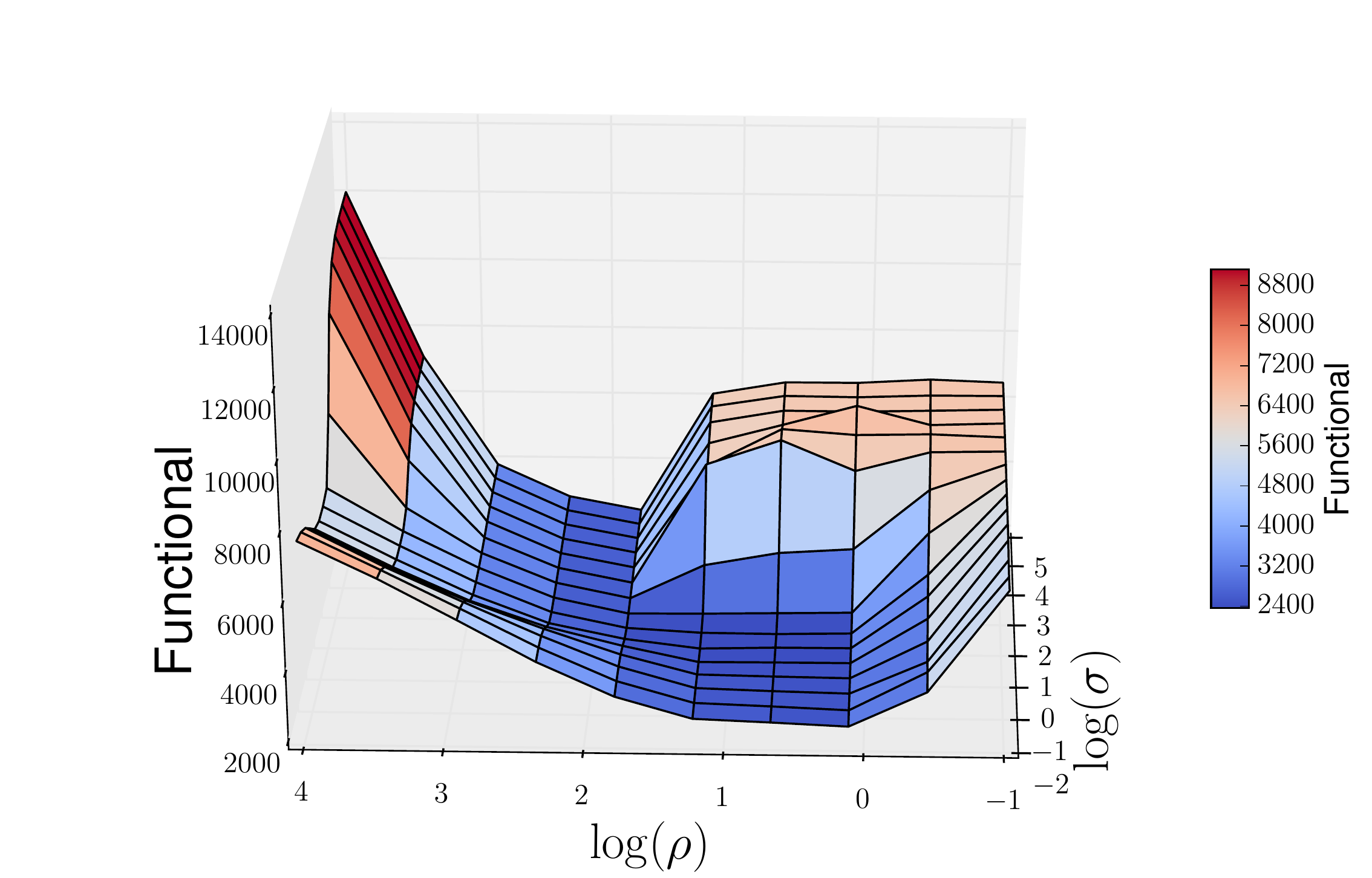}}}
}
\vfil
\centerline{
\subfigure[M-Cns]{\scalebox{0.18}{\includegraphics{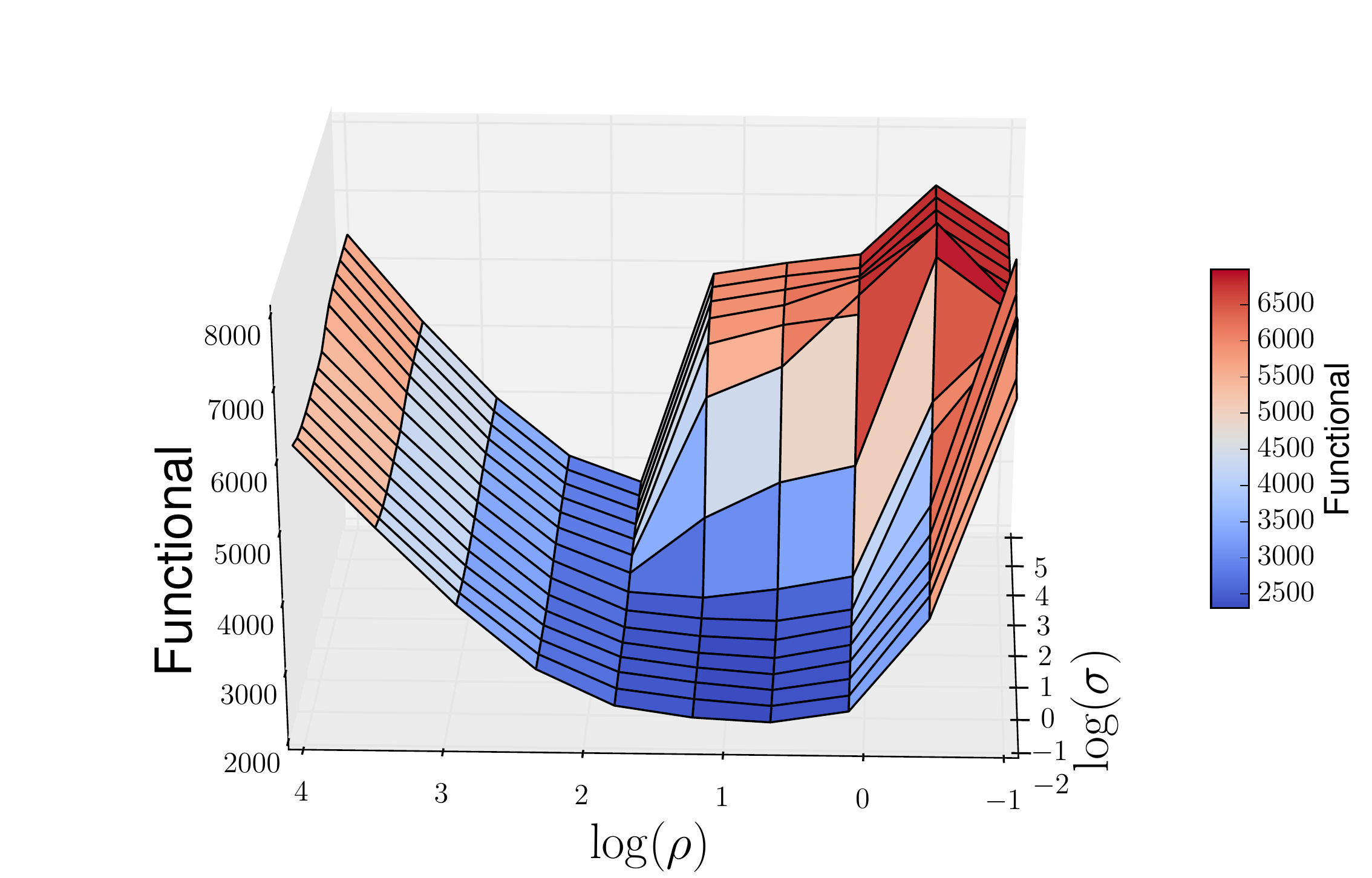}}}
\hfil
\subfigure[M-FISTA]{\scalebox{0.18}{\includegraphics{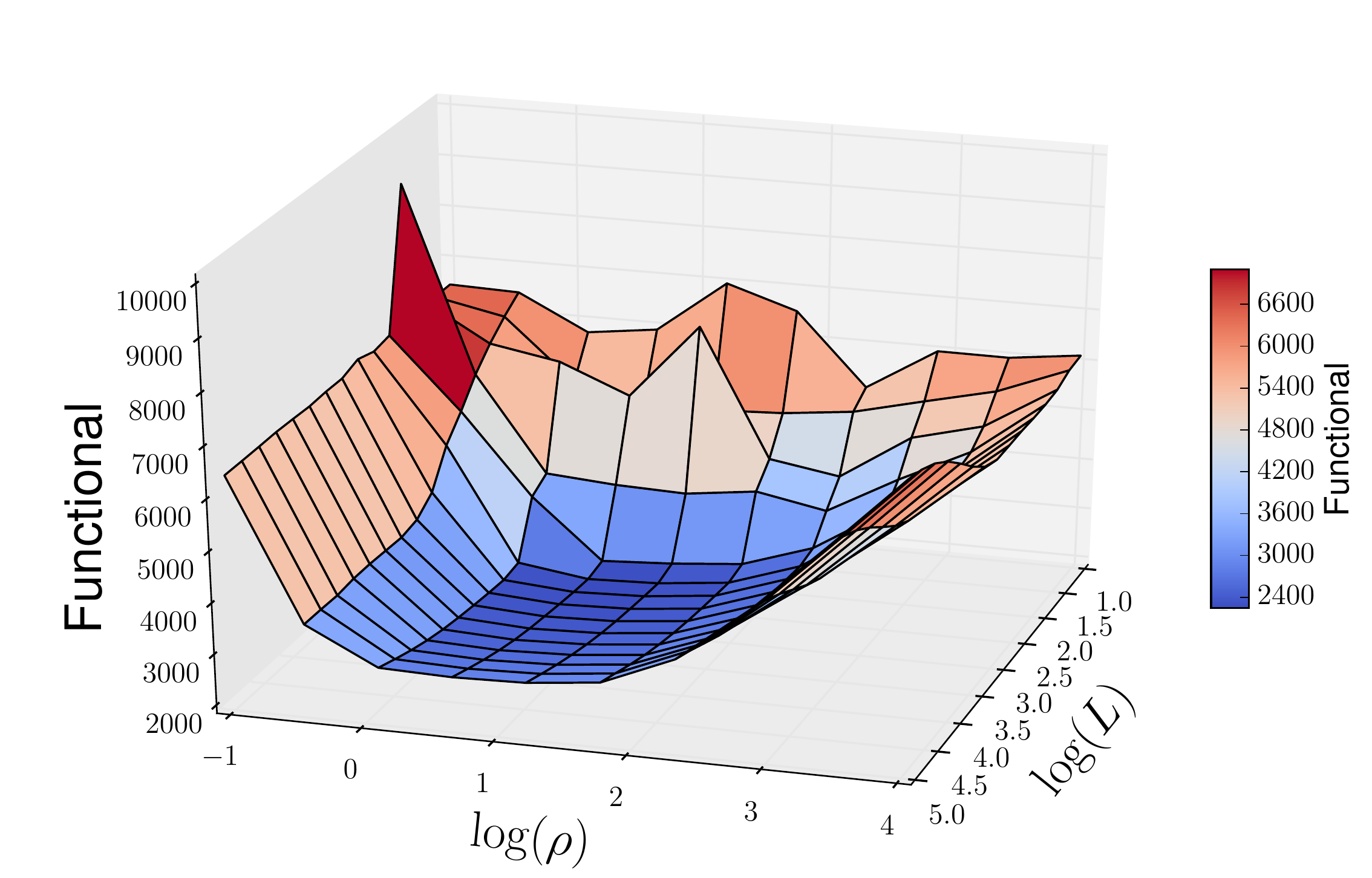}}}
}
        \caption{Grid search surfaces for masked conjugate gradient (M-CG), masked iterated Sherman-Morrison (M-ISM), masked consensus (M-Cns) and masked FISTA (M-FISTA) algorithms with $K=20$. Each surface represents the value of the masked CBPDN functional (\eq{cbpdnmsk} in the main document) after 100 iterations, for different parameters $\rho$, and $\sigma$ or $L$.}
        \label{fig:grid_c}
\end{figure}

The penalty parameter grid searches discussed in~\sctn{optimal_parameters} in the main document generate 2D surfaces representing the CDL functional value after a fixed number of iterations, plotted against the parameters for the sparse coding and dictionary update components of the dictionary learning algorithm. The surfaces corresponding to the coarse grids for the set of 20 training images are shown here in~\figs{grid_a}~--~\fign{grid_c}.

\section{Analytic Derivation of Penalty Parameter Scaling}
\label{sec:parameters_analytic_sm}

In order to estimate the scaling properties of the algorithm parameters with respect to the training set size, $K$, we consider the case in which the training set size is changed by replication of the same data. By removing the complexities associated with the characteristics of individual images, this simplified scenario allows analytic evaluation of the conditions under which an equivalent problem is obtained when the set size, $K$, is changed. In practice, changing $K$ involves introducing different training images, and we cannot expect that these scaling properties will hold exactly, but they represent the best possible estimate that depends only on $K$ and not on the properties of the training images themselves.

The following properties of the Frobenius norm, $\ell_2$ norm, and $\ell_1$ norm play an important role in these derivations:
\begin{align}
  \norm{\left(\begin{array}{cc} \mb{x} & \mb{y} \end{array} \right)}_F^2 = &
  \norm{\mb{x}}_2^2 + \norm{\mb{y}}_2^2 \\
  \norm{\left(\begin{array}{c} X \\ Y \end{array} \right)}_F^2 = &
  \norm{X}_F^2 + \norm{Y}_F^2 \\
  \norm{\left(\begin{array}{cc} \mb{x} & \mb{y} \end{array} \right)}_1 = &
  \norm{\mb{x}}_1 + \norm{\mb{y}}_1 \\
  \norm{\left(\begin{array}{c} X \\ Y \end{array} \right)}_1 = &
  \norm{X}_1 + \norm{Y}_1 \;.
\end{align}
We will also make use of the invariance  of the indicator function under scalar multiplication
\begin{equation}
 \alpha \iota_C (\mb{x} ) = \iota_C (\mb{x} ) \;\;\; \forall \alpha > 0 \;,
\end{equation}
which is due to the $\{0, \infty\}$ range of this function.

\subsection{ADMM Sparse Coding}

The augmented Lagrangian for the ADMM solution to CSC problem~\eq{bpdnmmvsplit} in the main document is
\begin{multline}
  L_{\rho}(X, Y, U) =  \\ \frac{1}{2} \norm{D X - S}_F^2  + \lambda  \norm{Y}_1 +  \frac{\rho}{2} \norm{X - Y + U}_F^2 \;,
  \label{eq:cbpdn_parameters}
\end{multline}
where we omit the final term, $-\frac{\rho}{2} \norm{U}_F^2$, which does not effect the minimizer of this functional.
For $K = 1$ we have $S = \mb{s}$, $X = \mb{x}$, $Y = \mb{y}$, and $U = \mb{u}$. If we construct the $K = 2$ case by replicating the training data, we have $S' = \left(\begin{array}{cc} \mb{s} & \mb{s} \end{array} \right)$, $X' = \left(\begin{array}{cc} \mb{x} & \mb{x} \end{array} \right)$, $Y' = \left(\begin{array}{cc} \mb{y} & \mb{y} \end{array} \right)$, and $U' = \left(\begin{array}{cc} \mb{u} & \mb{u} \end{array} \right)$, and the augmented Lagrangian is
\begin{multline}
  L_{\rho}(X', Y', U') = \\ \frac{1}{2} \norm{D X' - S'}_F^2  + \lambda  \norm{Y'}_1 + \frac{\rho}{2} \norm{X' - Y' + U'}_F^2  \\
  =  2 \frac{1}{2} \norm{D \mb{x} - \mb{s}}_2^2  + 2 \lambda \norm{\mb{y}}_1 +  2 \frac{\rho}{2} \norm{\mb{x} - \mb{y} + \mb{u}}_2^2 \\
  =  2 L_{\rho}(X, Y, U) \;.
\end{multline}
For this problem, the augmented Lagrangian for the $K=2$ case is just twice the augmented Lagrangian for the $K=1$ case, with the same penalty parameter $\rho$. Therefore we expect that the optimal penalty parameter should remain constant when changing the number of training images $K$.

\subsection{Equality Constrained ADMM Dictionary Update}

The augmented Lagrangian for the ADMM solution to the dictionary update problem~\eq{ccmodeqsplit} in the main document is
\begin{align}
   L_{\sigma}(\mb{d}, \mb{g}, \mb{h}) = & \frac{1}{2}  \normsz[\big]{X \mb{d} -
    \mb{s}}_2^2 + \iota_{C_{\text{PN}}} (\mb{g}) +  \nonumber \\
  & \frac{\sigma}{2} \norm{\mb{d} - \mb{g} + \mb{h}}_2^2 \;,
\label{eq:ccmod_parameters}
\end{align}
where we omit the final term, $-\frac{\sigma}{2} \norm{\mb{h}}_2^2$, which does not effect the minimizer of this functional. We assume that the variables in the above equation represent the $K=1$ case, and construct the $K = 2$ case by replicating the training data, \ie
\[
X' = \left(\begin{array}{c} X \\ X \end{array} \right) \; ,  \;\;\; \mb{s}' = \left(\begin{array}{c} \mb{s} \\ \mb{s} \end{array} \right) \;, \;\;\; \mb{d}' = \mb{d} \;,
\]
$\mb{g}' = \mb{g}$, and $\mb{h}' = \mb{h}$. The corresponding augmented Lagrangian is
\begin{multline}
  L_{\sigma}(\mb{d}', \mb{g}', \mb{h}') = \\ \frac{1}{2}  \normsz[\Big]{X' \mb{d}' -
    \mb{s}'}_2^2 + \iota_{C_{\text{PN}}} (\mb{g}')
   + \frac{\sigma}{2} \norm{\mb{d}' - \mb{g}' + \mb{h}'}_2^2  \\
  =  2 \frac{1}{2}  \norm{ X \; \mb{d} - \mb{s}}_2^2 + \iota_{C_{\text{PN}}} (\mb{g})
   + \frac{\sigma}{2} \norm{\mb{d} - \mb{g} + \mb{h}}_2^2 \\
  = 2 L_{2 \sigma}(\mb{d}, \mb{g}, \mb{h}) \;.
\end{multline}
For this problem, the augmented Lagrangian for the $K=2$ case is twice
the augmented Lagrangian for the $K=1$ case when the penalty parameter
is also twice the penalty parameter used for the $K=1$ case.
Therefore we expect that the optimal penalty parameter should scale
linearly when changing the number of training images $K$.

\subsection{Consensus ADMM Dictionary Update}

The augmented Lagrangian for the ADMM Consensus form of the dictionary update problem~\eq{dupprobcns} in the main document is
\begin{align}
   L_{\sigma}(\mb{d}, \mb{g}, \mb{h}) = & \frac{1}{2}  \normsz[\big]{X \mb{d} -
    \mb{s}}_2^2 + \iota_{C_{\text{PN}}} (\mb{g}) + \nonumber \\ &
   \frac{\sigma}{2} \norm{\mb{d} - E \mb{g} + \mb{h}}_2^2  \;,
\label{eq:ccons_parameters}
\end{align}
where we omit the final term, $-\frac{\sigma}{2} \norm{\mb{h}}_2^2$, which does not effect the minimizer of this functional, and
\begin{equation}
\addtolength{\arraycolsep}{-0.6mm}
E = \left( \begin{array}{c}  I \\ I\\
    \vdots  \end{array} \right) \;.
\end{equation}
We assume that the variables in the above equation represent the $K=1$ case, with $E = I$, and construct the $K = 2$ case by replicating the training data, \ie
\[
\addtolength{\arraycolsep}{-1mm}
X' = \left(\begin{array}{cc} X & 0 \\ 0 & X \end{array} \right) \,, \; \mb{s}' = \left(\begin{array}{c} \mb{s} \\ \mb{s} \end{array} \right) \,, \;
\mb{d}' = \left(\begin{array}{c} \mb{d} \\ \mb{d} \end{array} \right) \,, \;
\mb{h}' = \left(\begin{array}{c} \mb{h} \\ \mb{h} \end{array} \right) \,,
\]
$\mb{g}' = \mb{g}$, and $E' = \left(\begin{array}{cc} I & I \end{array} \right)^T$. The corresponding augmented Lagrangian is
\begin{multline}
  L_{\sigma}(\mb{d}', \mb{g}', \mb{h}') = \\
    \frac{1}{2}  \normsz[\big]{X' \mb{d}' -
    \mb{s}'}_2^2 + \iota_{C_{\text{PN}}} (\mb{g}')
   + \frac{\sigma}{2} \norm{\mb{d}' - E' \mb{g}' + \mb{h}'}_2^2 \\
  =  2 \frac{1}{2} \norm{ X \; \mb{d} - \mb{s}}_2^2 + \iota_{C_{\text{PN}}} (\mb{g})
   + 2 \frac{\sigma}{2} \norm{\mb{d} - E \mb{g} + \mb{h}}_2^2 \\
 = 2 L_{\sigma}(\mb{d}, \mb{g}, \mb{h}) \;.
\end{multline}
For this problem, the augmented Lagrangian for the $K=2$ case is just twice the augmented Lagrangian for the $K=1$ case, with the same penalty parameter $\sigma$. Therefore we expect that the optimal penalty parameter should remain constant when changing the number of training images $K$.

\subsection{FISTA Dictionary Update}

The FISTA solution to the dictionary update problem requires computing the gradient of the data fidelity term in the DFT domain (\eq{gradf} in the main document)
\begin{equation}
 \nabla_{\hat{\mb{d}}} \Big(\frac{1}{2} \normsz[\big]{\hat{X} \hat{\mb{d}} - \hat{\mb{s}}}_2^2 \Big) = \hat{X}^H \big(\hat{X} \hat{\mb{d}} - \hat{\mb{s}} \big) \;.
\end{equation}
We assume that the variables in the above equation represent the $K=1$ case, and construct the $K = 2$ case by replicating the training data, \ie
\[
\hat{X}' = \left(\begin{array}{c} \hat{X} \\ \hat{X} \end{array} \right)
\;, \;\; \hat{\mb{s}}' = \left(\begin{array}{c} \hat{\mb{s}} \\ \hat{\mb{s}} \end{array} \right) \;, \;\; \hat{\mb{d}}' = \hat{\mb{d}} \;,
\]
and the gradient in the DFT domain is
\begin{align}
 \nabla_{\hat{\mb{d}}'} \Big(\frac{1}{2} \normsz[\big]{\hat{X}' \hat{\mb{d}}' - \hat{\mb{s}}'}_2^2 \Big) = & \ \hat{X}'^H \big(\hat{X}' \hat{\mb{d}}' - \hat{\mb{s}}' \big) \nonumber \\
 = & \ 2 \hat{X}^H \big(\hat{X} \hat{\mb{d}} - \hat{\mb{s}} \big) \;.
\end{align}
For this problem, the gradient in the DFT domain for the $K=2$ case is just twice the gradient in the DFT domain for the $K=1$ case. To obtain the same solution we need the gradient step to be the same, which requires that the gradient step parameter be reduced by a factor of two to compensate for the doubling of the gradient. Therefore we expect that the optimal parameter $L$, which is the inverse of the gradient step size, should scale linearly when changing the number of training images $K$.

\subsection{Mask Decoupling ADMM Sparse Coding}

The augmented Lagrangian for the ADMM solution to the masked form of the MMV CBPDN problem~\eq{mdcsc} in the main document is
\begin{multline}
  L_{\rho}(X, Y_0, Y_1, U_0, U_1) =
   \frac{1}{2} \norm{W Y_1}_F^2  + \lambda  \norm{Y_0}_1 + \\
   \frac{\rho}{2} \norm{\left( \begin{array}{c}  \!Y_0\!\\ \!Y_1\! \\
     \end{array} \right) - \left [ \left( \begin{array}{c}  \!I\! \\ \!D\! \\
     \end{array} \right) X -  \left( \begin{array}{c}  0 \\ \!S\! \\
     \end{array} \right) \right ] + \left( \begin{array}{c}  \!U_0\!\\ \!U_1\! \\
     \end{array} \right)}_F^2 \:,
  \label{eq:mdcsc_parameters}
\end{multline}
where we omit the final term
\[
- \frac{\rho}{2} \norm{ \left( \begin{array}{c}  \!U_0\!\\ \!U_1\! \\  \end{array} \right)}_F^2 \;,
\]
which does not effect the minimizer of this functional. We assume that the variables in the above equation represent the $K=1$ case, and construct the $K = 2$ case by replicating the training data, \ie
$S' = \left(\begin{array}{cc} \mb{s} & \mb{s} \end{array} \right)$, $X' = \left(\begin{array}{cc} \mb{x} & \mb{x} \end{array} \right)$, $Y_{0}' = \left(\begin{array}{cc} Y_0 & Y_0 \end{array} \right)$, $Y_{1}' = \left(\begin{array}{cc} Y_1 & Y_1 \end{array} \right)$, $U_{0}' = \left(\begin{array}{cc} U_0 & U_0 \end{array} \right)$, $U_{1}' = \left(\begin{array}{cc} U_1 & U_1 \end{array} \right)$, and $0' = \left(\begin{array}{cc} \mb{0} & \mb{0} \end{array} \right)$.
The corresponding augmented Lagrangian is
\begin{multline}
   L_{\rho}(X', Y_{0}', Y_{1}', U_{0}', U_{1}') =
  \frac{1}{2} \norm{W Y_{1}'}_F^2  + \lambda  \norm{Y_{0}'}_1 +  \\
  \frac{\rho}{2} \norm{\left( \begin{array}{c}  Y_{0}' \\ Y_{1}' \\
     \end{array} \right) - \left [ \left( \begin{array}{c}  I \\ D \\
     \end{array} \right) X' - \left( \begin{array}{c}  0'  \\ S' \\
     \end{array} \right) \right ] + \left( \begin{array}{c}  U_{0}' \\ U_{1}' \\
     \end{array} \right)}_F^2  \\
   = 2 \frac{1}{2} \norm{W Y_1}_2^2  + 2 \lambda  \norm{Y_0}_1 +  \\
  2  \frac{\rho}{2} \norm{\left( \begin{array}{c}  Y_0 \\ Y_1 \\
     \end{array} \right) - \left [ \left( \begin{array}{c}  I \\ D \\
     \end{array} \right) X -  \left( \begin{array}{c}  \mb{0} \\ \mb{s} \\
     \end{array} \right) \right ] + \left( \begin{array}{c}  U_0\\ U_1 \\
     \end{array} \right)}_2^2 \nonumber \\
   = 2 L_{\rho}(X, Y_0, Y_1, U_0, U_1) \;.
\end{multline}
For this problem, the augmented Lagrangian for the $K=2$ case is just twice the augmented Lagrangian for the $K=1$ case, with the same penalty parameter $\rho$. Therefore we expect that the optimal penalty parameter should remain constant when changing the number of training images $K$.

\subsection{Mask Decoupling ADMM Dictionary Update}

The augmented Lagrangian for the Block-Constraint ADMM solution of the masked dictionary update problem~\eq{ccmodmsk_block} in the main document is
\begin{multline}
    L_{\sigma}(\mb{d}, \mb{g}_0, \mb{g}_1, \mb{h}_0, \mb{h}_1) =
    \frac{1}{2} \norm{W \mb{g}_1}_2^2  + \iota_{C_{\text{PN}}} (\mb{g}_0) +  \\
   \frac{\sigma}{2} \norm{\left( \begin{array}{c}  \!\mb{g}_0\!\\ \!\mb{g}_1\!\\
     \end{array} \right) - \left [ \left( \begin{array}{c}   \!I\! \\ \!X\! \\
     \end{array} \right) \mb{d} -  \left( \begin{array}{c}  \!\mb{0}\! \\ \!\mb{s}\! \\
     \end{array} \right) \right ] + \left( \begin{array}{c}  \!\mb{h}_0\!\\ \!\mb{h}_1\!\\
     \end{array} \right)}_2^2 \;,
  \label{eq:mddu_block_parameters}
\end{multline}
where we omit the final term
\[
- \frac{\sigma}{2} \norm{ \left( \begin{array}{c}  \!\mb{h}_0\! \\ \!\mb{h}_1\! \\
     \end{array} \right)}_2^2 \;,
\]
which does not effect the minimizer of this functional. We assume that the variables in the above equation represent the $K=1$ case, and construct the $K = 2$ case by replicating the training data, \ie
\[
\addtolength{\arraycolsep}{-1.1mm}
X' = \left(\begin{array}{c} X \\ X \end{array} \right) \,,\;
\mb{s}' = \left(\begin{array}{c} \mb{s} \\ \mb{s} \end{array} \right) \,,\;
\mb{g}_{1}' = \left(\begin{array}{c} \mb{g}_1 \\ \mb{g}_1 \end{array} \right) \,,\;
\mb{h}_{1}' = \left(\begin{array}{c} \mb{h}_1 \\ \mb{h}_1 \end{array} \right) \;,
\]
$\mb{d}' = \mb{d}$, $\mb{g}_{0}' = \mb{g}_0$, and $\mb{h}_{0}' = \mb{h}_0$.
The corresponding augmented Lagrangian is
\begin{multline}
   L_{\sigma}(\mb{d}', \mb{g}_{0}', \mb{g}_{1}', \mb{h}_{0}', \mb{h}_{1}') =
    \frac{1}{2} \norm{W \mb{g}_{1}'}_2^2  + \iota_{C_{\text{PN}}} (\mb{g}_{0}')  +  \\
    \qquad \frac{\sigma}{2} \norm{\left( \begin{array}{c}  \mb{g}_{0}' \\ \mb{g}_{1}' \\
     \end{array} \right) - \left [ \left( \begin{array}{c}   I \\ X' \\
     \end{array} \right) \mb{d}' - \left( \begin{array}{c}   \mb{0} \\ \mb{s}' \\
     \end{array} \right) \right ] + \left( \begin{array}{c}  \mb{h}_{0}' \\ \mb{h}_{1}' \\
     \end{array} \right)}_2^2  \\
    = 2 \frac{1}{2} \norm{W \mb{g}_1}_2^2  + \iota_{C_{\text{PN}}} (\mb{g}_0) + \frac{\sigma}{2} \norm{\mb{g}_0 - \mb{d} + \mb{h}_0}_2^2 \; + \\
    2 \frac{\sigma}{2} \norm{\mb{g}_1 - \left ( X \mb{d} - \mb{s} \right)+ \mb{h}_1}_2^2 \;.
\end{multline}
For this problem, the augmented Lagrangian for the $K=2$ case has terms that are twice the augmented Lagrangian for the $K=1$ case, as well as a term that is the same as for the $K=1$ case. Therefore, there is no simple rule to scale the optimal penalty parameter $\sigma$ when changing the number of training images $K$.

It is, however, worth noting that a scaling relationship could be obtained by replacing the constraint $\mb{g}'_0 = \mb{d}'$ with the equivalent constraint $2\mb{g}_0 = 2\mb{d}$ (or, more generally, $K\mb{g}_0 = K\mb{d}$) and appropriate rescaling of the scaled dual variable $\mb{h}_0$, so that the problematic term above, $(\sigma/2)\norm{\mb{g}'_0 - \mb{d}' + \mb{h}'_0}_2^2$, exhibits the same scaling as the other terms.

\subsection{Hybrid Consensus Masked Dictionary Update}
\label{sec:mcnskscale}

The augmented Lagrangian for the ADMM consensus solution of the masked dictionary update problem~\eq{ccmodmsk_cns} in the main document is
\begin{multline}
   L_{\sigma}(\mb{d}, \mb{g}_0, \mb{g}_1, \mb{h}_0, \mb{h}_1) =  \frac{1}{2} \norm{W \mb{g}_1}_2^2  + \iota_{C_{\text{PN}}} (\mb{g}_0) +  \\
    \frac{\sigma}{2} \left \lVert \left( \begin{array}{c}   \!\!I\!\! \\ \!\!X\!\! \\
     \end{array} \right) \mb{d} - \left(\begin{array}{cc} \!\!E\!\! & \!\!0\!\! \\ \!\!0\!\! & \!\!I\!\!
     \end{array} \right) \left( \begin{array}{c}  \!\!\mb{g}_0\!\!\\ \!\!\mb{g}_1\!\!\\
     \end{array} \right) - \left( \begin{array}{c}  \!\!\mb{0}\!\! \\ \!\!\mb{s}\!\! \\
     \end{array} \right) + \left( \begin{array}{c}  \!\!\mb{h}_0\!\!\\ \!\!\mb{h}_1\!\!\\
     \end{array} \right) \right \rVert_2^2 \;,
  \label{eq:mddu_cns_parameters}
\end{multline}
where we omit the final term
\[
- \frac{\sigma}{2} \norm{ \left( \begin{array}{c}  \mb{h}_0 \\ \mb{h}_1 \\ \end{array} \right)}_2^2 \;,
\]
which does not effect the minimizer of this functional. We assume that the variables in the above equation represent the $K=1$ case, with $E = I$, and construct the $K = 2$ case by replicating the training data, \ie
\[
X' = \left(\begin{array}{cc} X & 0 \\ 0 & X \end{array} \right) \;,\;\;
\mb{s}' = \left(\begin{array}{c} \mb{s} \\ \mb{s} \end{array} \right) \;,\;\;
\mb{d}' =  \left(\begin{array}{c} \mb{d} \\ \mb{d} \end{array} \right) \;,
\]
\[
\mb{g}_{1}' = \left(\begin{array}{c} \mb{g}_1 \\ \mb{g}_1 \end{array} \right) \;,\;\;
\mb{h}_{0}' = \left(\begin{array}{c} \mb{h}_0 \\ \mb{h}_0 \end{array} \right) \;,\;\;
\mb{h}_{1}' = \left(\begin{array}{c} \mb{h}_1 \\ \mb{h}_1 \end{array} \right) \;,
\]
$\mb{g}_{0}' = \mb{g}_0$, and $E' = \left(\begin{array}{cc} I & I \end{array} \right)^T$.
The corresponding augmented Lagrangian is
\begin{multline}
   L_{\sigma}(\mb{d}', \mb{g}_{0}', \mb{g}_{1}', \mb{h}_{0}', \mb{h}_{1}') =
   \frac{1}{2} \norm{W \mb{g}_{1}'}_2^2   + \iota_{C_{\text{PN}}} (\mb{g}_{0}')  +   \\
    \frac{\sigma}{2} \left \lVert \left( \begin{array}{c}   \!\!I\!\! \\ \!\!X'\!\! \\
     \end{array} \right) \mb{d}' - \left(\begin{array}{cc} \!\!E'\!\! & \!\!0\!\! \\ \!\!0\!\! & \!\!I\!\! \end{array} \right) \left( \begin{array}{c}  \!\!\mb{g}_{0}'\!\!\\ \!\!\mb{g}_{1}'\!\!\\
     \end{array} \right) - \left( \begin{array}{c}  \!\!0\!\! \\ \!\!\mb{s}'\!\! \\
     \end{array} \right) + \left( \begin{array}{c}  \!\!\mb{h}_{0}'\!\!\\ \!\!\mb{h}_{1}'\!\! \\
     \end{array} \right) \right \rVert_2^2 , \nonumber \\
     = 2 \frac{1}{2} \norm{W \mb{g}_1}_2^2  + \iota_{C_{\text{PN}}} (\mb{g}_0) \,+  \\
   2 \frac{\sigma}{2} \left \lVert \left( \begin{array}{c}   \!\!I\!\! \\ \!\!X\!\! \\
     \end{array} \right) \mb{d} - \left(\begin{array}{cc} \!\!E\!\! & \!\!0\!\! \\ \!\!0\!\! & \!\!I\!\! \end{array} \right) \left( \begin{array}{c}  \!\!\mb{g}_{0}\!\!\\ \!\!\mb{g}_{1}\!\!\\
     \end{array} \right)  - \left( \begin{array}{c}  \!\!\mb{0}\!\! \\ \!\!\mb{s}\!\! \\
     \end{array} \right) + \left( \begin{array}{c}  \!\!\mb{h}_{0}\!\! \\ \!\!\mb{h}_{1}\!\! \\
     \end{array} \right) \right \rVert_2^2 \\
= 2 L_{\sigma}(\mb{d}, \mb{g}_{0}, \mb{g}_{1}, \mb{h}_{0}, \mb{h}_{1}) \;.
\end{multline}
For this problem, the augmented Lagrangian for the $K=2$ case is just
twice the augmented Lagrangian for the $K=1$ case, with the same
penalty parameter $\sigma$.  Therefore we expect that the optimal
penalty parameter should remain constant when changing the number of
training images $K$.

\section{Experimental Sensitivity Analysis}
\label{sec:sensitivity_sm}

\begin{figure}[htb]
\centerline{
        \subfigure[CBPDN$(\rho)$ for best $\sigma$]{\scalebox{0.18}{\includegraphics{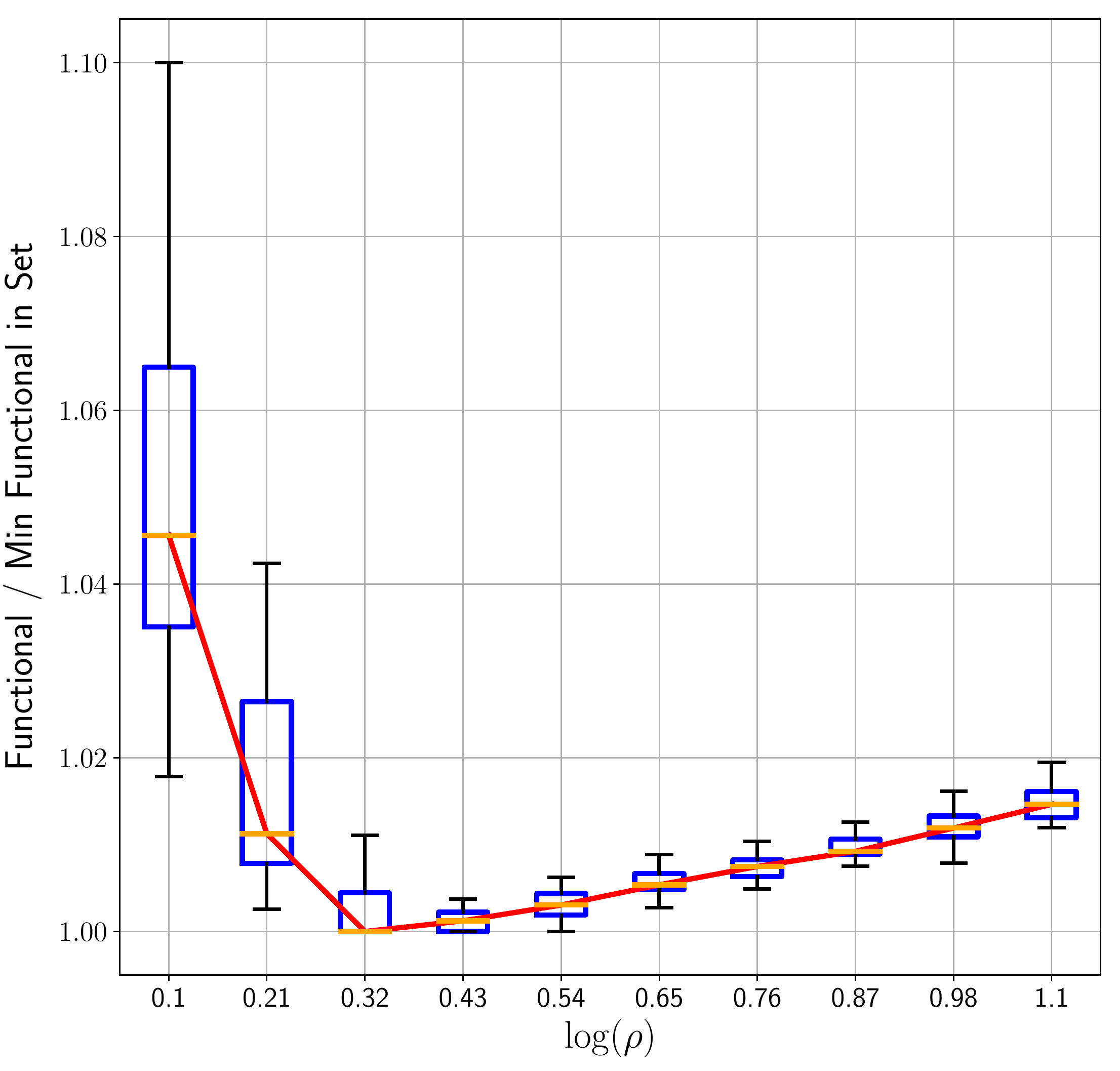}}}
        \hfil
        \subfigure[CBPDN$(\sigma)$ for best $\rho$]{\scalebox{0.18}{\includegraphics{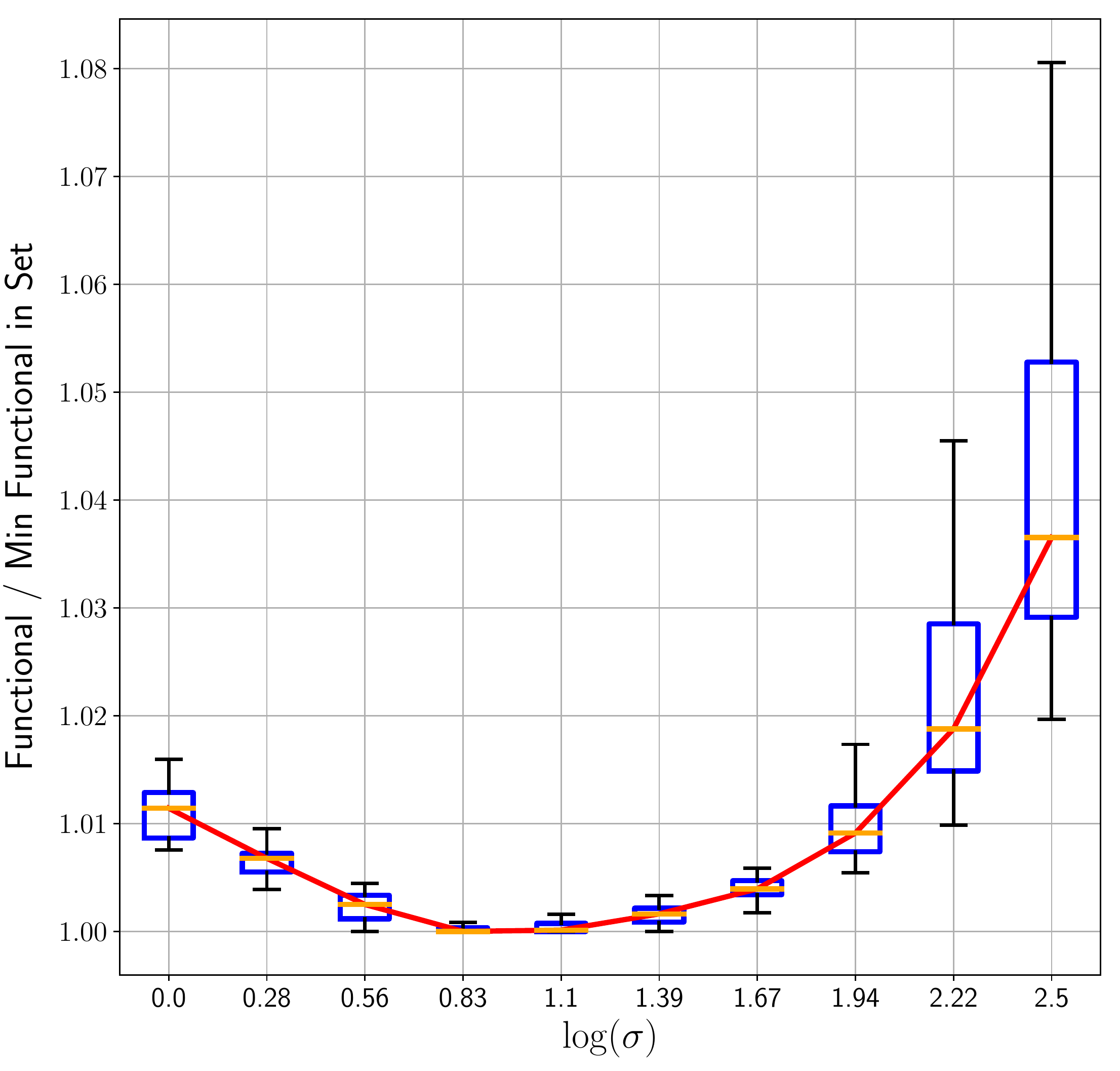}}}
}
        \caption{Distribution of normalized CBPDN functional (\eq{cbpdnmmv} in the main document) after 500 iterations, in the conjugate gradient (CG) grid search for 20 random selected sets of $K=5$ images.}
        \label{fig:cg_par_selection_K5}
\end{figure}

\begin{figure}[htb]
\centerline{
        \subfigure[CBPDN$(\rho)$ for best $\sigma$]{\scalebox{0.18}{\includegraphics{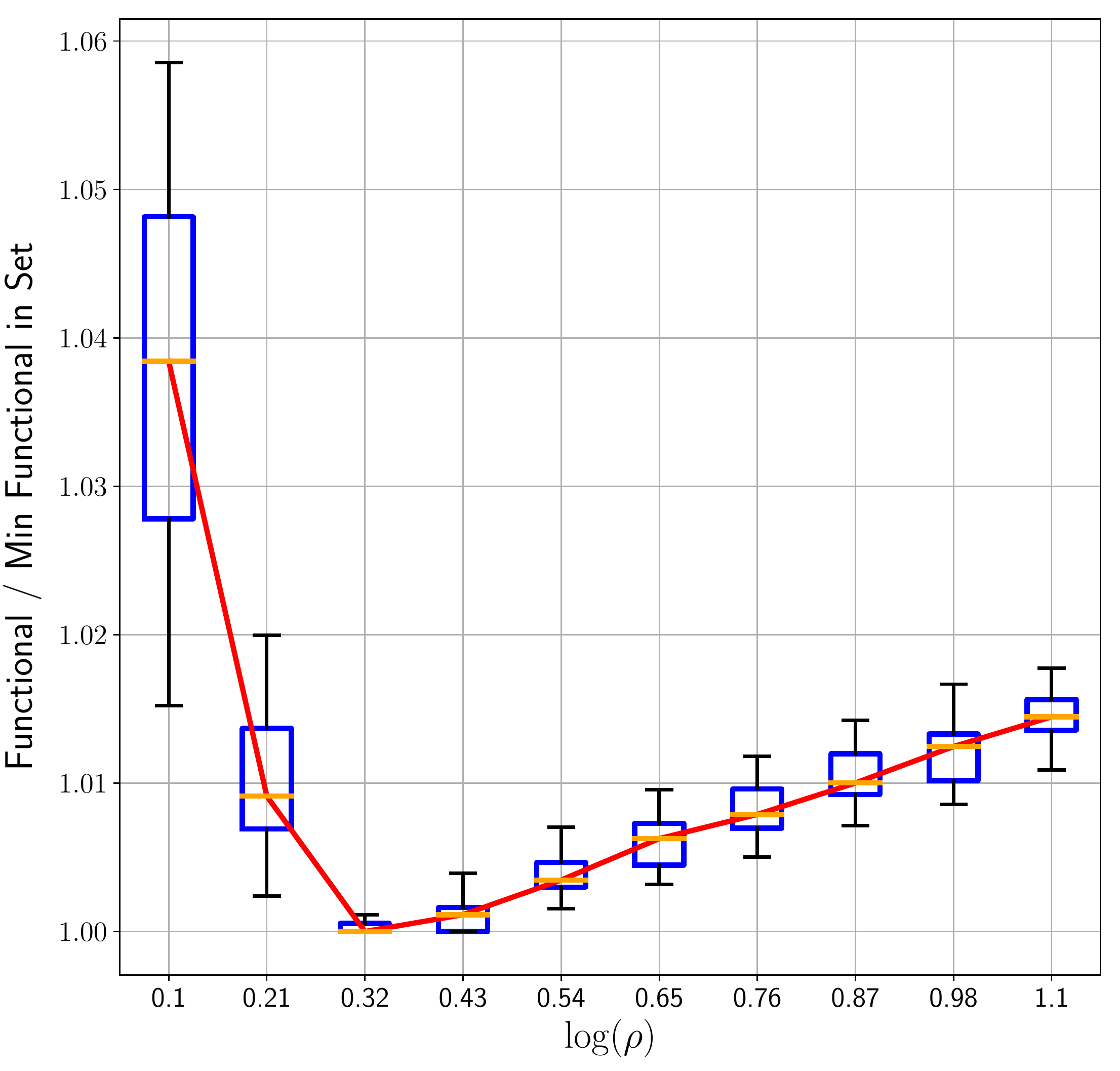}}}
        \hfil
        \subfigure[CBPDN$(\sigma)$ for best $\rho$]{\scalebox{0.18}{\includegraphics{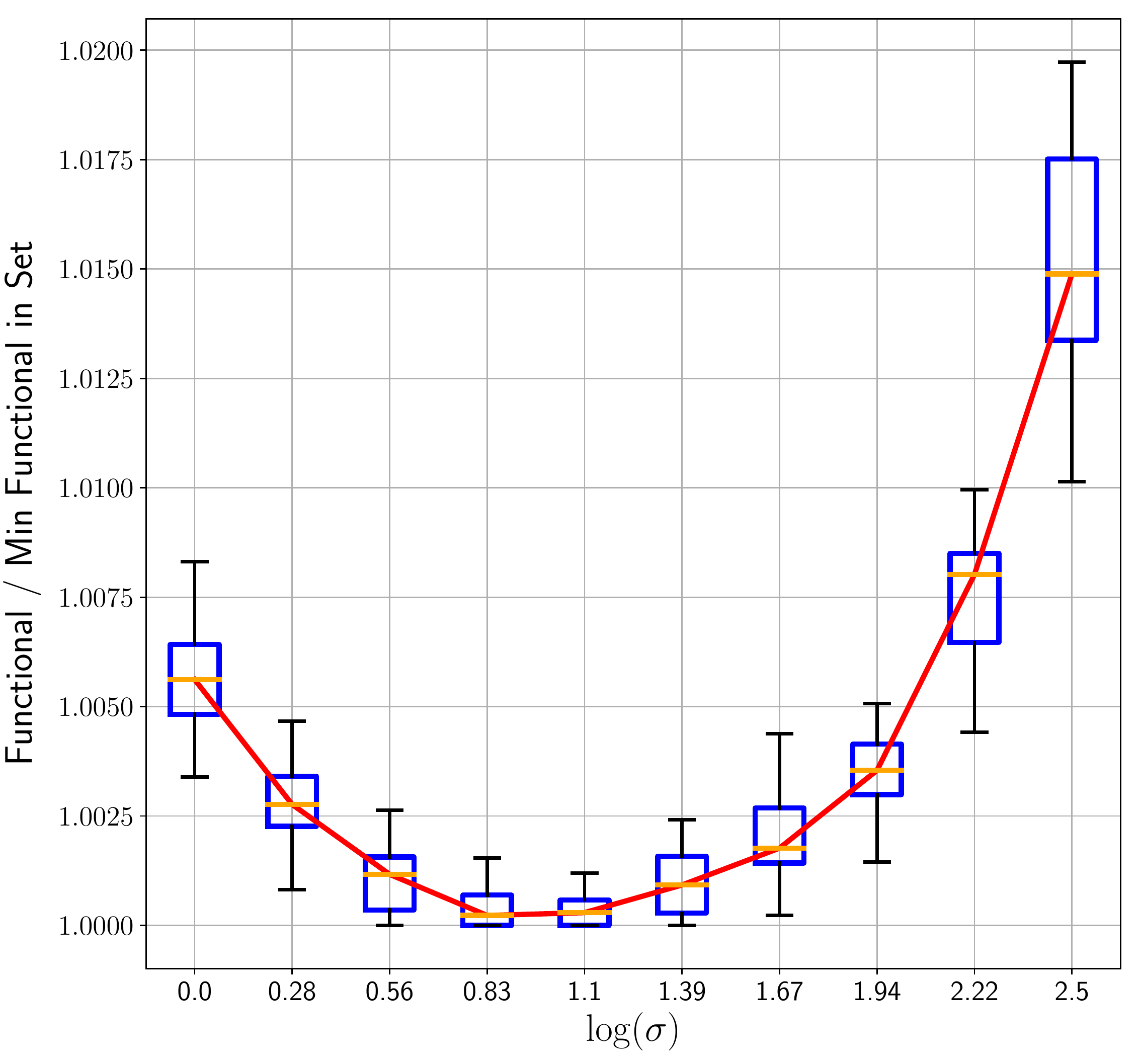}}}
}
        \caption{Distribution of normalized CBPDN functional (\eq{cbpdnmmv} in the main document) after 500 iterations, in the conjugate gradient (CG) grid search for 20 random selected sets of $K=10$ images.}
        \label{fig:cg_par_selection_K10}
\end{figure}

\begin{figure}[htb]
\centerline{
        \subfigure[CBPDN$(\rho)$ for best $\sigma$]{\scalebox{0.18}{\includegraphics{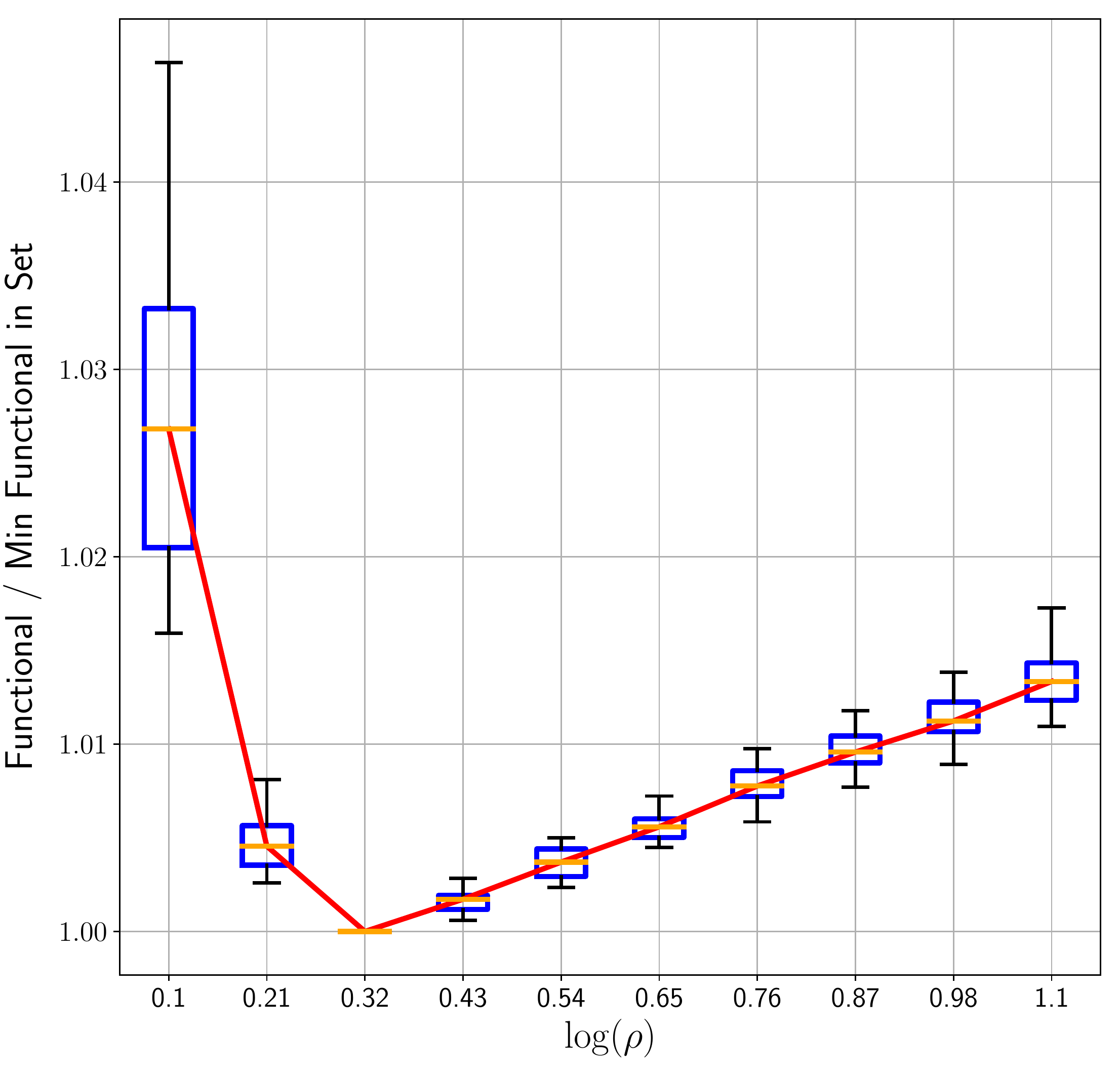}}}
        \hfil
        \subfigure[CBPDN$(\sigma)$ for best $\rho$]{\scalebox{0.18}{\includegraphics{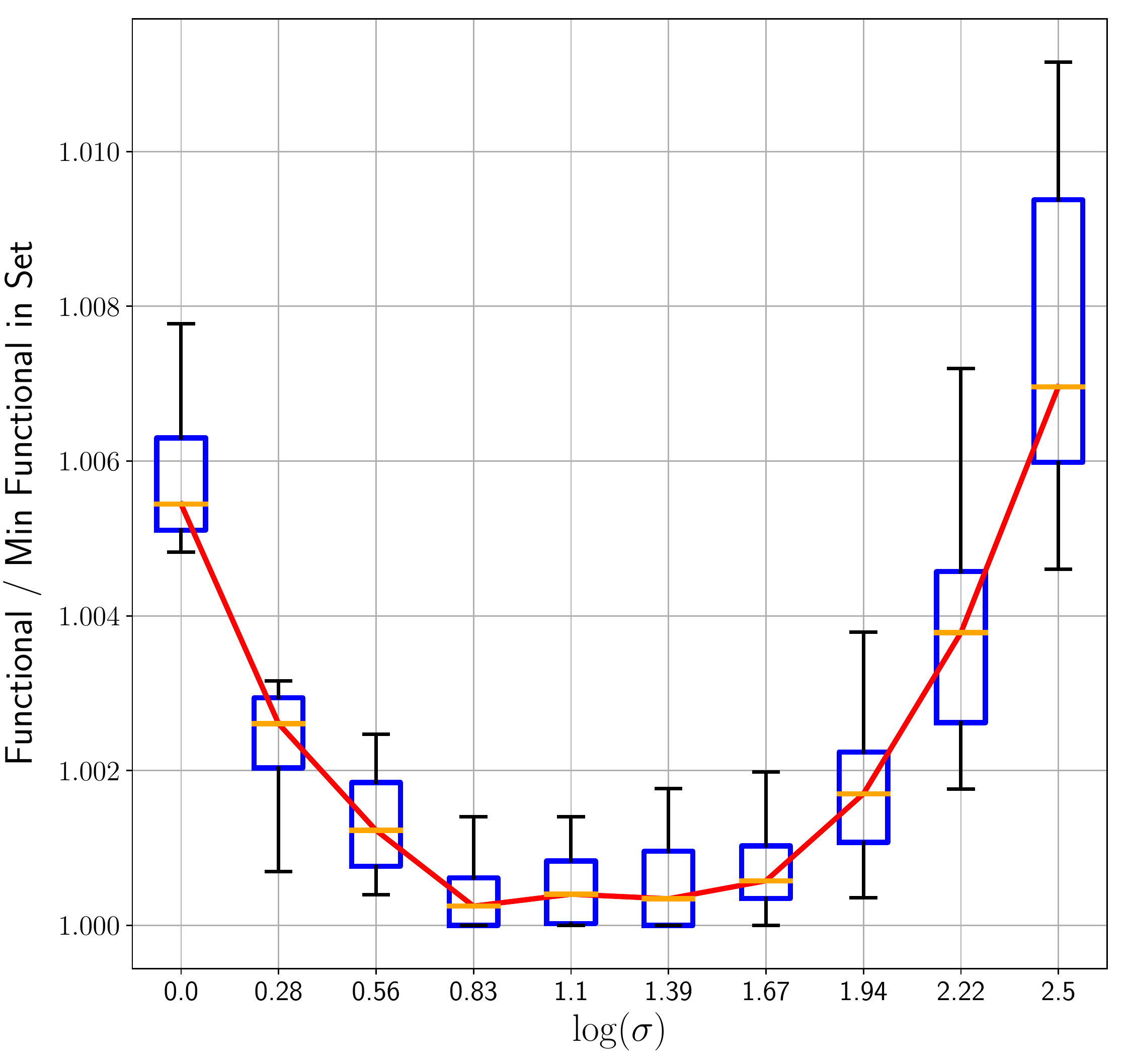}}}
}
        \caption{Distribution of normalized CBPDN functional (\eq{cbpdnmmv} in the main document) after 500 iterations, in the conjugate gradient (CG) grid search for 20 random selected sets of $K=20$ images.}
        \label{fig:cg_par_selection_K20}
\end{figure}

\begin{figure}[htb]
\centerline{
        \subfigure[CBPDN$(\rho)$ for best $\sigma$]{\scalebox{0.18}{\includegraphics{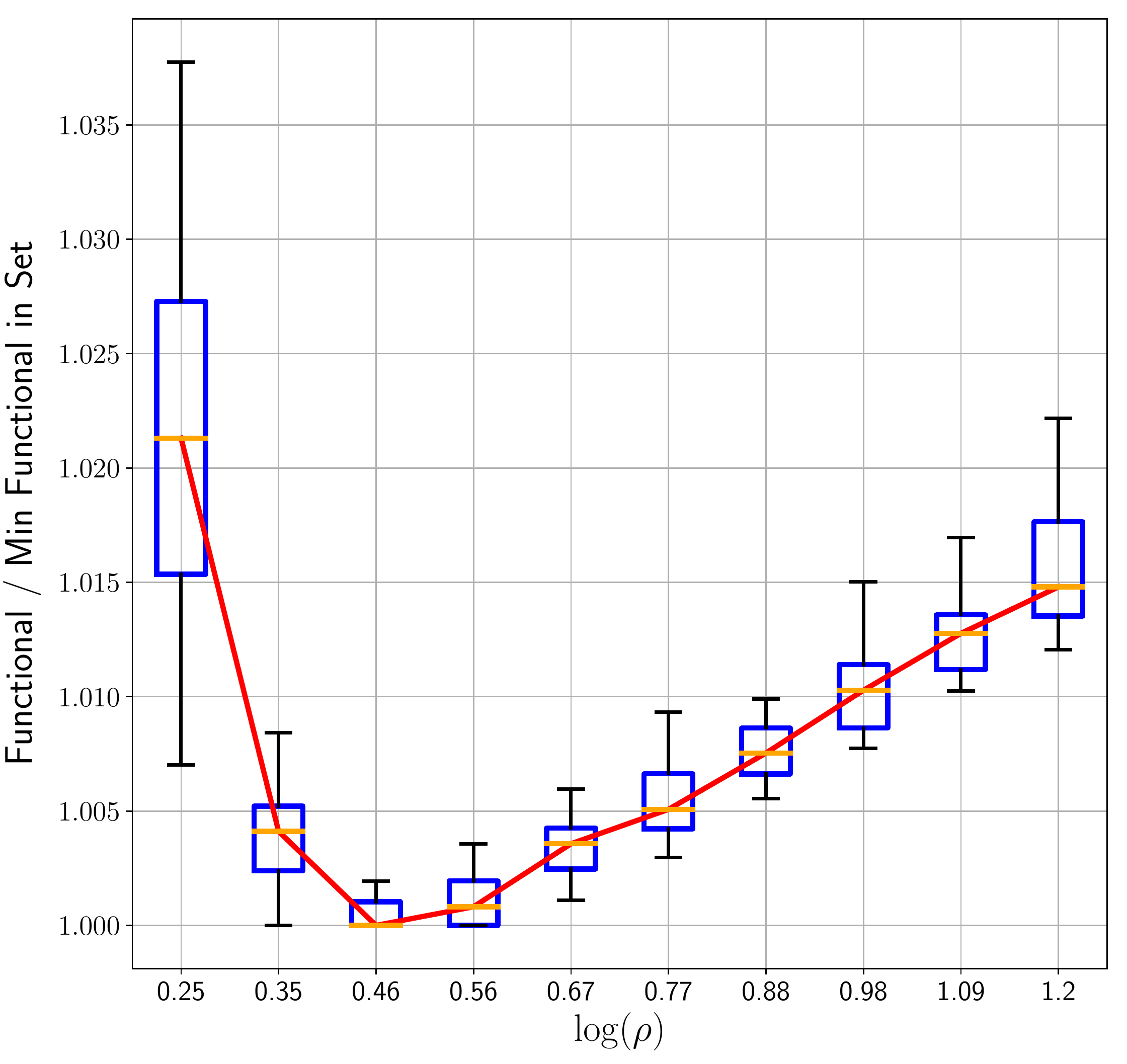}}}
        \hfil
        \subfigure[CBPDN$(\sigma)$ for best $\rho$]{\scalebox{0.18}{\includegraphics{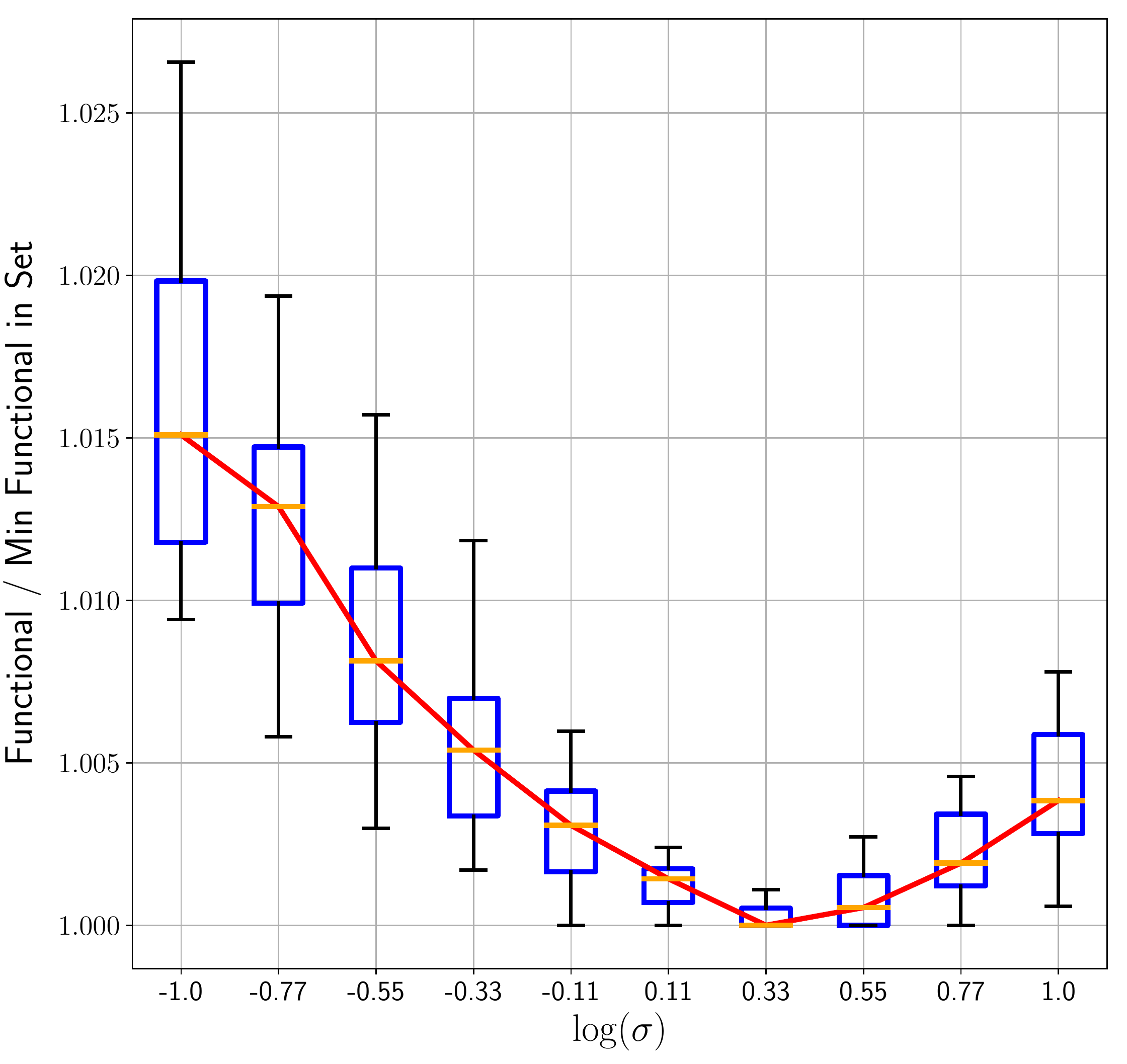}}}
}
        \caption{Distribution of normalized CBPDN functional (\eq{cbpdnmmv} in the main document) after 500 iterations, in the consensus (Cns / Cns-P) grid search for 20 random selected sets of $K=5$ images.}
        \label{fig:cns_par_selection_K5}
\end{figure}

\begin{figure}[htb]
\centerline{
        \subfigure[CBPDN$(\rho)$ for best $\sigma$]{\scalebox{0.18}{\includegraphics{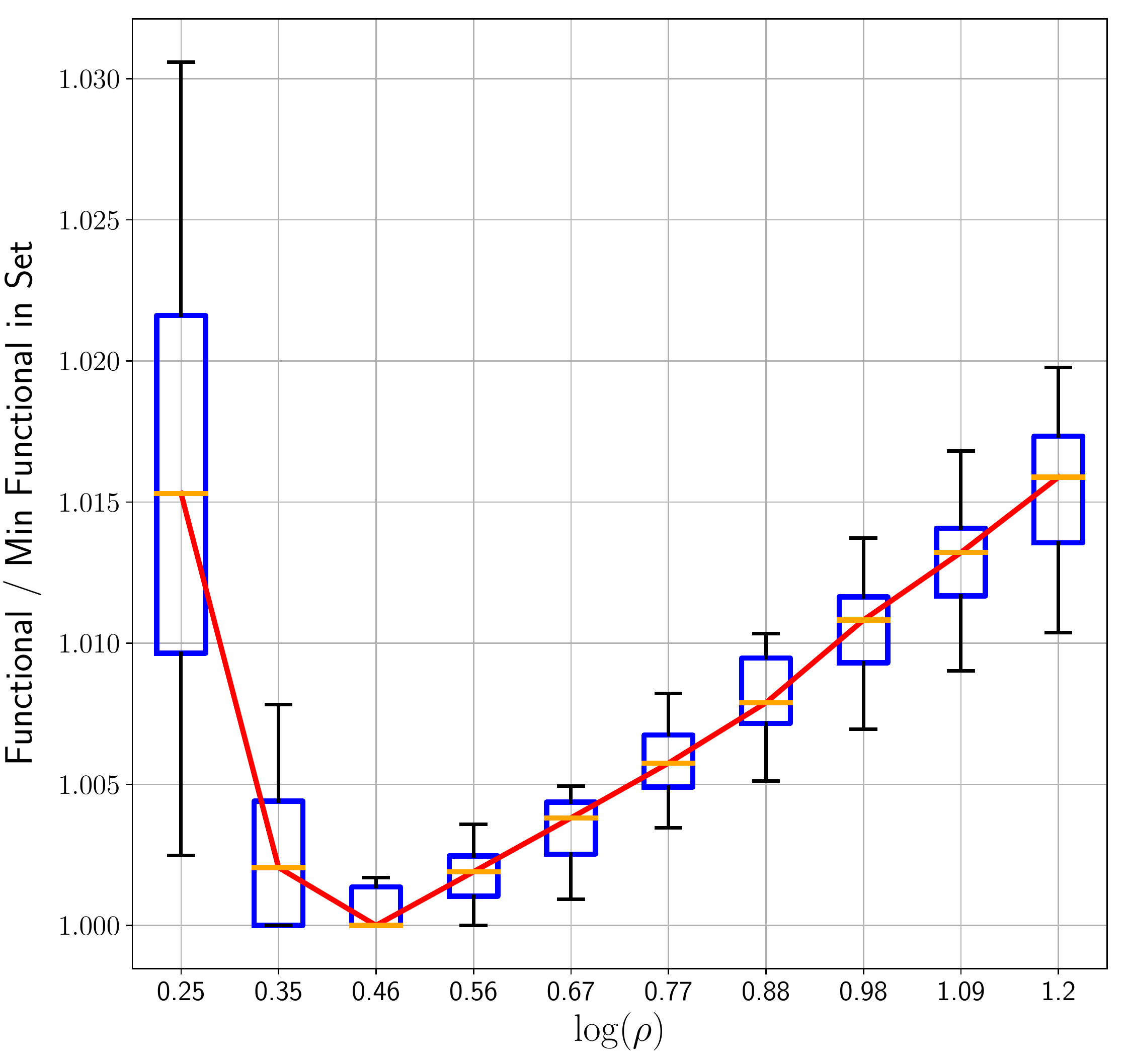}}}
        \hfil
        \subfigure[CBPDN$(\sigma)$ for best $\rho$]{\scalebox{0.18}{\includegraphics{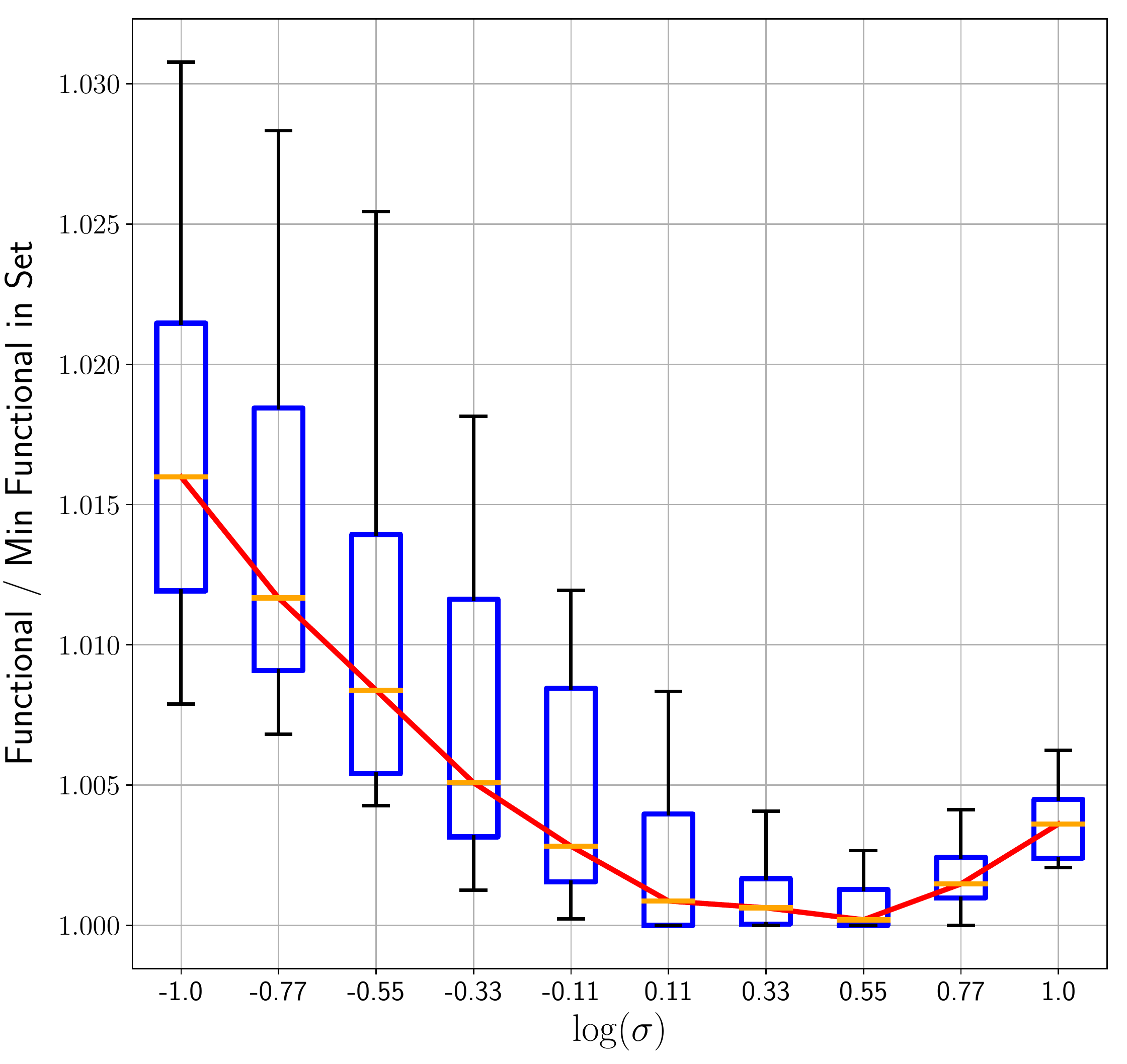}}}
}
        \caption{Distribution of normalized CBPDN functional (\eq{cbpdnmmv} in the main document) after 500 iterations, in the consensus (Cns / Cns-P) grid search for 20 random selected sets of $K=10$ images.}
        \label{fig:cns_par_selection_K10}
\end{figure}

\begin{figure}[htb]
\centerline{
        \subfigure[CBPDN$(\rho)$ for best $\sigma$]{\scalebox{0.18}{\includegraphics{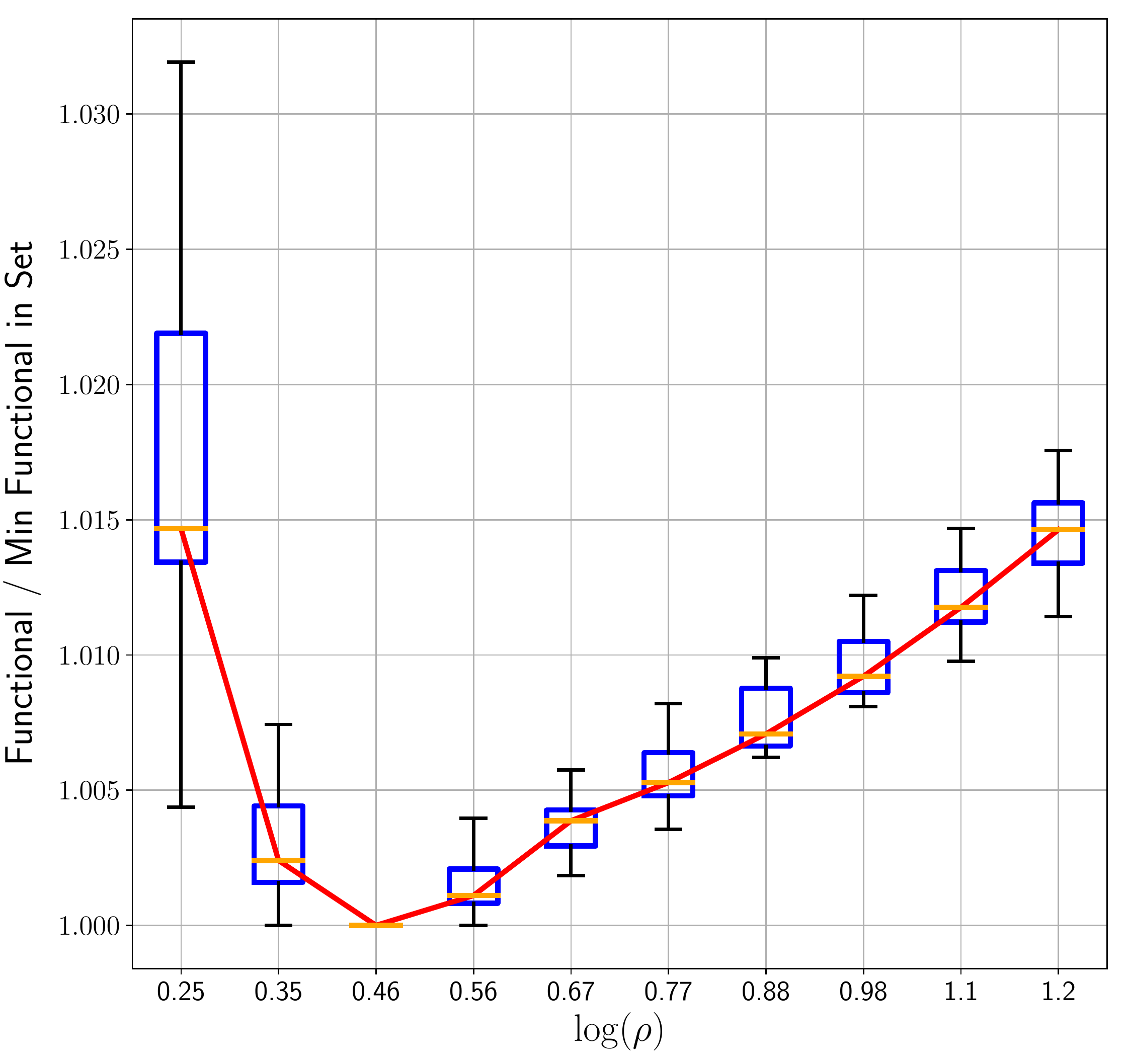}}}
        \hfil
        \subfigure[CBPDN$(\sigma)$ for best $\rho$]{\scalebox{0.18}{\includegraphics{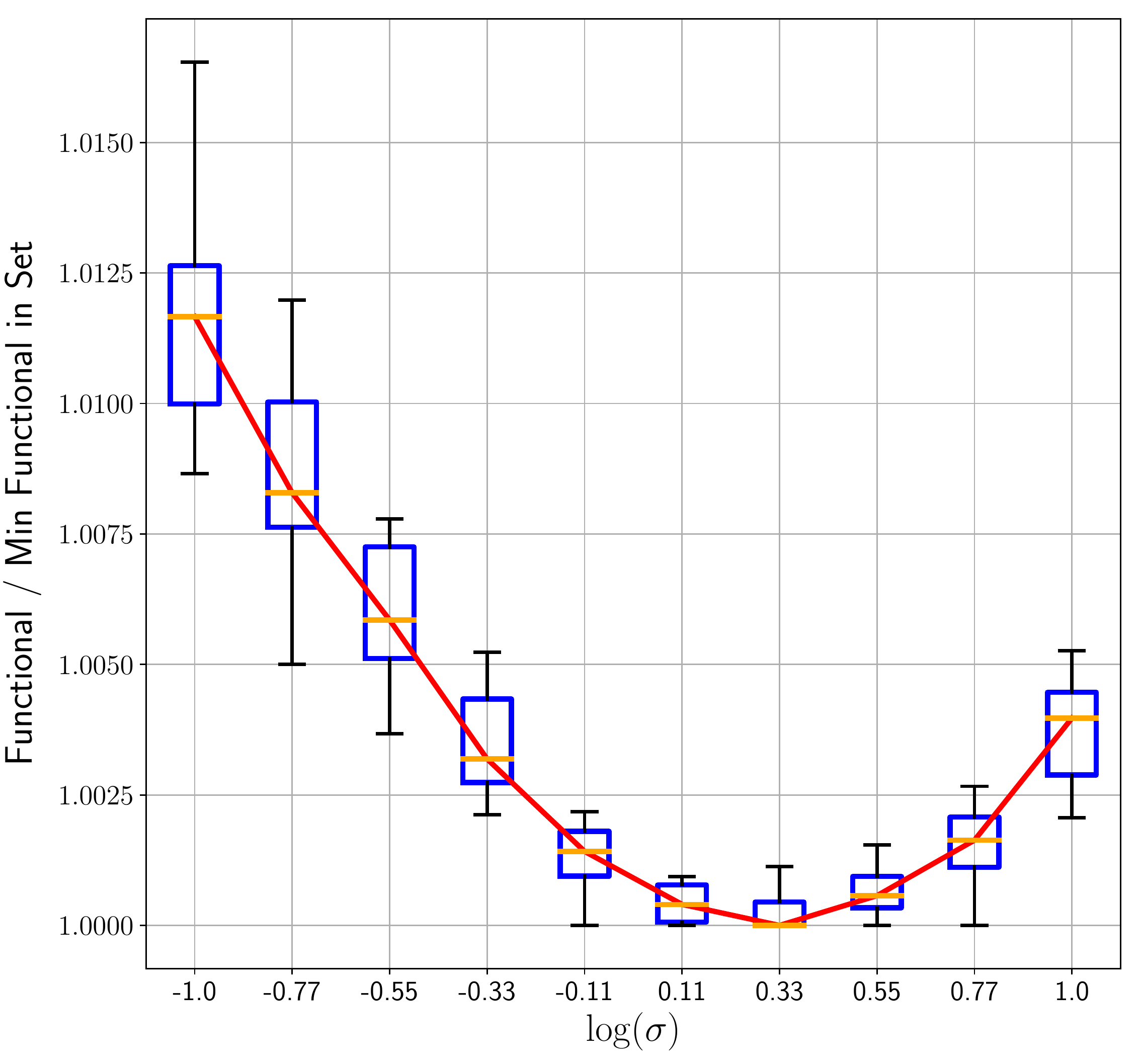}}}
}
        \caption{Distribution of normalized CBPDN functional (\eq{cbpdnmmv} in the main document) after 500 iterations, in the consensus (Cns / Cns-P) grid search for 20 random selected sets of $K=20$ images.}
        \label{fig:cns_par_selection_K20}
\end{figure}

\begin{figure}[htb]
\centerline{
        \subfigure[CBPDN$(\rho)$ for best $L$]{\scalebox{0.18}{\includegraphics{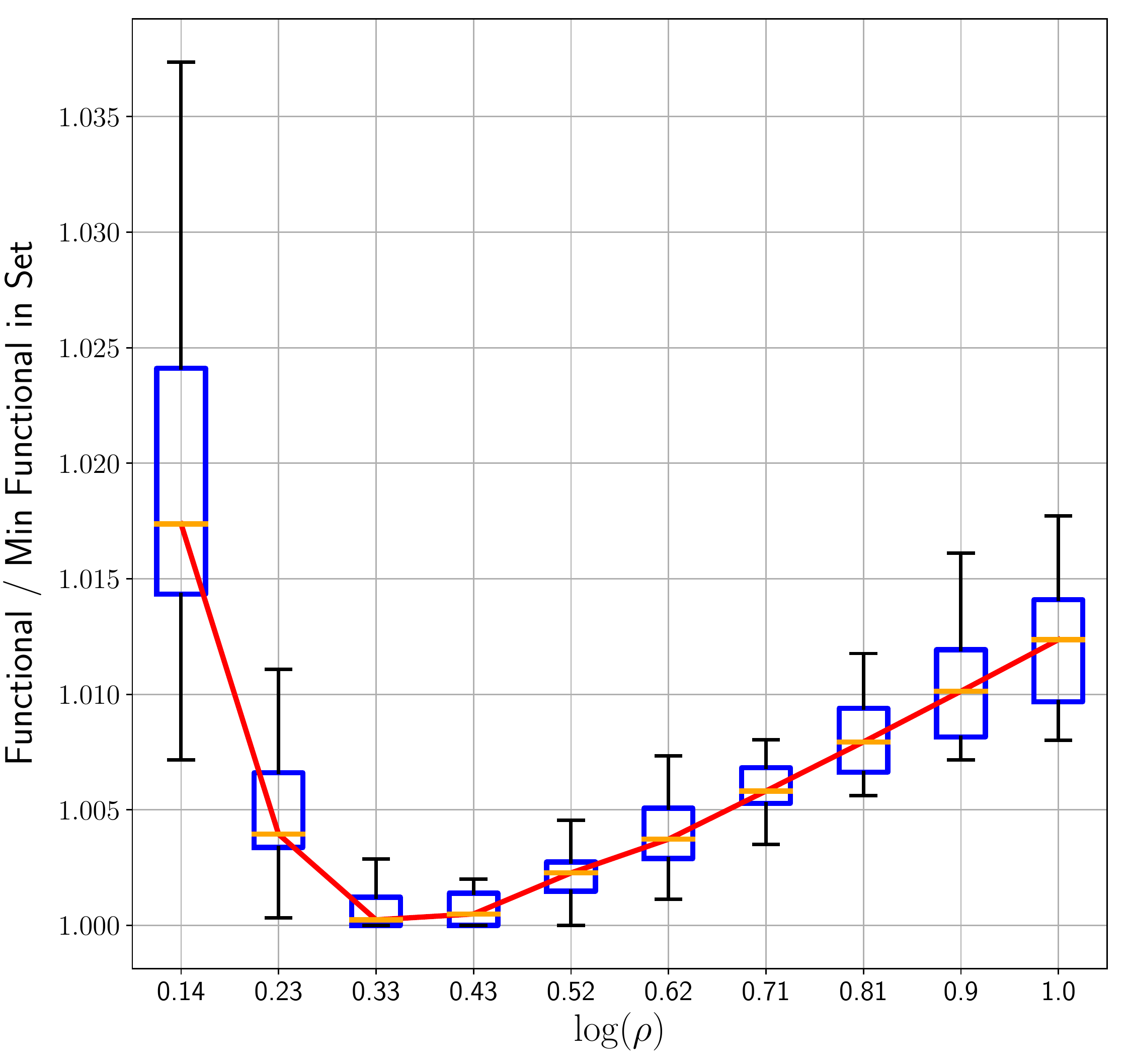}}}
        \hfil
        \subfigure[CBPDN$(L)$ for best $\rho$]{\scalebox{0.18}{\includegraphics{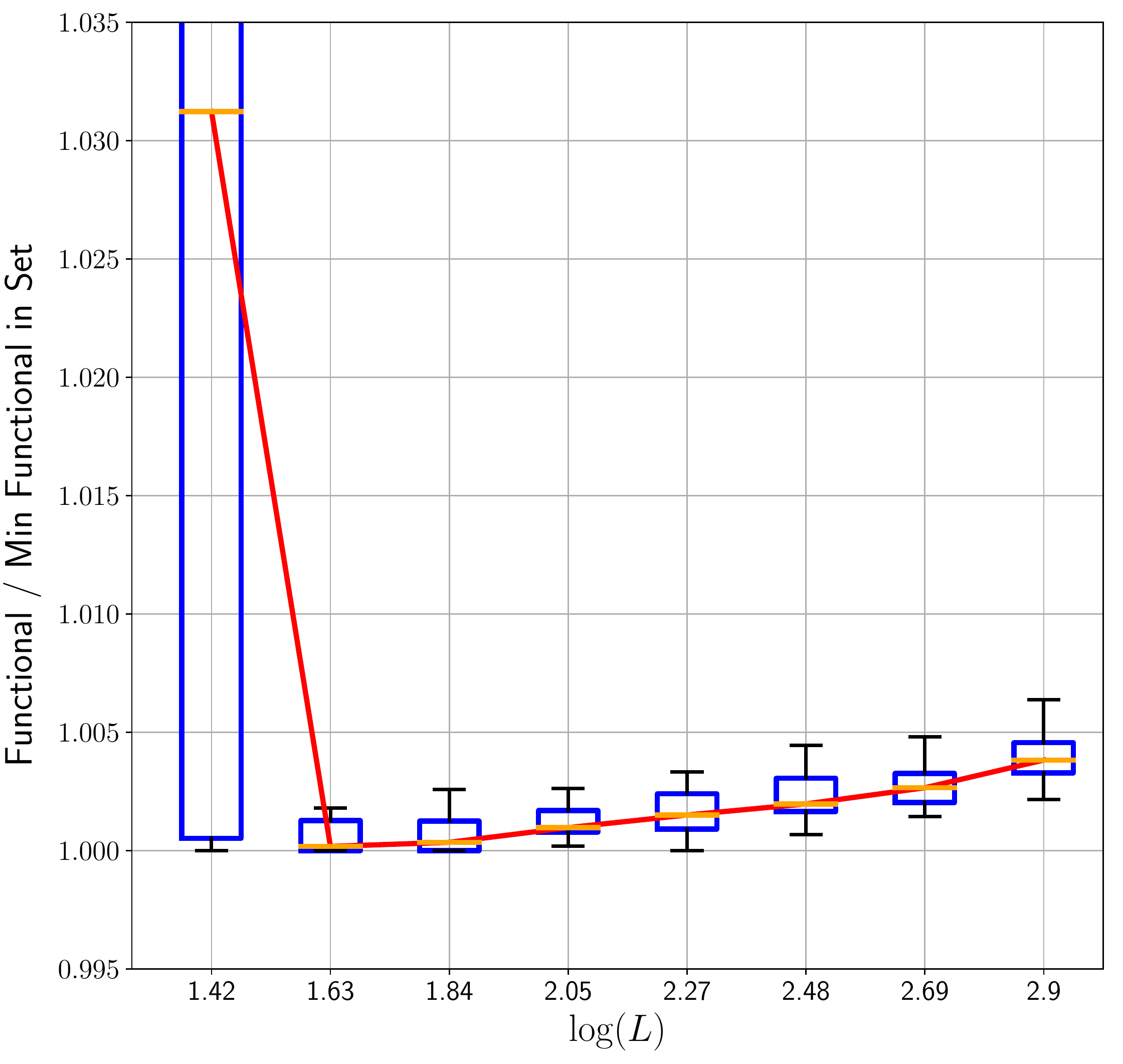}}
        \label{fig:fista_par_selection_K5_L}}
}
        \caption{Distribution of normalized CBPDN functional (\eq{cbpdnmmv} in the main document) after 500 iterations, in the FISTA grid search for 20 random selected sets of $K=5$ images.}
        \label{fig:fista_par_selection_K5}
\end{figure}

\begin{figure}[htb]
\centerline{
        \subfigure[CBPDN$(\rho)$ for best $L$]{\scalebox{0.18}{\includegraphics{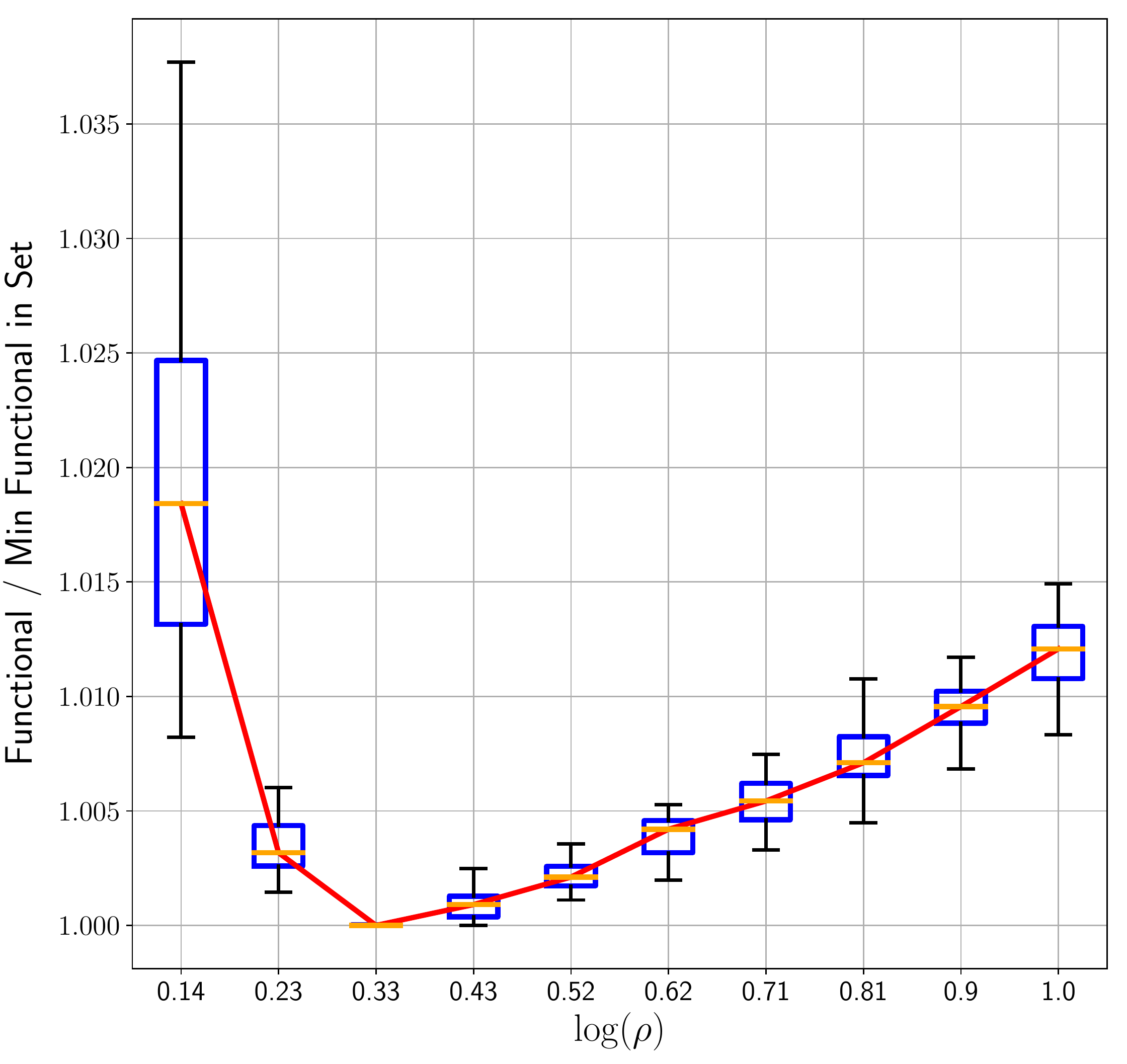}}}
        \hfil
        \subfigure[CBPDN$(L)$ for best $\rho$]{\scalebox{0.18}{\includegraphics{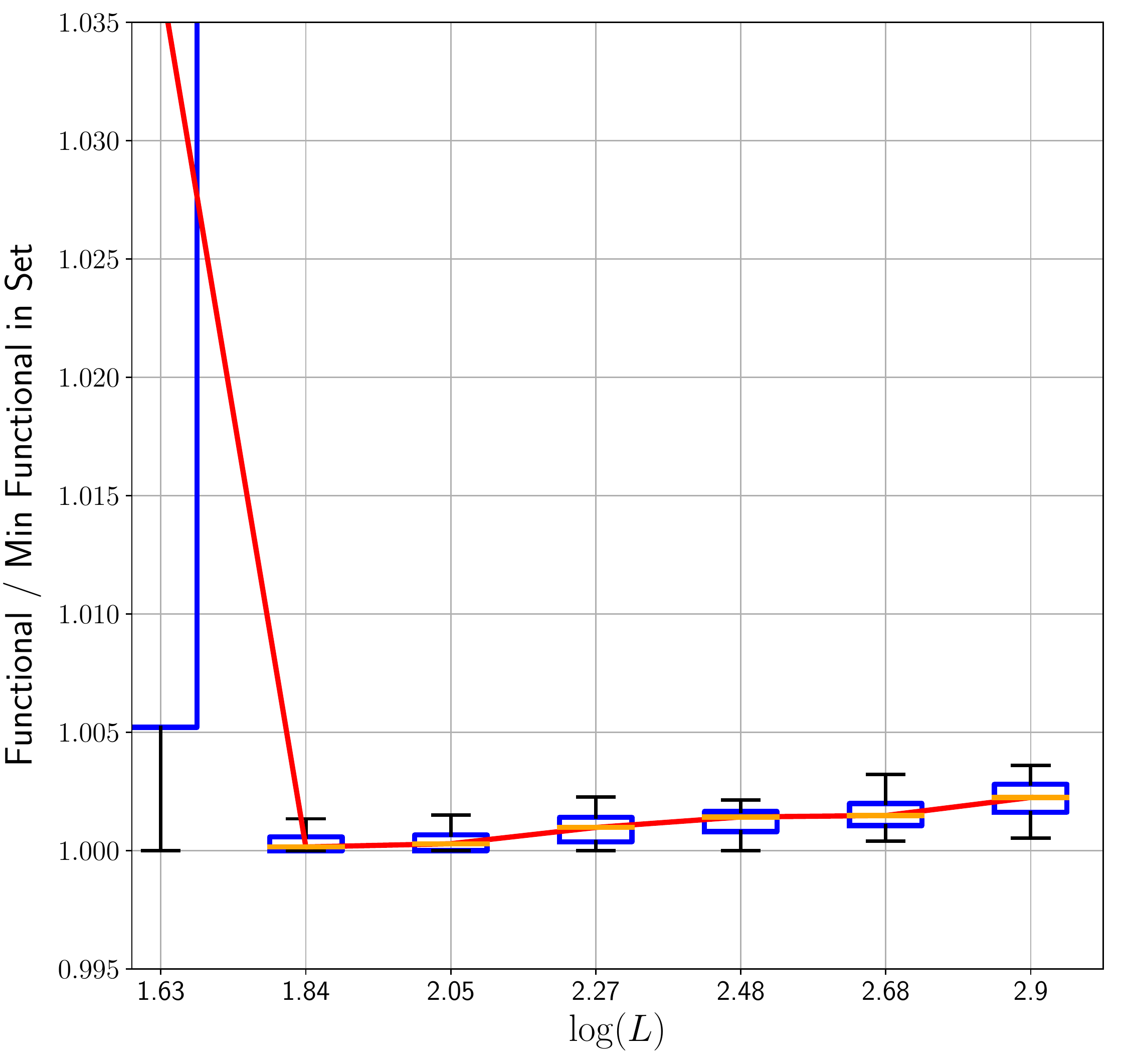}}
        \label{fig:fista_par_selection_K10_L}}
}
        \caption{Distribution of normalized CBPDN functional (\eq{cbpdnmmv} in the main document) after 500 iterations, in the FISTA grid search for 20 random selected sets of $K=10$ images.}
        \label{fig:fista_par_selection_K10}
\end{figure}

\begin{figure}[htb]
\centerline{
        \subfigure[CBPDN$(\rho)$ for best $L$]{\scalebox{0.18}{\includegraphics{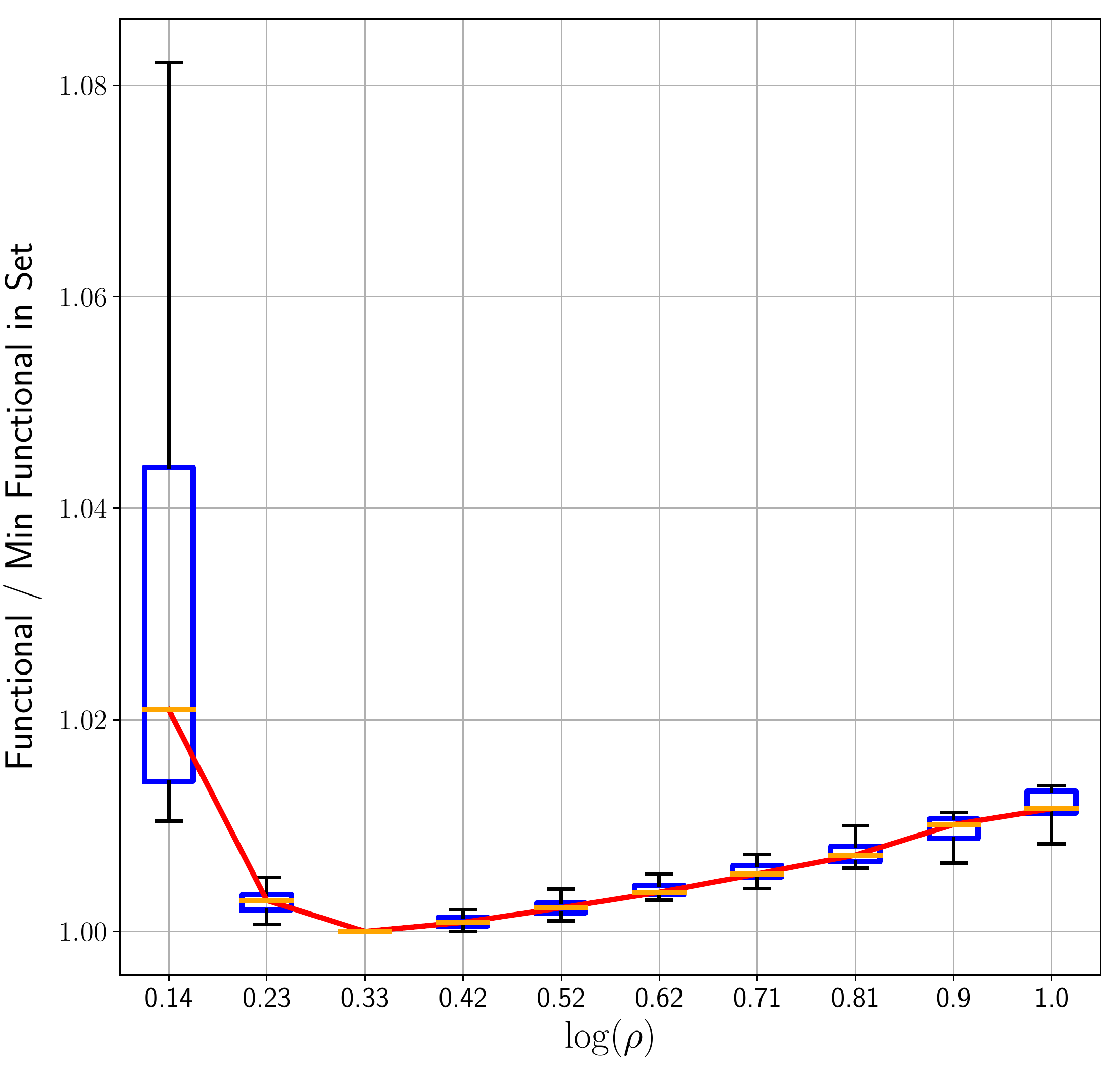}}}
        \hfil
        \subfigure[CBPDN$(L)$ for best $\rho$]{\scalebox{0.18}{\includegraphics{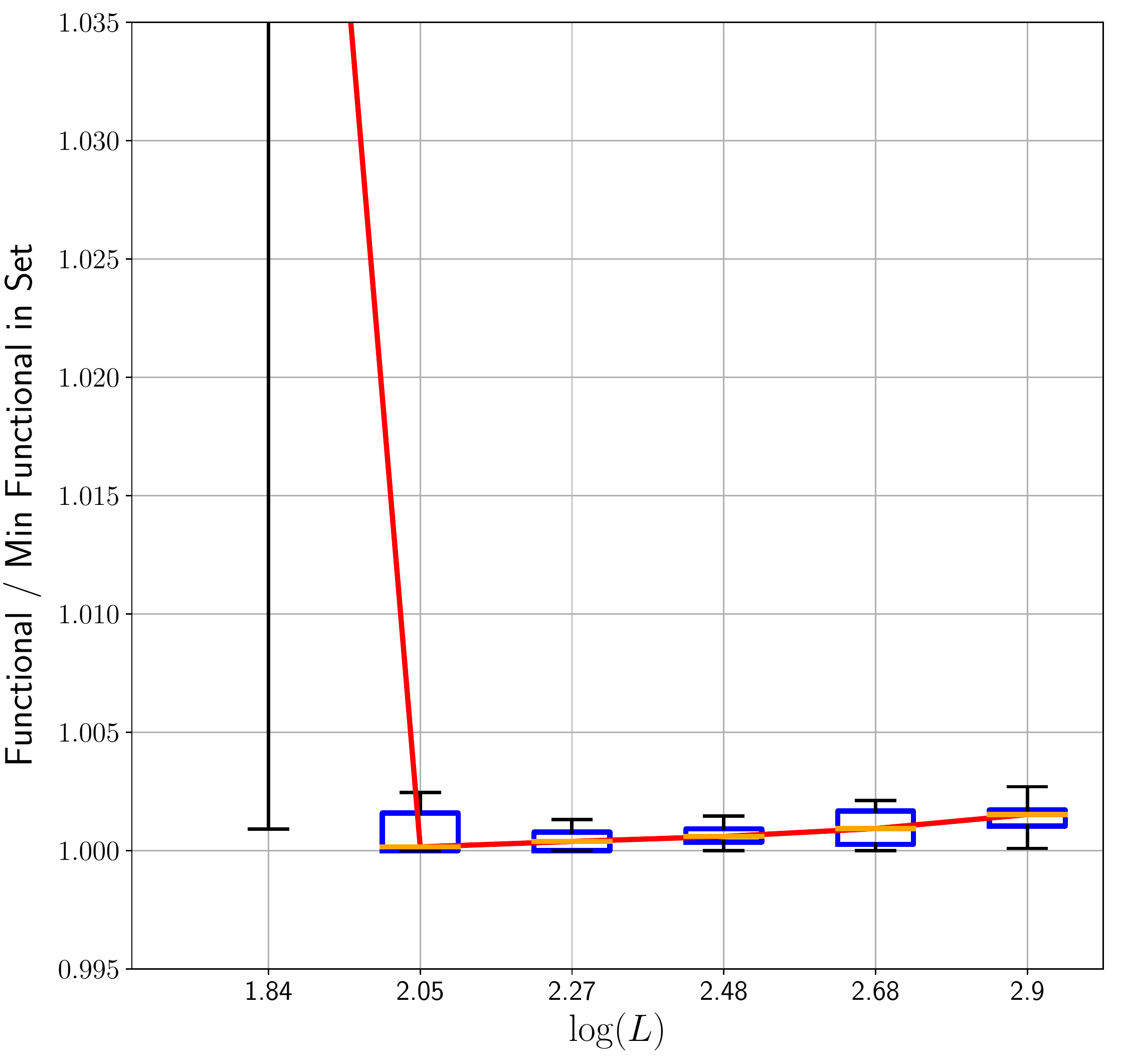}}
        \label{fig:fista_par_selection_K20_L}}
}
        \caption{Distribution of normalized CBPDN functional (\eq{cbpdnmmv} in the main document) after 500 iterations, in the FISTA grid search for 20 random selected sets of $K=20$ images.}
        \label{fig:fista_par_selection_K20}
\end{figure}

\begin{figure}[htb]
\centerline{
        \subfigure[CBPDN$(\rho)$ for best $\sigma$]{\scalebox{0.18}{\includegraphics{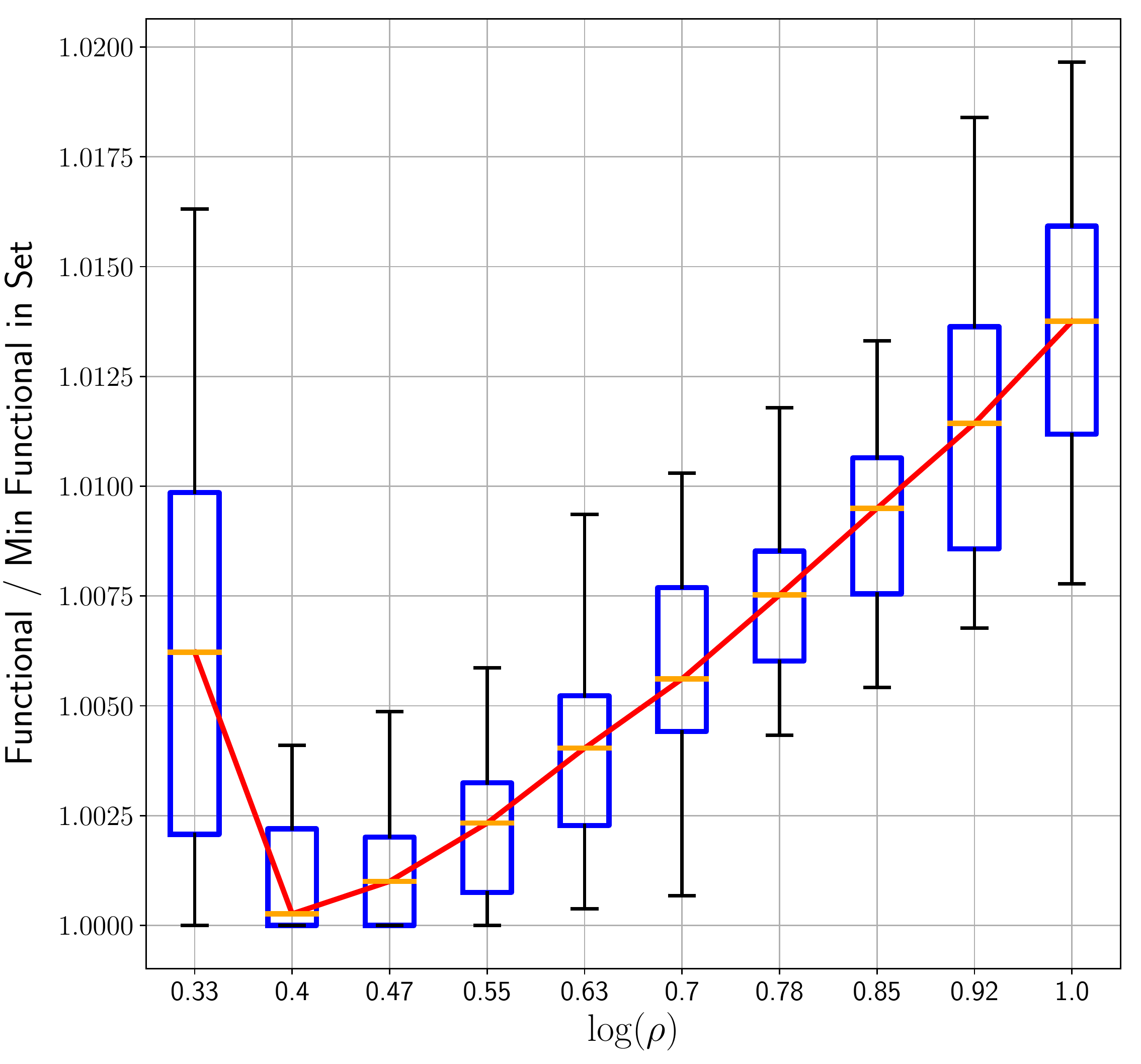}}}
        \hfil
        \subfigure[CBPDN$(\sigma)$ for best $\rho$]{\scalebox{0.18}{\includegraphics{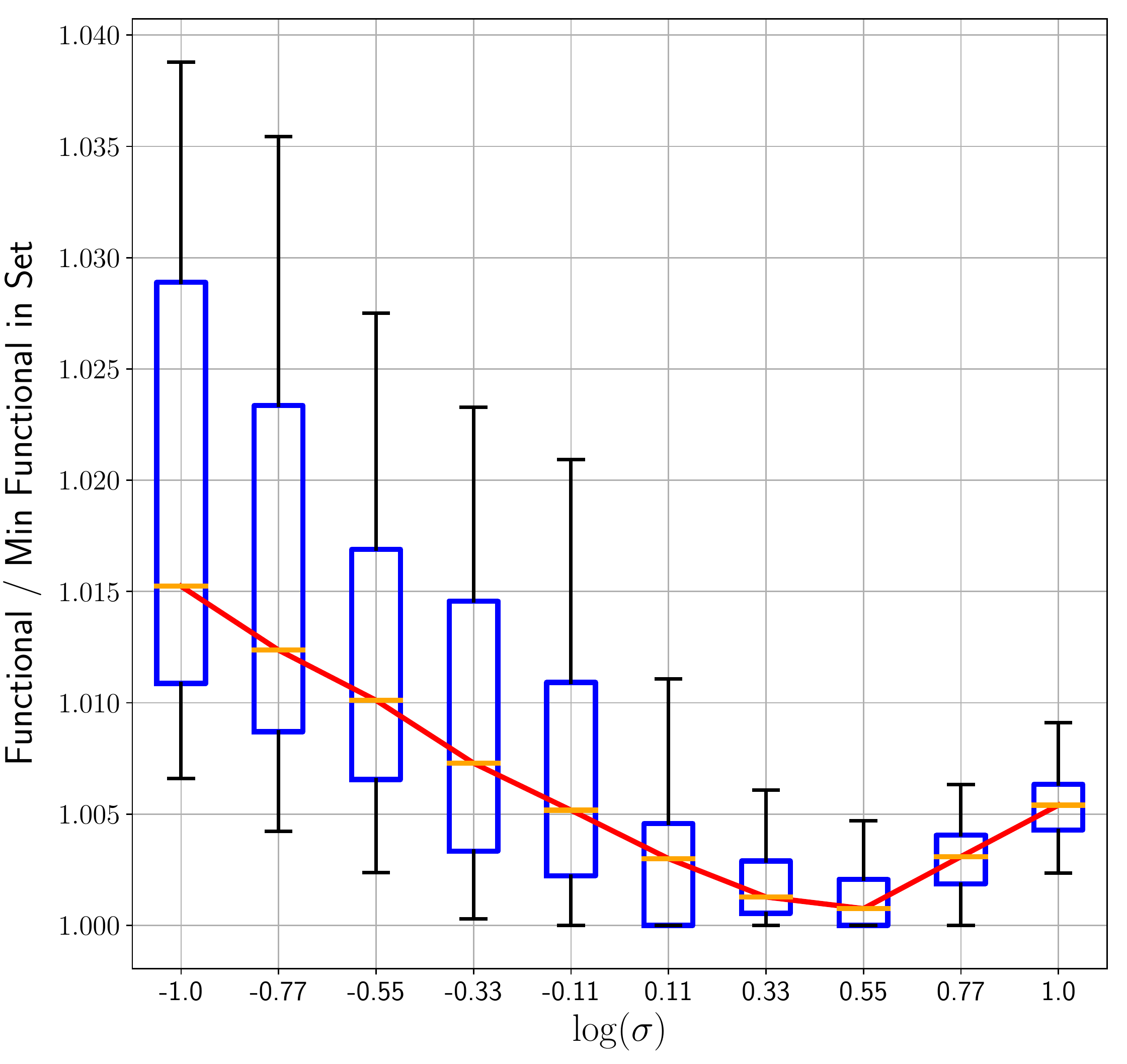}}}
}
        \caption{Distribution of normalized masked CBPDN functional (\eq{cbpdnmsk} in the main document) after 500 iterations, in the masked consensus (M-Cns / M-Cns-P) grid search for 20 random selected sets of $K=5$ images.}
        \label{fig:MDcns_par_selection_K5}
\end{figure}

\begin{figure}[htb]
\centerline{
        \subfigure[CBPDN$(\rho)$ for best $\sigma$]{\scalebox{0.18}{\includegraphics{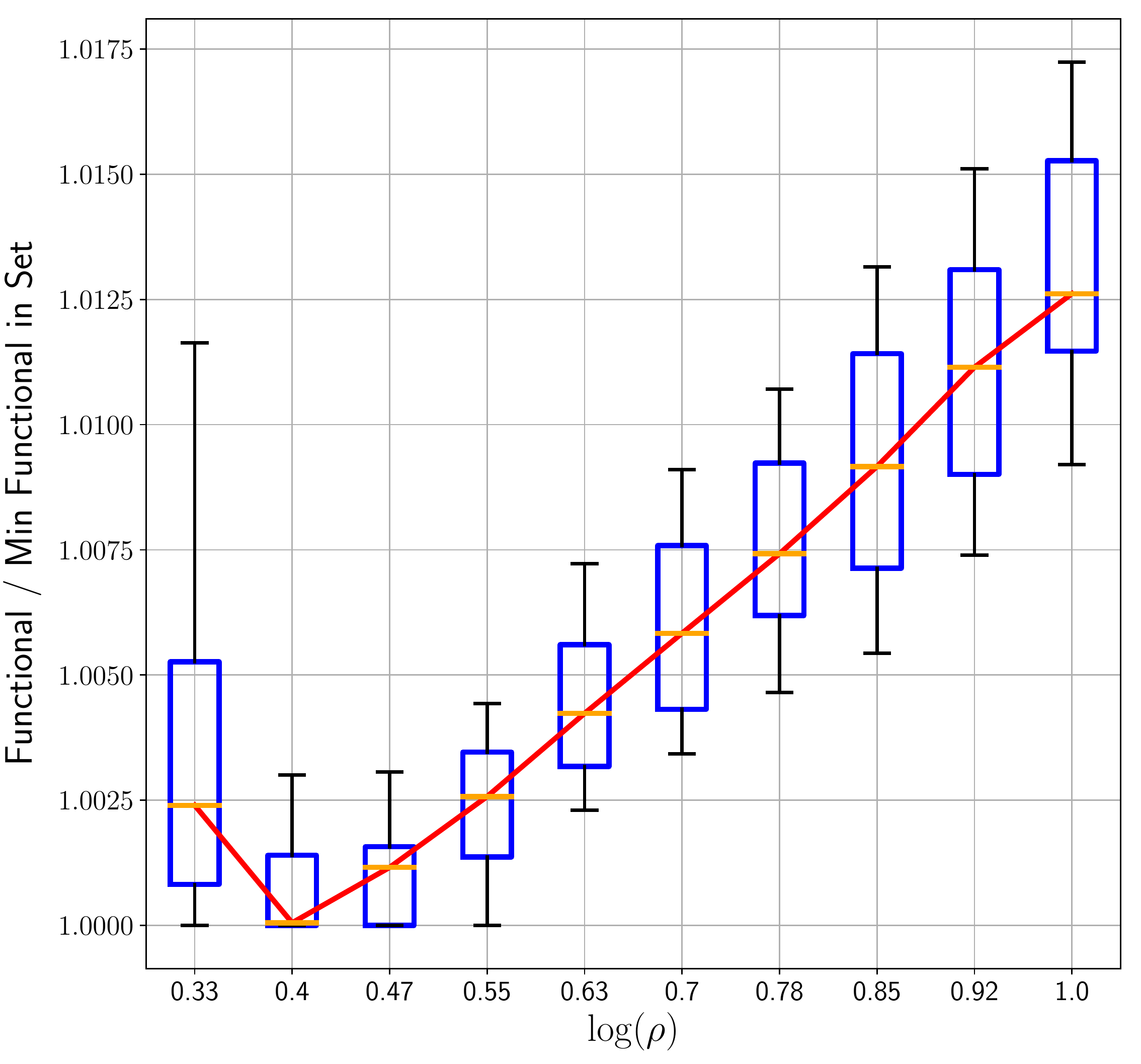}}}
        \hfil
        \subfigure[CBPDN$(\sigma)$ for best $\rho$]{\scalebox{0.18}{\includegraphics{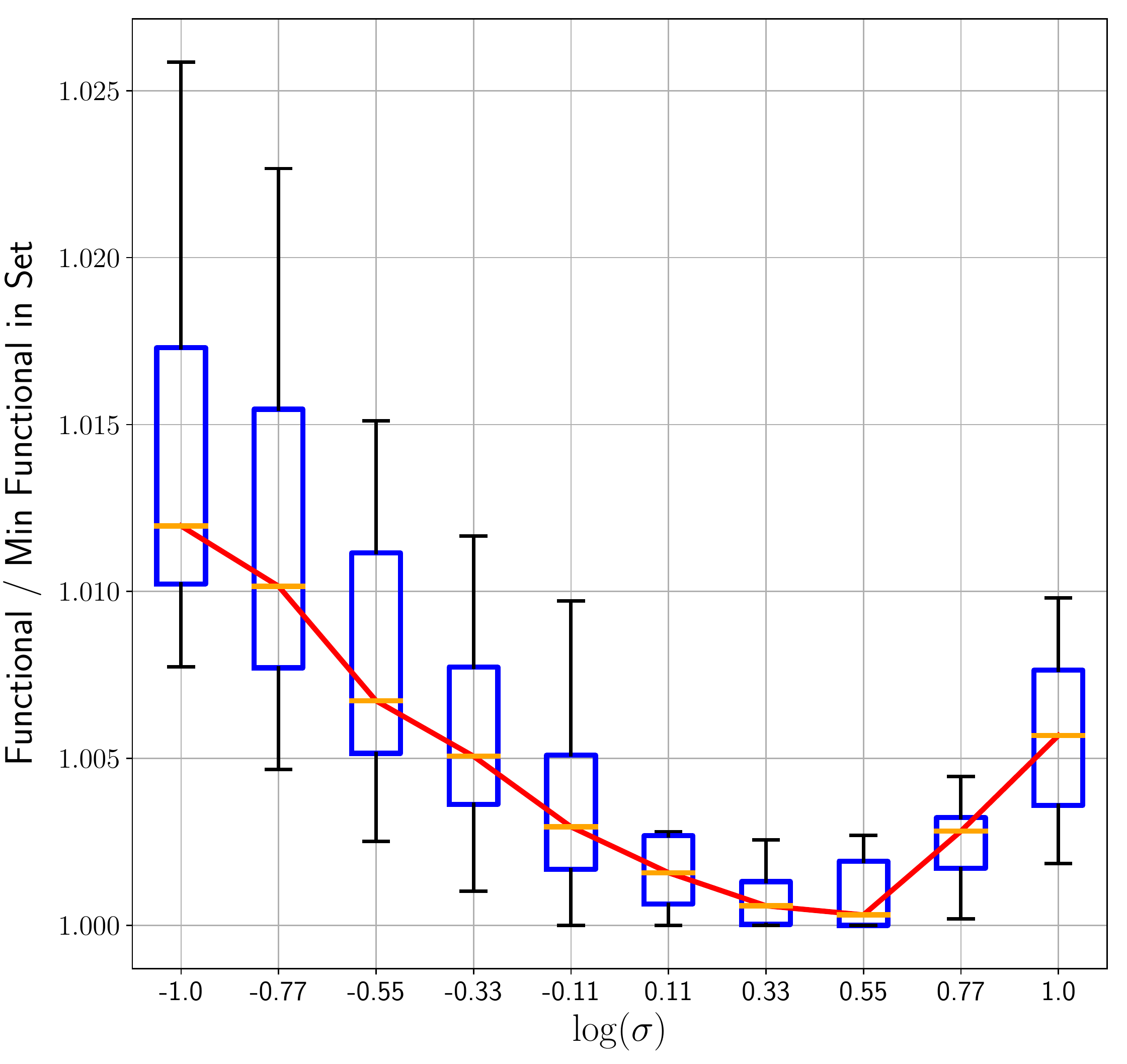}}}
}
        \caption{Distribution of normalized masked CBPDN functional (\eq{cbpdnmsk} in the main document) after 500 iterations, in the masked consensus  (M-Cns / M-Cns-P) grid search for 20 random selected sets of $K=10$ images.}
        \label{fig:MDcns_par_selection_K10}
\end{figure}

\begin{figure}[htb]
\centerline{
        \subfigure[CBPDN$(\rho)$ for best $\sigma$]{\scalebox{0.18}{\includegraphics{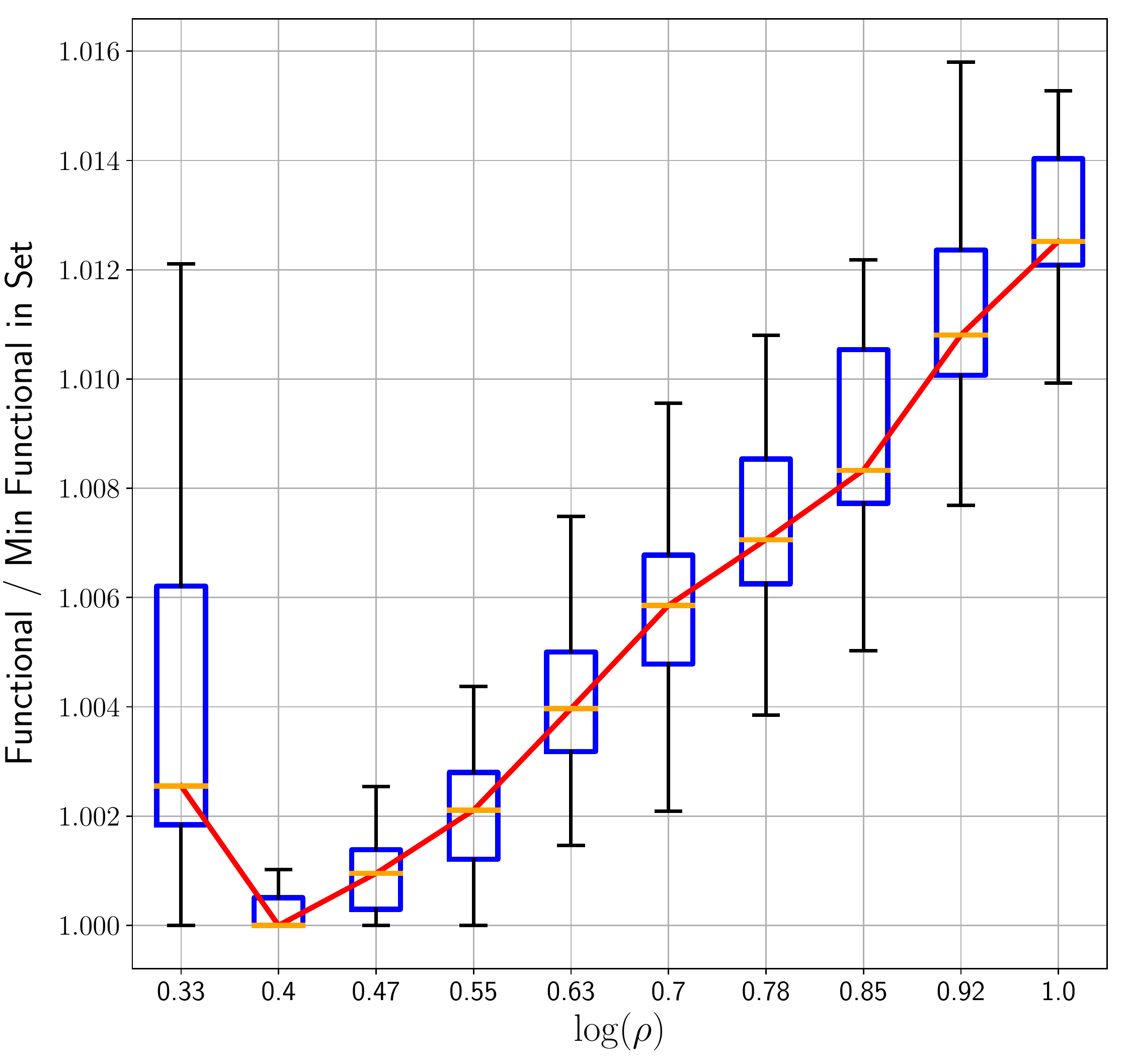}}}
        \hfil
        \subfigure[CBPDN$(\sigma)$ for best $\rho$]{\scalebox{0.18}{\includegraphics{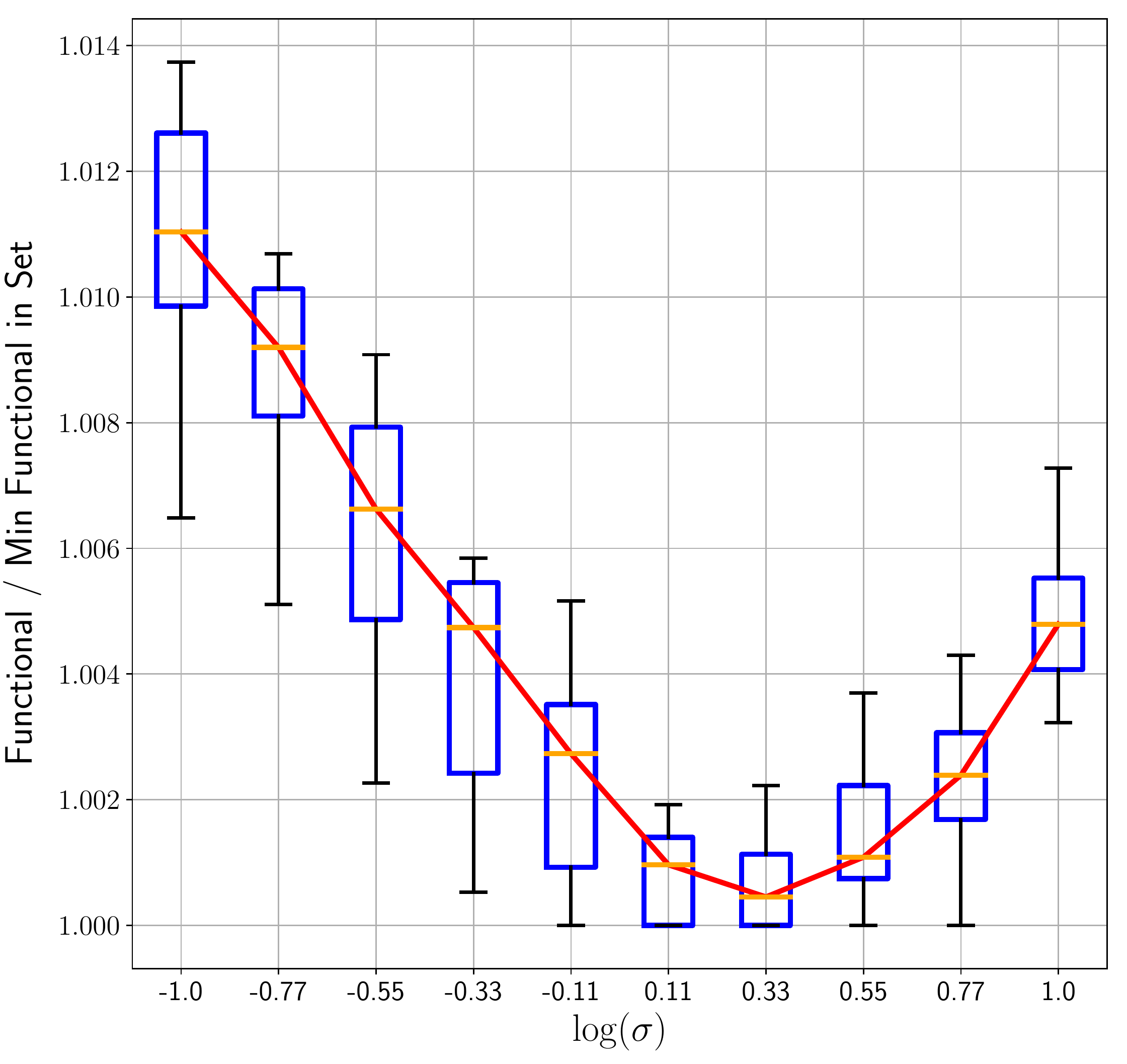}}}
}
        \caption{Distribution of normalized masked CBPDN functional (\eq{cbpdnmsk} in the main document) after 500 iterations, in the masked consensus (M-Cns / M-Cns-P) grid search for 20 random selected sets of $K=20$ images.}
        \label{fig:MDcns_par_selection_K20}
\end{figure}

Experiments to determine the median stability of the optimal parameters across an ensemble of training sets of the same size are discussed in~\sctn{parameter_sensitivity} in the main document. The corresponding results are plotted here in~\figs{cg_par_selection_K5}~--~\fign{MDcns_par_selection_K20}. The box plots represent median, quartiles, and the full range of variation of the normalized functional values obtained at each parameter value for the 20 different image subsets at each of the sizes $K \in \{5, 10, 20\}$. The red lines connect the medians of the distributions at each parameter value.

It can be seen in~\figs{fista_par_selection_K5_L}, \fign{fista_par_selection_K10_L}, and \fign{fista_par_selection_K20_L} that FISTA has very skewed sensitivity plots for $L$, the inverse of the gradient step size. This is related to the requirement, mentioned in the main document, that $L$ has to be greater than or equal to the Lipschitz constant of the gradient of the functional to guarantee convergence of the algorithm. Although this constant is not always computable~\cite{beck-2009-fast}, in these experiments we are able to estimate the threshold that indicates the change in behavior expected when $L$ becomes greater than the Lipschitz constant.  The variation of the normalized functional values is comparable to those for other methods and other parameters for values of $L$ greater than this threshold. However, for values of $L$ smaller than the threshold, the instability causes a much larger variance in the normalized functional values. We decided to clip the large vertical ranges resulting from the very large variances to the left of these plots in order to more clearly display the scaling in the useful range of $L$. As a result, some of the interquartile range boxes to the left are incomplete, or just the lower part of the full range of variation is visible.

\section{Large Training Set Experiments}
\label{sec:largecompare}

In order to evaluate the performance of the methods for larger training sets and images of different sizes, we performed additional experiments, including comparisons with the original implementations of competing algorithms. We used training sets of 25, 100 and 400 images of sizes 1024 $\times$ 1024 pixels, 512 $\times$ 512 pixels and 256 $\times$ 256 pixels, respectively. These combinations of number, $K$, and size, $N$, of images were chosen to maintain a constant number of pixels in the training set, which provides a useful way of simultaneously exploring performance variations with respect to both $N$ and $K$. All of these images were derived from images in the MIRFLICKR-1M dataset and pre-processed (scaling and highpass filtering) in the same way, as described in~\sctn{cdlexp} in the main document.

All the results using the methods discussed and analyzed in the main document were computed using the Python implementation of the SPORCO library~\cite{wohlberg-2016-sporco, wohlberg-2017-sporco} on a Linux workstation equipped with two Xeon E5-2690V4 CPUs. We also include comparisons with the method proposed by Papyan \etal~\cite{papayan-2017-convolutional}, using their publicly available Matlab and C implementation\footnote{Available from \url{http://vardanp.cswp.cs.technion.ac.il/wp-content/uploads/sites/62/2015/12/SliceBasedCSC.rar}}.

We tried to include the publicly available Matlab implementations of the methods proposed by \v{S}orel and \v{S}roubek\footnote{Available from
\url{https://github.com/michalsorel/convsparsecoding}}~\cite{sorel-2016-fast} and by Heide \etal\footnote{Available from \url{http://www.cs.ubc.ca/labs/imager/tr/2015/FastFlexibleCSC}}~\cite{heide-2015-fast} in these comparisons, but were unable to obtain acceptable results\footnote{The methods were very slow, with partial results after running for 4 days still being noisy and far from convergence.}. We therefore omit these methods from the comparisons here, including them only in a separate set of experiments on a smaller data set, reported in~\sctn{otheralgcompare} below.

In all of these experiments we learned a dictionary of 100 filters of size 11 $\times$ 11, setting the sparsity parameter $\lambda = 0.1$. We set the parameters for our methods according to the scaling rules discussed in~\sctn{parameter_sensitivity} in the main document, using fixed penalty parameters $\rho$ and $\sigma$ without any adaptation methods. In contrast to the experiments reported in the main document, relaxation methods~\cite[Sec. 3.4.3]{boyd-2010-distributed}\cite[Sec. III.D]{wohlberg-2016-efficient} were used, with $\alpha = 1.8$.

We used the default parameters from the demonstration scripts distributed with each of the publicly available Matlab implementations by the authors of~\cite{papayan-2017-convolutional}, \cite{heide-2015-fast}, and~\cite{sorel-2016-fast}. Our efforts to adjust the default parameters for the implementations of the methods of~\cite{heide-2015-fast}, and~\cite{sorel-2016-fast} to obtain better results were unsuccessful, at least in part due to the slow convergence of the methods and the absence of any parameter selection discussion or guidelines provided by the authors.

During training, the dictionaries were saved at 25 iteration intervals to allow evaluation on an independent test set, which consisted of the same additional set of 20 images, of size 256 $\times$ 256 pixels, that was used for this purpose for the experiments reported in the main document. This evaluation was performed by sparse coding of the images in the test set, for $\lambda = 0.1$, computing the evolution of the CBPDN functional over the series of dictionaries. This not only allows comparison of generalization performance, taking into account possible differences in overfitting effects between the different methods, but also allows for a fair comparison between the methods, avoiding the difficulty of comparing the training functional values that are computed differently by different implementations\footnote{All of our implementations calculate the functional values in the same way, but the implementations by other authors adopt slightly different approaches.}.

\subsection{CDL without Spatial Mask}

\begin{figure}[htb]
        \center
        \scalebox{0.26}{\includegraphics{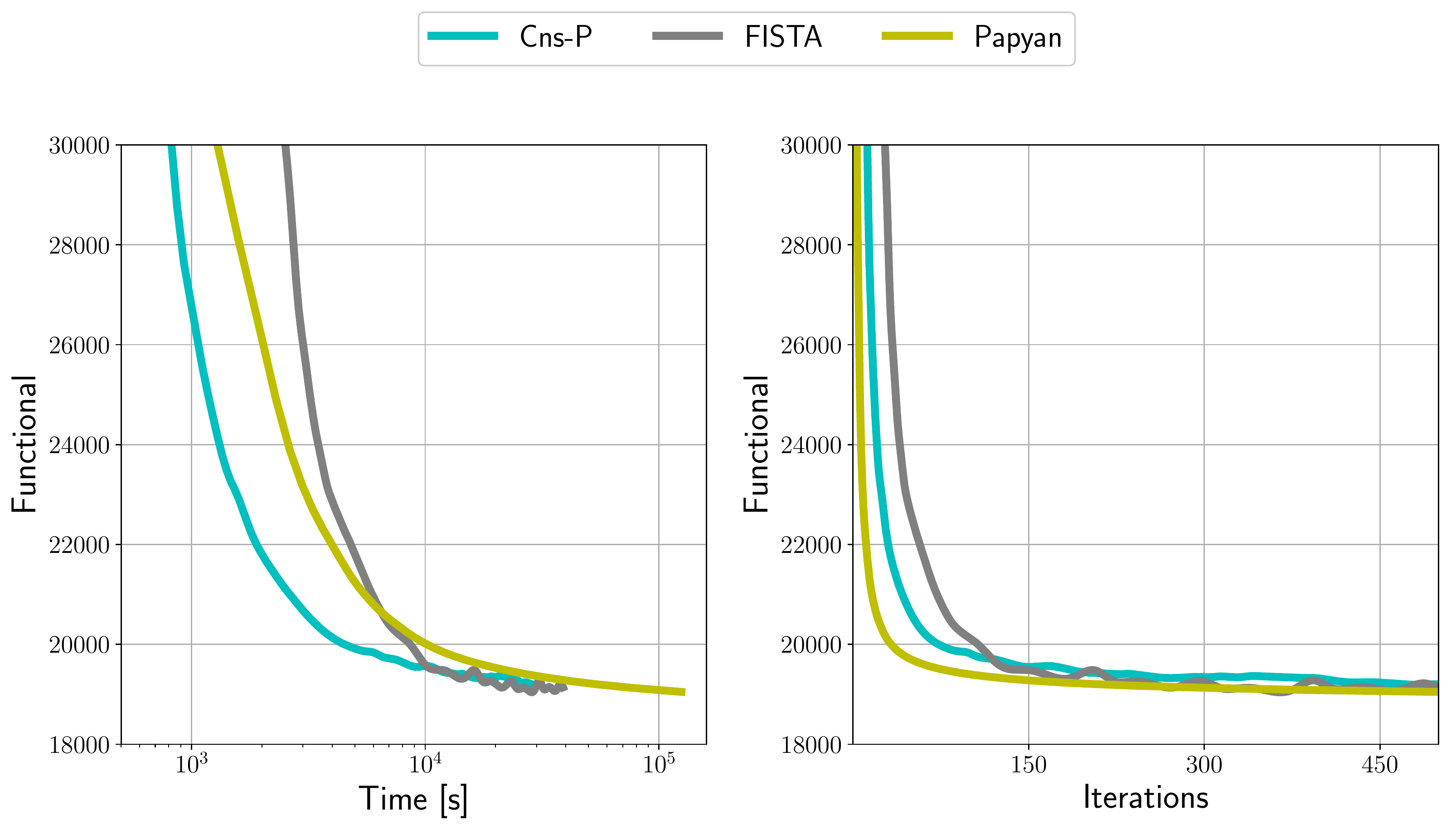}}
        \caption{Dictionary Learning ($K=25$): A comparison on a set of $K=25$ images, 1024 $\times$ 1024 pixels, of the decay of the value of the CPBDN functional~\eq{cbpdnmmv} with respect to run time and iterations.}
        \label{fig:fObj_k25}
\end{figure}

\begin{figure}[htb]
        \center
        \scalebox{0.26}{\includegraphics{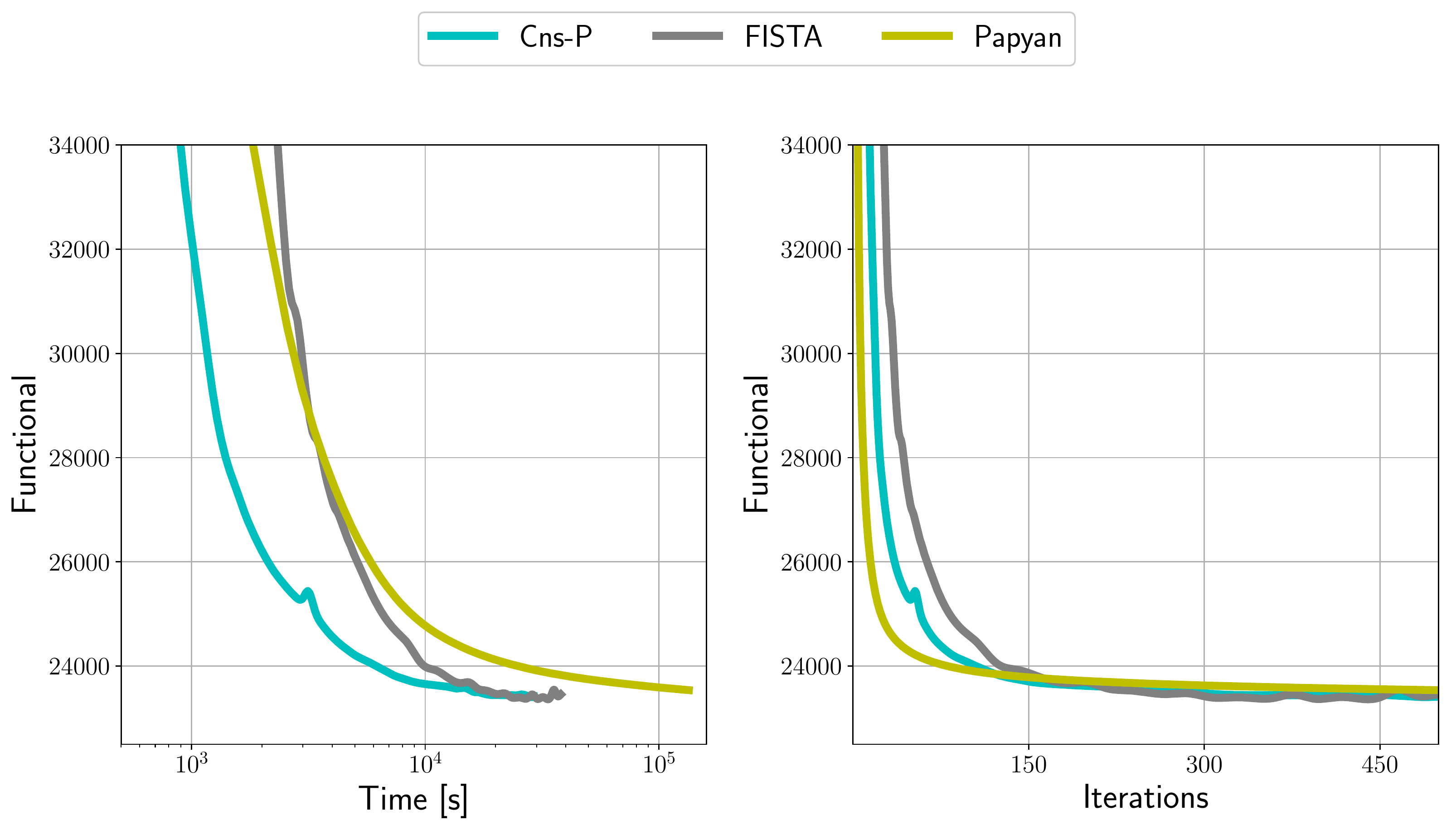}}
        \caption{Dictionary Learning ($K=100$): A comparison on a set of $K=100$ images, 512 $\times$ 512 pixels, of the decay of the value of the CBPDN functional~\eq{cbpdnmmv} with respect to run time and iterations.}
        \label{fig:fObj_k100}
\end{figure}

\begin{figure}[htb]
        \center
        \scalebox{0.26}{\includegraphics{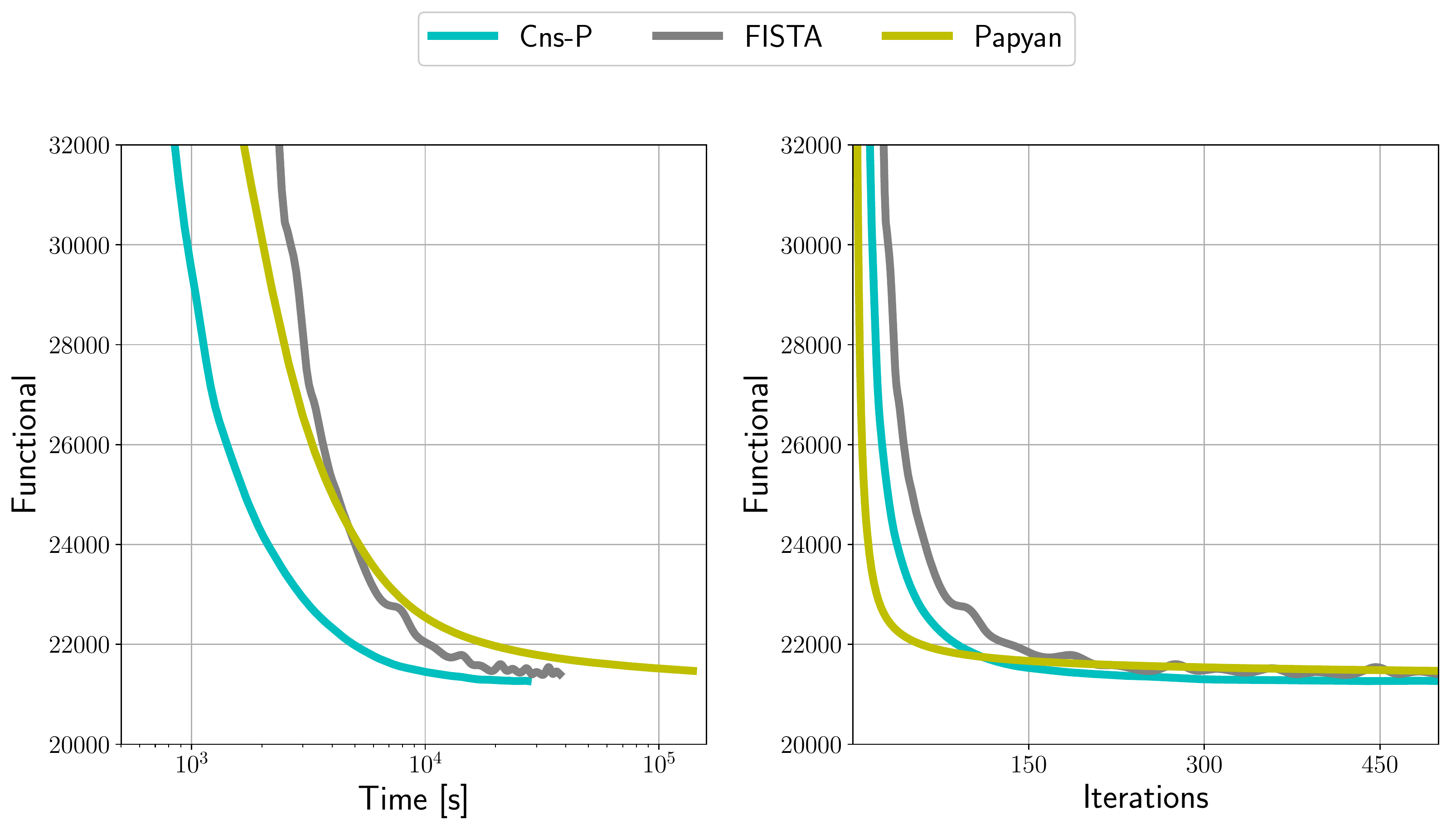}}
        \caption{Dictionary Learning ($K=400$): A comparison on a set of $K=400$ images, 256 $\times$ 256 pixels, of the decay of the value of the CBPDN functional~\eq{cbpdnmmv} with respect to run time and iterations.}
        \label{fig:fObj_k400}
\end{figure}

Results for the training objective function are shown in~\fig{fObj_k25} for $K=25$ with $1024 \times 1024$ images, in~\fig{fObj_k100} for $K=100$ with $512 \times 512$ images, and in~\fig{fObj_k400} for $K=400$ with 2$56 \times 256$ images. It is clear that Cns-P consistently achieves the best performance, converging smoothly to a slightly smaller functional value than the other two methods in all the cases except for~\fig{fObj_k25}.
It also exhibits the fastest convergence of the methods compared. In contrast, FISTA results are less stable, presenting some wild oscillations at the beginning and some small oscillations at the end, but nevertheless achieving similar final functional values to Cns-P. The method of Papyan \etal~\cite{papayan-2017-convolutional} has very rapid convergence in terms of iterations, but its time performance is the worst of the three methods.

The FISTA instability can be automatically corrected by using the backtracking step-size adaptation rule (see~\sctn{dufista} in main document).
However, due to the uni-directional correction of the backtracking rule that always increases $L$ (\ie it always decreases the gradient step size), the evolution of the functional is smooth, but also tends to converge to a larger functional value. A reasonable approach for methods that do not converge monotonically, such as FISTA, is to consider the solution at each time step as the best solution obtained until that step, as opposed to the solution specifically for that step, which has the effect of smoothing the functional value evolution. In all our experiments, we used a fixed $L$ value, set in accordance with the parameter rules described in the main document, and report actual convergence without any post processing since this more accurately illustrates the real FISTA behavior and the tradeoff between convergence smoothness and final functional value determined by parameter $L$.

\begin{figure}[htb]
        \center
        \scalebox{0.26}{\includegraphics{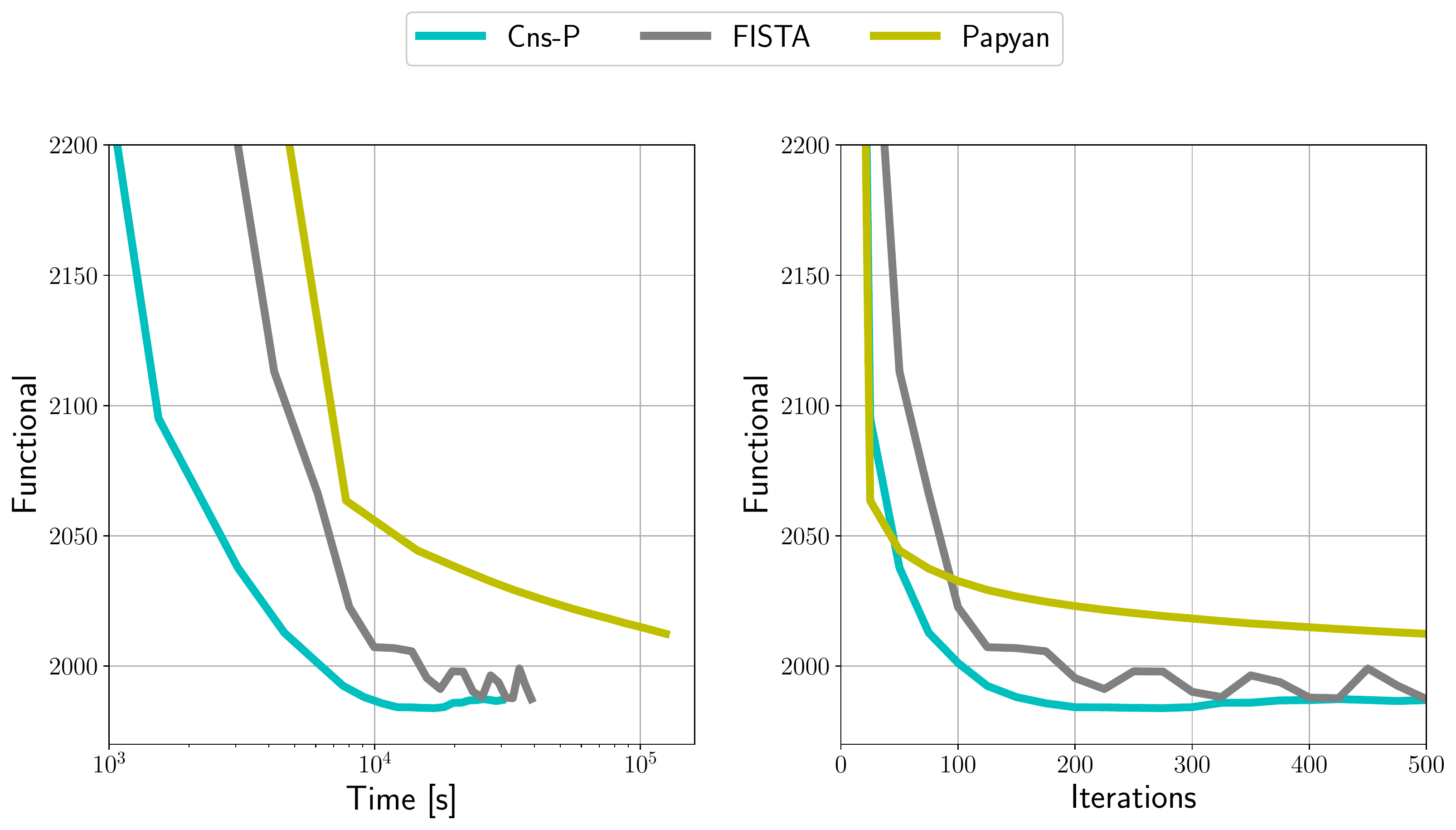}}
        \caption{Evolution of the CBPDN functional~\eq{cbpdnmmv} for the test set using the partial dictionaries obtained when training for $K=25$ images, 1024 $\times$ 1024 pixels, as in~\fig{fObj_k25}.}
        \label{fig:val_k25}
\end{figure}

\begin{figure}[htb]
        \center
        \scalebox{0.26}{\includegraphics{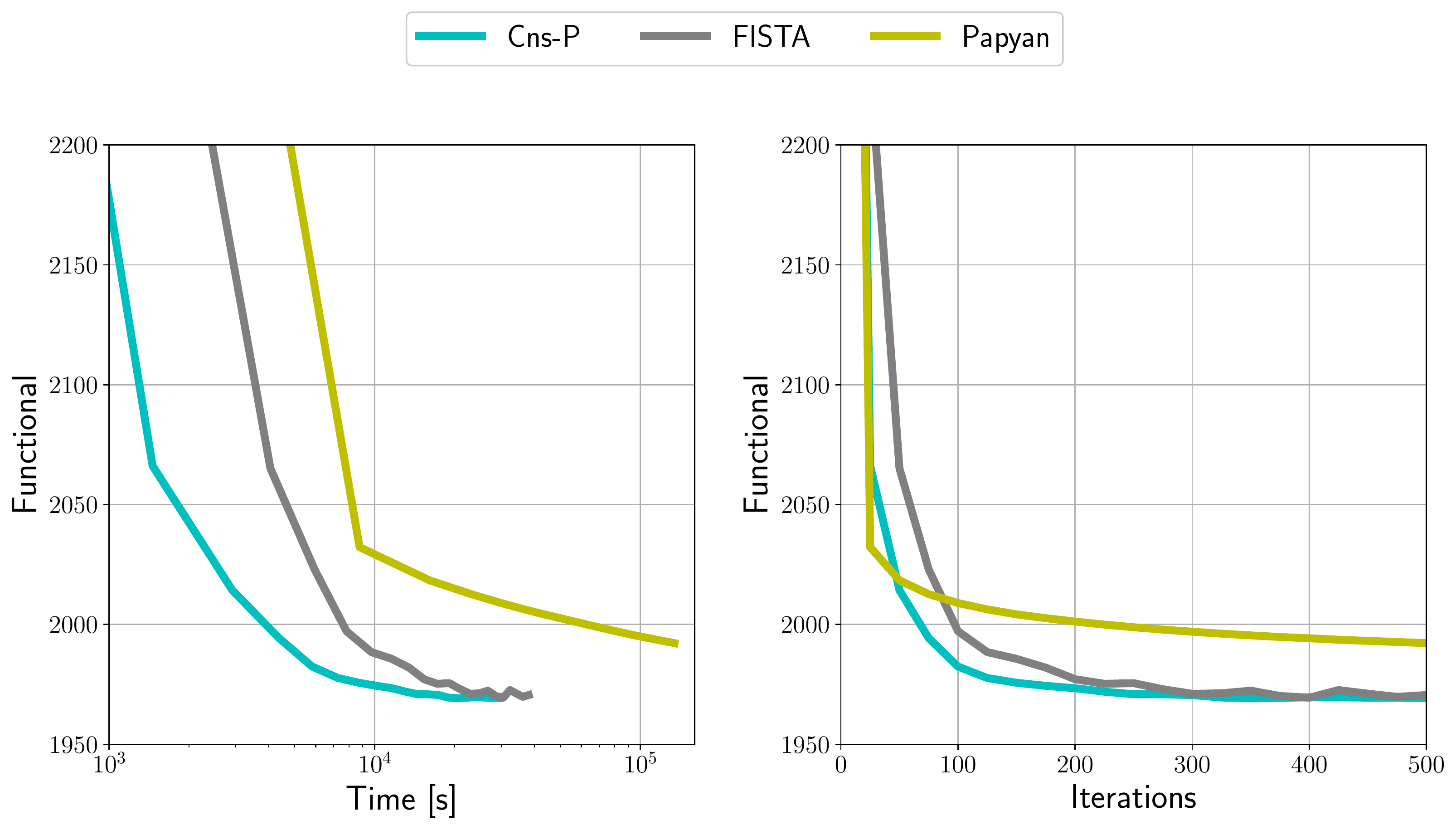}}
        \caption{Evolution of the CBPDN functional~\eq{cbpdnmmv} for the test set using the partial dictionaries obtained when training for $K=100$ images, 512 $\times$ 512 pixels, as in~\fig{fObj_k100}.}
        \label{fig:val_k100}
\end{figure}

\begin{figure}[htb]
        \center
        \scalebox{0.26}{\includegraphics{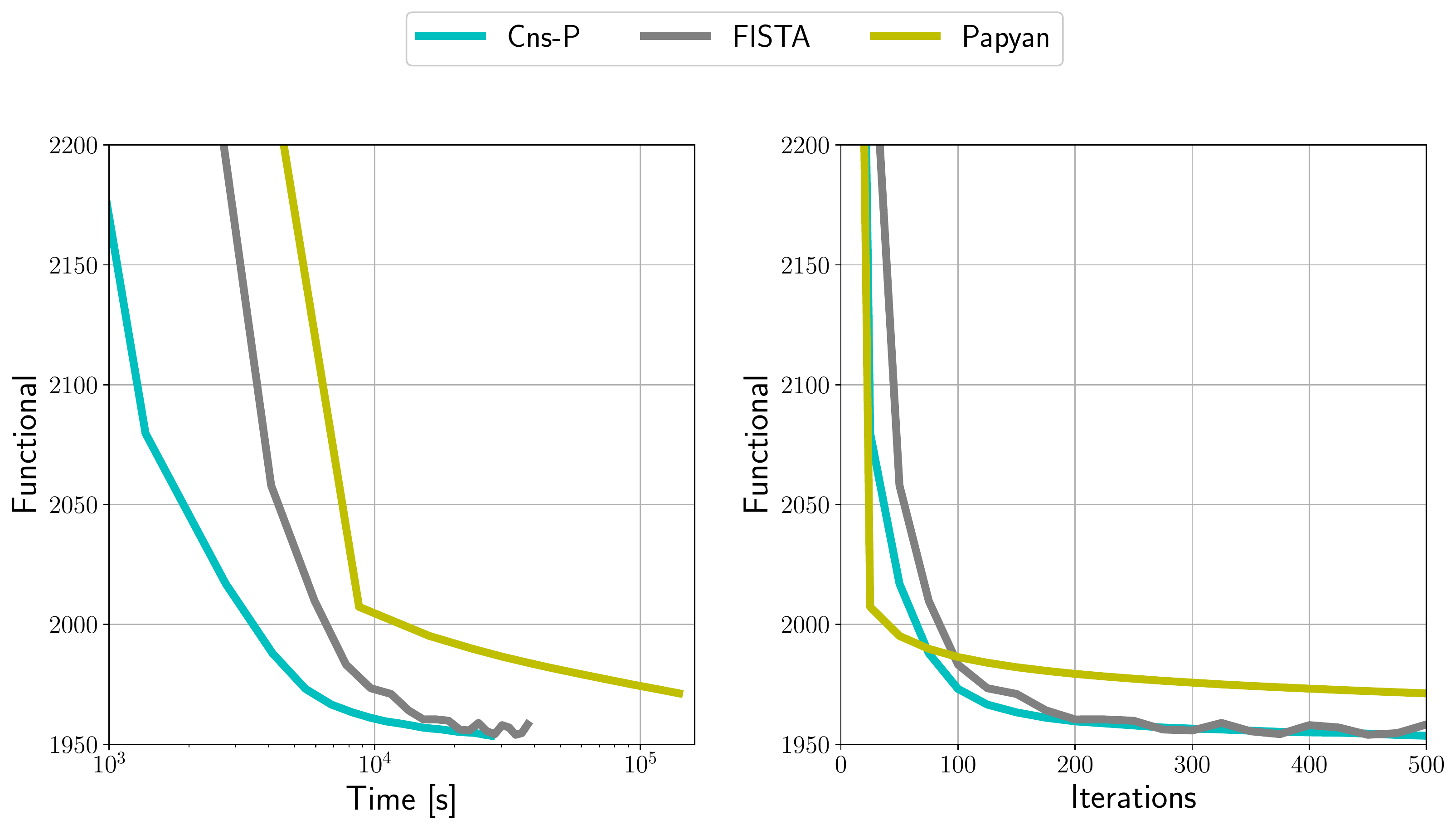}}
        \caption{Evolution of the CBPDN functional~\eq{cbpdnmmv} for the test set using the partial dictionaries obtained when training for $K=400$ images, 256 $\times$ 256 pixels, as in~\fig{fObj_k400}.}
        \label{fig:val_k400}
\end{figure}

Testing results obtained for the additional 20 images of size $256 \times 256$ are displayed in~\fig{val_k25}, for $K=25, 1024 \times 1024$ images, in~\fig{val_k100} for $K=100, 512 \times 512$ images and in~\fig{val_k400} for $K=400, 256 \times 256$ images. Note again that, as in the comparisons in the main document, the time axis in these plots refers to the run time of the dictionary learning code used to generate the relevant dictionary, and \emph{not} to the run time of the sparse coding on the test set.

All the testing plots show that the methods perform as expected from the training comparison, with Cns-P achieving better performance also in the test set, followed by FISTA. Results for the method of Papyan \etal are always worse, and do not match the functional values achieved either by Cns-P or FISTA. For all methods, testing results are better for the dictionary filters obtained when training with $K=400, 256 \times 256$ images (\fig{val_k400}), followed by the dictionary filters obtained when training with $K=100, 512 \times 512$ images (\fig{val_k100}), with the worst results obtained for the dictionary filters obtained when training with $K=25, 1024 \times 1024$ images (\fig{val_k25}). In particular, the Cns-P functional increases near the end of the evolution in~\fig{val_k25}. We believe that this is due to overfitting effects for the $K=100$ and $K=25$ cases, resulting from the mismatch between training and validation image sizes. Additional experiments (results not shown) confirmed that the functional decreases monotonically when the size of the images in the testing set corresponds to the size of the images in training set.  Nevertheless, we decided to use the same testing set for all of these experiments so that the corresponding functionals would be comparable across the different training sets.

\begin{figure}[htb]
\centerline{
        \subfigure[Without Spatial Mask]{\scalebox{0.24}{\includegraphics{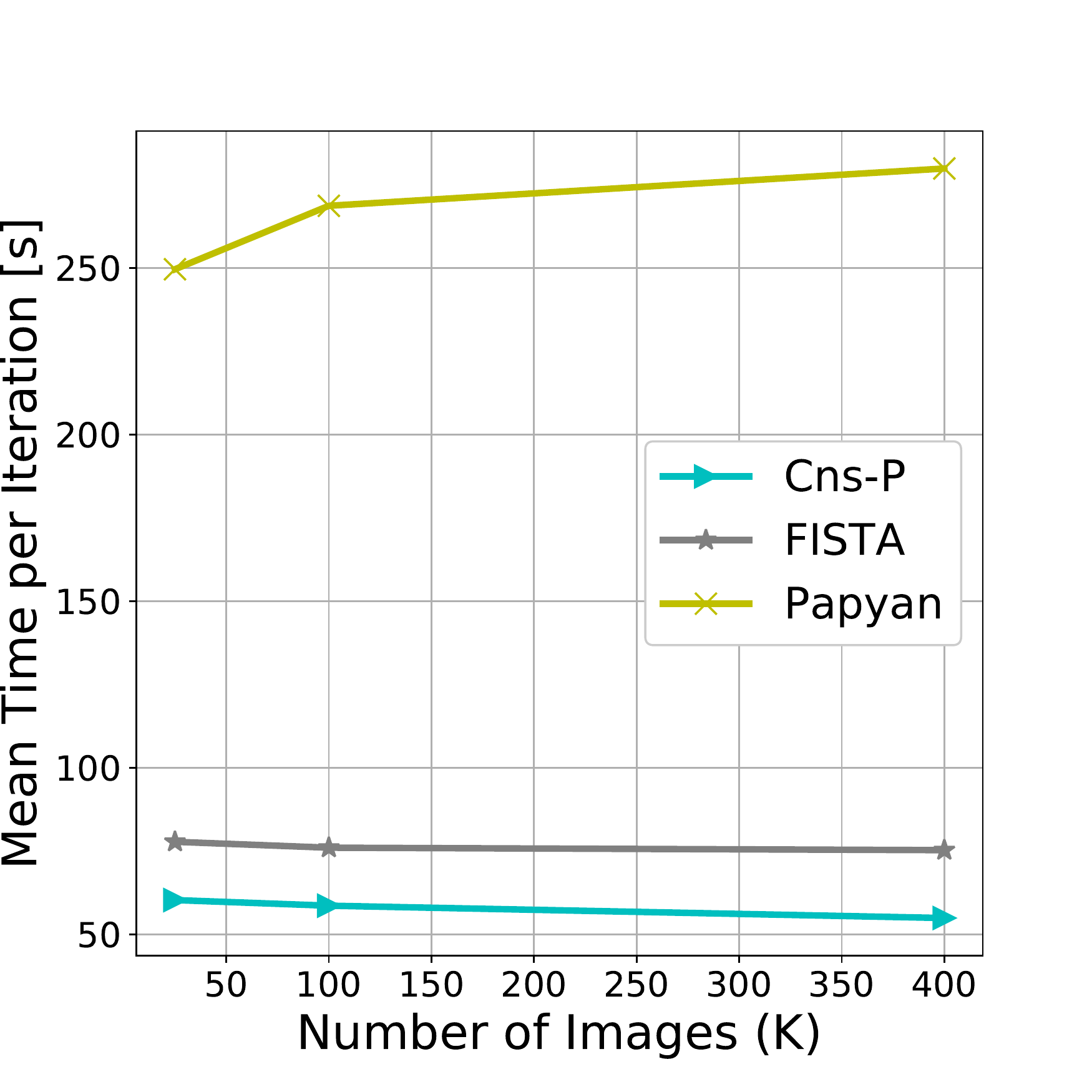}}
        \label{fig:per_it_scaling_unmasked}}
        \hfil
        \subfigure[With Spatial Mask]{\scalebox{0.24}{\includegraphics{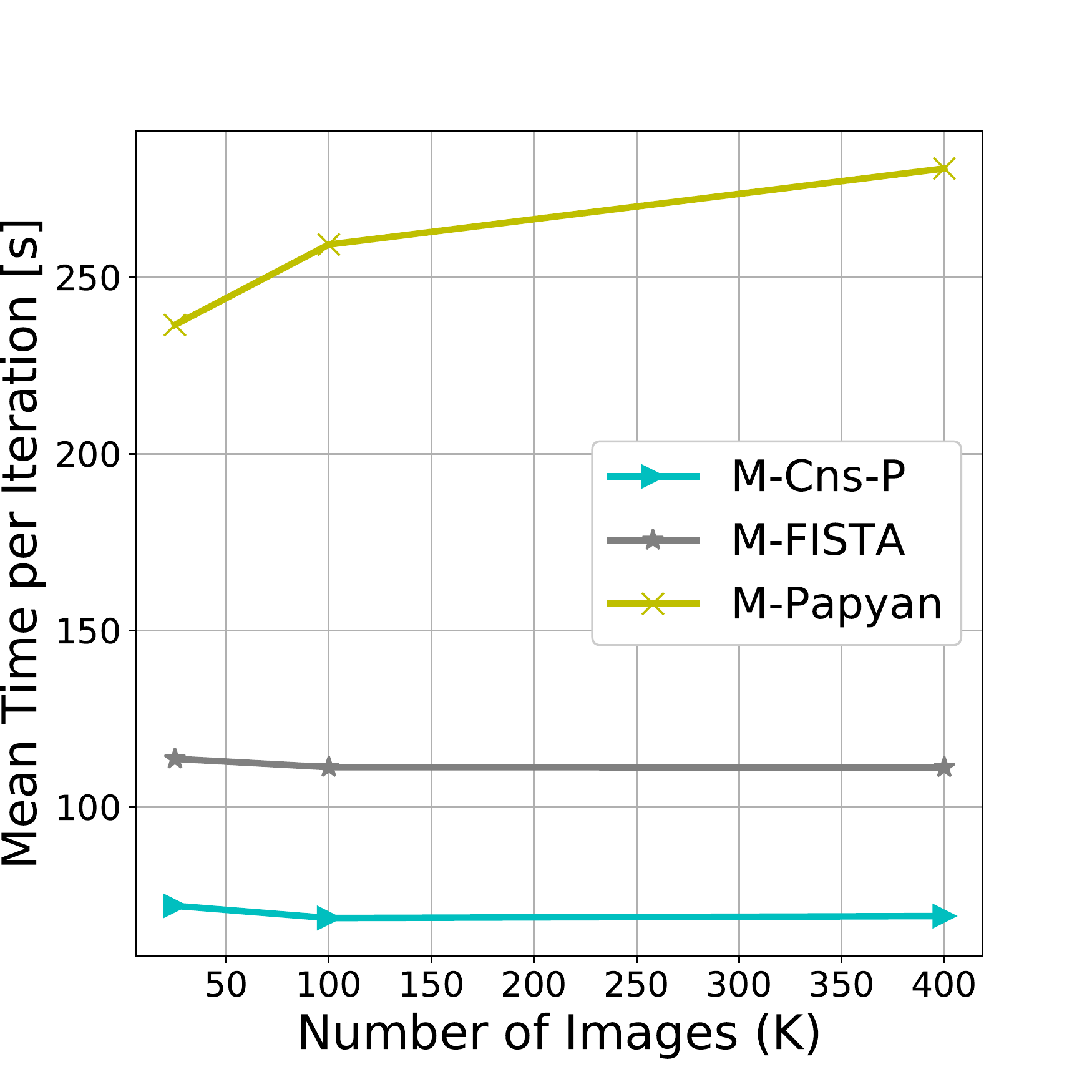}}
        \label{fig:per_it_scaling_mask}}
}
        \caption{Comparison of time per iteration for sets of 25, 100, and 400 images with size 1024 $\times$ 1024 pixels, 512 $\times$ 512 pixels and 256 $\times$ 256 pixels, respectively.}
        \label{fig:per_it_scaling}
\end{figure}

It can be see from~\fig{per_it_scaling_unmasked} that the time per iteration for both Cns-P and FISTA decreases very slowly with increasing $K$ and decreasing $N$, \ie it is roughly linear in $N K$, the number of pixels in the training image set. Since the results in~\fig{t_per_it_scaling} show that these algorithms scale linearly with $K$, this implies that the algorithms have approximately linear scaling with $N$ as well. The slight deviation from linearity can be attributed to the $N \log N$ complexity of the FFTs used in these algorithms (see the computational complexity analysis in~\tbl{complexity}
in the main document). The method of Papyan \etal seems to be more sensitive to the scaling in $K$, with time per iteration increasing as $K$ increases (which is not evident from the complexity analysis, see~\tbl{complexitySup} below), and requires more time per iteration than Cns-P or FISTA.

\subsection{CDL with Spatial Mask}

\begin{figure}[htb]
        \center
        \scalebox{0.26}{\includegraphics{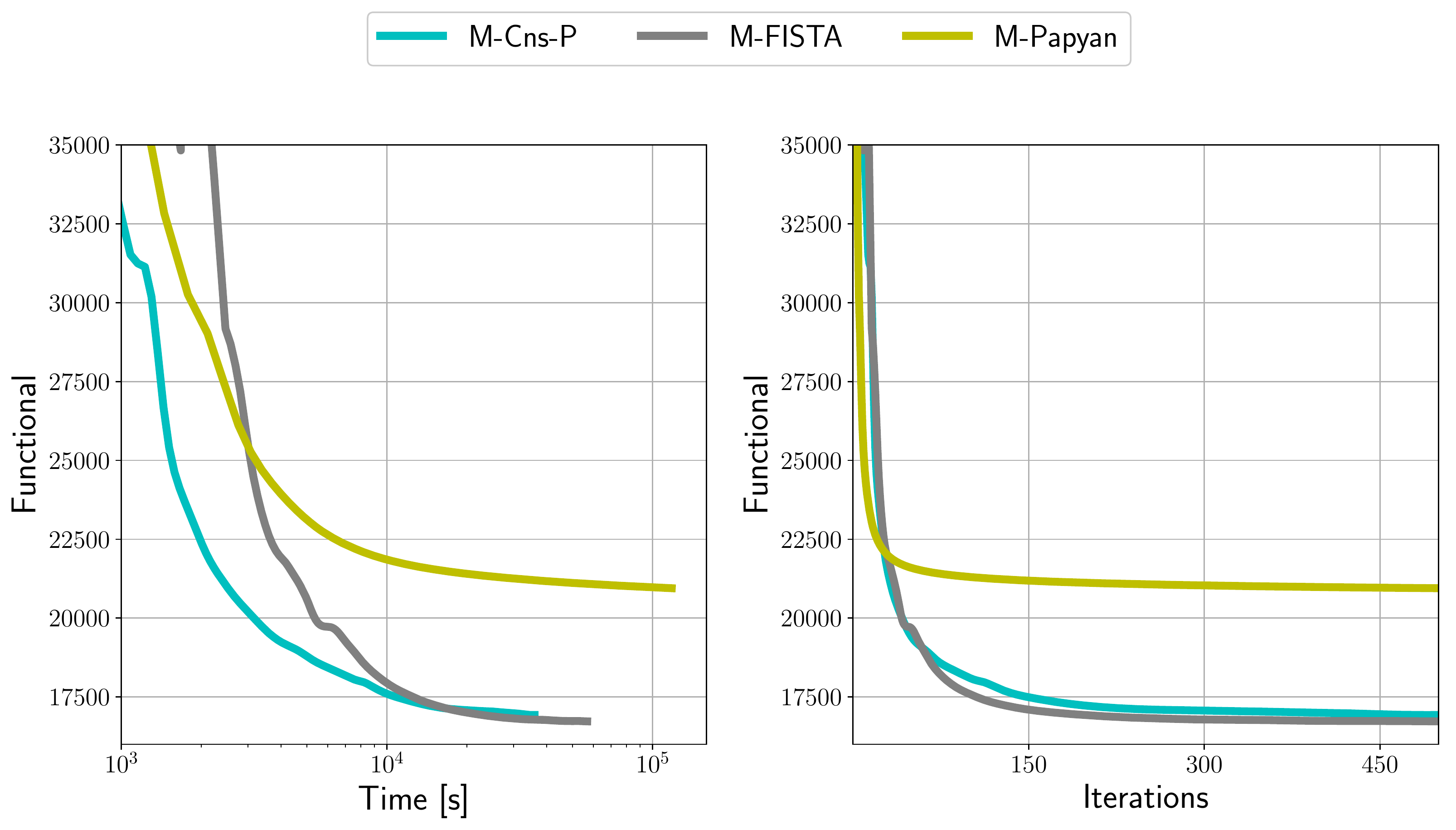}}
        \caption{Dictionary Learning with Spatial Mask ($K=25$): A comparison on a set of $K=25$ images, 1024 $\times$ 1024 pixels, of the decay of the value of the masked CBPDN functional~\eq{cbpdnmsk} with respect to run time and iterations for masked versions of the algorithms.}
        \label{fig:fObj_k25_mask}
\end{figure}

\begin{figure}[htb]
        \center
        \scalebox{0.26}{\includegraphics{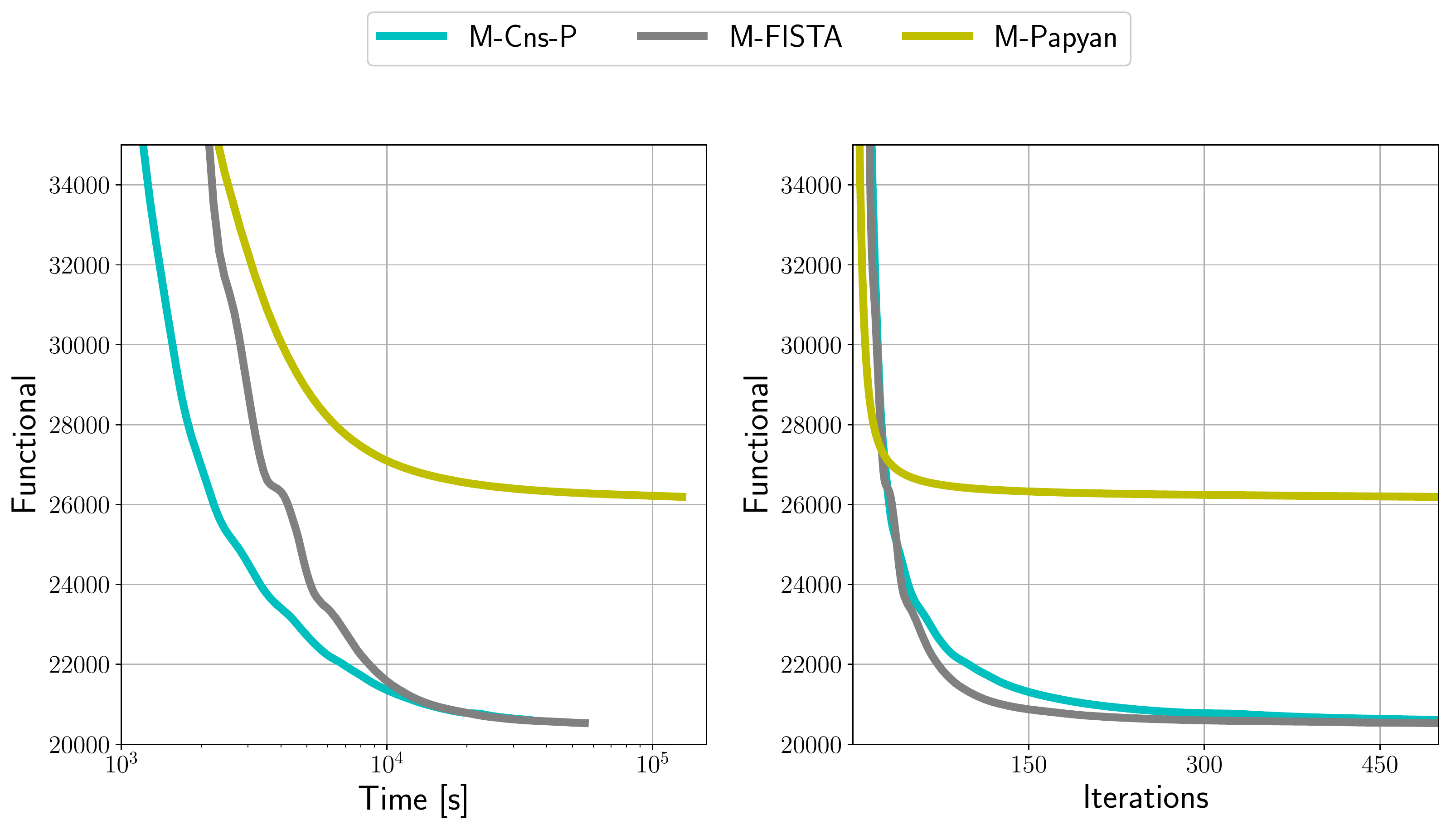}}
        \caption{Dictionary Learning with Spatial Mask ($K=100$): A comparison on a set of $K=100$ images, 512 $\times$ 512 pixels, of the decay of the value of the masked CBPDN functional~\eq{cbpdnmsk} with respect to run time and iterations for masked versions of the algorithms.}
        \label{fig:fObj_k100_mask}
\end{figure}

\begin{figure}[htb]
        \center
        \scalebox{0.26}{\includegraphics{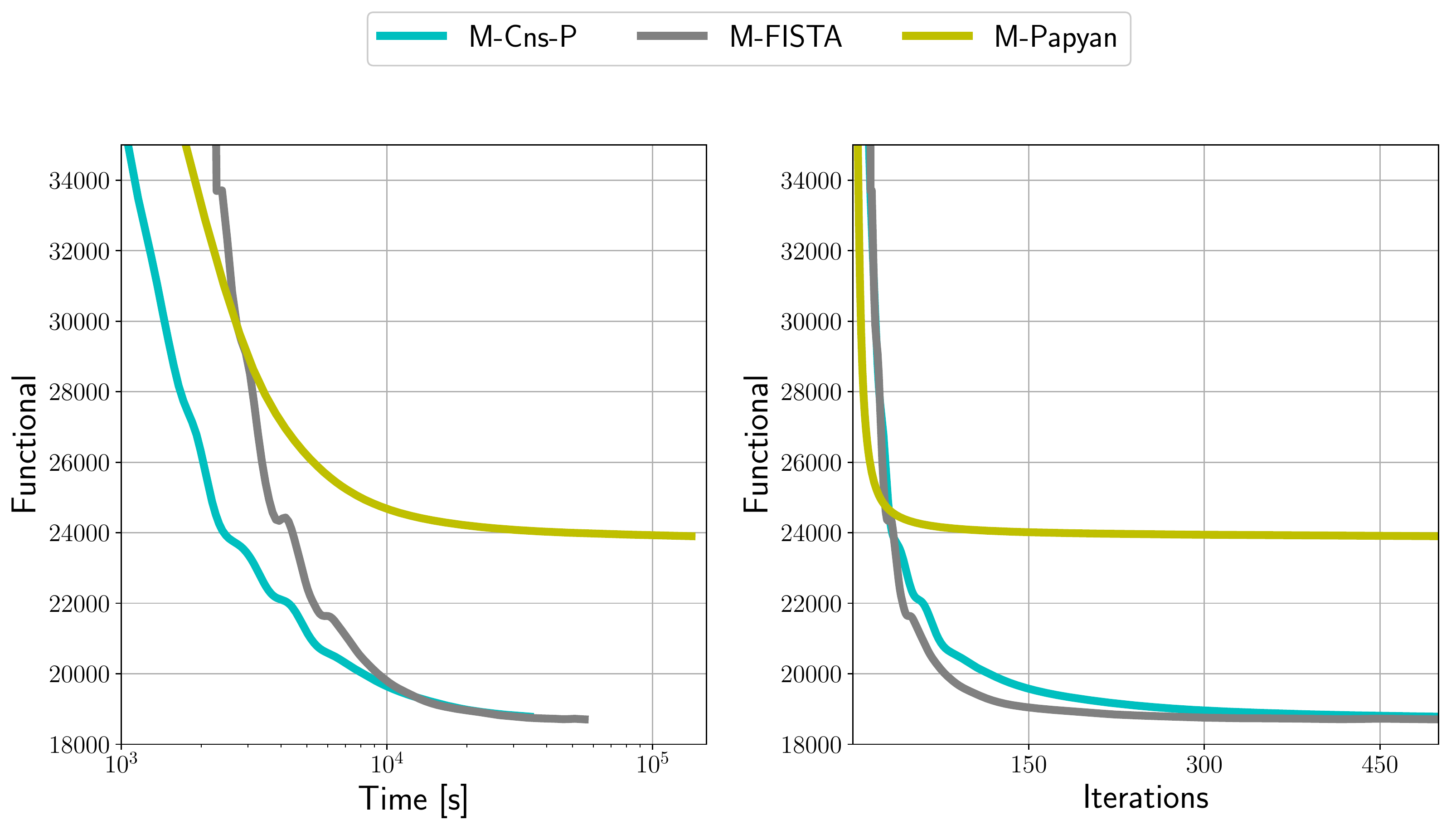}}
        \caption{Dictionary Learning with Spatial Mask ($K=400$): A comparison on a set of $K=400$ images, 256 $\times$ 256 pixels, of the decay of the value of the masked CBPDN functional~\eq{cbpdnmsk} with respect to run time and iterations for masked versions of the algorithms.}
        \label{fig:fObj_k400_mask}
\end{figure}

Comparisons for CDL with a spatial mask were performed with a random mask with values in $\{0,1\}$, with 25\% zero entries with a uniform random distribution. Three different random masks were generated, one for the set of images of 1024 $\times$ 1024 pixels, one for the set of 512 $\times$ 512 pixels, and one for the set of 256 $\times$ 256 pixels. All the methods used the same randomly generated masks. The corresponding results are shown in~\fig{fObj_k25_mask} for $K=25, 1024 \times 1024$ images, in~\fig{fObj_k100_mask} for $K=100, 512 \times 512$ images and in~\fig{fObj_k400_mask} for $K=400, 256 \times 256$ images. These resemble the results obtained for the unmasked variants, with M-Cns-P yielding the fastest convergence and smallest final masked CBPDN functional values, followed by M-FISTA. M-FISTA is still initially unstable in some cases, but its convergence becomes much smoother than the unmasked variant by the end of the learning. Since both M-Cns-P and M-FISTA converge to a similar functional value in learning, it is difficult to see the differences in computation time in the plots, but M-Cns-P is almost $2/3$ faster than M-FISTA. The functional values for the masked method of Papyan \etal~\cite{papayan-2017-convolutional} are inaccurate since the mask is not taken into account in the calculation.

\begin{figure}[htb]
        \center
        \scalebox{0.26}{\includegraphics{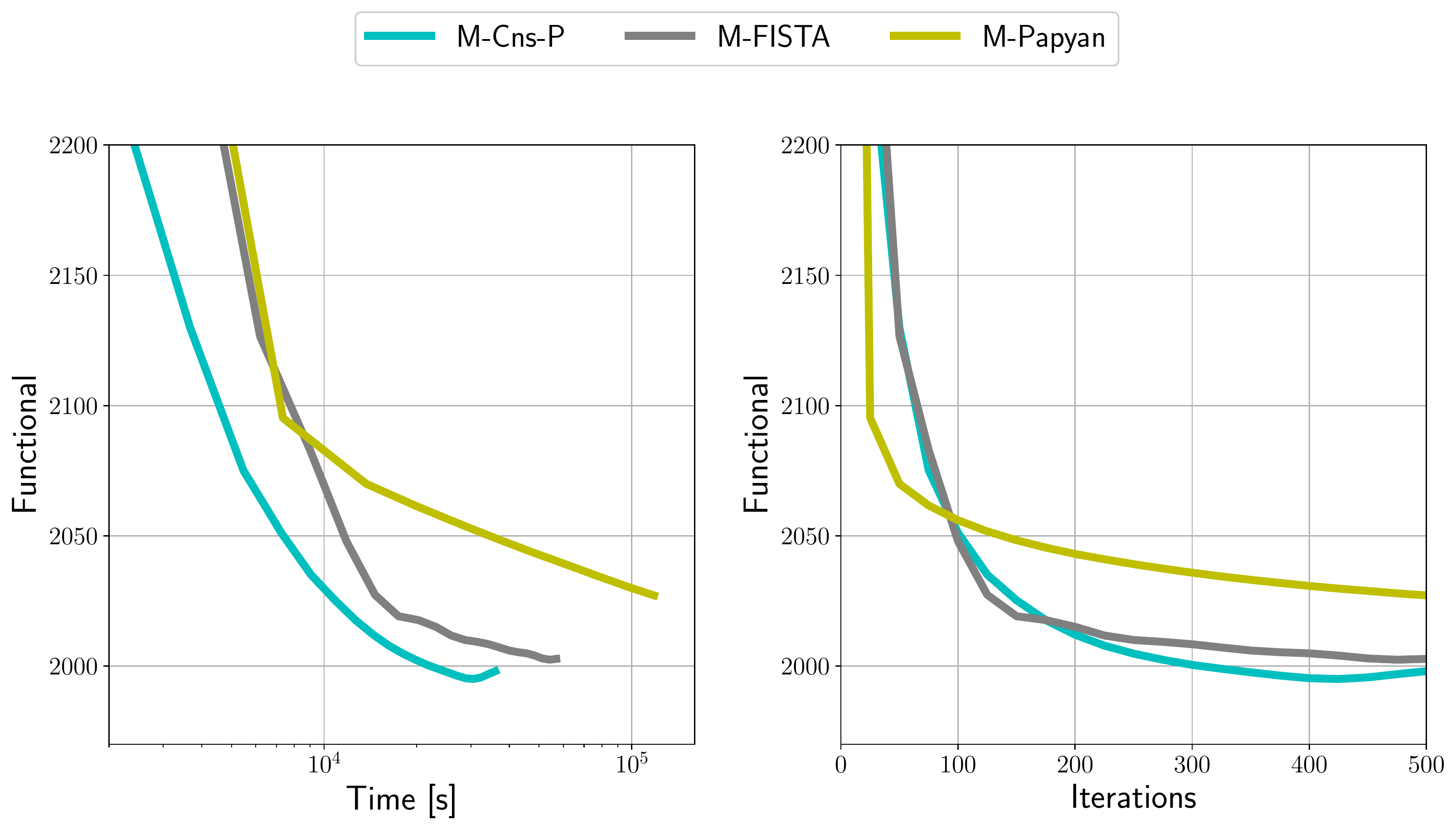}}
        \caption{Evolution of the CBPDN functional~\eq{cbpdnmmv} for the test set using the partial dictionaries obtained when training for $K=25$ images, 1024 $\times$ 1024 pixels, for masked versions of the algorithms, as in~\fig{fObj_k25_mask}.}
        \label{fig:val_k25_mask}
\end{figure}

\begin{figure}[htb]
        \center
        \scalebox{0.26}{\includegraphics{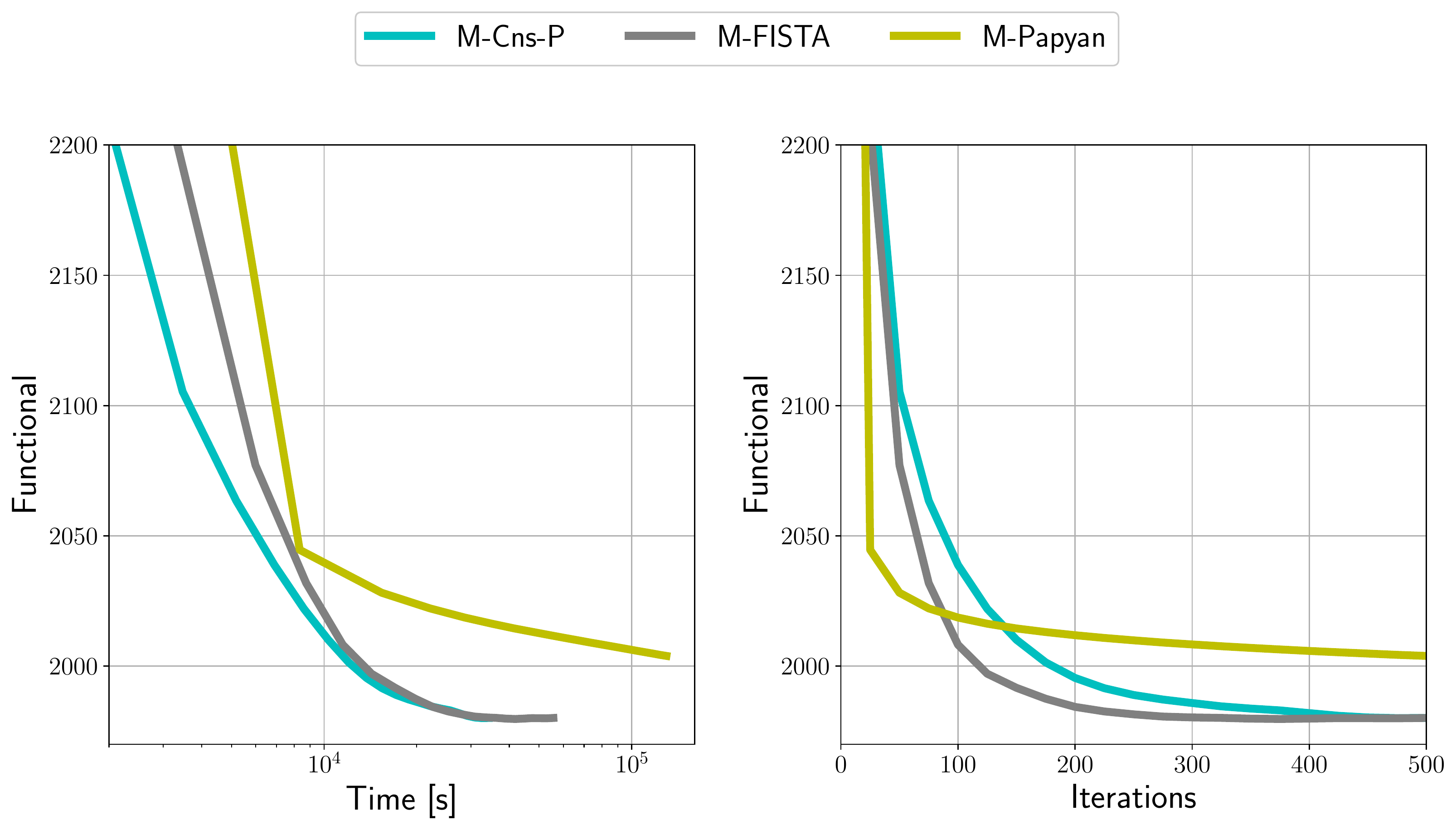}}
        \caption{Evolution of the CBPDN functional~\eq{cbpdnmmv} for the test set using the partial dictionaries obtained when training for $K=100$ images, 512 $\times$ 512 pixels, for masked versions of the algorithms, as in~\fig{fObj_k100_mask}.}
        \label{fig:val_k100_mask}
\end{figure}

\begin{figure}[htb]
        \center
        \scalebox{0.26}{\includegraphics{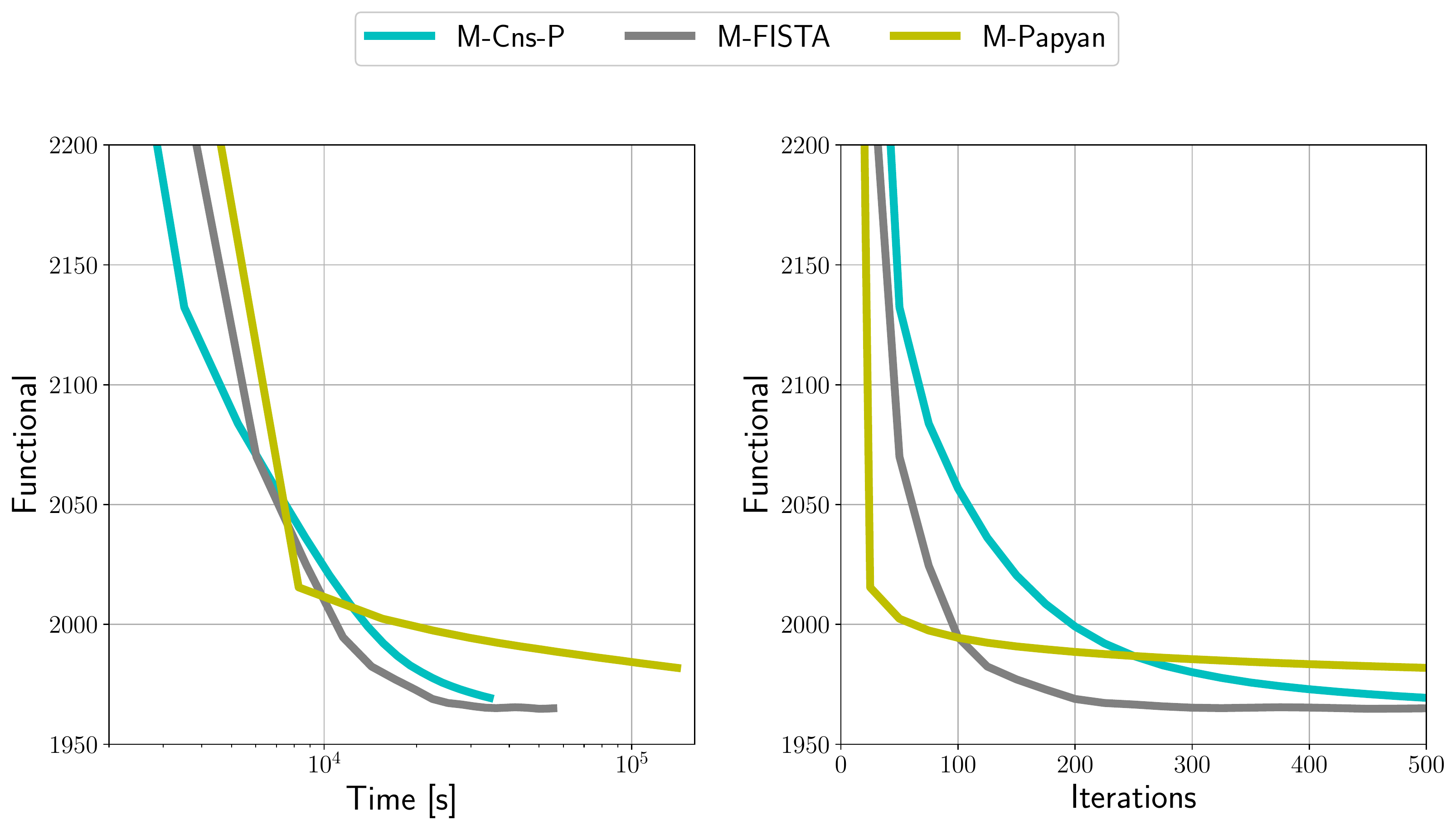}}
        \caption{Evolution of the CBPDN functional~\eq{cbpdnmmv} for the test set using the partial dictionaries obtained when training for $K=400$ images, 256 $\times$ 256 pixels, for masked versions of the algorithms, as in~\fig{fObj_k400_mask}.}
        \label{fig:val_k400_mask}
\end{figure}

A fair comparison can, however, be made by evaluating the CBPDN functional, \eq{cbpdnmmv}, when sparse coding the test set with the dictionary filters learned in training. The results are shown in~\fig{val_k25_mask}, for $K=25, 1024 \times 1024$ images, in~\fig{val_k100_mask} for $K=100, 512 \times 512$ images and in~\fig{val_k400_mask} for $K=400, 256 \times 256$ images. Again, note that testing results for the case of $K=400, 256 \times 256$ are better for all the methods, and that for our methods there are some overfitting effects for the $K=100$ and $K=25$ cases, although these are less significant than those for the unmasked ones. Also, it is clear that testing results for M-Cns-P and M-FISTA are much better than for the masked method of Papyan \etal~\cite{papayan-2017-convolutional}.

It can be seen from~\fig{per_it_scaling_mask} that M-Cns-P and M-FISTA exhibit similar behavior to the corresponding unmasked variants in that the time per iteration is almost constant when the product of $N$ and $K$ remains unchanged. The difference in the time per iteration between unmasked and masked variants is larger for M-FISTA than for M-Cns-P. Conversely, the time per iteration between unmasked and masked variants decreases for the method of Papyan \etal, for smaller $K$ and larger $N$, while it increases slightly for larger $K$ and smaller $N$. This behavior is not expected from the complexity analysis.

Finally, it is worth noting that, while we do not quantify the optimality of the parameters selected via the guidelines discussed in~\sctn{ppselect} of the main document, they do appear to provide good performance even for the substantially larger problems, considered here, than those used to develop these guidelines. In contrast, we found parameter selection to be problematic for the methods proposed by other authors discussed in~\sctn{otheralgcompare}.

\section{Scaling with Dictionary Size}
\label{sec:filtercompare}

\begin{figure}[htb]
        \center
 \scalebox{0.26}{\includegraphics{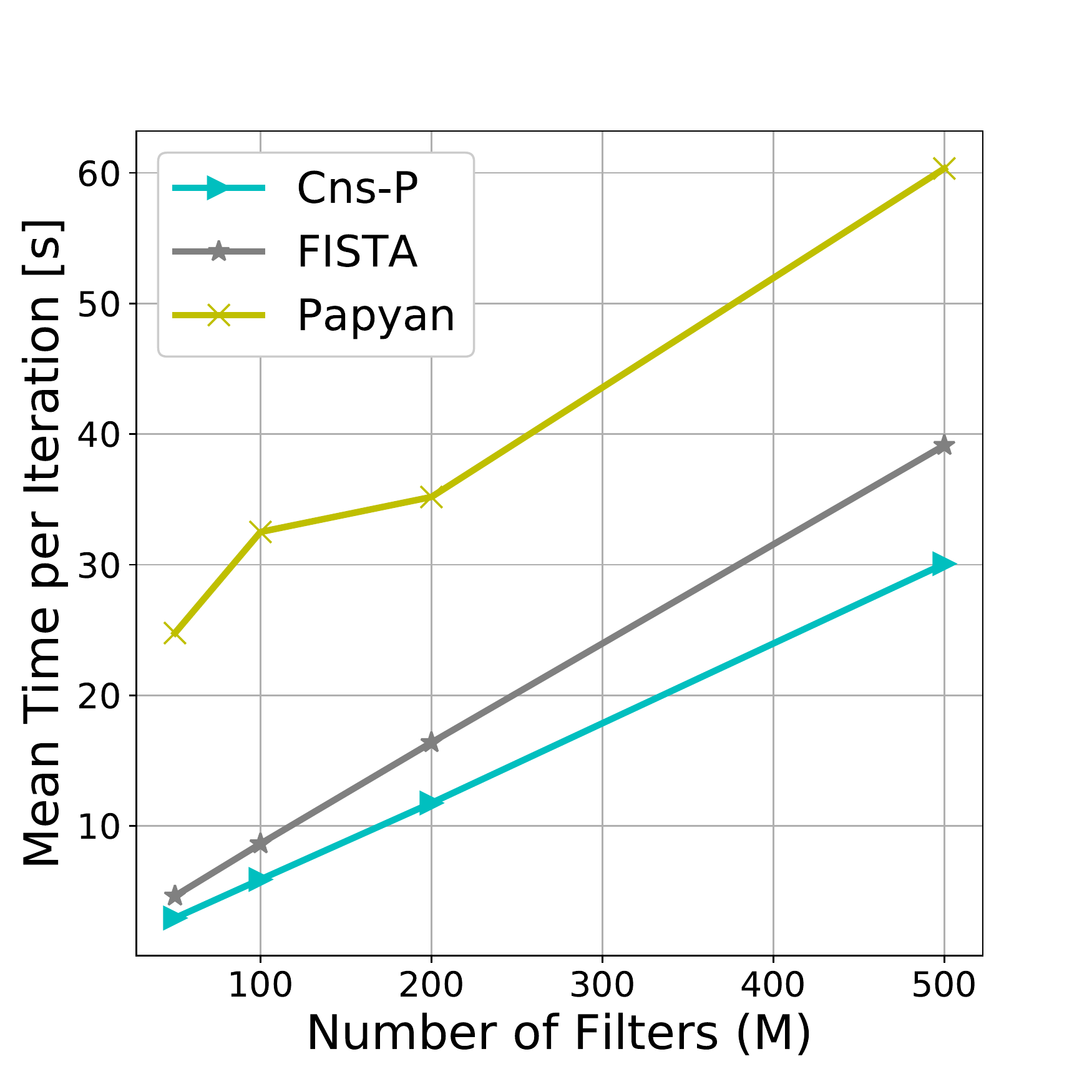}}
        \caption{Comparison of time per iteration for sets of $M \in \{50, 100, 200, 500\}$, with $11 \times 11$ dictionary filters and $K=40$ images of size $256 \times 256$ pixels.}
        \label{fig:filter_scaling}
\end{figure}

In this section we compare the scaling with respect to the number of filters, $M$, of our two leading methods (Cns-P and FISTA) and the method of Papyan \etal~\cite{papayan-2017-convolutional}. Dictionaries with $M \in \{50, 100, 200, 500\}$ filters of size 11 $\times$ 11 were learned, over 500 iterations, from the training set of $K=40$, 256 $\times$ 256 greyscale images described in the main document. The time per iteration for the three methods is compared in~\fig{filter_scaling}, which shows that all three methods exhibit linear scaling (modulo the outlier at $M=100$ for the method of Papyan \etal) with the number of filters.

These experiments do not address the issue of filter size. While the performance of the DFT-domain methods proposed here is roughly independent of the filter size, spatial domain methods such as that of Papyan \etal become more expensive as the filter size increases. In addition, multi-scale dictionaries are easily supported by the DFT-domain methods, but are much more difficult to support for spatial domain methods.

\section{Additional Algorithm Comparisons}
\label{sec:otheralgcompare}

We used the same training set as the previous section ($K=40$, 256 $\times$ 256 greyscale images) to compare the performance between our two leading methods (Cns-P and FISTA) and the competing methods proposed by Heide \etal~\cite{heide-2015-fast} and by Papyan \etal~\cite{papayan-2017-convolutional}, and the consensus method proposed by \v{S}orel and \v{S}roubek~\cite{sorel-2016-fast}. Our methods are implemented in Python, those of Heide \etal~\cite{heide-2015-fast}, and of \v{S}orel and \v{S}roubek~\cite{sorel-2016-fast} are implemented in Matlab, and that of Papyan \etal~\cite{papayan-2017-convolutional} is implemented in Matlab and C.

We compared the performance of the methods in learning a dictionary of 100 filters of size 11 $\times$ 11, setting the sparsity parameter $\lambda = 0.1$. We set the parameters for our methods according to the scaling rules discussed in~\sctn{parameter_sensitivity} in the main document, using fixed penalty parameters $\rho$ and $\sigma$ without any adaptation methods. Relaxation methods~\cite[Sec. 3.4.3]{boyd-2010-distributed}\cite[Sec. III.D]{wohlberg-2016-efficient} were used, with $\alpha = 1.8$. The parameters for the competing methods were set from the default parameters included in their respective demonstration scripts.  As before, the additional set of 20 images of size 256 $\times$ 256 pixels was used as a test set to evaluate the dictionaries learned. Again, we report the evolution of the CBPDN functional~\eq{cbpdnmmv} for the test set to provide a meaningful comparison, independent of the training functional evaluation implemented by each method, which use slightly different expressions, sometimes calculated with un-normalized dictionaries.

\begin{figure}[htb]
        \center
        \scalebox{0.26}{\includegraphics{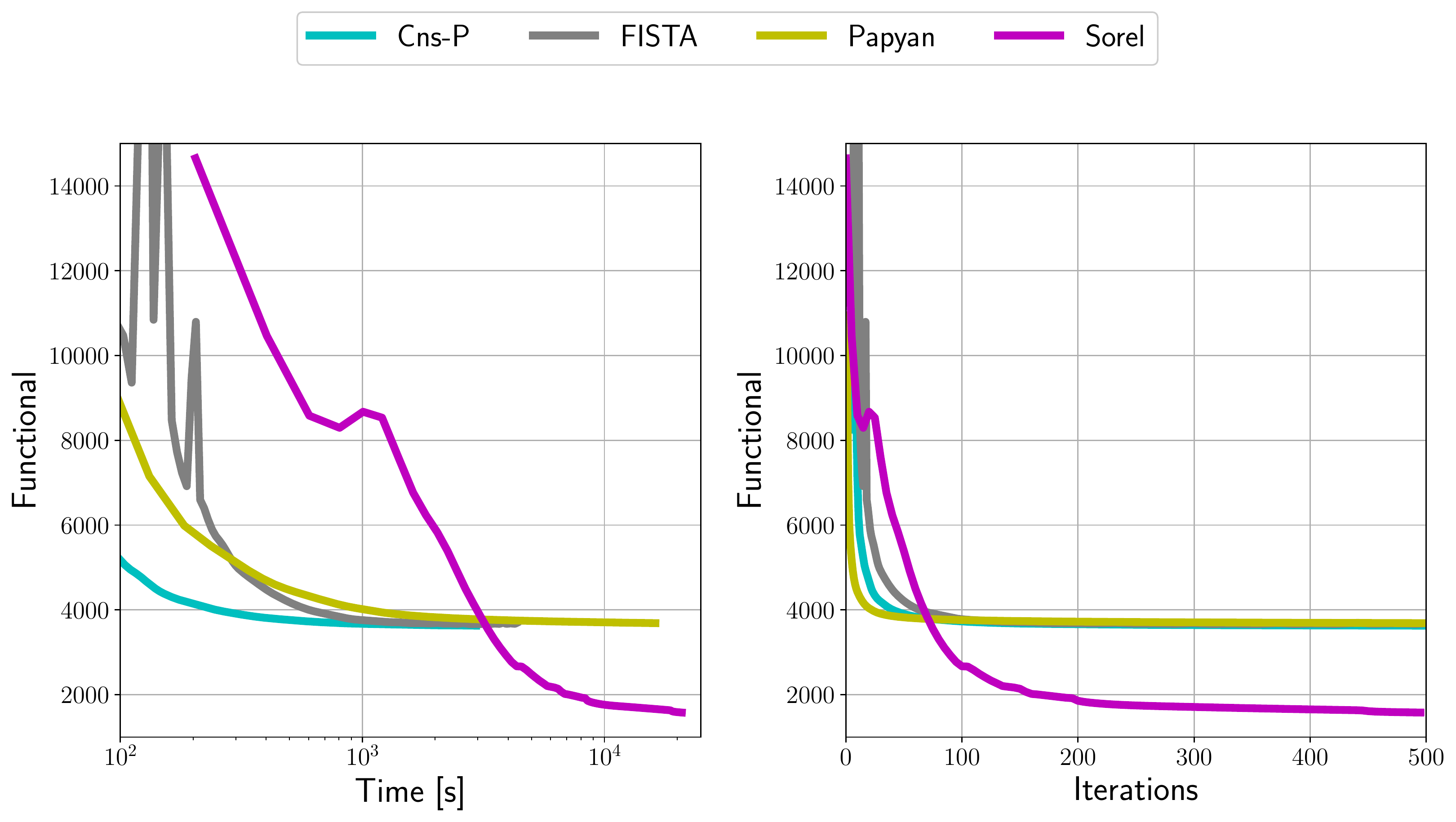}}
        \caption{Dictionary Learning ($K=40$): A comparison on a set of $K=40$ images, 256 $\times$ 256 pixels, of the decay of the functional value in training with respect to run time and iterations for Cns-P, FISTA, the method of Papyan \etal, and the consensus method of \v{S}orel and \v{S}roubek.}
        \label{fig:fObj_k40_extra_sorel}
\end{figure}

\begin{figure}[htb]
        \center
        \scalebox{0.26}{\includegraphics{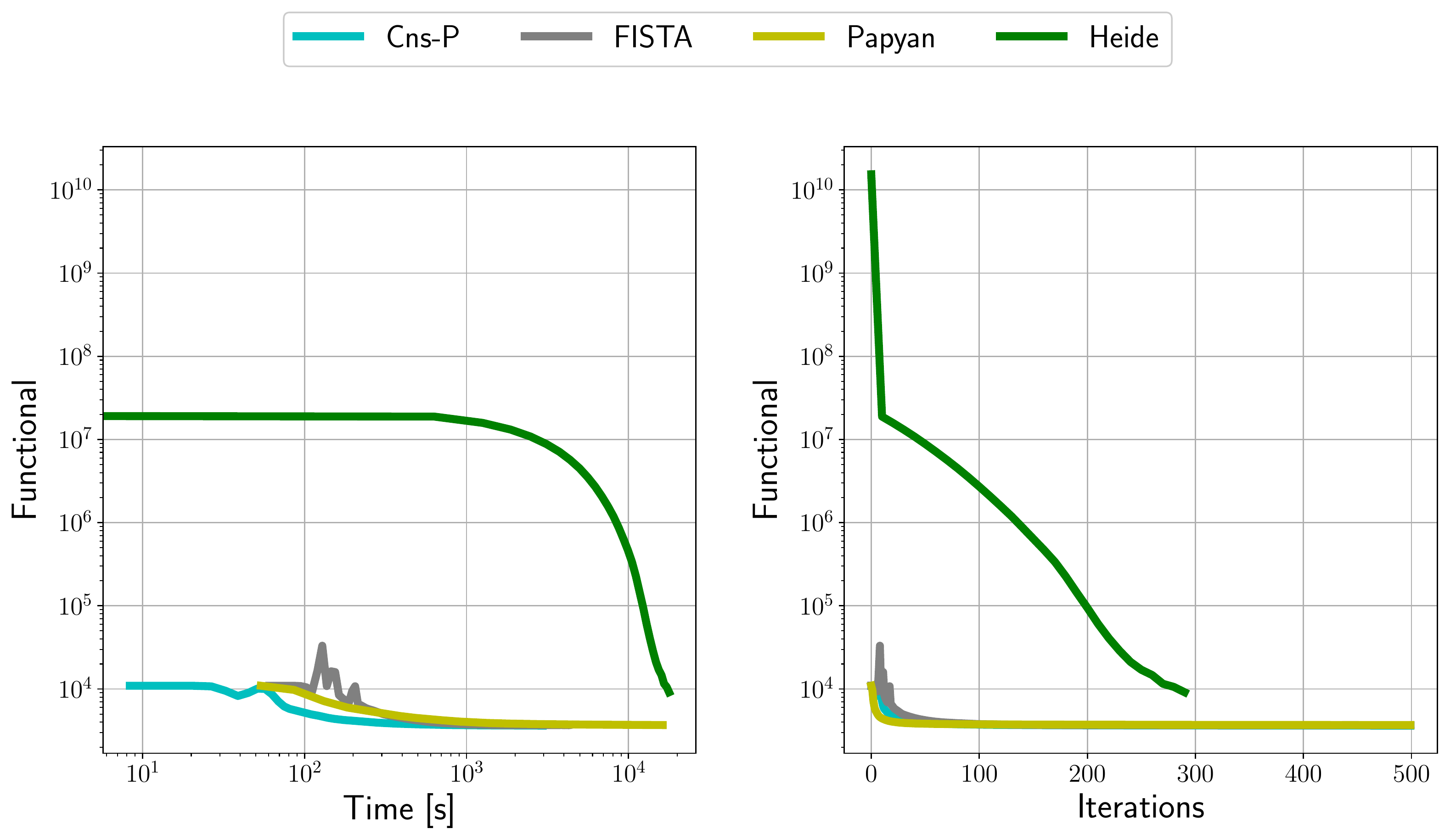}}
        \caption{Dictionary Learning ($K=40$): A comparison on a set of $K=40$ images, 256 $\times$ 256 pixels, of the decay of the functional value in training with respect to run time and iterations for Cns-P, FISTA, the method of Papyan \etal, and the method of Heide \etal.}
        \label{fig:fObj_k40_extra_heide}
\end{figure}

\begin{figure}[htb]
\begin{tabular}{cc}
\hspace{-3mm}\subfigure[Cns-P]{\includegraphics[trim={3mm 3mm 0mm 3mm},clip,width=5cm,]{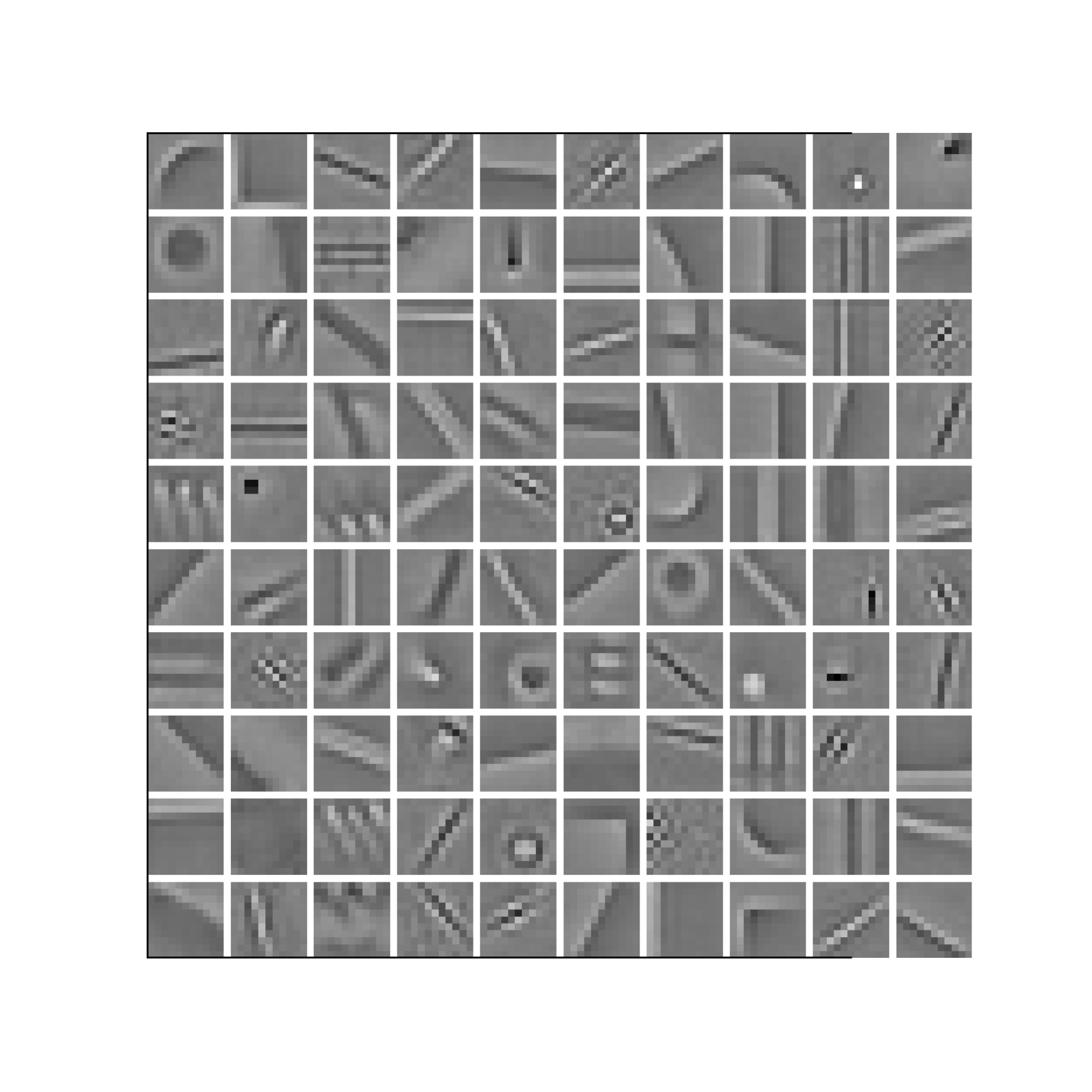}} &
\hspace{-10mm}\subfigure[FISTA]{\includegraphics[width=5cm,trim={3mm 3mm 0mm 3mm},clip]{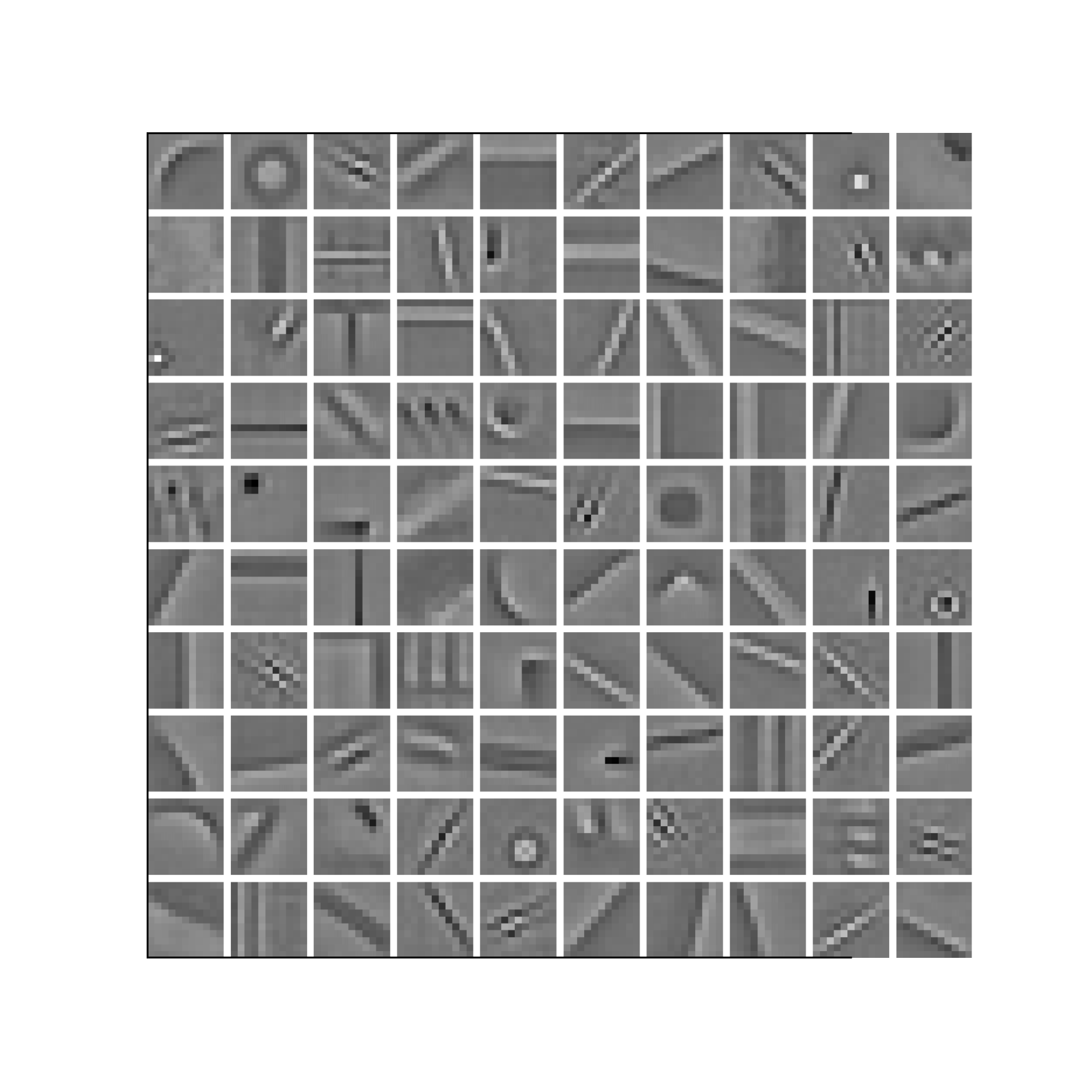}}
\\[-3.3mm]
\hspace{-3mm}\subfigure[Papyan \etal~\cite{papayan-2017-convolutional}]{\includegraphics[width=5cm,trim={3mm 3mm 0mm 3mm},clip]{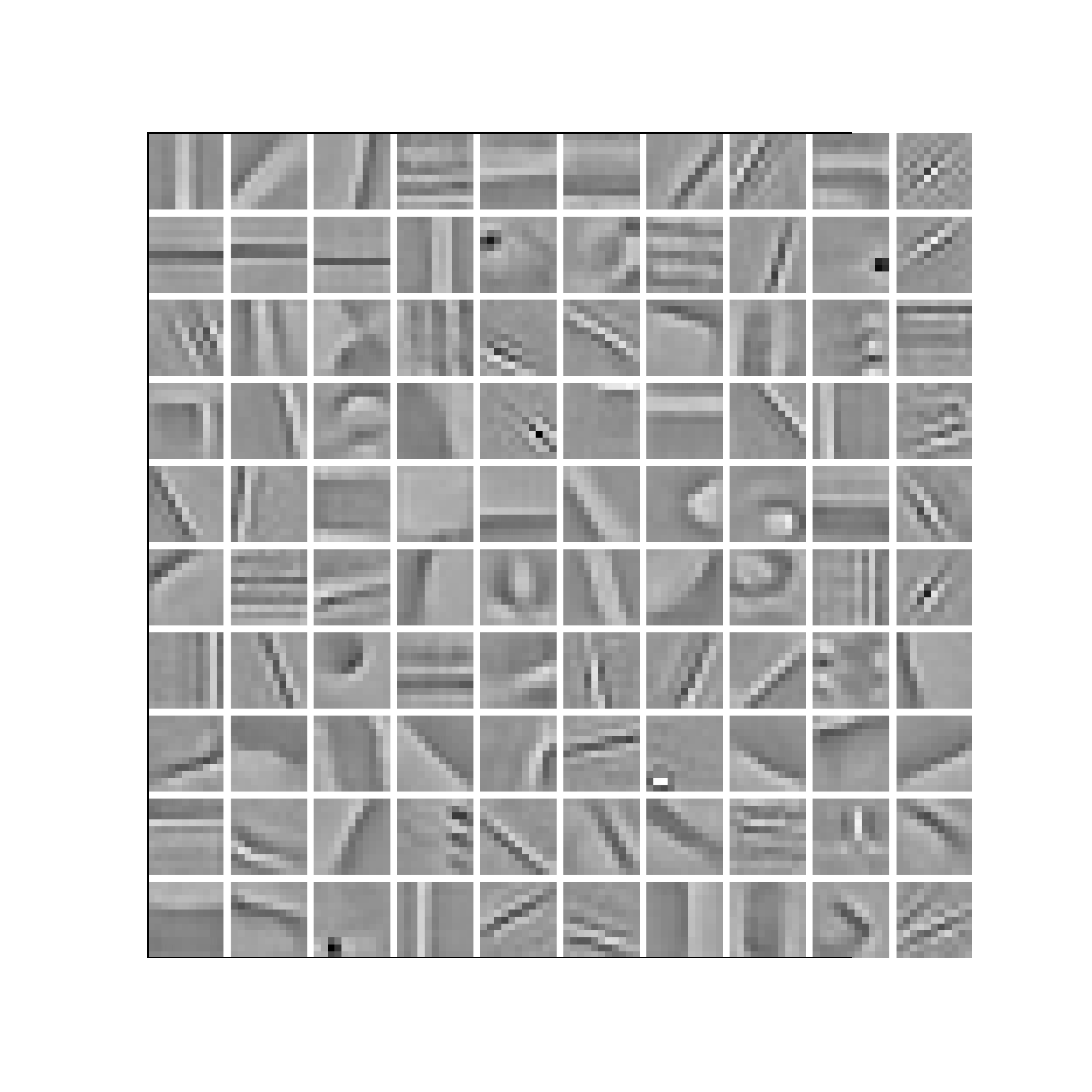}} &
\hspace{-10mm}\subfigure[Heide \etal~\cite{heide-2015-fast}]{\includegraphics[width=5cm,trim={3mm 3mm 0mm 3mm},clip]{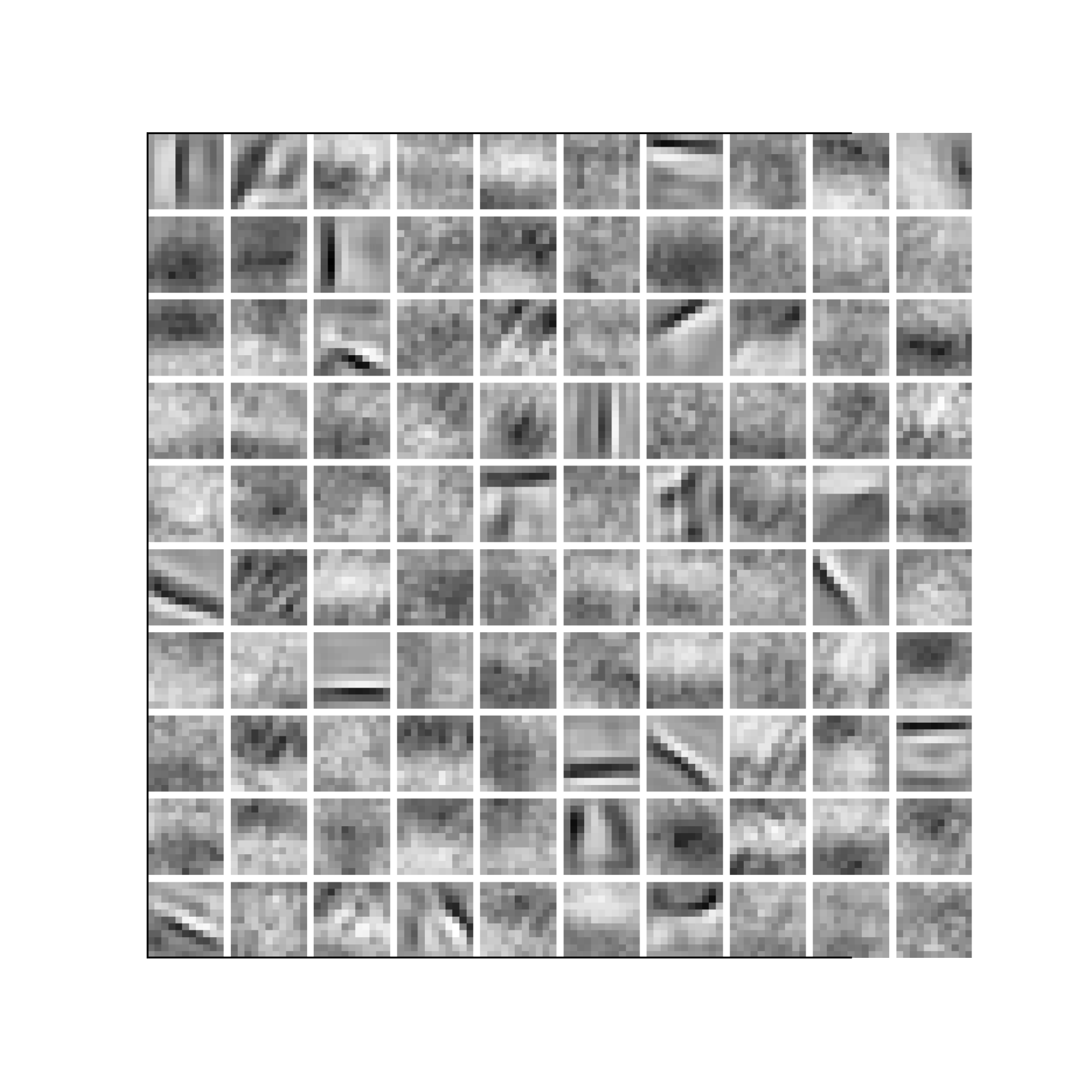}} \\[-3.3mm]
\hspace{-3mm}\subfigure[\v{S}orel and \v{S}roubek~\cite{sorel-2016-fast}]{\includegraphics[width=5cm,trim={3mm 0mm 3mm 3mm},clip]{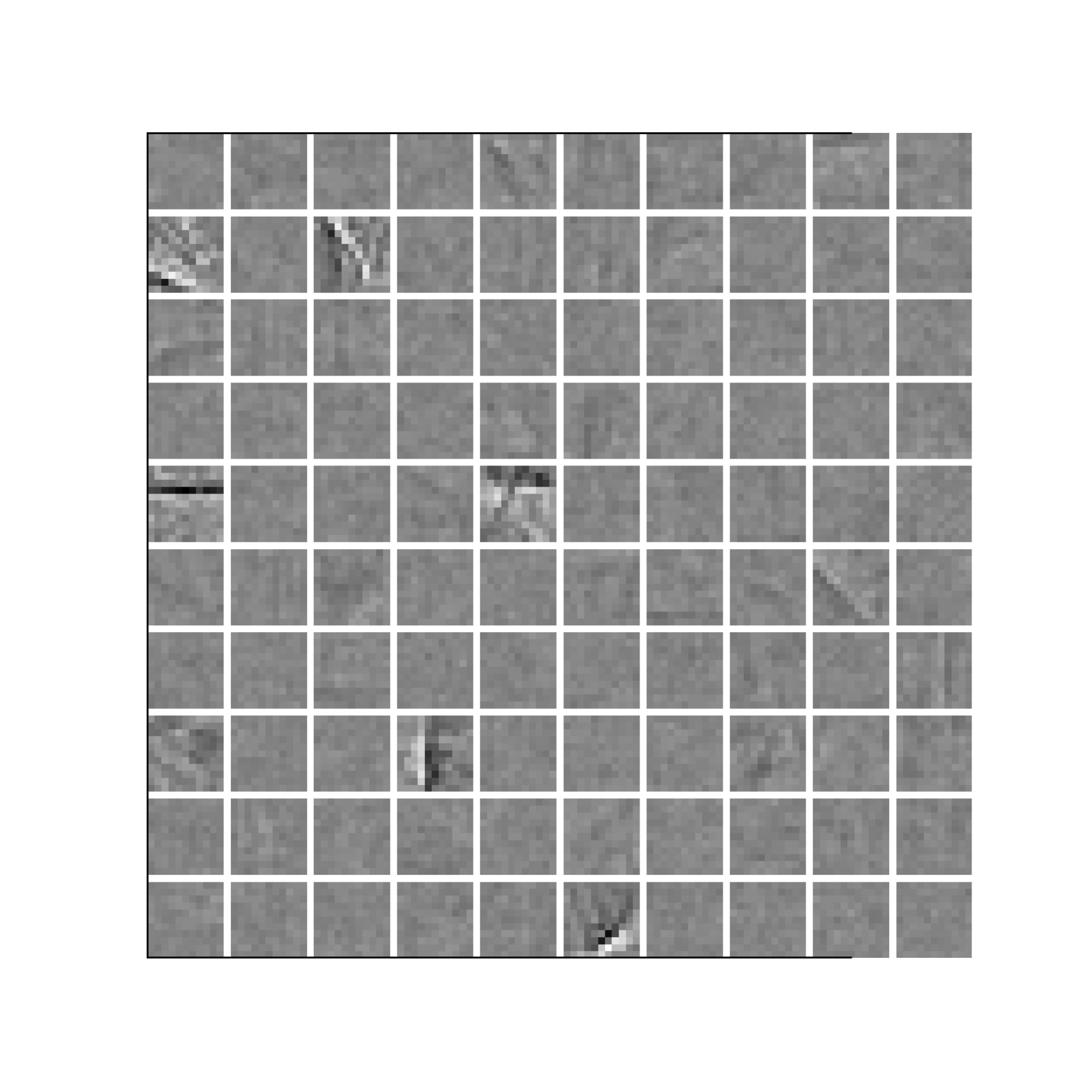}} &
\end{tabular}
        \caption{Dictionaries obtained for training with $K=40$ images, 256 $\times$ 256 pixels. These are the direct outputs: Cns-P, FISTA and the implementation of the method of Papyan \etal produce dictionaries normalized to 1; the implementation of the consensus method of \v{S}orel and \v{S}roubek produces dictionaries with most norms greater than 1; and the implementation of the method of Heide \etal produces dictionaries with most norms smaller than 1.}
        \label{fig:filters_k40}
\end{figure}

Comparisons for training are shown in~\figs{fObj_k40_extra_sorel} and~\fign{fObj_k40_extra_heide}. Performance is comparable for Cns-P, FISTA and the method of Papyan \etal, with FISTA initially exhibiting oscillatory behavior. Since the methods of \v{S}orel and \v{S}roubek, and of Heide \etal perform multiple inner iterations\footnote{Set to 10 and 5 inner iterations in the demonstration scripts provided by Heide \etal, and \v{S}orel and \v{S}roubek respectively.}  of the sparse coding and dictionary learning subproblems for each outer iteration, the iteration counts for these methods are reported as the product of inner and outer iterations. The method of Heide \etal starts with a very large functional value and is slow to converge\footnote{We were unable to coerce this code to run for a full 500 iterations (50 outer iterations with 10 inner iterations) by any adjustment of stopping conditions and tolerances.}.  The consensus method of \v{S}orel and \v{S}roubek appears to achieve significantly lower functional values than the other methods, but these results are not comparable since their dictionary filters are not properly normalized.  The final dictionaries computed are displayed in~\fig{filters_k40}.

\begin{figure}[htb]
        \center
        \scalebox{0.26}{\includegraphics{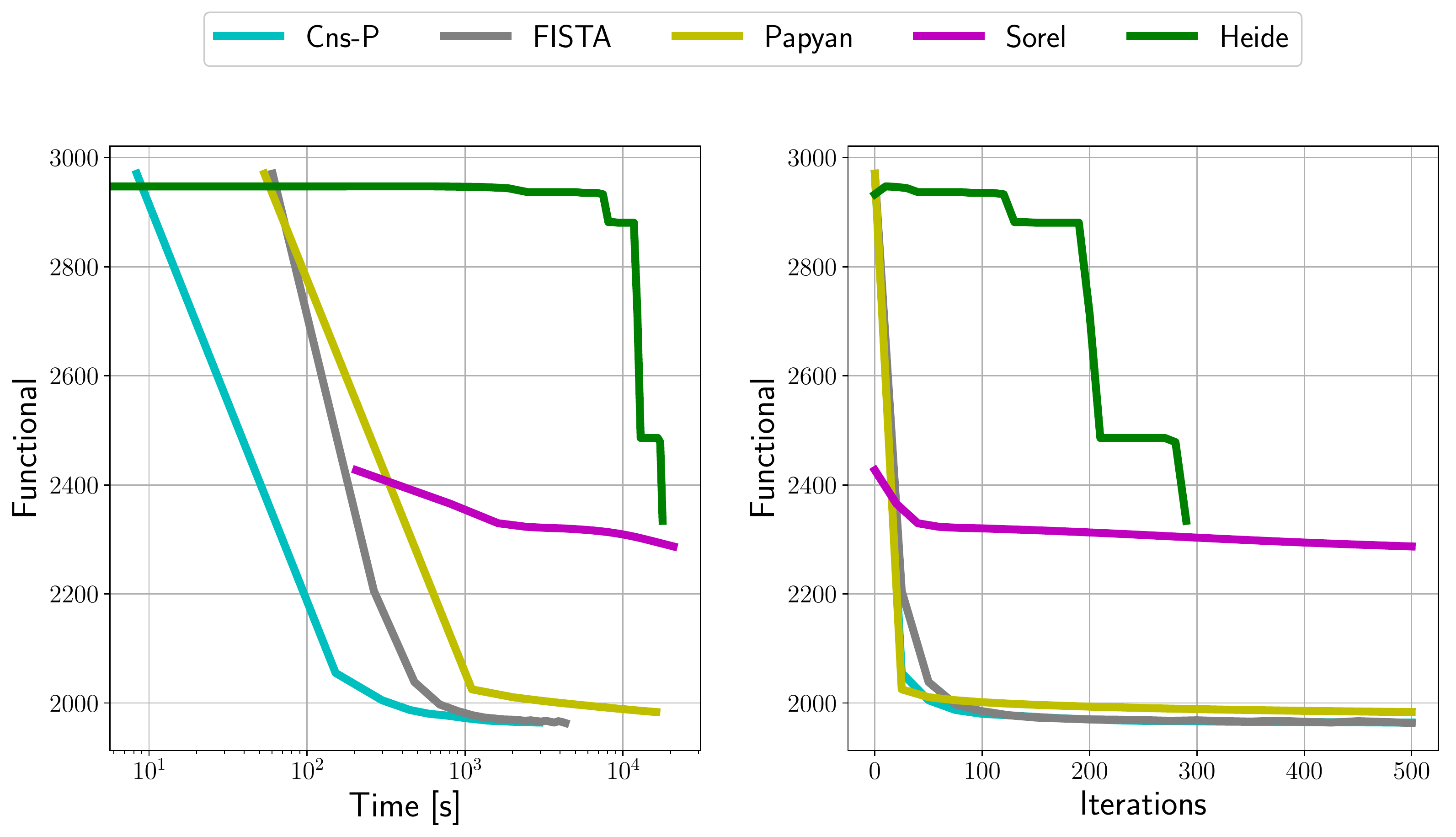}}
        \caption{Evolution of the CBPDN functional~\eq{cbpdnmmv} for the test set using the partial dictionaries obtained when training for $K=40$ images, 256 $\times$ 256 pixels.}
        \label{fig:val_k40_extra}
\end{figure}

Sparse coding results on the test set are shown in~\fig{val_k40_extra}. Note that Cns-P and FISTA produce the smallest CBPDN functional values, followed by the method of Papyan \etal, while results for the methods of \v{S}orel and \v{S}roubek as well as Heide \etal are much worse. Since the functional value evolution for the method of Heide \etal is highly oscillatory, at each iteration we plot the best functional value obtained up until that point instead of the functional value for that iteration. In terms of time evolution, it is clear that Cns-P is the fastest to converge, followed by FISTA and the method of Papyan \etal. The methods of \v{S}orel and \v{S}roubek and of Heide \etal are slow even for this relatively small dataset.

\begin{table}[ht]
\renewcommand{\arraystretch}{1.15}
\caption{Computational complexities per iteration of CDL algorithms. The number of pixels in the training images, the number of dictionary filters, and the number of training images are denoted by $N$, $M$, and $K$ respectively. Additionally, $n$ represents the local filter support, $\alpha$ the maximum number of non-zeros in a needle~\cite{papayan-2017-convolutional} and $P$ the number of internal ADMM iterations.}
\label{tab:complexitySup}
\centering
\setlength{\tabcolsep}{3.5pt}
\begin{tabular}{| l | l |}
\hline
\multicolumn{1}{|c|}{ {\bf Algorithm} } & \multicolumn{1}{|c|}{ {\bf Complexity} } \\
\hline
\hline
{\bf Cns-P}, {\bf FISTA} & $\co(K M N \log N + K M N)$  + \\
& $\co(KMN \log N + KMN + M N)$ \\
\hline
{\bf Papyan \etal~\cite{papayan-2017-convolutional}} & $\co(K M N n  + K N (\alpha^3 + M \alpha^2) + n M^2)$ + \\
  & $\co(KN n \alpha + KNM \alpha + nM^2)$ \\
\hline
{\bf \v{S}orel and \v{S}roubek~\cite{sorel-2016-fast}} & $\co(P K M N \log N + P K M N)$ + \\
{ADMM consensus} & $\co(P K M N \log N + P K M N)$ \\
\hline
{\bf Heide \etal}~\cite{heide-2015-fast} & $\co(M K^2 N + (P - 1) M K N)$ + \\
 $M > K$ & $\co(P K M N \log N) + \co(P K M N)$ \\
\hline
 {\bf Heide \etal}~\cite{heide-2015-fast} &  $\co(M^3 N + (P - 1) M^2 N)$ + \\
 $M \leq K$ & $\co(P K M N \log N) + \co(P K M N)$ \\
\hline
\end{tabular}
\end{table}

The per-iteration computational complexities of the methods, including both sparse coding and convolutional dictionary learning subproblems, are summarized in~\tbl{complexitySup}. The complexity expressions for the methods of Papyan \etal~\cite{papayan-2017-convolutional} and Heide \etal~\cite{heide-2015-fast} are reproduced from those provided in those works, and the expression provided by \v{S}orel and \v{S}roubek~\cite{sorel-2016-fast} is modified to make explicit the dependence on the number of images $K$ (for the sparse coding subproblem) and the internal ADMM iterations $P$. Our methods have mostly linear scaling in the problem size variables, with the exception of the image size, $N$, for which the scaling is $N \log N$, which is shared by all of the methods that compute convolutions in the frequency domain. The corresponding scaling of the spatial domain method of Papyan \etal is $N n$, where $n$ is the number of samples in each filter kernel, \ie the additional $\log N$ scaling with image size of the frequency domain methods is replaced with a linear scaling with filter size. This suggests that frequency domain methods are to be preferred for images of moderate size and moderate to large filter kernels, while spatial domain methods have an advantage for very large images and small filter kernels.

\section{Multi-Channel Experiments}
\label{sec:colorcompare}

In this section we report on an experiment intended to demonstrate the multi-channel CDL capability discussed in~\sctn{mltchn} of the main document.  We only provide results for the two leading approaches proposed here (Cns-P and FISTA), and do not compare with the algorithms of Heide \etal~\cite{heide-2015-fast}, \v{S}orel and \v{S}roubek~\cite{sorel-2016-fast}, or Papyan \etal~\cite{papayan-2017-convolutional} since none of the corresponding publicly available implementations support multi-channel CDL.  All of the color images used for these experiments were derived from images in the MIRFLICKR-1M dataset and pre-processed (cropping, scaling and highpass filtering per channel) in the same way (except for conversion to greyscale) as described in~\sctn{cdlexp} in the main document. The parameters of the Cns-P method were set using the parameter selection rules for the single channel problem, without any additional tuning. These rules were
also used to set the parameters of the FISTA method, but the rule for $L$ was multiplied by 3 for a more stable convergence.

\begin{figure}[htb]
        \center
        \scalebox{0.26}{\includegraphics{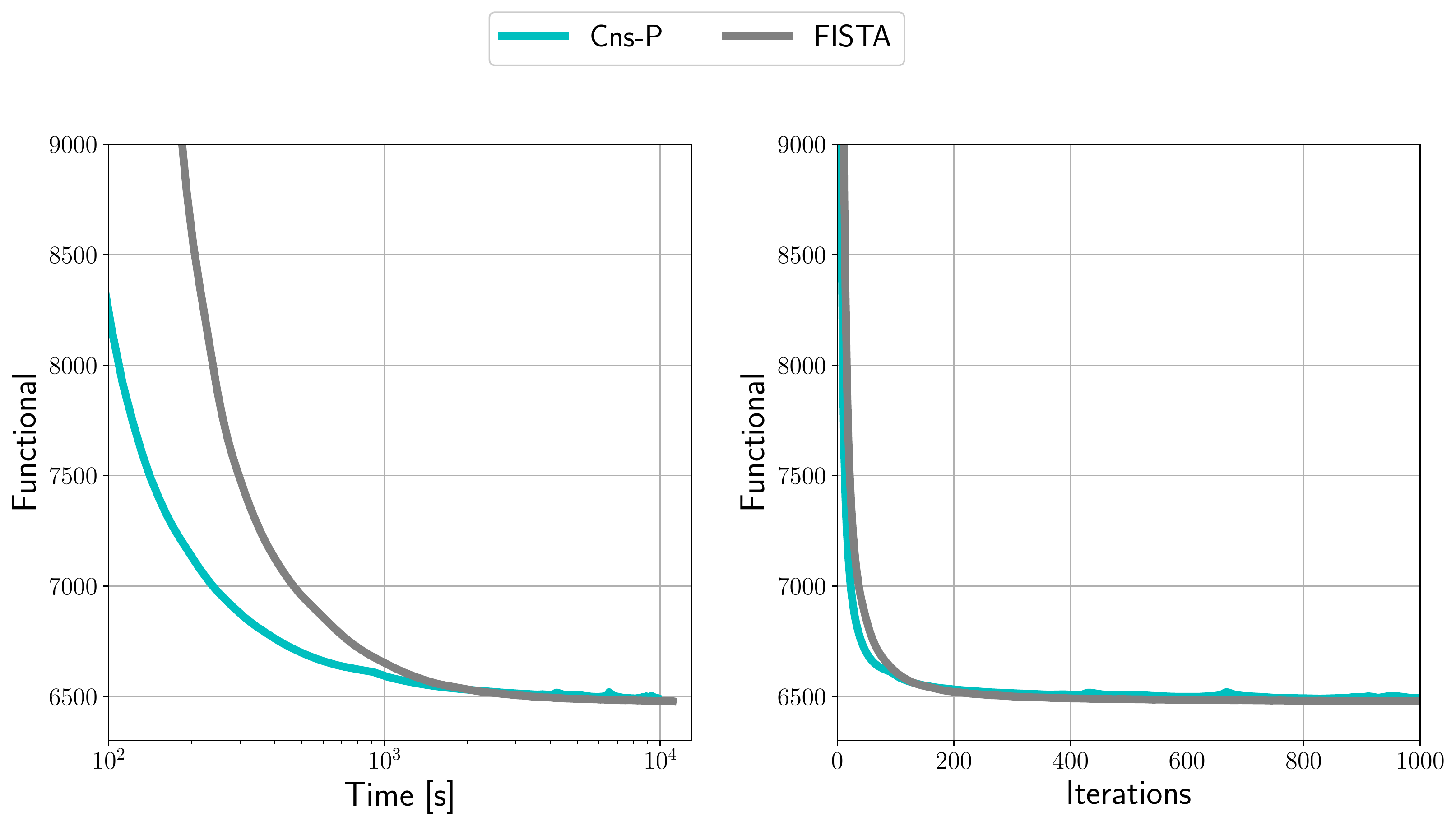}}
        \caption{Dictionary Learning ($K=40$): A comparison on a set of $K=40$ color images, 256 $\times$ 256 pixels, of the decay of the value of the multi-channel CPBDN functional~\eq{cscmchn} with respect to run time and iterations.}
        \label{fig:fObj_k40_color}
\end{figure}

\begin{figure}[htb]
        \center
        \scalebox{0.26}{\includegraphics{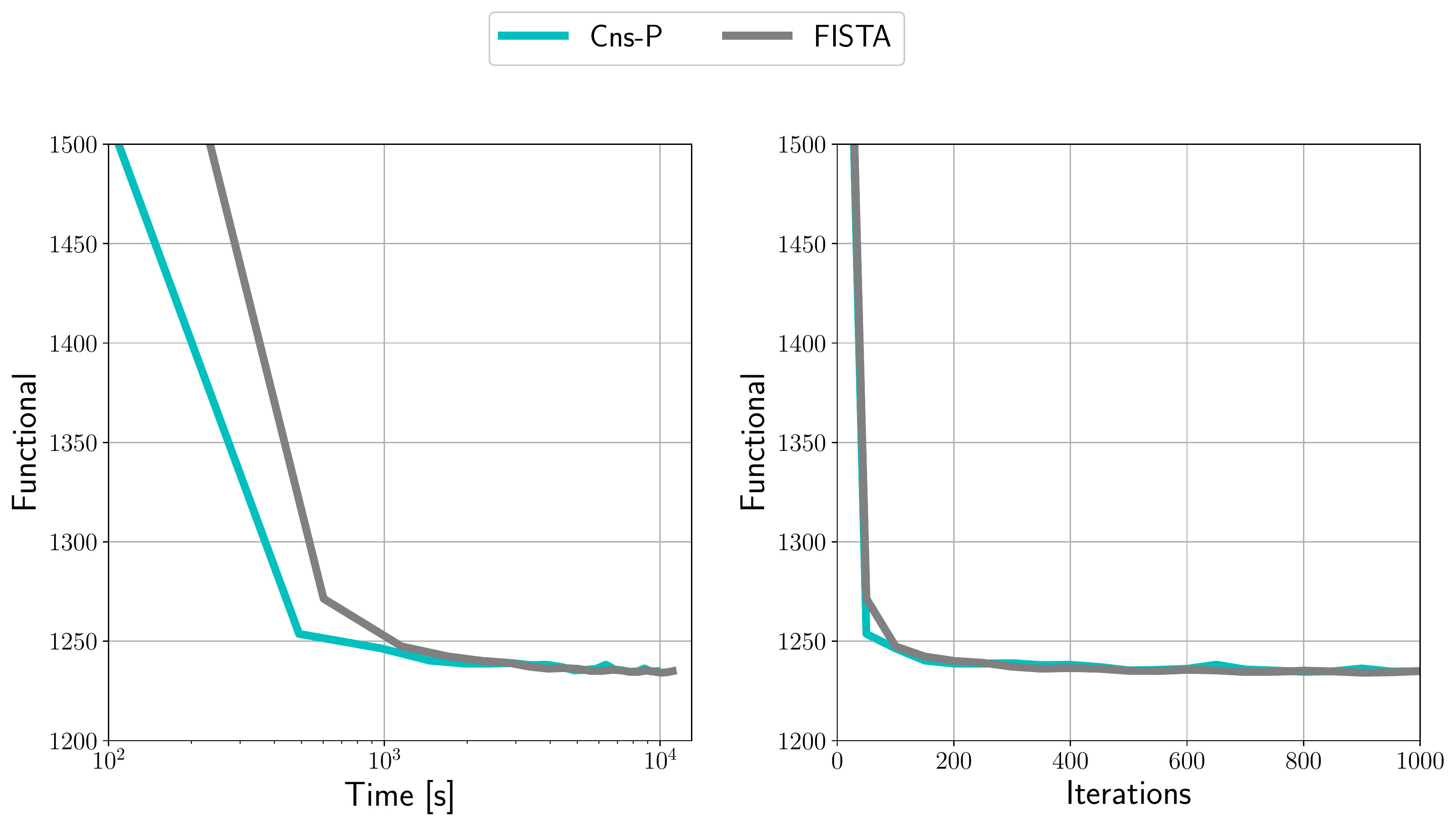}}
        \caption{Evolution of the multi-channel CBPDN functional~\eq{cscmchn} for the test set using the partial dictionaries obtained when training for $K=40$ color images, 256 $\times$ 256 pixels, as in~\fig{fObj_k40_color}.}
        \label{fig:val_k40_color}
\end{figure}

A dictionary of $M=64$ filters of size $8 \times 8$ and $C=3$ channels was learned from a set of $K=40$ color images of size $256 \times 256$, using a sparsity parameter setting $\lambda = 0.1$. The results for this experiment are reported in~\fig{fObj_k40_color}. Comparing with single-channel dictionary learning results for a dictionary of the same size, and a training image set of the same number of images of the same size, reported in~\fig{fObj_k40} in the main document, it can be seen that Cns-P requires about $2/3$ of time to compute the greyscale result compared to the color result, while FISTA requires about $3/4$ of time to compute the greyscale result compared to the color result. This additional cost for learning a color dictionary from color images is quite moderate considering that three times more training data is used.

Similarly to the other experiments, we saved the dictionaries at regular intervals during training and used an additional set of 10 color images, of size 256 $\times$ 256 pixels and from the same source, for testing. We compared the methods by sparse coding the color images in the test set, with $\lambda = 0.1$, and computing the evolution of the CBPDN functional over the series of multi-channel dictionaries. \fig{val_k40_color} shows that Cns-P performs slightly better than FISTA in testing too, although, Cns-P convergence is less smooth in the final stages compared to the single-channel cases, perhaps due to suboptimal parameter selection. Further evaluation of the multi-channel performance, including parameter selection guidelines, is left for future work.

\end{document}